\documentclass[twoside,11pt]{article}
\usepackage{float}

\usepackage[preprint]{jmlr2e}

\usepackage{amsmath,amssymb,mathtools,bm}
\usepackage{booktabs,tabularx,array,multirow}
\usepackage[shortlabels]{enumitem}
\usepackage{xcolor}
\usepackage{tikz}
\usetikzlibrary{arrows.meta,calc,positioning}
\newcounter{algorithm}
\renewcommand{\thealgorithm}{\arabic{algorithm}}
\newenvironment{algorithm}[1][tb]{%
  \begin{figure}[!#1]\centering
  \begin{minipage}{0.97\linewidth}\small
  \setlength{\parindent}{0pt}\setlength{\parskip}{.75pt}%
  \hrule height 0.8pt\relax
}{%
  \par\smallskip\hrule height 0.8pt\relax
  \end{minipage}\end{figure}}
\newcommand{\algcaption}[1]{%
  \par\smallskip\refstepcounter{algorithm}%
  \textbf{Algorithm~\thealgorithm:}~#1\par\smallskip
  \hrule height 0.4pt\relax\smallskip}
\newcommand{\DontPrintSemicolon}{}
\newcommand{\SetKwInOut}[2]{\expandafter\def\csname #1\endcsname##1{\par\hangindent=1.5em\hangafter=1\textbf{#2:}~##1\par}}
\newcommand{\BlankLine}{\par\smallskip}
\newcommand{\tcp}[1]{\par\hfill\textit{// #1}\par}
\newcommand{\tcps}[1]{\hfill\textit{(#1)}\par}
\newcommand{\For}[2]{%
  \par\noindent\textbf{for }#1\textbf{ do}\par
  \par\noindent
  \vrule width 0.4pt\relax
  \hspace{0.75em}%
  \begin{minipage}[t]{\dimexpr\linewidth-0.75em-0.4pt\relax}
    \setlength{\parindent}{0pt}%
    \setlength{\parskip}{.75pt}%
    #2\par
  \end{minipage}%
  \par}
\usepackage{placeins}
\usepackage{needspace}
\usepackage{microtype}
\newlength\figpanelw
\newlength\figlegw

\graphicspath{{figures/}}

\allowdisplaybreaks[1]
\numberwithin{equation}{section}
\numberwithin{theorem}{section}

\newcommand{\R}{\mathbb R}
\newcommand{\E}{\mathbb E}
\newcommand{\PP}{\mathbb P}
\newcommand{\M}{\mathcal M}
\newcommand{\T}{\mathsf T}
\newcommand{\Cov}{\operatorname{Cov}}
\newcommand{\Var}{\operatorname{Var}}
\newcommand{\Ran}{\operatorname{Ran}}
\newcommand{\reach}{\operatorname{reach}}
\newcommand{\polylog}{\operatorname{polylog}}
\newcommand{\II}{\mathrm{II}}
\newcommand{\ind}{\mathbf 1}
\newcommand{\Ptop}{P_{\mathrm{Top}_m}}
\newcommand{\Pbot}{P_{\mathrm{Bot}_m}}
\newcommand{\Sfit}{\mathcal S^{\rm fit}_k}
\newcommand{\Spool}{\mathcal S^{\rm pool}_k}
\newcommand{\Srd}{\mathcal S_{\rm pred}}
\newcommand{\Sanc}{\mathcal S_{\rm anc}}
\newcommand{\Sdes}{\mathcal S_{\rm des}}
\newcommand{\Hscore}{\mathcal H}
\newcommand{\Wwin}{\mathcal W}
\newcommand{\Aref}[1]{\hyperref[ass:A#1]{(A#1)}}
\DeclareMathOperator{\clip}{clip}
\DeclareMathOperator*{\argmin}{arg\,min}

\makeatletter
\let\MFM@makecaption\@makecaption
\long\def\@makecaption#1#2{\small\MFM@makecaption{#1}{#2}}
\makeatother

\newenvironment{informaltheorem}
  {\par\medskip\noindent\textbf{Theorem (informal).}\ \itshape}
  {\par\medskip}

\ShortHeadings{The Manifold-Index Model}{Wu and Maggioni}
\firstpageno{1}

\begin{document}

\title{The Manifold-Index Model: \\ provably and efficiently avoiding the statistical and computational curse of dimensionality}

\author{\name Yantao Wu \email ywu212@jhu.edu \\
       \addr Department of Mathematics\\
       Johns Hopkins University\\
       Baltimore, MD 21218, USA
       \AND
       \name Mauro Maggioni \email mauromaggionijhu@icloud.com \\
       \addr Department of Mathematics and Department of Applied Mathematics \& Statistics\\
       Johns Hopkins University\\
       Baltimore, MD 21218, USA}

\editor{}

\maketitle

\begin{abstract}
We study vector-valued regression with predictors filling a tube around an
unknown $m$-dimensional manifold $\M$ in $\R^D$. In the Manifold-Index Model,
the mean response is $F=f\circ\Pi_\M$, where $\Pi_\M$ is the nearest-point
projection onto $\M$ and $f:\M\to\R^p$ is a possibly non-injective link function.
Multiscale ANisotropic lOcal Regression (MANOR) estimates tangent planes by multiscale PCA of predictors with
similar responses, then refines them by pooling outer products of local slopes.
The neighborhoods narrow tangentially while retaining complete normal fibers.
We assume a Lipschitz link differential and bounded response noise.
For $p\ge m$, under our geometric and sampling conditions, $n$ independent observations
give mean tangent-space error of order $n^{-2/(m+4)}$ and integrated squared
prediction error at the $m$-dimensional nonparametric rate $n^{-4/(m+4)}$, up to
logarithmic factors. In a scaling regime with fixed intrinsic dimension, model and scale parameters,
and ratio of manifold volume to the $m$-th power of its reach, the required
sample size grows polynomially in $D+p$, and computation is polynomial in $n$, $D$ and $p$.
Numerical experiments on curves with non-monotone links and surfaces with vector-valued responses
compare observed convergence rates with the theoretical bounds; a Langevin dynamics example tests prediction of hitting probabilities under an
approximate manifold-index model.
\end{abstract}



\section{Introduction}\label{sec:intro}

We seek to estimate a function $F$ from $n$ independent observations
\[
  (X_i,Y_i),\qquad Y_i=F(X_i)+\zeta_i,\qquad i=1,\dots,n,
\]
where the predictors $X_i\in\R^D$ are independent copies of a random vector
$X$, the responses $Y_i$ take values in $\R^p$, and the noise vectors
$\zeta_i$ are centered and independent of each other and of the predictors.
We write $(X,Y,\zeta)$ for a generic observation and its noise.
Suppose that all we know about $F$ is that it is
of H\"older smoothness $s$ on a full-dimensional region of $\R^D$.
The best achievable accuracy is then governed by a simple bias--variance trade-off.
A local estimator $\widehat F$ of suitable order on a neighborhood of width $b$ has a squared bias
of order $b^{2s}$, and its variance is inversely proportional to the number of
observations in the neighborhood, which is of order $nb^D$
at a typical interior point when the predictors have a locally bounded positive ambient density.
Balancing the two terms gives
\begin{equation}\label{eq:intro-ambient}
  b^{2s}\asymp\frac1{nb^D}
  \quad\Longrightarrow\quad
  b\asymp n^{-1/(2s+D)},
  \qquad
    \E\|\widehat F(X)-F(X)\|^2\asymp n^{-2s/(2s+D)} ,
\end{equation}
or $n^{-s/(2s+D)}$ for the root mean squared error. This is the minimax rate
over the class of $s$-smooth functions
\citep{Stone82,GKKW02,Tsybakov2009}, and kernel, Fourier and multiscale
estimators attain it
\citep{GKKW02,DeVore:UniversalAlgorithmsLearningTheoryI,DeVore:UniversalAlgorithmsLearningTheoryII,DeVore:OptimalLearning}.
The learning rate $2s/(2s+D)$ degrades quickly as $D$ grows. This is an instance of the \emph{curse of dimensionality}. Since the
rate is minimax, it cannot be improved uniformly over all $s$-smooth
functions; it can only be improved for regression problems with additional
structure. 
Even when additional assumptions on $F$ give a learning rate with exponent independent
of $D$, the constants in the bounds on the estimation error may grow exponentially with $D$, so that the sample size
sufficient to guarantee a prescribed accuracy still grows exponentially with $D$: this is also part of the curse of dimensionality.
When the existence of an estimator that attains a learning rate independent of $D$ is proven, the next problem becomes to produce an algorithm for constructing the estimator at computational cost not exponential in $D$.

It is natural to ask which kind of special structure in $F$ is both plausible in applications
and exploitable by estimators so that the statistical and computational costs do
not grow exponentially with $D$. This of course has been subject of much research, and we discuss some of the ideas most related to the present work in section \ref{sec:intro-structure}.

As we discuss at length in section \ref{sec:intro-structure}, we are interested in understanding when additional structure in $F$ helps avoid the curse of dimensionality; and in particular when $F$ has compositional structure.
We study the \emph{Manifold-Index Model}: the predictors fill a tube around an unknown $m$-dimensional manifold in $\R^D$, and their mean responses depend only on
their nearest points on the manifold. We discuss the detail momentarily in Section~\ref{sec:intro-model}. This model extends multi-index
regression by replacing the linear projection onto a fixed linear subspace with
the nonlinear nearest-point projection onto the manifold.
For curves, this construction is essentially the Nonlinear Single-Variable Model
of \citet{WuMaggioniNSVM}. From $n$ observations, Multiscale ANisotropic
lOcal Regression (MANOR) attains the rate of $m$-dimensional nonparametric
regression without knowledge of the manifold  or the link. At fixed model
scales, normalized response range and ratio of the manifold's volume to
the $m$-th power of its reach, which are natural scalings of the problem, the sample
requirement grows polynomially with $D$; the computational cost is polynomial
in $D$ and $n$ (Section~\ref{sec:intro-result}).
MANOR initializes the tangent spaces by multiscale local principal component
analysis (PCA) \citep{LittleJungMaggioni2009,LittleLeeJungMaggioni2009,ChenLittleMaggioniRosasco2011,LittleMaggioniRosasco2017} of predictors with similar responses, blending inverse regression \citep{Li91,DL91,Cook00} ideas with geometric analysis of the (approximate) level sets. It refines these estimated tangent spaces by pooling outer products of local regression slopes to form adapted, anisotropic neighborhoods \citep{HJS01,XiaTongLiZhu2002,DalalyanJuditskySpokoiny2008}, and finally uses them for prediction with highly anisotropic local kernels.

We next define the Manifold-Index Model and state our main results informally;
Section~\ref{sec:intro-related} reviews related models and estimators and puts this work in a broad context.

\subsection{The Manifold-Index Model}\label{sec:intro-model}

\begin{figure}[!t]
\centering
\includegraphics[width=.99\linewidth]{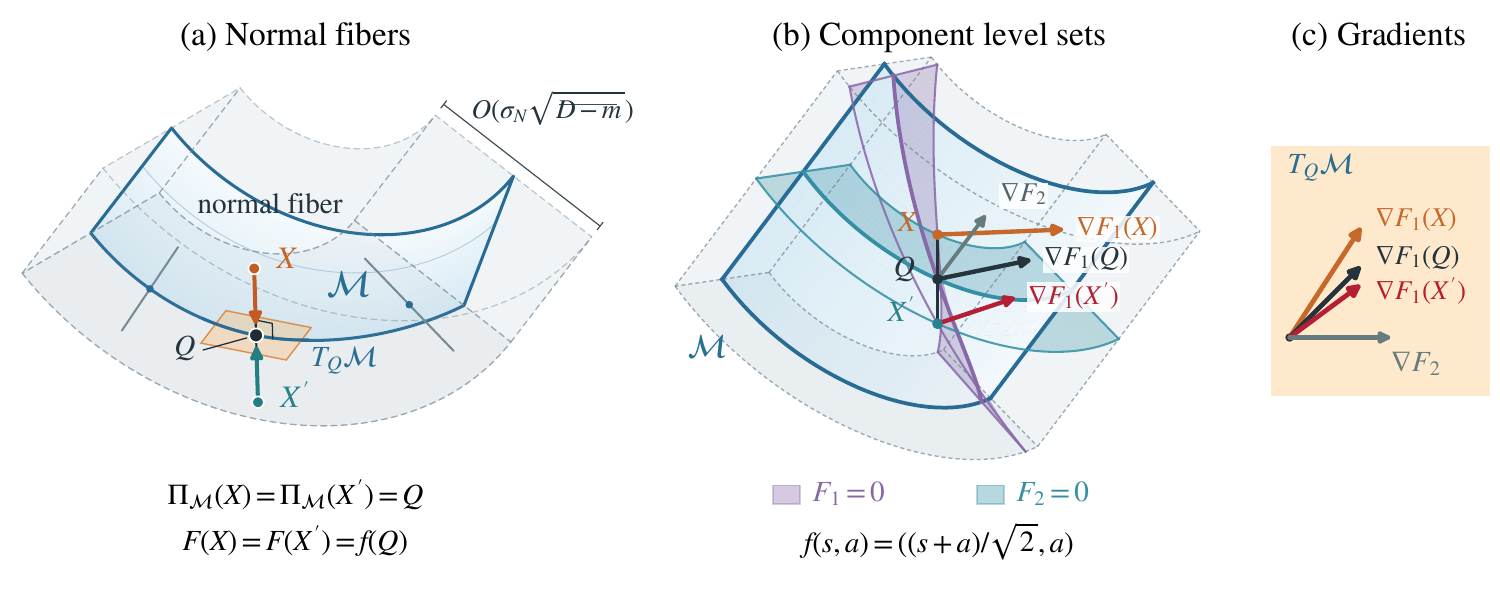}
\caption{Geometry, response values and gradients in the Manifold-Index
Model. All panels use a cylindrical surface patch $\M$ with $m=p=2$,
$D=3$, and $f(s,a)=((s+a)/\sqrt2,a)$, where $s$ is circumferential
arc length and $a$ is the axial coordinate.
(a) Normal fibers and nearest-point projection.
The arrows project $X=Q+Z$ and $X'=Q-Z$ to $Q$, so
$F(X)=F(X')=f(Q)$. The orange plane is $T_Q\M$; the dashed grey boundary
encloses a tube of diameter $O(\sigma_N\sqrt{D-m})$ (Assumption~\Aref{2}).
(b) Component level sets. The purple sheet $F_1=0$ and teal
sheet $F_2=0$ meet $\M$ along the curves $s+a=0$ and $a=0$,
respectively, and extend along normal fibers.
Their intersection is the normal fiber through $Q$,
where $F=(0,0)$. The orange, black and red arrows show $\nabla F_1$
at $X,Q,X'$, respectively; the grey-green arrow shows $\nabla F_2$.
(c) The same gradients translated to a common origin in $T_Q\M$ (orange).
For a unit-radius cylinder and signed outward normal offset $r$,
$\nabla F_1=(e_s/(1+r)+e_a)/\sqrt2$ and $\nabla F_2=e_a$, where
$e_s,e_a$ are the unit circumferential and axial tangent vectors at $Q$.
Thus $\nabla F_1$ changes along the fiber, while $\nabla F_2$ is unchanged;
each gradient is orthogonal to its component level surface in (b), and
the two gradients span $T_Q\M$ at each location.}
\label{fig:model}
\end{figure}

Let $\M\subset\R^D$ be an $m$-dimensional submanifold whose reach $\reach_\M$
is positive \citep{Federer1959,NSV}. Every point at distance less than
$\reach_\M$ from $\M$ has a unique nearest point on $\M$, so the nearest-point
projection
\[
  \Pi_\M(x)=\argmin_{q\in\M}\|x-q\|
\]
is well defined on the tube of radius $\reach_\M$ around $\M$. Every $X$ in this
tube decomposes as
\begin{equation}\label{eq:intro-decomposition}
  X=Q+Z,\qquad Q=\Pi_\M X\in\M,
\end{equation}
where the \emph{manifold index} $Q$ is the \emph{base point}, the nearest
point of $\M$ to $X$, and $Z=X-Q$ is the displacement from it
(Figure~\ref{fig:model}). Write $T_Q\M$ for the tangent space at $Q$
and $N_Q\M=(T_Q\M)^\perp$ for its normal space.

\begin{definition}[Manifold-Index Model]\label{def:mim}
A random pair $(X,Y)\in\R^D\times\R^p$ follows the \emph{Manifold-Index
Model} if
\[
  \E[Y\mid X]=F(X)=f(\Pi_\M X),
\]
where $\M\subset\R^D$ is an $m$-dimensional submanifold with
$\reach_\M>0$, $f:\M\to\R^p$ is the link function, and
$X$ has distance at most $R$ from $\M$, for some $R<\reach_\M$.
\end{definition}

Throughout, the \emph{normal fiber} over $q\in\M$ is the affine disk
$\{q+z:z\in N_q\M,\ \|z\|\le R\}$ in the predictor tube.
Write $Df$ for the differential of $f$, represented by a $p\times m$
matrix in orthonormal tangent coordinates, and $\nabla F(x)\in\R^{D\times p}$
for the matrix whose columns are the gradients of the $p$ components of $F$.
We call $Df(q)$ rank deficient when $\operatorname{rank}Df(q)<m$;
isolation of such points always refers to the base manifold $\M$.
It is partially rank deficient if $0<\operatorname{rank}Df(q)<m$.
For a matrix $M$, $\Ran M$ denotes its range and $M^\T$ its transpose.

\paragraph{Problem statement.} \emph{Under the Manifold-Index Model, the manifold $\M$, the link $f$ and the distribution of $X$ are unknown. Our goal is to estimate $F$ from independent observations $(X_i,Y_i)$ with $Y_i=F(X_i)+\zeta_i$, where the centered noise vectors are independent of the predictors and of one another; they are sub-Gaussian, and on the initialization sample their norms are at most one eighth of the response-window width (Assumption~\Aref{5}).}
\vskip0.25cm

Three features of the model deserve emphasis.

\emph{The predictors are full-dimensional.} The normal displacement $Z$
ranges over a $(D-m)$-dimensional ball of radius $R$, and the
distribution of $X$ does not concentrate on $\M$ as $n$ grows. If $X$ has
a locally bounded positive ambient density, a Euclidean
ball of sufficiently small radius $b$ around a typical interior data point contains of order
$nb^D$ observations, and the covariance of the predictors in such a ball does
not reveal the tangent space of $\M$. This contrasts with models in which the data concentrate near a
low-dimensional manifold, and it parallels the single-index model,
in which the predictors typically have large variance orthogonally to the
index.

\emph{The regression function is intrinsically $m$-dimensional.} It is
constant on each normal fiber and is
determined by its values on $\M$. The responses reveal tangential directions wherever $Df$ has full rank.

\emph{The model interpolates between index models and general compositions.}
If $\M$ is an affine subspace, $\Pi_\M$ is an affine orthogonal projection and
the model is a multi-index model: $F$ depends only on the projection
onto that affine subspace. Its index space is the direction space of $\M$. For a curved $\M$, the inner map $\Pi_\M$ is
nonlinear as well, but it has a precise geometric form.
Section~\ref{sec:core} shows that this form is what makes the problem
learnable: the gradient of $F$ at $Q+Z$ lies in the tangent space $T_Q\M$, so
the fixed index space is replaced by a field of tangent spaces
that moves with the base point.

%

\subsection{Main result}\label{sec:intro-result}

For $Q\in\M$, let $P_Q$ denote the orthogonal projector onto
$T_Q\M$. At an independent query $X$, we assess estimates $\widehat P(x)$ of
$P_{\Pi_\M x}$ and $\widehat F$ of the regression function by
\begin{equation}\label{eq:intro-metrics}
  \mathcal E_{\rm tan}(\widehat P):=\E\bigl\|\widehat P(X)-P_{\Pi_\M X}\bigr\|,
  \qquad
  \mathcal E_{\rm pred}(\widehat F):=\frac{\E\|\widehat F(X)-F(X)\|^2}{\E\|F(X)-\E F(X)\|^2}.
\end{equation}
The expectation averages over the training sample, the random split used for
initialization, and an independent query. The first error is the mean sine of
the largest principal angle between the estimated and true tangent spaces;
the second is the mean squared prediction error relative to the variance of
the signal.

\begin{informaltheorem}
Under \Aref{1}--\Aref{5}, let $p\ge m$, $D>m$, with a
tube-to-reach ratio $0<R/\reach_\M<2/3$. The differential $Df$ may be rank deficient at finitely many points.
With the scales and construction of Theorem~\ref{thm:isolated-rank},
and positive signal variance,
\[
\mathcal E_{\rm tan}=\widetilde O\bigl(n^{-2/(m+4)}\bigr) \quad \text{and} \quad \mathcal E_{\rm pred}=\widetilde O\bigl(n^{-4/(m+4)}\bigr).
\]
Here $\widetilde O$ suppresses fixed powers of $\log n$ for each fixed model.
At fixed intrinsic dimension, model scales and $\operatorname{vol}(\M)/\reach_\M^m$,
the constants grow at most polynomially in $D+p$. For a fixed grid of
scan radii and $n\ge D$, constructing the estimator costs
$O(n^2D(D+p)\log n)$ operations.
\end{informaltheorem}

Both exponents are governed by $m$. Constancy on normal fibers allows a
neighborhood of tangential width $b$ to retain about $nb^m$ observations
without incurring normal approximation bias. The model itself requires only
$R<\reach_\M$ for the nearest-point projection to be defined. The
analysis also requires the bound \eqref{eq:bend-margin} on root mean square
tangential stretch and the uniform bound \eqref{eq:opening-margin} on
variation of the tangential derivative along normal fibers near
rank-deficient points.

At a regular base point the component gradients span the tangent space. At
an isolated point where $Df$ is rank deficient, slopes at neighboring base points supply the missing
directions. The geometric conditions control the typical tangential stretch
of normal offsets in the regular region \Aref{2} and the variation of the
gradient of $f\circ\Pi_\M$ along each fiber near a point where $Df$ is rank deficient \Aref{3}.

\subsection{Contributions}\label{sec:intro-contributions}

The Manifold-Index Model is a nonlinear
generalization of the multi-index model in which the index space is replaced
by the tangent spaces of an unknown manifold, the predictors fill a tube of
radius below the reach, and the link function may be non-injective. It lies between
semiparametric index models and general nonparametric compositions.
\begin{enumerate}[(i)]
\item \emph{An estimator.} Multiscale Anisotropic lOcal Regression (MANOR) estimates the tangent field by multiscale
local PCA on response preimages, the sets of predictors whose responses lie within a small distance of a given response, refines it by pooling outer products of local slopes, and predicts with an anisotropic kernel smoother. These
steps use local covariances and affine least squares.
\item \emph{Guarantees.} With $s=2$, Theorem~\ref{thm:isolated-rank} gives
intrinsic-dimensional rates for mean tangent-space error and squared
prediction risk. At fixed intrinsic dimension, model scales and
$\operatorname{vol}(\M)/\reach_\M^m$, the leading risk coefficient and the sample threshold grow
at most polynomially in $D+p$.
\item \emph{Relation to the curve case.} For curves and coarsely monotone
links of smoothness $s\in[1/2,1]$, the rate of \citet{WuMaggioniNSVM}
has a noise-dependent resolution limit. Here the prediction error continues
to decrease with sample size at a fixed noise level, for manifolds of any
intrinsic dimension and links that may have several preimages. The
refinement schedule allows $1<s\le2$; the theorem and experiments use $s=2$.
\item \emph{Multiscale initialization.} Restricting a response preimage to a
small patch of the manifold gives local PCA, performed at the right scale \citep{LittleJungMaggioni2009,LittleMaggioniRosasco2017}, a spectral gap even when the predictors fill a tube of positive radius. This supplies the initial tangent
spaces for anisotropic refinement \citep{HJS01}.
\item \emph{Rank-deficient $Df$ at isolated points.} Neighboring slopes can have opposite signs
in a direction missing at the central point. Pooling their outer products
preserves the signal in that direction; the fiberwise variation bound
keeps it aligned with the local tangent space.
\item \emph{Tangent-space estimation.} We consistently estimate tangent planes when
predictors fill a tube whose radius is comparable to the reach. The estimator
uses the noisy responses and the regularity of the link to recover this geometry.
\end{enumerate}

Experiments on a line segment, a curved index and two surfaces examine the
dependence of both errors on sample size. A diffusion example tests prediction
when the manifold-index structure is approximate.
%

%
\subsection{Context and related work}\label{sec:intro-related}\label{sec:intro-structure}

We review three kinds of structure that can avoid the curse of
dimensionality: low-dimensional data, regularity assumptions, and
compositional models. We discuss the last in more detail, distinguishing
linear and nonlinear inner maps and the estimators used to learn them.

\paragraph{Low-dimensional data.}
The distribution of $X$ may be concentrated on, or near, a low-dimensional
set, say an $m$-dimensional manifold with $m\ll D$, while $F$ is a generic
$s$-smooth function on that set. For data supported on the manifold
with density bounded above and below, a small Euclidean ball of radius $b$
contains about $nb^m$ observations instead of $nb^D$. With noise in the
normal directions, the $nb^m$ count applies only at scales larger than the
noise thickness. For manifold-supported data, local estimators adapt to the
intrinsic dimension and attain the rate $n^{-2s/(2s+m)}$
\citep{BL07,Kpotufe11,KG13,LMV16,LMV22ME,liu2024deepneuralnetworksadaptive,RosaRousseau2025};
some of these estimators also come with fast algorithms and adapt to
unknown regularity that may vary across locations and scales
\citep{LMV16,LMV22ME}.
Here the reduction comes from the geometry of the distribution of $X$, not
from the structure of $F$.
Local PCA estimates tangent spaces when the predictors lie on or near a
manifold, and regression on manifold-supported data is well developed
\citep{AamariLevrard2019,PuchkinSpokoiny2022,MukherjeeWuZhou2010,AswaniBickelTomlin2011,ChengWu2013}.
Those results do not directly provide the coarse tangent estimate needed
in a fixed-radius tube. Proposition~\ref{prop:pilot} establishes it from
response-preimage local covariances, after which anisotropic neighborhoods
aligned with these estimates
support finer estimation.

\paragraph{Highly regular functions.}
One may restrict $F$ to classes simpler than $s$-smooth functions on $\R^D$: let the smoothness grow proportionally to $D$; impose
mixed smoothness \citep{stromberg,Marshall2019MixedHolder,Marshall2019HigherDimensions},
or anisotropic smoothness with ordinary smoothness
along $O(1)$ coordinate directions and smoothness of order $D$ along all
others, for which neural networks avoid the curse of dimensionality in
approximation \citep{10.5555/3540261.3540537}; or impose integrability
conditions on the Fourier transform, as in Barron spaces
\citep{Barron93,10.5555/3122009.3122028}. In general, these restrictions can
become demanding as the ambient dimension grows.

For a fixed index space, Barron-type conditions can exploit concentration of
the gradient in that space. Curvature removes this fixed concentration, and
Fourier transforms of measures supported on curved sets have different decay
behavior \citep{ACK87,BGGIST07}. The argument that places index models in
Barron spaces therefore does not extend even to a curve $\M$ of dimension $1$;
see \citet{WuMaggioniNSVM} for discussion.

\paragraph{Compositional structure.}
A third approach assumes that $F$ is a composition
\begin{equation}\label{eq:intro-composition}
  F=f\circ g,\qquad g:\R^D\to\R^m,\qquad f:\R^m\to\R^p,\qquad m<D,
\end{equation}
in which $g$ extracts a small number of features and the link function $f$ acts on
them. Classical instances for which the curse of dimensionality can provably
be avoided include additive and generalized additive models
\citep{Stone85Additive,doi:10.1080/01621459.1985.10478157,Hastie:GeneralizedAdditiveModels,Horowitz_generalclassnonparametric}
and the single- and multi-index models discussed below
\citep{Ichimura93,HHI93,Li91,XiaTongLiZhu2002}. Composition alone, however, does not
remove the curse of dimensionality: \citet{JLT09} show that the smoothness of
$g$ can improve the minimax rate for general compositions, but the improved rate
still depends on $D$. Results for deep networks
\citep{10.1214/19-AOS1875,shen2021deepquantileregressionmitigating}, for
hierarchical interaction models \citep{KohlerKrzyzak2017,BauerKohler2019} and
for architectures inspired by the Kolmogorov--Arnold representation
\citep{Lai2021TheKS,liu2024kankolmogorovarnoldnetworks} quantify approximation
and estimation over classes of compositions of smooth functions. Some
of these classes restrict the innermost functions to small subsets of the
given coordinates, whereas generalized hierarchical interaction models also
allow arbitrary low-dimensional linear projections at the innermost level
\citep[Definition~2]{BauerKohler2019}. On the computational side, these results often concern
approximation, or the risk of an estimator, and neither is typically proved to be computable
efficiently.
 Moreover, their constants, if not the rates, often depend exponentially on $D$ \citep{Shamir_discussion}: this is also part of the curse of dimensionality, as the sufficient sample size supplied by such a bound to guarantee a prescribed accuracy can grow exponentially with $D$. Even recent works that provably avoid the curse of dimensionality  still have constants, often left implicit, that grow exponentially with $D$.
When $g$ is a polynomial and $f$ belongs to a known finite-dimensional space, specially trained networks provably avoid the curse
\citep{ChenMekaPolys,wang2024learning,lee2024neuralnetworklearnslowdimensional}, both in the rates and in the constants.
For a generic nonlinear $g$, learning $g$ is itself a $D$-dimensional
nonparametric problem: a small range dimension alone does not remove the
ambient degrees of freedom.

\paragraph{Linear inner maps: index models.}
The best understood compositional model is the one in which $g$ is linear: for
a matrix $G\in\R^{D\times m}$ of rank $m$,
\begin{equation}\label{eq:intro-multiindex}
  F(x)=f(G^\T x),\qquad f:\R^m\to\R^p.
\end{equation}
Here $F$ depends on $x$ only through its projection onto the \emph{index
space} $\Ran G$, and the chain rule shows that every gradient lies in this
fixed subspace:
\begin{equation}\label{eq:intro-multiindex-gradient}
  \nabla F(x)=G\,Df(G^\T x)^\T,
  \qquad
  \Ran\nabla F(x)\subseteq\Ran G\quad\text{for every }x.
\end{equation}
If $\Ran G$ were known, a neighborhood of width $b$ along $\Ran G$,
unrestricted orthogonally, would contain about $nb^m$ observations, and
\eqref{eq:intro-ambient} would hold with $D$ replaced by $m$:
\begin{equation}\label{eq:intro-intrinsic}
  b^{2s}\asymp\frac1{nb^m}
  \quad\Longrightarrow\quad
  b\asymp n^{-1/(2s+m)},
  \qquad
  \E\|\widehat F(X)-F(X)\|^2\asymp n^{-2s/(2s+m)}.
\end{equation}
\citet{Stone82} conjectured that this intrinsic rate is optimal for
multi-index models. For the single-index model ($m=1$) it is attained by kernel
estimators \citep[Theorem~3.3]{HS89}, \citep[Section~2.5]{Horowitz98},
\citep[Corollary~22.1]{GKKW02}, and it is minimax \citep[Theorem~2]{GL07};
under suitable conditions the index itself can be estimated at the parametric
rate $n^{-1/2}$. Index models also provide test beds for feature learning by neural networks
\citep{arous2021online,bietti2022learning,bietti2023learninggaussianmultiindexmodels,pmlr-v247-damian24a,lee2024neuralnetworklearnslowdimensional,mousavi2024learning,ren2024learningorthogonalmultiindexmodels};
trained networks appear to exploit such linear structure
\citep{doi:10.1126/science.adi5639}.

Semiparametric least squares, $M$-estimators and average-derivative methods
estimate a fixed index space under varying smoothness and optimization
conditions \citep{Ichimura93,HHI93,DH99,DHH97,DHP06,CFGW97,Stoker86,HS89}.
The minimum average variance estimator and outer-product-of-gradients methods
use local affine fits and pooled gradient information
\citep{XiaTongLiZhu2002,Xia2007}; structural adaptation alternates between
estimating an index space and adapting elongated neighborhoods
\citep{HJS01,DalalyanJuditskySpokoiny2008}. Inverse regression methods, such as
sliced inverse regression \citep{DL91,Li91,CLS13}, sliced average variance
estimation \citep{Cook00}, and contour and directional regression
\citep{LZC05,LW07}, use statistics of $X$ within slices of the response.
Smallest Vector Regression gives an explicit conditional construction with polynomial dependence on $D$
\citep{LMV22B}. These methods provide the index-model precedents for our
response preimages, gradient pooling and anisotropic refinement; here the
estimated space varies with the base point rather than being global.

\paragraph{Nonlinear inner maps.}
Compositional models are well understood when $g$ is linear, but the
corresponding guarantees are more limited when $g$ is nonlinear. Nonlinear
sufficient dimension reduction, in kernel form
\citep{FukumizuBachJordan2009,WuMukherjeeLiang2008} or through nonlinear
extensions of inverse regression
\citep{LiArtemiouLi2011,LeeLiChiaromonte2013}, and active manifolds
\citep{BridgesEtAl2019} seek nonlinear features that capture the dependence of
$Y$ on $X$. Within the settings considered here, they do not provide the
intrinsic-dimension guarantees with polynomial dependence on $D$ that we seek.
In previous work~\citep{WuMaggioniNSVM} we studied the simplest nonlinear case,
in which $g$ is the nearest-point coordinate on an unknown curve
$\gamma\subset\R^D$, so that $F(x)=f(\Pi_\gamma x)$ (the Nonlinear
Single-Variable Model, originally introduced in \citet{manuscriptNSIM2018}). Response preimages approximate the preimages of $F$,
local principal component analysis of a response preimage recovers the direction of the
curve, and a one-dimensional local-polynomial fit estimates $f$. For coarsely monotone link
functions of smoothness $s\in[1/2,1]$, that estimator attains the rate
$n^{-2s/(2s+1)}$ up to logarithmic factors, down to an explicit saturation
level determined by curvature, noise and the scale below which $f$ may fail to
be monotone. The estimator is constructed in time $O(D^2n\log n)$.

The present paper extends the curve case of \citet{WuMaggioniNSVM} to manifolds of any intrinsic dimension, with prediction rates whose exponent matches, up to logarithms, the minimax exponent over the broader class of Appendix~\ref{app:learnability}, and explicit algorithms for constructing the estimator, and constants with no exponential dependence on the ambient dimension $D$. The sample threshold separates the ambient dimension from the probability of observing a local patch. Both enter polynomially.

Models inspired by the nonlinear single-variable model of
\citet{WuMaggioniNSVM} have also been considered by
\citet{ShenEtAl2025TaskManifold}, who establish transformer approximation and
empirical risk generalization bounds under a manifold-projection model. Their approximation result relies on an oracle network construction, and they do not provide a computationally tractable learning algorithm whose output satisfies the stated guarantees. Moreover, none of the numerical tests (with transformers) in that work are actually performed for regression with a nonlinear underlying manifold. They consider the case of $s$-H\"older link functions $f$, for $s\le1$: first-order projection accuracy suffices at this regularity, whereas exploiting $s=2$ requires controlling the shape operator and the variation of normal spaces. They also consider completely noiseless responses $Y$. For $s=2$, we give a constructive estimator with noisy responses whose exponent in $n$ matches, up to logarithms, the minimax exponent over the broader regularity class in Appendix~\ref{app:learnability}, at fixed positive noise.

Also building on \citet{WuMaggioniNSVM},
\citet{KokotEtAl2026ContinuousIndex} analyze a localized expected-gradient-
outer-product (EGOP) construction under a supervised noisy-manifold hypothesis
and state a learning rate whose exponent depends on the intrinsic dimension.
The supplied proofs (in version arXiv:2601.07061v4), however, do not establish this guarantee: the covariance
scaling asserted in their Theorem~2 is false under its stated hypotheses, while
the proof of their manifold result, Theorem~3, inverts a compressed Hessian that
is necessarily singular under the stated geometric and Hessian-rank hypotheses.\footnote{Full Hessian invertibility is incompatible with the noisy-manifold model when
$D>2m$; the compressed invertibility used in the proof contradicts the
hypotheses of Theorem~3.} Moreover, even when the
manifold geometry is held fixed as $D$ increases, Gaussian localization can
produce variance factors that grow exponentially in $D$, and a coefficient
used in the published bias proof can grow faster than exponentially in $D$.
The published analysis therefore does not establish that the constants in
the error bounds grow more slowly than exponentially in $D$.
We control our rates and constants, both in the theory and the algorithm, showing that they are all controlled by low-degree polynomials in $D$, thereby fully avoiding the curse of dimensionality.

\paragraph{Connections to multi-task learning.}
Multi-task learning is a framework for learning multiple related tasks simultaneously \citep{Caruana1997,Baxter2000,AndoZhang2005}. It is a generalization of single-task learning, where the goal is to learn a single task, but the tasks are related and may share a representation \citep{ArgyriouEvgeniouPontil2008,MaurerPontilRomeraParedes2016}.
In our setting, the tasks $F_j=f_j\circ\Pi_\M$, $j=1,\dots,p$, share the manifold index $\Pi_\M X$. The functions $F_j$ are defined on the tube, while the links $f_j$ are defined on $\M$. The predictors fill a $D$-dimensional tube, so their distribution alone need not reveal this representation. Pooling gradients across response components and nearby base points allows us to estimate the common tangent spaces and attain the intrinsic prediction rate.

\paragraph{Connections to the estimator.}
Phase~1 groups observations with similar responses and uses local
predictor covariances to obtain initial tangent-space estimates. Phase~2 pools
outer products of local gradients to update an anisotropic metric, connecting
the construction to outer-product-of-gradients and structural-adaptation
methods \citep{XiaTongLiZhu2002,HJS01,DalalyanJuditskySpokoiny2008} and to
recursive feature machines \citep{doi:10.1126/science.adi5639}. Response selection is used for initialization; subsequent updates use responses through local regression slopes and use the current tangent estimates to select neighborhoods. The ambient-dimension dependence remains explicit.
%
%
%

\subsection{Organization and notation}\label{sec:intro-notation}

Section~\ref{sec:core} describes MANOR and its mechanism, ending with
simplified pseudocode (Algorithm~\ref{alg:manor}). Section~\ref{sec:theory}
states the assumptions, scales, estimator and sample-size condition;
Appendix~\ref{app:defs} gives the exact definitions. Section~\ref{sec:numerics}
reports the experiments, and Section~\ref{sec:discussion} discusses what
the results imply for prediction and tangent-space recovery.
Appendix~\ref{app:proofs} proves Theorem~\ref{thm:isolated-rank},
Appendix~\ref{app:numerics-details} gives the
experiment-specific constructions, and
Appendix~\ref{app:learnability} establishes the minimax exponent for the
broader regularity class defined there and states conditions for its transfer
to \Aref{1}--\Aref{5}. Table~\ref{tab:notation} collects the core
notation used in the main text; proof-specific symbols are introduced
where used in Appendix~\ref{app:proofs}.

Rate comparisons in Sections~\ref{sec:intro} and~\ref{sec:core} may
suppress logarithmic factors in $n$.
We use $a_+=\max\{a,0\}$, $I$ for the identity on the space in question,
and $P^\perp=I-P$ for the complementary orthogonal projector.
The ball $B(x,r)$ is Euclidean; $B_\M(q,r)$ uses the geodesic distance
$d_\M$ on $\M$. An average written $\operatorname*{avg}$ is an arithmetic
average over the indicated finite set. For a symmetric matrix $M$, the
operators $\Ptop(M)$ and $\Pbot(M)$ return the orthogonal projectors onto
the eigenspaces of its $m$ largest and $m$ smallest eigenvalues, respectively;
all spectral ties are resolved by a fixed deterministic rule.
For a subspace $V$, $P_V$ denotes its orthogonal projector.
The notation $\widetilde O$ suppresses fixed powers of $\log n$.

\section{The core idea: tangential gradients and anisotropic localization}\label{sec:core}

%
%
The intrinsic rate in Theorem~\ref{thm:isolated-rank} rests on two geometric
facts. The compositional structure places every gradient of $F$ in a tangent space of $\M$
(Section~\ref{sec:core-gradient}). If these tangent spaces were known,
anisotropic neighborhoods would localize at the intrinsic scale
(Section~\ref{sec:core-oracle}). Phase~1 constructs a first
approximation of such neighborhoods from the responses
(Section~\ref{sec:core-phase1}); Phase~2 sharpens them over $O(\log n)$ steps
until they reach the statistically optimal scale
(Section~\ref{sec:core-phase2}); and Phase~3 uses them to predict
(Section~\ref{sec:core-phase3}). Algorithm~\ref{alg:manor} summarizes the
procedure; Section~\ref{sec:theory} gives its precise definition and
finite-sample guarantees.
For readability, the formulas in this section omit the spatial restriction
\eqref{eq:local-cylinder} and the sample splitting specified in
Section~\ref{sec:theory-estimator} and Appendix~\ref{app:defs}.
%

\subsection{Gradients of the composition lie in tangent spaces}\label{sec:core-gradient}

Fix a point $x=Q+z$ of the tube, with base point
$Q=\Pi_\M x$ and normal
displacement $z\in N_Q\M$. Let $E_Q\in\R^{D\times m}$ have orthonormal columns
spanning $T_Q\M$, so that $P_Q=E_QE_Q^\T$; we identify tangent vectors
with their coordinates in $\R^m$.
Let $\II_Q:T_Q\M\times T_Q\M\to N_Q\M$ be the second fundamental
form. The shape operator $S_{Q,z}$, abbreviated $S_z$ at a fixed base,
is the symmetric matrix defined by
$\langle S_z a,b\rangle=\langle\II_Q(E_Qa,E_Qb),z\rangle$.
The Weingarten equation \citep[Proposition~8.4]{Lee2018Riemannian}
gives its meaning: if the base point moves along a tangent direction $a$
and the normal vector $z$ is carried along, the tangential part of the
change in $z$ is $-E_QS_za$. The point $x=Q+z$ therefore moves tangentially
by $E_Q(I-S_z)a$.
When $Df(Q)$ acts on ambient vectors, we extend it by zero on $N_Q\M$;
in coordinates this extension is $Df(Q)E_Q^\T$.
The differential of $\Pi_\M$ at $x$ is (Lemma~\ref{lem:dPi})
\begin{equation}\label{eq:dPi}
  D\Pi_\M(Q+z)=E_Q(I-S_z)^{-1}E_Q^\T.
\end{equation}
Moving along a normal fiber fixes the base point, so
$D\Pi_\M(x)$ annihilates $N_Q\M$.
Moving the base point along a tangent vector $a$, carrying the normal
displacement along, moves $x$ to first order by $E_Q(I-S_z)a$ plus a normal
vector (the Weingarten equation \citep[Proposition~8.4]{Lee2018Riemannian}).
Thus a tangential displacement of $x$ is amplified by $(I-S_z)^{-1}$ at the
base point. The shape operator satisfies
$\|S_z\|\le\|z\|/\reach_\M$ \citep{NSV}, so $I-S_z$ is invertible in the tube;
if $\|z\|\le R=c\reach_\M$ with the tube ratio $c$ of
Assumption~\Aref{1}, then
$(1+c)^{-1}I\preceq(I-S_z)^{-1}\preceq(1-c)^{-1}I$. The chain rule
applied to $F=f\circ\Pi_\M$ gives
\begin{equation}\label{eq:core-gradient}
  \nabla F(Q+z)=E_Q(I-S_z)^{-1}Df(Q)^\T\in\R^{D\times p},
\end{equation}
where $Df(Q)\in\R^{p\times m}$ is the differential of $f$ at $Q$ in the basis
$E_Q$.
Since $E_Q$ has full column rank and $I-S_z$ is invertible,
$\operatorname{rank}DF(Q+z)=\operatorname{rank}Df(Q)$, where
$DF=(\nabla F)^\T$. Thus each point where $Df$ is rank deficient gives
an entire normal fiber where $DF$ has rank below $m$.

The geometric point is that the normal displacement changes the coordinates of
a tangent derivative without creating a normal derivative. Thus the ambient
gradients can vary along a fiber; at a full-rank base point they span the same tangent space at every offset.

Formula \eqref{eq:core-gradient} contains three facts that the rest of the
paper uses.
\begin{enumerate}[(a)]
\item \emph{Normal derivatives vanish:} $P_Q^\perp\nabla F(Q+z)=0$, where $P_Q^\perp=I-P_Q$, by constancy along normal fibers.
\item \emph{The gradients span the tangent space:} if $Df(Q)$ has rank $m$,
then $\Ran\nabla F(Q+z)=T_Q\M$ for every $z$ in the
fiber. The multi-index inclusion $\Ran\nabla F(x)\subseteq\Ran G$ in
\eqref{eq:intro-multiindex-gradient} becomes $\Ran\nabla F(x)=T_{\Pi_\M x}\M$:
the index space becomes a moving tangent space.
\item \emph{The subspace is invariant at full rank.} As the normal offset
varies, $(I-S_z)^{-1}$ can change the component gradients within $T_Q\M$.
When $Df(Q)$ has rank $m$, their span remains $T_Q\M$ at every offset; its
projector $P_Q$ is the object we estimate.
\end{enumerate}
On a curve the shape factor is scalar: wherever the derivative is nonzero,
one gradient identifies the tangent line along the entire fiber, as in
\citet{WuMaggioniNSVM}. For $m>1$, a single nonzero gradient spans a line
that can rotate within the tangent space. With $p\ge m$ response components,
the gradients can identify the whole plane at a regular base point
(Figure~\ref{fig:model}).

\subsection{Localizing along the manifold}\label{sec:core-oracle}

To provide intuition and motivation for our construction, we briefly discuss here an idealized scenario.
If $\M$ were known, base points, geodesic distances and tangent spaces
would be available. To estimate $F$ near $x_0$ with base point $q_0$,
consider the neighborhood
\[
  \mathcal N_b(x_0)=\{q+z:q\in B_\M(q_0,b),\ z\in N_q\M,\ \|z\|\le R\},
\]
which consists of the entire normal fibers over a geodesic ball of radius $b$
in $\M$.
A local linear (or local constant) estimator on this neighborhood has two properties:
\begin{itemize}
\item \emph{Bias of order $b^s$.} On $\mathcal N_b(x_0)$, the function
$F(x)=f(\Pi_\M x)$ depends only on the base point, which stays within intrinsic
distance $b$ of $q_0$. For $1<s\le2$, the first-order expansion of $f$ around $q_0$ has remainder $O(b^s)$. A local linear fit in tangent coordinates reproduces the linear term. A local average with an even leading kernel has the same bias order under the density regularity in \Aref{2}, since the contribution of the linear term is $O(b^2)$. The extent of the neighborhood in the normal directions
does not contribute to the bias, because $F$ does not vary along fibers.
\item \emph{Variance of order $1/(nb^m)$.} Observations in $\mathcal N_b(x_0)$ correspond to base points in a geodesic
ball of radius $b$, giving about
$n\,\PP\{\Pi_\M X\in B_\M(q_0,b)\}\asymp nb^m$ observations if the base points
have a density bounded above and below on $\M$. Complete fibers contribute no factor $(b/R)^{D-m}$.
\end{itemize}
The tube is therefore statistically wide in directions along which $F$ is
constant and locally narrow only in the $m$ directions that affect the response.
An effective estimator should use the wide directions for replication and spend
its spatial resolution on the tangent directions.

Balancing the two gives \eqref{eq:intro-intrinsic}, with
$b\asymp n^{-1/(2s+m)}$. With a locally bounded positive ambient density,
at a typical interior point a Euclidean ball of
radius $b<R$ contains about $nb^D$ observations and gives \eqref{eq:intro-ambient}.

Near $x_0$, the neighborhood $\mathcal N_b(x_0)$ is approximated by an
ellipsoid aligned with $T_{q_0}\M$; building the ellipsoid requires only the
tangent space $T_{q_0}\M$, not $\M$ itself. For a rank-$m$ projector $P$ and
widths $0<b\le h$, define the anisotropic distance
\begin{equation}\label{eq:metric}
  d_{P,b,h}(v)^2=\frac{\|v\|^2}{h^2}+\frac{\|Pv\|^2}{b^2},
  \qquad v\in\R^D.
\end{equation}
The set $\{v:\ d_{P,b,h}(v)\le1\}$ is an ellipsoid with $m$ short semi-axes of
length $b/\sqrt{1+\rho^2}$ along $\Ran P$ and $D-m$ long semi-axes of length
$h$ along $\Ran P^\perp$, where $\rho=b/h\le1$. We call $b$ the
\emph{tangential width}, $h$ the \emph{normal width} and $\rho$ the
\emph{aspect ratio}. With $P=P_{q_0}$ and a normal width $h$ slightly larger
than the fiber diameter $2R$, the ellipsoid $\{x:\ d_{P,b,h}(x-x_0)\le1\}$
contains the complete normal fibers over a tangential ball of radius
comparable to $b$.
The diameter, not the radius, is needed because $x_0$ may lie at normal
distance $R$ from $\M$ on one side of its fiber, while the fiber extends to
distance $R$ on the other side.

Two difficulties separate this ideal picture from an estimator. The first is
that $P_{q_0}$ is unknown and changes with the base point. The second is that an
anisotropic neighborhood is only as thin as the tangent estimate is accurate.
If $\|\widehat P-P_{q_0}\|=\Delta$, then a vector $v$ with
$\|\widehat Pv\|\le b$ and $\|v\|\le h$ has a true tangential component
$\|P_{q_0}v\|\le b+\Delta h=(\rho+\Delta)h$. A neighborhood of aspect ratio
$\rho$ therefore requires a tangent estimate with error at most a fixed fraction of
$\rho$ when $h/R$ is fixed. More precisely, the proof controls $R\Delta/b$,
the displacement across a fiber in units of the short width. This coupling drives the
estimator. A coarse tangent estimate allows a moderately thin neighborhood;
local regression in that neighborhood yields a better tangent estimate; the
better estimate allows a thinner neighborhood; and so on. Phase~1 starts this
process, Phase~2 runs it for $O(\log n)$ steps, and Phase~3 uses the final
neighborhoods to predict.
\subsection{Phase 1: a warm start from response preimages}\label{sec:core-phase1}

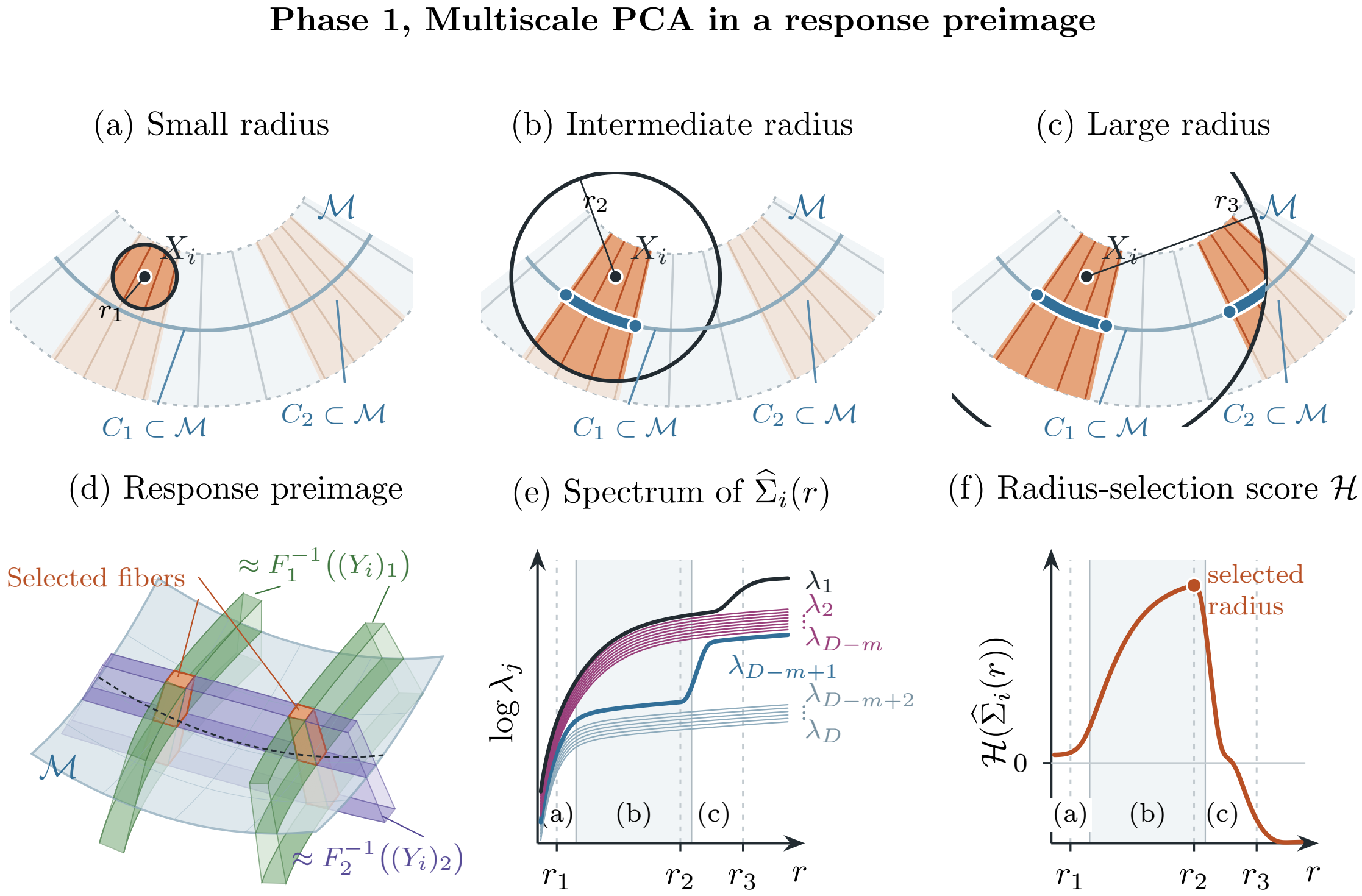
\begin{figure}[!t]%
\centering
\resizebox{.98\linewidth}{!}{%
\begingroup%
\begin{tikzpicture}[x=.028cm,y=.028cm,>=Stealth,
  declare function={
    eigsat(\r,\L,\s)=\L*(\r/\s)^2/sqrt(1+(\r/\s)^4)*(1+.012*\r);
    eignormal(\r)=eigsat(\r,72,32);
    eigtangent(\r)=eigsat(\r,5.57,12);
    eigstep(\u)=(min(1,max(0,\u)))^3
      *(10-15*min(1,max(0,\u))+6*(min(1,max(0,\u)))^2);
    eigjoin(\r)=eigtangent(\r)+150*eigstep((\r-46)/30)*(1+.012*\r);
    eigsoft(\a,\b)=ln(1+exp(-6*abs(ln(\a)-ln(\b))))/6;
    eigfirst(\r)=max(ln(eignormal(\r)),ln(eigjoin(\r)))
      +eigsoft(eignormal(\r),eigjoin(\r));
    eiggap(\r)=min(ln(eigjoin(\r)),ln(.43*eignormal(\r)))
      -eigsoft(eigjoin(\r),.43*eignormal(\r));
    eigscore(\r)=2*(ln(eignormal(\r))-.3993)-eigfirst(\r)-eiggap(\r);
    eigy(\l)=13*(\l+1.9);
  }]
\definecolor{respblue}{RGB}{50,111,152}
\definecolor{respnormal}{RGB}{153,63,123}
\definecolor{respthinblue}{RGB}{142,171,188}
\definecolor{resplightblue}{RGB}{241,245,247}
\definecolor{respblueedge}{RGB}{174,185,192}
\definecolor{respfiber}{RGB}{195,203,208}
\definecolor{resporange}{RGB}{230,164,123}
\definecolor{resporangeedge}{RGB}{184,80,37}
\definecolor{resptan}{RGB}{241,229,220}
\definecolor{resptanedge}{RGB}{217,192,173}
\definecolor{respblack}{RGB}{34,43,49}
\definecolor{respfone}{RGB}{62,118,62}%
\definecolor{respftwo}{RGB}{86,74,150}

\newcommand{\respannulus}[3]{%
  \path[fill=#3,draw=none]
    ({99.75*cos(#1)},{99.75*sin(#1)})
    arc[start angle=#1,end angle=#2,radius=99.75]
    -- ({42.75*cos(#2)},{42.75*sin(#2)})
    arc[start angle=#2,end angle=#1,radius=42.75] -- cycle;
}

\newcommand{\resppacketfibers}[2]{%
  \foreach \aa in {#1}
    \draw[#2,line width=.55pt]
      ({99.75*cos(\aa)},{99.75*sin(\aa)}) --
      ({42.75*cos(\aa)},{42.75*sin(\aa)});
}

\newcommand{\resppackets}[2]{%
  \respannulus{-125.5}{-102.5}{#1}
  \respannulus{-64.5}{-41.5}{#1}
  \resppacketfibers{-124.5,-117.2,-109.8,-103.5}{#2}
  \resppacketfibers{-63.5,-56.2,-48.8,-42.5}{#2}
}

\newcommand{\respsection}{%
  \begin{scope}
    \clip (-73.03,-107.23) rectangle (77.31,-12.47);
    \respannulus{-142.1}{-29.8}{resplightblue}
    \draw[respblueedge,dash pattern=on 1.3pt off 1.82pt,line width=.65pt]
      ({99.75*cos(-142.1)},{99.75*sin(-142.1)})
      arc[start angle=-142.1,end angle=-29.8,radius=99.75];
    \draw[respblueedge,dash pattern=on 1.3pt off 1.82pt,line width=.65pt]
      ({42.75*cos(-142.1)},{42.75*sin(-142.1)})
      arc[start angle=-142.1,end angle=-29.8,radius=42.75];
    \foreach \aa in {-138.7,-91.7,-76.8,-32.7}
      \draw[respfiber,line width=.65pt]
        ({99.75*cos(\aa)},{99.75*sin(\aa)}) --
        ({42.75*cos(\aa)},{42.75*sin(\aa)});
    \resppackets{resptan}{resptanedge}
  \end{scope}
}

\newcommand{\respmanifold}{%
  \draw[respthinblue,line width=1.15pt]
    ({71.25*cos(-142.1)},{71.25*sin(-142.1)})
    arc[start angle=-142.1,end angle=-29.8,radius=71.25];
  \node[respblue,font=\small] at (49,-25) {$\mathcal M$};
  \draw[respblue!82,line width=.65pt] (-18,-99) -- (-8,-71);
  \node[respblue,font=\scriptsize] at (-19,-108) {$C_1\subset\mathcal M$};
  \draw[respblue!82,line width=.65pt] (52,-91) -- (49,-60);
  \node[respblue,font=\scriptsize] at (48,-103) {$C_2\subset\mathcal M$};
}

\newcommand{\resppacketmanifold}{%
  \foreach \lo/\hi in {-125.5/-102.5,-64.5/-41.5}{%
    \draw[white,line width=4.7pt]
      ({71.25*cos(\lo)},{71.25*sin(\lo)})
      arc[start angle=\lo,end angle=\hi,radius=71.25];
    \draw[respblue,line width=3pt]
      ({71.25*cos(\lo)},{71.25*sin(\lo)})
      arc[start angle=\lo,end angle=\hi,radius=71.25];
    \filldraw[fill=respblue,draw=white,line width=.7pt]
      ({71.25*cos(\lo)},{71.25*sin(\lo)}) circle[radius=2.1pt];
    \filldraw[fill=respblue,draw=white,line width=.7pt]
      ({71.25*cos(\hi)},{71.25*sin(\hi)}) circle[radius=2.1pt];
  }
}

\newcommand{\respball}[1]{%
  \begin{scope}
    \clip (-73.03,-107.23) rectangle (77.31,-12.47);
    \begin{scope}
      \clip (-22.80,-51.20) circle[radius=#1];
      \resppackets{resporange}{resporangeedge}
    \end{scope}
    \draw[respblack,line width=1.15pt]
      (-22.80,-51.20) circle[radius=#1];
  \end{scope}
  \respmanifold
  \begin{scope}
    \clip (-22.80,-51.20) circle[radius=#1];
    \resppacketmanifold
  \end{scope}
  \filldraw[fill=respblack,draw=white,line width=.7pt]
    (-22.80,-51.20) circle[radius=2pt];
  \node[respblack,font=\small,anchor=south west,inner sep=1pt]
    at (-19,-48) {$X_i$};
}

\newcommand{\resppanel}[3]{%
  \begin{scope}[shift={(#1,0)}]
    \respsection
    \respball{#3}
    \node[font=\small,fill=none,draw=none] at (2,5) {#2};
  \end{scope}
}

  \resppanel{0}{(a) Small radius}{12}
  \draw[respblack,line width=.5pt] (-22.80,-51.20) -- (-30.5,-60.4);
  \node[font=\scriptsize] at (-35,-65) {$r_1$};

  \resppanel{176}{(b) Intermediate radius}{39}
  \draw[respblack,line width=.5pt] (153.20,-51.20) -- (139.86,-14.55);
  \node[font=\scriptsize] at (146,-24) {$r_2$};

  \resppanel{352}{(c) Large radius}{67}
  \draw[respblack,line width=.5pt] (329.20,-51.20) -- (392.16,-28.29);
  \node[font=\scriptsize] at (382,-24) {$r_3$};

  \newcommand{\respregimes}[2]{%
    \fill[resplightblue] (14.4,0) rectangle (57.6,106);
    \draw[respblueedge,line width=.4pt] (14.4,0) -- (14.4,106);
    \draw[respblueedge,line width=.4pt] (57.6,0) -- (57.6,106);
    \foreach \xx/\rr in {7.2/1,53.4/2,76.8/3} {
      \draw[respfiber,dash pattern=on 1.5pt off 2pt,line width=.55pt]
        (\xx,0) -- (\xx,106);
      \draw[respblack,line width=.6pt] (\xx,0) -- (\xx,-3);
      \node[font=\small,anchor=north] at (\xx,-5) {$r_{\rr}$};
    }
    \draw[respblack,->,line width=.75pt] (0,0) -- (100,0);
    \draw[respblack,->,line width=.75pt] (0,0) -- (0,110);
    \foreach \xx/\lab/\bg in {7.2/a/#1,36/b/resplightblue,#2/c/none}
      \node[font=\scriptsize,anchor=south,inner sep=1pt,fill=\bg]
        at (\xx,4) {(\lab)};
    \node[font=\small,anchor=north] at (98,-4) {$r$};
  }
  \begin{scope}[shift={(124,-263)}]
    \node[font=\small] at (50,132) {(e) Spectrum of $\widehat\Sigma_i(r)$};
    \respregimes{none}{66}%
    \node[font=\small,rotate=90] at (-12,53) {$\log\lambda_j$};
    \foreach \ff in {.45,.497,.548,.605,.668,.737}
      \draw[respthinblue,line width=.4pt,line cap=round]
        plot[domain=3:80,samples=90,smooth,variable=\rad]
          ({1.2*(\rad-2)},{eigy(ln(\ff*eigtangent(\rad)))});
    \draw[respblue,line width=1.3pt,line cap=round]
      plot[domain=3:80,samples=140,smooth,variable=\rad]
        ({1.2*(\rad-2)},{eigy(eiggap(\rad))});
    \foreach \ff in {.5,.5437,.5911,.6427,.6988,.7598,.8261,.9}
      \draw[respnormal,line width=.45pt,line cap=round]
        plot[domain=3:80,samples=90,smooth,variable=\rad]
          ({1.2*(\rad-2)},{eigy(ln(\ff*eignormal(\rad)))});
    \draw[respblack,line width=1.15pt,line cap=round]
      plot[domain=3:80,samples=140,smooth,variable=\rad]
        ({1.2*(\rad-2)},{eigy(eigfirst(\rad))});
    \node[respblack,font=\scriptsize,anchor=west] at (95,98) {$\lambda_1$};
    \node[respnormal,font=\scriptsize,anchor=west] at (95,89) {$\lambda_2$};
    \foreach \yy in {80.6,82.8,85}
      \fill[respnormal] (99.6,\yy) circle[radius=.42pt];
    \node[respnormal,font=\scriptsize,anchor=west] at (95,76) {$\lambda_{D-m}$};
    \node[respblue,font=\scriptsize,anchor=north,inner sep=1pt]
      at (92,72) {$\lambda_{D-m+1}$};
    \node[respthinblue!80!respblack,font=\scriptsize,anchor=west]
      at (95,55) {$\lambda_{D-m+2}$};
    \foreach \yy in {46.6,48.8,51}
      \fill[respthinblue!80!respblack] (99.6,\yy) circle[radius=.42pt];
    \node[respthinblue!80!respblack,font=\scriptsize,anchor=west]
      at (95,42) {$\lambda_D$};
  \end{scope}
  \begin{scope}[shift={(316,-263)}]
    \node[font=\small] at (38,132) {(f) Radius-selection score $\Hscore$};
    \respregimes{white}{64}%
    \node[font=\small,rotate=90] at (-22,53)
      {$\Hscore(\widehat\Sigma_i(r))$};
    \draw[respblack,line width=.6pt] (0,30) -- (-3,30);
    \node[font=\scriptsize,anchor=east] at (-4,30) {$0$};
    \draw[respfiber,line width=.5pt] (0,30) -- (95,30);
    \draw[resporangeedge,line width=1.35pt,line cap=round]
      plot[domain=3:80,samples=160,smooth,variable=\rad]
        ({1.2*(\rad-2)},{30+40*eigscore(\rad)});
    \filldraw[fill=resporangeedge,draw=white,line width=.6pt]
      (53.4,96.4) circle[radius=2.2pt];
    \node[resporangeedge,font=\scriptsize,anchor=west,align=left,
      inner sep=1pt] at (57,95) {selected\\radius};
  \end{scope}
  \node[font=\small] at (11,-131) {(d) Response preimage};%
  \begin{scope}[shift={(-76,-280)},x=4.9cm,y=3.548cm]
    \node[anchor=south west,inner sep=0] at (0,0)
      {\includegraphics[width=4.9cm]{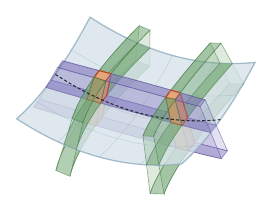}};
    \node[respblue,font=\small] at (.12,.36) {$\mathcal M$};
    \node[respfone,font=\scriptsize,inner sep=1pt] (S1) at (.69,.955)
      {$\approx F_1^{-1}\bigl((Y_i)_1\bigr)$};
    \draw[respfone,line width=.4pt]
      (S1.south west) ++(.1,0) -- (.54,.87);
    \draw[respfone,line width=.4pt]
      (S1.south east) ++(-.06,0) -- (.79,.785);
    \node[respftwo,font=\scriptsize,anchor=east,inner sep=1pt] (S2) at (1.0,.09)
      {$\approx F_2^{-1}\bigl((Y_i)_2\bigr)$};
    \draw[respftwo,line width=.4pt]
      (S2.north east) ++(-.1,0) -- (.81,.215);
    \node[resporangeedge,font=\scriptsize] (Ui) at (.20,.92) {Selected fibers};
    \draw[resporangeedge,line width=.4pt] (Ui.south east) -- (.375,.625);
    \draw[resporangeedge,line width=.4pt] (Ui.east) -- (.633,.527);
  \end{scope}
  \node[font=\small] at (current bounding box.north |- 0,44)
    {{\bf{Phase 1, Multiscale PCA in a response preimage}}};
\end{tikzpicture}%
\endgroup
}
\caption{Phase 1: Multiscale local PCA in a response preimage: schematic with noiseless
responses and a locally bounded positive ambient density. The orange regions depict the population set $\{x:\|F(x)-Y_i\|\le\delta_Y\}$ within the tube; $U_i$ consists of the sampled predictors in this set. (a--c) A normal section of a curved manifold $\mathcal{M}$, with fixed $Y_i,X_i$ and increasing radius. Dark orange marks the part inside $B(X_i,r)$, pale orange the rest, and dark blue the retained base points. The small ball truncates fibers, and at this scale the curse of dimensionality applies;
the intermediate ball isolates $C_1$; the large ball also reaches $C_2$.
(d) A $2$-dimensional $\mathcal M\subset\R^3$ with $p=2$: the slab
$\{x:|F_k(x)-(Y_i)_k|\le\delta_Y\}$ is the union of the normal fibers over a
thin strip of $\mathcal M$ around the level set $\{f_k=(Y_i)_k\}$, two curves
for $k=1$ (green) and one for $k=2$ (purple); $U_i$ lies in the intersection
of the slabs, a union of cylinders (orange) over the packets $C_1$ and $C_2$
around the points of $f^{-1}(Y_i)$. The dashed curve is the section of (a--c).
(e) The eigenvalues $\lambda_1\ge\cdots\ge\lambda_D$ of the covariance
$\widehat\Sigma_i(r)$ of the predictors in $U_i\cap B(X_i,r)$, in the regimes of (a), (b) and (c) (bands): all eigenvalues
grow together; the $m$ tangential ones $\lambda_{D-m+1},\dots,\lambda_D$
saturate at a level set by the packet diameter, while the $D-m$ normal ones
keep growing; the direction joining $C_1$ and $C_2$ acquires a large variance,
so that $\lambda_1$ rises and $\lambda_{D-m+1}$ joins the normal eigenvalues.
(f) The score $\Hscore$ of \eqref{eq:hscore} is maximal at the selected radius
$r_2$ and decreases once $C_2$ enters the ball. Dashed lines mark $r_1$, $r_2$,
$r_3$.}
\label{fig:m2-response}
\end{figure}

Phase~1, the \emph{pilot}, estimates tangent spaces from predictors with
similar responses. Since $Y=f(Q)+\zeta$, conditioning on the response at a
given base point leaves the normal displacement law unchanged
(Lemma~\ref{lem:response-events}). Fix a response-window width
$\delta_Y>0$ (this $\delta_Y$ is independent of $n$). For an observation $i$, called the
\emph{anchor}, the \emph{response preimage}
\[
  U_i=\{X_j:\ \|Y_j-Y_i\|\le\delta_Y\}
\]
collects predictors with responses close to $Y_i$. Phase~1 computes local
covariances on $U_i\cap B(X_i,r)$ as the spatial radius $r$ varies.
The base points can form several distinct groups, or \emph{packets}, when $f$ is non-injective. The
packet containing the anchor's base point $Q_i=\Pi_\M X_i$ can be close
in ambient distance to another packet with similar responses.
A useful radius retains the normal spread while confining the selected
base points to one small patch; Figure~\ref{fig:m2-response} illustrates
this choice. The spatial extent of the patch, rather than the number of
connected components in it, determines the spread along the manifold.
This is a local use of inverse regression (Section~\ref{sec:intro-related}).

\Needspace{8\baselineskip}
\paragraph{Multiscale local PCA.}
Write $\widehat\Sigma_i(r)$ for the empirical centered covariance of the predictors in
$U_i\cap B(X_i,r)$. Figure~\ref{fig:m2-response}(a--c) illustrates three
regimes of the radius scan.
\begin{enumerate}[(1)]
\item At a typical interior point, a sufficiently small ball lies inside
the packet in every direction. With a locally bounded positive ambient
density, the rescaled cloud becomes nearly uniform as the radius tends to zero,
and its covariance eigenvalues are comparable.
\item For intermediate radii, the ball isolates one packet and meets no other
packet; it may still mildly truncate the outer ends of the normal fibers.
The selected predictors have a small spread along $T_{Q_i}\M$ (the
packet diameter) and a large spread across it (the retained fiber length), so the $m$ smallest covariance eigenvalues separate from the $D-m$ largest,
and their eigenvectors approximately span $T_{Q_i}\M$.
\item For large radii, the ball reaches other packets of the same response preimage.
The selected predictors form a mixture, the covariance acquires a large
eigenvalue in the direction joining the packets, and the separation
deteriorates.
\end{enumerate}
The scan is therefore a three-way balance: a radius that is too small hides
the normal spread, an intermediate radius exposes the tangent--normal split,
and a large radius mixes packets. The score below rewards the middle regime
while penalizing both failures.
Figure~\ref{fig:m2-response}(e,f) shows the spectral separation and the
resulting score.
For a covariance matrix $\Sigma$, let
$\lambda_1\ge\cdots\ge\lambda_D$ be its eigenvalues.
To detect the second regime, Phase~1 uses the score
\begin{equation}\label{eq:hscore}
  \Hscore(\Sigma)=
  \log\frac{(\lambda_2\cdots\lambda_{D-m})^{2/(D-m-1)}}{\lambda_1\lambda_{D-m+1}}
  \quad(D-m\ge2),
  \qquad
  \Hscore(\Sigma)=\log\frac{\lambda_1}{\lambda_2}\quad(D-m=1),
\end{equation}
For $D-m\ge2$, let $\bar\lambda=(\lambda_2\cdots\lambda_{D-m})^{1/(D-m-1)}$ be the
geometric mean of the eigenvalues $\lambda_2,\dots,\lambda_{D-m}$; then
\[
  \Hscore(\Sigma)=\log\frac{\bar\lambda}{\lambda_{D-m+1}}+\log\frac{\bar\lambda}{\lambda_1}.
\]
The first term is large when $\lambda_{D-m+1}$, the largest of the $m$
tangential eigenvalues, is small compared with $\bar\lambda$ (the cloud is a
thin packet). The second term is never positive, and it is close to zero
unless $\lambda_1$ is much larger than $\bar\lambda$ (a second packet has
entered the ball).
For $D-m=1$ the score only measures how thin the cloud is. Phase~1 scans a finite grid of radii $\mathcal R$,
selects the smallest maximizer $r_i$ of $r\mapsto\Hscore(\widehat\Sigma_i(r))$,
and records an orthonormal basis $V_i$ of the eigenspace of the $m$ smallest
eigenvalues of $\widehat\Sigma_i(r_i)$. Under \Aref{4}, the theoretical
scan grid lies in the part of regime~(2) that retains complete fibers.
The numerical scan may also select a radius that truncates their ends, as in
Figure~\ref{fig:m2-response}(b).

\paragraph{Aggregation.}
The scan supplies many local bases, but each basis can be distorted
when the selected ball truncates the normal fibers, or by response noise in the response preimage.
The ball is centered at $X_i$, not at its base point, so the retained
part of a fiber is not centered on $\M$ and its length varies with the base
point; this tilts the eigenvectors of $\widehat\Sigma_i(r_i)$ away from
$T_{Q_i}\M$. Aggregation turns these local estimates
into one projector field by averaging projectors rather than basis vectors.
Phase~1 therefore averages the projectors $V_jV_j^\T$ over the observations
$j$ in the selected neighborhood of $i$ and keeps the dominant $m$-dimensional
eigenspace of the average. With
$\Omega_i=\{j:X_j\in U_i\cap B(X_i,r_i)\}$,
\[
  \widehat P_{\rm pil}(X_i)=\Ptop\Bigl(\frac1{|\Omega_i|}\sum_{j\in\Omega_i}V_jV_j^\T\Bigr).
\]
We average projectors because a basis is only defined up to an orthogonal
transformation. The average is stable: if all
projectors in the collection are within operator-norm distance $\Delta$ of
$P_{Q_i}$, so is their average, and replacing a fraction $\alpha$ of them by
arbitrary projectors adds at most $2\alpha$. Once the averaged projector
$\overline P$ satisfies $\|\overline P-P_{Q_i}\|\le 1/4$, extracting its
dominant $m$-dimensional eigenspace cannot inflate the error by more than a
fixed factor: $\|\Ptop(\overline P)-P_{Q_i}\|\le C\|\overline P-P_{Q_i}\|$
for an absolute constant $C$ (Lemma~\ref{lem:block-spectral}).

\paragraph{Output.}
Phase~1 returns the pooled projectors $\widehat P_{\rm pil}$, from which the initial projector $\widehat P_0$ of Phase~2 is obtained by the assignment of Section~\ref{sec:theory-estimator}, and the selected radii,
whose average, possibly multiplied by a constant factor $h_{\rm rescale}$,
is the initial normal width $h_0$ of
Phase~2. Quantities attached to Phase~1 carry the subscript ``pil'' (for ``pilot'').
The analysis uses the same multiscale PCA on a grid within the one-packet
interval of \Aref{4}. Proposition~\ref{prop:pilot} bounds the angle after
aggregation at the scale required for the first refinement.

A small Euclidean ball localizes the base point by cutting the normal
fibers short (Section~\ref{sec:core-oracle}). The initial tangent estimate
lets Phase~2 choose tangential and normal widths separately.

\subsection{Phase 2: shrink tangentially, expand normally}\label{sec:core-phase2}

Phase~2 performs $k_*$ refinement steps with decreasing tangential widths
$b_k=\rho_kh_k$, where $\rho_k$ is the aspect ratio and $h_k$ the normal
width at step $k$. Each center fits an affine function using its incoming
projector $\widehat P_{k-1}$. The next tangent field is computed from the
outer products of these slopes.
\begin{enumerate}[(a)]
\item \emph{Local gradients.} At each center $X_j$, fit once using its own
incoming projector $\widehat P_{k-1}(X_j)$ and the kernel $K(t)=(1-t)_+^2$.
The optimization variables are an intercept $\beta_0\in\R^p$ and a
slope matrix $G\in\R^{D\times p}$:
\[
 (\widehat\beta_{0,j,k},\widehat G_{j,k})\in\argmin_{\beta_0,G}\sum_\ell
 K\bigl(d_{\widehat P_{k-1}(X_j),b_k,h_k}(X_\ell-X_j)^2\bigr)
 \bigl\|Y_\ell-\beta_0-G^\T(X_\ell-X_j)\bigr\|^2.
\]
The slope matrix $\widehat G_{j,k}\in\R^{D\times p}$ estimates the gradients
of the response components near $X_j$.
\item \emph{Pooling.} At a target $x$, use its incoming projector to select
\[
 \Wwin_{x,k}
 =\{u:\ d_{\widehat P_{k-1}(x),b_k,h_k}(u-x)\le1\}.
\]
Average the outer products and retain the leading $m$-dimensional eigenspace:
\[
 \widehat C_k(x)=\operatorname*{avg}_{j:\,X_j\in\Wwin_{x,k}}
       \widehat G_{j,k}\widehat G_{j,k}^\T,
 \qquad \widehat P_k(x)=\Ptop\bigl(\widehat C_k(x)\bigr).
\]
\end{enumerate}
Averaging outer products of local gradient estimates is the device of the
outer product of gradients method for index models
(see Section~\ref{sec:intro-related} and references therein). Here each fit follows the estimated
tangent space at its own center, and pooling refines the moving tangent
field from step to step.

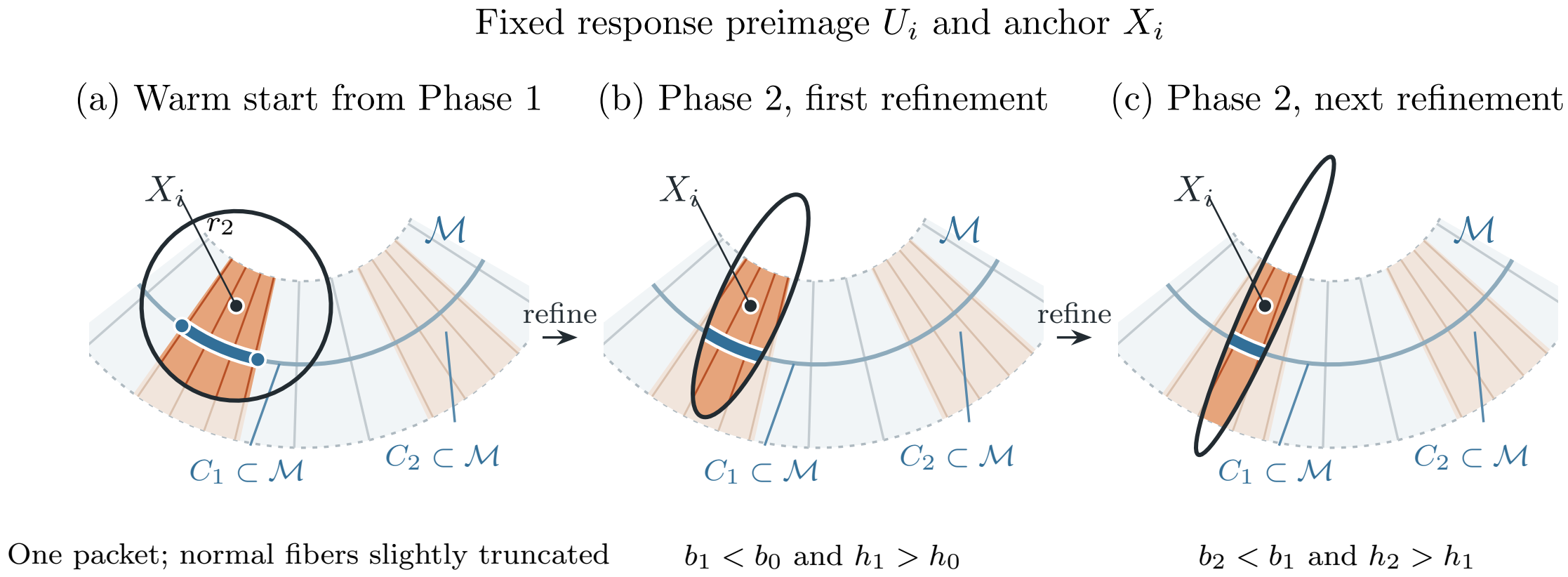
\begin{figure}[!htbp]
\centering
\resizebox{.98\linewidth}{!}{%
\begingroup%
\begin{tikzpicture}[x=.028cm,y=.028cm,>=Stealth]
\definecolor{phaseblue}{RGB}{50,111,152}
\definecolor{phasethinblue}{RGB}{142,171,188}
\definecolor{phaselightblue}{RGB}{241,245,247}
\definecolor{phaseblueedge}{RGB}{174,185,192}
\definecolor{phasefiber}{RGB}{195,203,208}
\definecolor{phaseorange}{RGB}{230,164,123}
\definecolor{phaseorangeedge}{RGB}{184,80,37}
\definecolor{phasetan}{RGB}{241,229,220}
\definecolor{phasetanedge}{RGB}{217,192,173}
\definecolor{phaseblack}{RGB}{34,43,49}

\newcommand{\phaseannulus}[3]{%
  \path[fill=#3,draw=none]
    ({99.75*cos(#1)},{99.75*sin(#1)})
    arc[start angle=#1,end angle=#2,radius=99.75]
    -- ({42.75*cos(#2)},{42.75*sin(#2)})
    arc[start angle=#2,end angle=#1,radius=42.75] -- cycle;
}

\newcommand{\phasepacketfibers}[2]{%
  \foreach \aa in {#1}
    \draw[#2,line width=.55pt]
      ({99.75*cos(\aa)},{99.75*sin(\aa)}) --
      ({42.75*cos(\aa)},{42.75*sin(\aa)});
}

\newcommand{\phasepackets}[2]{%
  \phaseannulus{-125.5}{-102.5}{#1}
  \phaseannulus{-64.5}{-41.5}{#1}
  \phasepacketfibers{-124.5,-117.2,-109.8,-103.5}{#2}
  \phasepacketfibers{-63.5,-56.2,-48.8,-42.5}{#2}
}

\newcommand{\phasesection}{%
  \begin{scope}
    \clip (-73.03,-107.23) rectangle (77.31,-12.47);
    \phaseannulus{-142.1}{-29.8}{phaselightblue}
    \draw[phaseblueedge,dash pattern=on 1.3pt off 1.82pt,line width=.65pt]
      ({99.75*cos(-142.1)},{99.75*sin(-142.1)})
      arc[start angle=-142.1,end angle=-29.8,radius=99.75];
    \draw[phaseblueedge,dash pattern=on 1.3pt off 1.82pt,line width=.65pt]
      ({42.75*cos(-142.1)},{42.75*sin(-142.1)})
      arc[start angle=-142.1,end angle=-29.8,radius=42.75];
    \foreach \aa in {-138.7,-91.7,-76.8,-32.7}
      \draw[phasefiber,line width=.65pt]
        ({99.75*cos(\aa)},{99.75*sin(\aa)}) --
        ({42.75*cos(\aa)},{42.75*sin(\aa)});
    \phasepackets{phasetan}{phasetanedge}
  \end{scope}
}

\newcommand{\phasemanifold}{%
  \draw[phasethinblue,line width=1.15pt]
    ({71.25*cos(-142.1)},{71.25*sin(-142.1)})
    arc[start angle=-142.1,end angle=-29.8,radius=71.25];
  \node[phaseblue,font=\small] at (49,-25) {$\mathcal M$};
  \draw[phaseblue!82,line width=.65pt] (-18,-99) -- (-8,-71);
  \node[phaseblue,font=\scriptsize] at (-19,-108) {$C_1\subset\mathcal M$};
  \draw[phaseblue!82,line width=.65pt] (52,-91) -- (49,-60);
  \node[phaseblue,font=\scriptsize] at (48,-103) {$C_2\subset\mathcal M$};
}

\newcommand{\phasepacketmanifold}{%
  \foreach \lo/\hi in {-125.5/-102.5,-64.5/-41.5}{%
    \draw[white,line width=4.7pt]
      ({71.25*cos(\lo)},{71.25*sin(\lo)})
      arc[start angle=\lo,end angle=\hi,radius=71.25];
    \draw[phaseblue,line width=3pt]
      ({71.25*cos(\lo)},{71.25*sin(\lo)})
      arc[start angle=\lo,end angle=\hi,radius=71.25];
    \filldraw[fill=phaseblue,draw=white,line width=.7pt]
      ({71.25*cos(\lo)},{71.25*sin(\lo)}) circle[radius=2.1pt];
    \filldraw[fill=phaseblue,draw=white,line width=.7pt]
      ({71.25*cos(\hi)},{71.25*sin(\hi)}) circle[radius=2.1pt];
  }
}

\newcommand{\phaseneighborhood}[2]{%
  \begin{scope}
    \clip[rotate around={-24.1:(-22.80,-51.20)}]
      (-22.80,-51.20) ellipse[x radius=#1,y radius=#2];
    \begin{scope}
      \clip (-73.03,-107.23) rectangle (77.31,-12.47);
      \phasepackets{phaseorange}{phaseorangeedge}
    \end{scope}
  \end{scope}
  \phasemanifold
  \begin{scope}
    \clip[rotate around={-24.1:(-22.80,-51.20)}]
      (-22.80,-51.20) ellipse[x radius=#1,y radius=#2];
    \phasepacketmanifold
  \end{scope}
  \draw[phaseblack,line width=1.15pt,
        rotate around={-24.1:(-22.80,-51.20)}]
    (-22.80,-51.20) ellipse[x radius=#1,y radius=#2];
  \filldraw[fill=phaseblack,draw=white,line width=.7pt]
    (-22.80,-51.20) circle[radius=2pt];
  \draw[phaseblack,line width=.5pt] (-22.80,-51.20) -- (-42,-14);
  \node[phaseblack,font=\small] at (-47,-12) {$X_i$};
}

\newcommand{\phasepanel}[5]{%
  \begin{scope}[shift={(#1,0)}]
    \node[font=\small] at (2,19) {#2};
    \phasesection
    \phaseneighborhood{#3}{#4}
    \node[font=\scriptsize,align=center] at (2,-137) {#5};
  \end{scope}
}

  \node[font=\small] at (177,45)
    {Fixed response preimage $U_i$ and anchor $X_i$};%
  \phasepanel{0}{(a) Warm start from Phase~1}{32.4}{32.4}
    {One packet; normal fibers slightly truncated}
  \node[font=\scriptsize] at (-28,-24) {$r_2$};
  \phasepanel{176}{(b) Phase~2, first refinement}{11.25}{41.4}
    {$b_1<b_0$ and $h_1>h_0$}
  \phasepanel{352}{(c) Phase~2, next refinement}{6.75}{55.8}
    {$b_2<b_1$ and $h_2>h_1$}
  \draw[-{Stealth[length=2mm]},phaseblack,line width=.65pt]
    (82,-62) -- node[above,font=\scriptsize] {refine} (94,-62);
  \draw[-{Stealth[length=2mm]},phaseblack,line width=.65pt]
    (258,-62) -- node[above,font=\scriptsize] {refine} (270,-62);
\end{tikzpicture}%
\endgroup
}
\caption{Phase~2 refinement in the normal section of Figure~\ref{fig:m2-response}.
The orange fibers mark the noiseless response selection region depicted in Figure~\ref{fig:m2-response}; the black outlines
show the predictor neighborhoods about the same anchor $X_i$. Dark orange lies
inside the displayed neighborhood and pale orange outside. Panel~(a)
reproduces the intermediate-radius selection in
Figure~\ref{fig:m2-response}(b), which here truncates the ends of the fibers.
In panels~(b) and~(c), $b_{k+1}<b_k$ narrows the range of base points while
$h_{k+1}>h_k$ retains a longer portion of each nearby fiber. Unlike
Phase~1, Phase~2 does not restrict a neighborhood to the response preimage: it
selects predictors by position alone, and their responses enter the local
affine fits.}
\label{fig:m2-phase2}
\end{figure}

\paragraph{What a neighborhood contributes.}
At a regular base point, the responses already distinguish every tangent
direction. At a point where $Df$ is rank deficient, some of those directions are invisible on the
central fiber. The surrounding ball matters: a missing direction whose
slope grows linearly away from the central point contributes energy of order
$b_k^2$ to the pooled slope outer products. Taking the outer product before averaging retains this energy even
when the slopes at opposite sides of the point have different signs. Meanwhile the tangent
planes at neighboring centers rotate. Assumption~\Aref{3} limits how much
this rotation can change the response differential on the complete fibers
near a point where $Df$ is rank deficient.

\paragraph{Why the next neighborhood retains complete fibers.}
Across a normal fiber of length $2R$, a tilt $\Delta$ displaces the
estimated tangential coordinates by at most $2R\Delta$. If this displacement
fits within the next tangential width, the neighborhood retains complete
fibers above a common ball of base points. Thus refinement must control
the angle \emph{relative to the next width}. Near a point where $Df$ is rank deficient the local
fits recover a missing tangent slope of order $b_k$, while their normal
slope is of order $b_k^2$ in fixed physical units. Cancellation of the
first-order tilt in the fit gives the error bound needed to retain complete fibers at the next step (Lemma~\ref{lem:quadratic-update}). Away from points where $Df$ is rank deficient, curvature transfers
part of a center's tangent error to its neighbors. By \eqref{eq:regular-memory},
the root mean square stretch in \Aref{2} controls this transfer over two
successive updates.

\paragraph{The schedule and where to stop.}
The aspect ratio shrinks by $a_\rho=e^{-1/[2(2s+m)]}$, and the normal
width grows by $a_h=\exp\{1/[2(D+4)k_*]\}$ per step, see \eqref{eq:schedule}.
Starting from $b_0=\rho_0h_0$, $b_k=b_0(a_\rho a_h)^k$ decreases.
The statistical results use $s=2$. The warm start supplies the first
aligned neighborhood. Through the ensuing rounds the largest displacement due to
tilt remains smaller than the next tangential width
(Proposition~\ref{prop:finite}). Narrower neighborhoods reduce tangential
bias. At fixed local sampling density, a neighborhood contains order
$n_f b_k^m$ observations from a sample of size $n_f$ assigned to that round.
Refinement stops while these observations still resolve a normal slope of
order $b_k^2$. At fixed model scales the terminal tangential width is of
order $n^{-1/(m+4)}$, up to logarithms, with complete normal fibers.

\subsection{Phase 3: prediction}\label{sec:core-phase3}

The regression estimate is an anisotropic kernel smoother.
For a fixed positive multiplier $c_\rho$, set
$h_*=h_{k_*}$, $\rho_*=c_\rho\rho_{k_*}$ and $b_*=\rho_*h_*$.
For a query $x$,
\begin{equation}
  \label{e:Fhat_def}
  \widehat F(x):=\frac{\sum_i\varpi_i(x)Y_i}{\sum_i\varpi_i(x)}\quad,\quad
  \varpi_i(x):=K\Bigl(\frac{\|x-X_i\|^2}{h_*^2}
     +\frac{\|\widehat P_{k_*}(X_i)(x-X_i)\|^2}{b_*^2}\Bigr),
\end{equation}  
where the sum is over the observations reserved for prediction; their mean
is used if the denominator vanishes. Each predictor supplies its own
tangent plane. With the true planes the kernel is even to
first order in the base displacement, giving bias of order $b_*^2$. The
local count is proportional to the size of the prediction sample times
$b_*^m$ at fixed local density. Near a point where $Df$ vanishes, even an
imperfectly aligned neighborhood has little response variation. If $Df$ is only partially rank deficient, the other responses still vary, but the least
accurate planes occupy a shrinking region. Integrating their contribution
preserves the $m$-dimensional prediction rate (Proposition~\ref{prop:readout}).

\FloatBarrier

\subsection{Algorithm}\label{sec:core-algorithm}

Algorithm~\ref{alg:manor} gives simplified pseudocode for the three
phases of MANOR. It shows how each fitted slope enters the pools and how
the widths change between rounds. Section~\ref{sec:theory-estimator}
specifies the assignment of an anchor star to each target, the local
cylinder, the tests for retaining a fit and the independent samples used in the theorem.
There, an anchor star $S$ consists of an eligible anchor and the eligible
anchors linked to it by the response and spatial tests of
Appendix~\ref{app:defs}; $P_S$ projects orthogonally onto
the leading $m$ eigenspace of the average of their pooled projectors.
Appendix~\ref{app:defs} determines the stopping step $k_*$.
In the version analysed in the theorem, $k_*$, $\rho_0$, $h_{\rm rescale}$,
the Phase~1 filters and the local cylinder are set from $m$, $\delta_Y$, the
radii $\mathcal R$ and the model scales $R$, $\sigma_N$, $r_{\rm sing}$,
$w_Y$, $p_0$, $\pi_-$, $\sigma_\zeta$ and $\sigma_{f,+}$ of
Section~\ref{sec:theory-assumptions}, together with fixed multipliers.

\begin{algorithm}[t]
\algcaption{MANOR (simplified pseudocode)}
\label{alg:manor}
\DontPrintSemicolon
\SetKwInOut{Input}{Input}\SetKwInOut{Output}{Output}
\Input{data $(X_i,Y_i)_{i=1}^n\subset\R^D\times\R^p$; intrinsic dimension $m$;
kernel $K(t)=(1-t)_+^2$; response resolution $\delta_Y$; finite set of radii
$\mathcal R$; initial aspect ratio $\rho_0\in(0,1)$; factors
$a_\rho<1\le a_h$; number of steps $k_*$; constants $h_{\rm rescale},c_\rho$}
\Output{field of tangent spaces $x\mapsto\widehat P_{k_*}(x)$; regression estimate
$x\mapsto\widehat F(x)$}
\BlankLine
\tcp{{\bf{Phase 1}}: warm start by multiscale local PCA}

\For{$i=1,\dots,n$}{
  $U_i\leftarrow\{X_j:\ \|Y_j-Y_i\|\le\delta_Y\}$ \tcps{response preimage}
  \For{$r\in\mathcal R$}{
    $\Omega_i(r)\leftarrow\{j:\ X_j\in U_i,\ \|X_j-X_i\|\le r\}$;\par
    $\widehat\Sigma_i(r)\leftarrow$ covariance of $\{X_j\}_{j\in\Omega_i(r)}$;\par
  }
  $r_i\leftarrow$ smallest maximizer of $r\mapsto\Hscore(\widehat\Sigma_i(r))$;\par
  $V_i\leftarrow$ eigenvectors of the $m$ smallest eigenvalues of $\widehat\Sigma_i(r_i)$;\par
}
\For{$i=1,\dots,n$}{
  $\widehat P_{\rm pil}(X_i)\leftarrow\Ptop\bigl(|\Omega_i(r_i)|^{-1}\sum_{j\in\Omega_i(r_i)}V_jV_j^\T\bigr)$ \tcps{pool at anchor $i$}
}
$h_0\leftarrow h_{\rm rescale}\cdot n^{-1}\sum_ir_i$ \tcps{average selected radius, rescaled}
\BlankLine
\tcp{{\bf{Phase 2}}: centerwise local fits and refinement}
For each target $x$, set $\widehat P_0(x)\leftarrow P_S$ for its assigned anchor star $S$ (Appendix~\ref{app:defs});\par
\For{$k=1,\dots,k_*$}{
  $\rho_k\leftarrow a_\rho\rho_{k-1}$,\quad $h_k\leftarrow a_hh_{k-1}$,\quad $b_k\leftarrow\rho_kh_k$;\par
  \For{$j=1,\dots,n$}{
    $\widehat G_{j,k}\leftarrow$ slope of $\min_{\beta_0,G}\sum_\ell
    K\bigl(d_{\widehat P_{k-1}(X_j),b_k,h_k}(X_\ell-X_j)^2\bigr)
    \|Y_\ell-\beta_0-G^\T(X_\ell-X_j)\|^2$;\par
    $\widehat A_{j,k}\leftarrow\widehat G_{j,k}\widehat G_{j,k}^\T$ \tcps{store once per center}
  }
  \For{each target $x$ (data points and queries)}{
    $\Wwin_{x,k}\leftarrow\{u:\ d_{\widehat P_{k-1}(x),b_k,h_k}(u-x)\le1\}$;\par
    $\mathcal I_{x,k}\leftarrow\{j:\ X_j\in\Wwin_{x,k}\}$;\par
    $\widehat P_k(x)\leftarrow\Ptop\bigl(|\mathcal I_{x,k}|^{-1}\sum_{j\in\mathcal I_{x,k}}\widehat A_{j,k}\bigr)$;\par
  }
}
\BlankLine

\tcp{{\bf{Phase 3}}: prediction at a query point $x$}
$\rho_*\leftarrow c_\rho\rho_{k_*}$,\quad $h_*\leftarrow h_{k_*}$,\quad $b_*\leftarrow\rho_*h_*$;\par
$\varpi_i(x)\leftarrow K\bigl(h_*^{-2}\|x-X_i\|^2+b_*^{-2}\|\widehat P_{k_*}(X_i)(x-X_i)\|^2\bigr)$, $i=1,\dots,n$;\par
$\widehat F(x)\leftarrow\sum_i\varpi_i(x)Y_i\big/\sum_i\varpi_i(x)$; \tcps{prediction-fold mean if the denominator is zero}
\end{algorithm}

\paragraph{Computational cost.}
Phase~1 costs $O(n^2p+n|\mathcal R|(nD^2+D^3))$ using direct covariance
computations and dense eigendecompositions. In Phase~2, each center is
fitted once, at cost $O(N'(D^2+Dp)+D^3+D^2p)$ for $N'$ local observations;
its slope outer product is reused across pools. At the terminal scale,
$N'$ is of order $n^{2s/(2s+m)}$ up to logarithms, while earlier fits may
use $N'\le n$. Direct neighborhood searches and pooling add
$O(n^2D^2)$ per iteration. Theorem~\ref{thm:isolated-rank} gives the full
training bound, which for fixed dimensions and $k_*=O(\log n)$ is
$O((|\mathcal R|+\log n)n^2)$. With tangent bases stored as $D\times m$
matrices, prediction at one test point costs $O(n(Dm+p))$.

\begin{table}[p]
\centering
\footnotesize
\renewcommand{\arraystretch}{0.90}
\begin{tabularx}{\linewidth}{@{}l X@{}}
\toprule
\textbf{Symbol} & \textbf{Meaning in the main text}\\
\midrule
$b_k$ & Phase~2 tangential width, $b_k=\rho_kh_k$ (Section~\ref{sec:core-phase2}).\\
$b_*$ & Prediction tangential width, $b_*=\rho_*h_*$ (Section~\ref{sec:core-phase3}).\\
$c$ & Tube-to-reach ratio, $R=c\reach_\M$ (Assumption~\Aref{1}).\\
$c_{\rm op}$ & Constant in the lower bound \eqref{eq:rank-opening}.\\
$\widehat C_k$ & Pooled outer-product matrix whose leading eigenspace gives $\widehat P_k$ (Section~\ref{sec:core-phase2}).\\
$d_\M$ & Geodesic distance on $\M$ (Section~\ref{sec:intro-notation}).\\
$d_{P,b,h}$ & Anisotropic distance with tangent width $b$, normal width $h$ and projector $P$ \eqref{eq:metric}.\\
$D$ & Ambient dimension of the predictors (Definition~\ref{def:mim}).\\
$Df$ & Differential of $f$ on a tangent space (Section~\ref{sec:intro-model}).\\
$\mathcal E_{\rm pred}$ & Relative mean squared prediction error \eqref{eq:intro-metrics}.\\
$\mathcal E_{\rm tan}$ & Mean sine of the largest principal angle at a random query \eqref{eq:intro-metrics}.\\
$f$ & Link function mapping a base point to its mean response (Definition~\ref{def:mim}).\\
$F$ & Regression function on the tube, $F=f\circ\Pi_\M$ (Definition~\ref{def:mim}).\\
$\widehat F$ & Anisotropic kernel estimate of the regression function (Section~\ref{sec:core-phase3}).\\
$\widehat G_{j,k}$ & Local affine-fit slope stored at center $X_j$ in Phase~2 (Section~\ref{sec:core-phase2}).\\
$h_k$ & Phase~2 normal width at step $k$ (Section~\ref{sec:core-phase2} and \eqref{eq:schedule}).\\
$h_*$ & Normal width used for prediction (Section~\ref{sec:core-phase3}).\\
$\Hscore$ & Covariance shape score used to select a Phase~1 radius \eqref{eq:hscore}.\\
$k_*$ & Number of Phase~2 refinement steps (Section~\ref{sec:core-phase2} and Appendix~\ref{app:defs}).\\
$m$ & Intrinsic dimension of the manifold $\M$ (Definition~\ref{def:mim}).\\
$\M$ & The $m$-dimensional predictor manifold (Definition~\ref{def:mim}).\\
$n$ & Number of training observations (Section~\ref{sec:intro}).\\
$n_0$ & Training sample threshold \eqref{eq:n0}.\\
$N_k$ & Effective local count, $N_k=n_f\pi_-b_k^m$ (Section~\ref{sec:theory-estimator}).\\
$p$ & Response dimension; the theorem assumes $p\ge m$ (Definition~\ref{def:mim}).\\
$p_0$ & Probability lower bound in \eqref{eq:packet-core}.\\
$P_Q$ & Orthogonal projector onto $T_Q\M$ (Section~\ref{sec:intro-result}).\\
$\widehat P_k$ & Estimated tangent projector after Phase~2 step $k$ (Section~\ref{sec:core-phase2}).\\
$\Ptop(M)$ & Projector onto the $m$ largest-eigenvalue eigenspace of a symmetric matrix $M$ (Section~\ref{sec:intro-notation}).\\
$Q$ & Base point $Q=\Pi_\M X$ of a predictor \eqref{eq:intro-decomposition}.\\
$r_{\rm cyl}$ & Tangential radius of the local cylinder (Assumption~\Aref{5} and \eqref{eq:local-cylinder}).\\
$r_{\rm sing}$ & Radius scale of the neighborhoods around points where $Df$ is rank deficient (Assumption~\Aref{3}).\\
$\reach_\M$ & Reach of $\M$, the geometric scale used to measure curvature (Section~\ref{sec:intro-model}).\\
$R$ & Radius of the normal tube containing the predictors (Definition~\ref{def:mim}).\\
$\mathcal R$ & Finite grid of spatial radii scanned in Phase~1 (Section~\ref{sec:core-phase1}).\\
$s$ & Smoothness index used in the refinement schedule; the theorem sets $s=2$ \eqref{eq:schedule}.\\
$T_Q\M$ & Tangent space of $\M$ at $Q$ (Section~\ref{sec:intro-model}).\\
$U_i$ & Predictors selected by proximity of their responses to $Y_i$ (Section~\ref{sec:core-phase1}).\\
$w_Y$ & Base resolution of the pilot scan: packets visible to the scan have diameter at most $C_{\rm pkt}w_Y$ (Assumption~\Aref{4}).\\
$Z$ & Normal displacement $Z=X-Q$ \eqref{eq:intro-decomposition}.\\
$\delta_Y$ & Response-window width used by the pilot scan (Section~\ref{sec:core-phase1}).\\
$\zeta$ & Centered response noise in $Y=F(X)+\zeta$ (Section~\ref{sec:intro} and Assumption~\Aref{5}).\\
$\pi$ & Density of the base point with respect to manifold volume (Assumption~\Aref{2}).\\
$\pi_-$ & Positive lower bound for the base-point density (Assumption~\Aref{2}).\\
$\Pi_\M$ & Nearest-point projection from the tube to $\M$ (Section~\ref{sec:intro-model}).\\
$\rho_k$ & Phase~2 aspect ratio at step $k$ (Section~\ref{sec:core-phase2} and \eqref{eq:schedule}).\\
$\rho_*$ & Prediction aspect ratio, $\rho_*=c_\rho\rho_{k_*}$ (Section~\ref{sec:core-phase3}).\\
$\sigma_N$ & Standard-deviation scale in a single normal direction (Assumption~\Aref{2}).\\
$\sigma_{f,+}$ & Uniform upper bound on the largest singular value of $Df$ (Assumption~\Aref{3}).\\
$\Sigma_{\rm sing}$ & Finite set of base points where $Df$ has rank less than $m$ (Assumption~\Aref{3}).\\
$\tau$ & Confidence exponent controlling the probability of exceptional events (Section~\ref{sec:theory}).\\
\bottomrule
\end{tabularx}
\caption{Core notation used in the main text. Proof-specific symbols are introduced where used in Appendix~\ref{app:proofs}.}
\label{tab:notation}
\end{table}

\section{Assumptions and main results}\label{sec:theory}

The link differential $Df$ may be rank deficient at finitely many points.
Appendix~\ref{app:proofs} gives the proofs in the order of the three phases.

\Needspace{6\baselineskip}

\paragraph{Role of the assumptions.}
\Aref{1}
sets the reach geometry and tube ratio; \Aref{2} specifies how the full normal
fibers are sampled and how they affect tangent coordinates; \Aref{3} supplies
link smoothness and local rank; \Aref{4} localizes the response preimage
within the scan; and \Aref{5} matches the response window to the noise and
normal-variance scales.

\paragraph{Conventions.}
The confidence exponent $\tau>0$ is fixed;
a quantity is called \emph{fixed} if it does not depend on $n$.
Write
$\|M\|_F=(\sum_{i,j}M_{ij}^2)^{1/2}$ for the Frobenius norm of a matrix
$M$, $\sigma_j(M)$ for its $j$th largest singular value,
$\sigma_{\min}(M)$ for its smallest singular value, and
$\lambda_{\min}(M)$ for the smallest eigenvalue of a symmetric matrix.
For random vectors, $\Cov(U,V)=\E[(U-\E U)(V-\E V)^\T]$ and $\Cov(U)=\Cov(U,U)$.
The symbol $\ind_E$ denotes the indicator of
a set or event $E$, and $|S|$ is the cardinality of a finite set $S$.
Write $B_d(r)=B_{\R^d}(0,r)$ for the Euclidean ball in $\R^d$ centered
at the origin with radius $r$.
For a
scalar random variable $U$, write
$\|U\|_{\psi_2}=\inf\{t>0:\E\exp(U^2/t^2)\le2\}$.
The letter $C$
denotes positive constants that depend only on the fixed model parameters,
the kernel and $\tau$, and may change from line to line; it never depends on
$n,D,p$. Lower bounds are written with $C^{-1}$.
We write $a\lesssim b$ for $a\le Cb$, and $a\asymp b$ when both $a\lesssim b$ and $b\lesssim a$.
Subscripts on $C$, $\lesssim$ or $O$ indicate additional allowed
constant dependence.
The geometric constants depend on $m,s,c$, the dimensionless chart and
packet-diameter bounds, the design constants in \Aref{2}, and the link bounds in \Aref{3}, including $c_{\rm op}$ in \eqref{eq:rank-opening}. In particular, the population pilot
constant and the multipliers used to choose the initial widths are independent of $p_0$ in \eqref{eq:packet-core}, which enters the pilot sample requirement explicitly. The
constants in Propositions~\ref{prop:finite} and~\ref{prop:readout} and in
Theorem~\ref{thm:isolated-rank} may also depend on the ratio of $\sigma_N$
to the pilot base resolution $w_Y$ of \Aref{4}. Local mass, response resolution and the units for
each statistical bound are specified below.

\subsection{Assumptions}\label{sec:theory-assumptions}

The five assumptions below specify the conditions for
Theorem~\ref{thm:isolated-rank} and Propositions~\ref{prop:pilot}--\ref{prop:readout}.
Section~\ref{sec:numerics} also studies link functions and designs beyond these
conditions.

The tube ratio $c$, the smoothness exponent, the chart bounds, the
design bounds in \Aref{2}, packet-diameter bounds and constants in \eqref{eq:rank-opening} are independent of $n,D,p$. The dimensional
quantities below are in the original predictor and response units.

\begin{description}[leftmargin=0pt,labelindent=0pt,itemsep=4pt]
\item[(A1) Geometry in reach units.]\phantomsection\label{ass:A1}
$D>m\ge1$ and $\M\subset\R^D$ is a compact $C^3$ submanifold of dimension
$m$ without
boundary, with $\reach_\M>0$. Put $R=c\reach_\M$ with a fixed
$0<c<2/3$. In charts of radius $\reach_\M/5$, write the
manifold as $q+w+g(w)$ for $w\in T_q\M$, with $g(w)\in N_q\M$.
The graph map $g$ satisfies
$\|Dg\|\le1/4$, and the dimensionless chart bound
\[
 \reach_\M\|D^2g\|+\reach_\M^2\|D^3g\|
\]
is at most a constant independent of $n,D,p$.

\item[(A2) Sampling the normal fibers.]\phantomsection\label{ass:A2}
The base point $Q$ has a $C^1$ density $\pi$ with respect to manifold volume.
Write $\pi_-=\inf_\M\pi>0$.
Write $\mu_q$ for the conditional law of $Z$ given $Q=q$. It is centered
and has support the full normal ball of radius $R$.
Its covariance $\Sigma_N(q)$, restricted to $N_q\M$, has eigenvalues
in $[\sigma_N^2,C_N\sigma_N^2]$, where
$\sigma_N>0$ is the normal standard-deviation scale and $C_N\ge1$ is fixed,
with $\sigma_N^2\asymp R^2/(D-m+2)$.
As an ambient operator, $\Sigma_N(q)$ is extended by zero on $T_q\M$.
The ratio $\sup_\M\pi/\pi_-$ and the scaled Riemannian gradient
$\sigma_N\|\nabla_\M\log\pi\|$ are bounded.
For unit normals $e$ and scalar Lipschitz functions $\phi$, with Lipschitz
seminorm $\operatorname{Lip}(\phi)$,
\begin{equation}\label{eq:normalcov}
 \E_{\mu_q}|\langle e,Z\rangle|^4\le C_N\sigma_N^4,\qquad
 \Var_{\mu_q}\phi(Z)\le C_N\sigma_N^2\operatorname{Lip}(\phi)^2 .
\end{equation}
The law may vary along the manifold. If $d_\M(q,q')\le\sigma_N/8$,
identify the two normal spaces by the isometry given by the polar factor
of orthogonal projection from $N_q\M$ to $N_{q'}\M$. The law $\mu_{q'}$
pulled back by this isometry, denoted by $\widetilde\mu_{q'}$, satisfies
\begin{equation}\label{eq:fiber-variation}
 \left\|\log\frac{d\widetilde\mu_{q'}}{d\mu_q}\right\|_{L^\infty(\mu_q)}
       \le C_N\,d_\M(q,q')/\sigma_N .
\end{equation}
Thus local changes in the sampling weights are measured on the scale of
the spread in one normal direction. Write
$\lambda(q)=\sigma_m(Df(q))$ for the smallest singular value of the
differential of the link and
$\Sigma_{\rm sing}=\{q\in\M:\lambda(q)=0\}$ for the set of points where
$Df$ is rank deficient. Let $r_{\rm sing}$, the \emph{singular-chart
radius}, be a radius with $r_{\rm sing}/\sigma_N$ a fixed positive ratio at
most $1/8$ when $\Sigma_{\rm sing}$ is nonempty. The \emph{regular region} is
$\{q:d_\M(q,\Sigma_{\rm sing})\ge r_{\rm sing}/2\}$, and the
\emph{regular condition number} $\kappa_f$ of $Df$ is the supremum of
$\sigma_1(Df(q))/\sigma_m(Df(q))$ over it. On the regular region, define
the root mean square tangential stretch
\begin{equation}\label{eq:bend}
 \omega_{\rm bend}:=\sup_{q\ {\rm in\ this\ region},\,v\in T_q\M,\,\|v\|=1}
       \big(\E_{\mu_q}\langle \II_q(v,v),Z\rangle^2\big)^{1/2}.
\end{equation}
Let $C_{\rm reg}$ be an admissible constant in \eqref{eq:regular-memory},
depending only on $m$, $c$, $C_N$ and $\kappa_f$
(Remark~\ref{rem:regular-constant}), and recall $a_\rho=e^{-1/[2(m+4)]}$ for $s=2$.
Assume that the stretch is small:
\begin{equation}\label{eq:bend-margin}
 \frac{c+C_{\rm reg}\omega_{\rm bend}}{a_\rho^2}
 +\frac{C_{\rm reg}\omega_{\rm bend}}{a_\rho^4}<1,
\end{equation}
with a gap independent of $n,D,p$. For each fixed $c<2/3<a_\rho^2$,
this permits sufficiently small positive $\omega_{\rm bend}$, which
measures curvature times the normal spread.

\item[(A3) Isolated rank deficiency of $Df$.]\phantomsection\label{ass:A3}
The link $f:\M\to\R^p$, $p\ge m$, is $C^{1,1}$. Its differential satisfies
$\|Df\|\le\sigma_{f,+}$ and is Lipschitz in parallel frames, with constant
at most $C\sigma_{f,+}/\sigma_N$. The set $\Sigma_{\rm sing}$ of points where $Df$ is rank deficient,
defined in \Aref{2}, is finite, with cardinality bounded independently of
$n$, $D$ and $p$. The balls $B_\M(q_0,2r_{\rm sing})$,
$q_0\in\Sigma_{\rm sing}$, which we call the \emph{singular charts}, are
pairwise disjoint. Since $r_{\rm sing}\le\sigma_N/8$ and $\sigma_N\le R$,
each singular chart has radius at most $R/4$, less than the chart radius
$\reach_\M/5$ of \Aref{1}.
For a fixed constant $c_{\rm op}>0$ and every $q\in\M$,
\begin{equation}\label{eq:rank-opening}
 \lambda(q)\ge c_{\rm op}\,\sigma_{f,+}
 \min\bigl\{1,\ d_\M(q,\Sigma_{\rm sing})/\sigma_N\bigr\},
\end{equation}
and at each $q_0\in\Sigma_{\rm sing}$ the nonzero singular values of
$Df(q_0)$ are at least $c_{\rm op}\sigma_{f,+}$. Thus $\lambda$ is at least
a fixed multiple of $\sigma_{f,+}$ outside the balls of radius
$r_{\rm sing}/2$, and the directions in $\ker Df(q_0)$ become visible
linearly with distance inside the singular charts.
At each $q_0$ the differential has a linear expansion in parallel frames.
For a common modulus $\omega(t)\to0$ as $t\downarrow0$,
its remainder at distance $t$ is bounded by
$(\sigma_{f,+}/\sigma_N)t\,\omega(t/\sigma_N)$. This requires a second differential at
the singular points only.

In each singular chart the tangential derivative of $F=f\circ\Pi_\M$ varies
along a normal fiber by at most a fixed fraction $\varepsilon_{\rm sing}>0$
of the weakest tangent signal:
\begin{equation}\label{eq:opening-margin}
 \sup_{z\in N_q\M,\ \|z\|\le R}
 \big\|Df(q)(I-S_{q,z})^{-1}-Df(q)\big\|
 \ \le\ \varepsilon_{\rm sing}\,\lambda(q).
\end{equation}
At $q_0$ this means $Df(q_0)S_{q_0,z}=0$; at nearby regular points it implies
$\|S_{q,z}\|\le(1+c)\varepsilon_{\rm sing}$. Thus \eqref{eq:opening-margin} restricts the shape operators throughout each singular chart. The constant $\varepsilon_{\rm sing}$ depends on the bounds in \eqref{eq:rank-opening} and the link, chart and design bounds
as in Appendix~\ref{app:propagation}. When $Df$ has full rank everywhere, $\Sigma_{\rm sing}$ is empty and
\eqref{eq:rank-opening} reads $\lambda\ge c_{\rm op}\sigma_{f,+}$ on $\M$;
set $r_{\rm sing}=\reach_\M$ in the radius $r_{\rm cyl}$ of the local cylinder \Aref{5}.

\item[(A4) Response packets visible to the scan.]\phantomsection\label{ass:A4}
Fix a packet-diameter constant $C_{\rm pkt}>0$, the response-window width
$\delta_Y$ of the scan and a base width $w_Y$. These are the response and
base resolutions of the pilot scan: by \eqref{eq:visible-packets} below,
base points within the scan radius whose true responses differ by at most
$2\delta_Y$ lie within $C_{\rm pkt}w_Y$ of one another on $\M$.
Assume there is $p_0>0$ such that, for every base point $q$,
\begin{equation}\label{eq:packet-core}
 \PP\{d_\M(Q,q)\le w_Y\,,\quad \|f(Q)-f(q)\|\le\delta_Y/4\}
 \ge p_0.
\end{equation}
Each scan radius lies strictly between
$\max\{2.05R,2R+2C_{\rm pkt}w_Y\}$ and
$3\reach_\M/2$. For base points visible to the scan assume
\begin{equation}\label{eq:visible-packets}
 \left.\begin{aligned}
  \|q'-q\|&\le\max\mathcal R+2R,\\
  \|f(q')-f(q)\|&\le2\delta_Y
 \end{aligned}\right\}
 \quad\Longrightarrow\quad
 d_\M(q',q)\le C_{\rm pkt}w_Y.
\end{equation}
Thus equal responses may occur on spatially separated packets.
The diameter constant $C_{\rm pkt}$ is fixed before choosing the
resolution. After choosing $w_Y$ and $\delta_Y$, the probability lower bound $p_0$ is
determined; in units of the local mass $\pi_-w_Y^m$ it need not remain
bounded below as the resolution decreases.
Proposition~\ref{prop:pilot} uses conditional moments, whose bounds are
independent of $p_0$.

\item[(A5) Response noise and pilot resolution.]\phantomsection\label{ass:A5}
The noise vectors are centered, independent of one another and of the predictors, with
$\|\langle v,\zeta_i\rangle\|_{\psi_2}\le\sigma_\zeta$ for unit $v$ and
bounded $\sigma_\zeta/\delta_Y$.
Set $r_{\rm cyl}=\gamma_{\rm cyl}r_{\rm sing}$,
the tangential radius of the local cylinder of
Section~\ref{sec:theory-estimator},
where $0<\gamma_{\rm cyl}<1$ is a small fixed geometric multiplier, and let
$\varepsilon_0>0$ be the fixed resolution margin. At the pilot resolution assume
\begin{equation}\label{eq:A5}
 \begin{aligned}
 \|\zeta_i\|&\le\delta_Y/8\quad\text{a.s. for all pilot observations},\\
 w_Y&\le\varepsilon_0\min\{\sigma_N,r_{\rm cyl}\}.
 \end{aligned}
\end{equation}
The resolution margin $\varepsilon_0$ is chosen after the chart and kernel bounds, $c_{\rm op}$ in \eqref{eq:rank-opening}, and pilot multipliers
(Appendix~\ref{app:propagation}); it also depends on the constants for
Phase~2 chosen there, on the gap in
\eqref{eq:bend-margin} and on the modulus $\omega$ of \Aref{3}. Neither noise bound depends on $n$ or $D$.

In Phase~1 the pilot noise bound keeps the true responses of an
anchor and each observation it selects within $2\delta_Y$ of
one another, as required by \Aref{4}. The pilot
blocks are reserved before sampling, and their random split into anchor
and design folds is independent of the data. The response-window width
$\delta_Y$ is fixed as $n$ grows. Concentration for refinement and
prediction uses the directional sub-Gaussian bound.

Together with \Aref{4}, the almost-sure bound is a signal-to-noise
condition. At a regular base point, \eqref{eq:visible-packets} applied to
base points at intrinsic distance just above $C_{\rm pkt}w_Y$ forces
$2\delta_Y\le\sigma_{f,+}C_{\rm pkt}w_Y$, so the pilot noise
norm is at most a fixed fraction, of order $C_{\rm pkt}\varepsilon_0$, of the
tangential response scale $\sigma_{f,+}\sigma_N$. Near a point where $Df$ is
rank deficient the admissible fraction is smaller, of order
$(C_{\rm pkt}\varepsilon_0)^2\,\omega(C_{\rm pkt}\varepsilon_0)$, as for the
cusp model in Remark~\ref{rem:rank-loss-scope}. For fixed packet and
resolution margins and link modulus, these fractions are independent of
$D$. The corresponding bounds in response units scale with
$\sigma_{f,+}\sigma_N$, where
$\sigma_N\asymp R/\sqrt{D-m+2}$. At fixed $R$ and $\sigma_{f,+}$ this
scale decreases with codimension. For isotropic Gaussian noise the
$\sqrt p$ term in \eqref{eq:gaussian-resolution} also limits the
per-coordinate standard deviation at a given response resolution.
\end{description}

\paragraph{Units and local mass.}
The width $\sigma_N$ measures displacement in \emph{one} normal direction.
Taking $R=c\reach_\M\asymp\sqrt{D-m}$ in physical predictor units keeps
this variance of order one as $D$ increases. The response windows and a
positive sub-Gaussian noise scale may then be fixed in physical response
units. If $p$ varies, \Aref{5} still controls the norm of the response noise
at the chosen resolution. Under $X\mapsto a_XX$ and $Y\mapsto a_YY$, the two
sides of \eqref{eq:opening-margin} and of \eqref{eq:rank-opening} scale
by $a_Y/a_X$. The fiber-law ratio in \eqref{eq:fiber-variation}, the noise relative
to the response windows, the normalized risk and the local probabilities
$\pi_-b^m$ are unchanged. Since $\pi$ integrates to one against
manifold volume and $\sup_\M\pi/\pi_-$ is bounded,
$\pi_-\sigma_N^m\asymp\sigma_N^m/\operatorname{vol}(\M)$: the local mass is
the inverse volume of $\M$ in units of $\sigma_N^m$. The size of $\M$ enters
the bounds only through this ratio; for connected $\M$ the response diameter
$\operatorname{diam}(f(\M))$ in the last term of \eqref{eq:rank-profile-risk}
is also bounded by it (Appendix~\ref{app:main}).

\subsection{Choice of scales}\label{sec:theory-scales}
\phantomsection\label{par:theory-normal-widths}
The normal width starts at a fixed multiple of the mean selected radius.
Its multiplier keeps $h_k/R$ above the minimum normal width (in units of $R$)
in Lemma~\ref{lem:quadratic-update} and below a fixed upper bound.
The initial tangential width $b_0$ is comparable to a sufficiently large
fixed multiple of $w_Y$, so the pilot tilt fits within $b_1$.
The local cylinder of Section~\ref{sec:theory-estimator} keeps the
base points of the observations used at a target in one graph chart of \Aref{1}, even
when $h_k$ exceeds the reach. Thereafter $b_k$ shrinks geometrically. Refinement stops at the finest
width for which the local count $n_f\pi_- b_k^m$ times $(b_k/\sigma_N)^4$
covers the sampling cost in \eqref{eq:sample-condition}. Appendix~\ref{app:defs}
gives the integer rule, and Appendix~\ref{app:n0} the sample size needed to start it.

\subsection{The estimator used in the theorem}\label{sec:theory-estimator}

The three phases use the scales of Section~\ref{sec:theory-scales}.
We specify here which observations enter each step and how the pilot
restricts a neighborhood to one chart. Appendix~\ref{app:defs} gives the
thresholds for the anchor stars and the filters for degenerate fits.

\paragraph{Sample splitting.}
The Phase~1 observations are split uniformly at random between the design
fold $\Sdes$, which selects the scan radii, and the anchor fold
$\Sanc$, which supplies the projectors to be pooled. Each refinement step $k$ uses fresh fit and pool folds $\Sfit,\Spool$, and prediction uses a separate fold $\Srd$ of size $n_{\rm pred}$ (Appendix~\ref{app:defs}).

\paragraph{Phase 1.}\phantomsection\label{par:theory-phase1}
For each anchor in the anchor fold, the scan uses the design-fold observations
in \(U_i\cap B(X_i,r)\).  It retains radii passing the count and covariance
filters of Appendix~\ref{app:defs} and chooses the smallest maximizer of
\eqref{eq:hscore}.  An anchor is \emph{eligible} if at least one
radius is retained. Each eligible anchor stores the mean of the observations
selected at its chosen radius, its \emph{selected mean}, and a \emph{pooled
projector} \eqref{eq:pilotpool}, obtained by averaging the projectors onto
the bottom $m$ covariance eigenspaces of nearby eligible anchors.

\paragraph{Anchor stars and the local cylinder.}
Two eligible anchors are linked if their responses and their selected
means are close, at the thresholds of Appendix~\ref{app:defs}. An
\emph{anchor star} $S$ consists of an eligible anchor and the eligible
anchors linked to it. There is one such star for each eligible anchor,
and the stars may overlap. Its \emph{center} $x_S$ is the average of
their selected means, and $P_S$ projects onto the leading $m$ eigenspace of
the average of their pooled projectors, so that $x_S+\Ran P_S$ estimates a
tangent plane. Each target $x$ is assigned an anchor star minimizing the
tangential residual $\|P_S(x-x_S)\|$ among the anchor stars with
$\|x-x_S\|\le2R$ that pass the size threshold of Appendix~\ref{app:defs},
and we set $\widehat P_0(x)=P_S$.

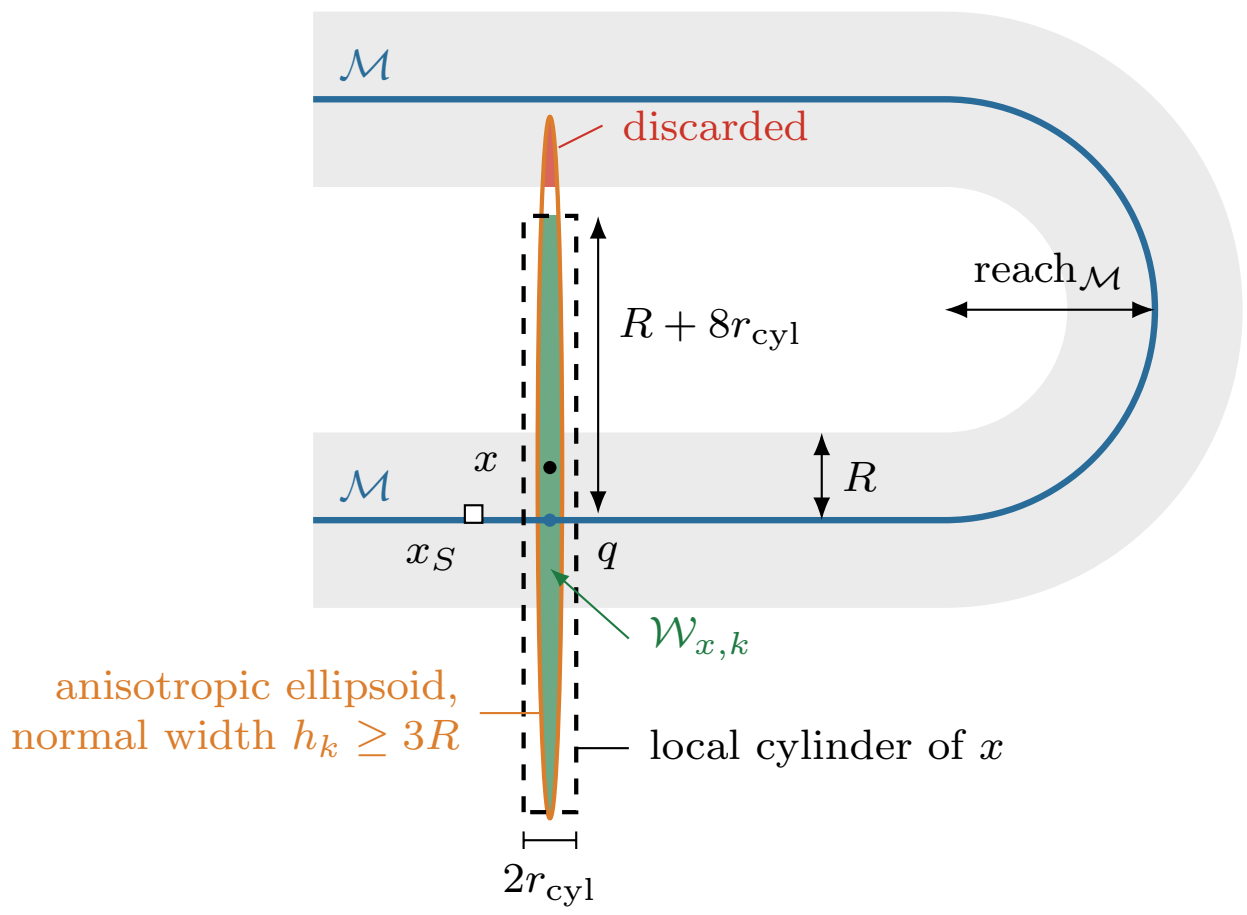
\begin{figure}[t]
\centering
\begin{tikzpicture}[scale=0.58, font=\scriptsize]
  \definecolor{lcblue}{rgb}{0.16,0.42,0.60}
  \definecolor{lcorange}{rgb}{0.86,0.49,0.17}
  \definecolor{lcred}{rgb}{0.80,0.20,0.15}
  \definecolor{lcgreen}{rgb}{0.12,0.48,0.26}
  \fill[black!8] (-4.2,-1) -- (3,-1) arc[start angle=-90, end angle=90, radius=3.4]
     -- (-4.2,5.8) -- (-4.2,3.8) -- (3,3.8) arc[start angle=90, end angle=-90, radius=1.4]
     -- (-4.2,1) -- cycle;
  \begin{scope}
    \clip (-1.8,-3.33) rectangle (-1.2,3.47);
    \fill[lcgreen!65] (-1.5,0.6) ellipse [x radius=0.14, y radius=4.0];
  \end{scope}
  \begin{scope}
    \clip (-4.2,3.8) rectangle (3,5.8);
    \fill[lcred!75] (-1.5,0.6) ellipse [x radius=0.14, y radius=4.0];
  \end{scope}
  \draw[lcorange, thick] (-1.5,0.6) ellipse [x radius=0.14, y radius=4.0];
  \draw[lcblue, very thick] (-4.2,0) -- (3,0) arc[start angle=-90, end angle=90, radius=2.4] -- (-4.2,4.8);
  \node[lcblue, above] at (-3.6,0) {$\M$};
  \node[lcblue, above] at (-3.6,4.8) {$\M$};
  \draw[dashed, thick] (-1.8,-3.33) rectangle (-1.2,3.47);
  \fill (-1.5,0.6) circle [radius=0.075];
  \node[anchor=east] at (-1.85,0.66) {$x$};
  \fill[lcblue] (-1.5,0) circle [radius=0.075];
  \node[anchor=north west] at (-1.2,-0.02) {$q$};
  \draw[fill=white] (-2.47,-0.03) rectangle (-2.27,0.17);
  \node[anchor=north east] at (-2.3,-0.05) {$x_S$};
  \draw (-1.8,-3.55) -- (-1.8,-3.75) (-1.2,-3.55) -- (-1.2,-3.75) (-1.8,-3.65) -- (-1.2,-3.65);
  \node[below] at (-1.5,-3.7) {$2r_{\rm cyl}$};
  \draw[<->, >={Latex[length=1.5mm]}] (-0.95,0.07) -- (-0.95,3.47)
     node[pos=0.62, right] {$R+8r_{\rm cyl}$};
  \draw[<->, >={Latex[length=1.5mm]}] (1.6,0) -- (1.6,1) node[midway, right] {$R$};
  \draw[<->, >={Latex[length=1.5mm]}] (3,2.4) -- (5.4,2.4) node[midway, above] {$\reach_\M$};
  \node[lcorange, anchor=east, align=right] (nb) at (-2.3,-2.2) {anisotropic ellipsoid,\\normal width $h_k\ge3R$};
  \draw[lcorange, thin] (nb.east) -- (-1.62,-2.2);
  \node[lcgreen, anchor=west] (pool) at (-0.6,-1.35) {$\Wwin_{x,k}$};
  \draw[lcgreen, thin, -{Latex[length=1.3mm]}] (pool.west) -- (-1.5,-0.55);
  \node[anchor=west] (lc) at (-0.6,-2.6) {local cylinder of $x$};
  \draw[thin] (lc.west) -- (-1.2,-2.6);
  \node[lcred, anchor=west] (dc) at (-0.9,4.5) {discarded};
  \draw[lcred, thin] (dc.west) -- (-1.4,4.25);
\end{tikzpicture}
\caption{Local cylinder and pooling neighborhood for a curve in the plane
($m=1$, $D=2$); the drawing is not to scale. Grey band: the predictor
tube of radius $R$ around $\M$ (blue). Orange ellipse:
$\{v:d_{\widehat P_{k-1}(x),b_k,h_k}(v-x)\le1\}$, with the metric
\eqref{eq:metric}, centered at the target $x$ above $q$.
Its normal width $h_k\ge3R$ can exceed the reach and admit predictors
from a distant part of the tube (red). The dashed rectangle is the
local cylinder \eqref{eq:local-cylinder}, oriented by the pilot projector
$P_S$: its tangential half-width $r_{\rm cyl}$ is measured from $x$,
and its normal half-height $R+8r_{\rm cyl}$ from $x_S+\Ran P_S$.
The green region is the pooling neighborhood $\Wwin_{x,k}$ in
\eqref{eq:window}, the intersection of the ellipsoid and cylinder;
only its intersection with the grey tube contains predictors.
It retains complete normal fibers near $q$ and excludes the distant red
part (Lemma~\ref{lem:cylinder-localization}). The cylinder is fixed by
the pilot, whereas the ellipsoid uses the step-$k$ widths and the
current projector $\widehat P_{k-1}(x)$; their axes are drawn parallel here.}
\label{fig:local-cylinder}
\end{figure}

The neighborhoods of Phases~2 and~3 have normal width at least $3R$
(Appendix~\ref{app:defs}), which may exceed the reach, so they can contain
predictors whose base points lie on a distant part of $\M$
(Figure~\ref{fig:local-cylinder}). They are therefore intersected with the
\emph{local cylinder} of the target $x$, the set of points $x'$ with
\begin{equation}\label{eq:local-cylinder}
 \|P_S(x'-x)\|\le r_{\rm cyl},\qquad
 \|P_S^\perp(x'-x_S)\|\le R+8r_{\rm cyl},\qquad
 \|P_S(x'-x_S)\|\le8r_{\rm cyl},
\end{equation}
where $r_{\rm cyl}$ is the radius fixed in \Aref{5}. The first two conditions
form a cylinder with tangential radius $r_{\rm cyl}$ measured from the
target and normal radius $R+8r_{\rm cyl}$ measured from the estimated
plane $x_S+\Ran P_S$. The third bounds the tangential displacement from
the star center; on the event of Proposition~\ref{prop:pilot}, it follows
from the first
(Appendix~\ref{app:pilot}). Proposition~\ref{prop:pilot} shows that the
local cylinder contains the complete normal fibers over
$B_\M(\Pi_\M x,r_{\rm cyl}/4)$ and that the points it contains have base
points in a single graph chart of \Aref{1}.

\paragraph{Phase 2.}
The normal width $h_k$ and the aspect ratio $\rho_k$ follow the
multiplicative schedule, with per-step factors $a_h$ and $a_\rho$,
\begin{equation}\label{eq:schedule}
\begin{gathered}
  h_0=h_{\rm rescale}\operatorname*{avg}_{i\ {\rm eligible}}r_i,\qquad
  h_k=a_hh_{k-1},\qquad \rho_k=a_\rho\rho_{k-1},\qquad 1\le k\le k_*,\\
  a_\rho=e^{-1/[2(2s+m)]},\qquad
  a_h=\exp\Bigl\{\frac1{2(D+4)k_*}\Bigr\}.
\end{gathered}
\end{equation}
The normal width remains comparable to the reach, and $b_k=\rho_kh_k$ is
the tangential width. Let $\chi_0(x,v)$ indicate membership of $v$ in
the local cylinder \eqref{eq:local-cylinder} of $x$. The pooling neighborhood
at $x$ and the fitting weights at a center $u$ are
\begin{equation}\label{eq:window}
\begin{aligned}
 \Wwin_{x,k}
 &=\{v:\ \chi_0(x,v)=1,\ d_{\widehat P_{k-1}(x),b_k,h_k}(v-x)\le1\},\\
 \omega_{u,k}(v)
 &=\chi_0(u,v)K\bigl(d_{\widehat P_{k-1}(u),b_k,h_k}(v-u)^2\bigr).
\end{aligned}
\end{equation}
Each pool center $u$ fits a slope on the fit fold with weights $\omega_{u,k}$.
Pooling the stored outer products over $\Wwin_{x,k}$ gives
$\widehat C_k(x)$ and hence $\widehat P_k(x)=\Ptop(\widehat C_k(x))$ on
the concentration event. The effective local count is $N_k:=n_f\pi_-b_k^m$,
where $n_f$ is the size of a sample block up to integer rounding
(Appendix~\ref{app:defs}).

\paragraph{Phase 3.}
The prediction estimator $\widehat F$ in \eqref{eq:readout}
uses the prediction fold, with widths $h_*$ and $b_*=\rho_*h_*$,
and restricts the kernel to the local cylinder \eqref{eq:local-cylinder}. The widths prescribed in Appendix~\ref{app:defs} ensure that,
at a query $x=q+z$, the prediction kernel is at least $1/16$ on every normal
fiber above $B_\M(q,b_*/16)$, and place the restricted support strictly inside a local base chart
(Lemma~\ref{lem:cylinder-localization}).

\subsection{Results}\label{sec:theory-results}
At a target $x=q+z$, write $d(q)=\min\{1,d_\M(q,\Sigma_{\rm sing})/\sigma_N\}$,
with $d=1$ if the singular set is empty. This distance distinguishes the
tangent directions recovered at the center from those recovered through its
neighbors. The following results use \Aref{1}--\Aref{5}, the estimator
above, and $n\ge n_0$, where the sample threshold is
\begin{equation}\label{eq:n0}
\begin{aligned}
 n_0&=C\,\Theta\log^2(2+\Theta),\\
 \Theta&=\max\Bigl\{
 \frac{(1+D+p)^4(1+R/\sigma_N)^{16}}{\pi_-\sigma_N^m\,(w_Y/\sigma_N)^{m+4}}
 \Bigl(1+\frac{\sigma_\zeta}{\sigma_{f,+}\sigma_N}\Bigr)^2,\\
 &\qquad\quad\ \frac{D+\log(2|\mathcal R|/p_0)}{p_0}
 \Bigl(1+\frac{\max\mathcal R}{w_Y}
       +\frac{(\max\mathcal R)^2\reach_\M}{\sigma_N^2w_Y}\Bigr)^2\Bigr\},
\end{aligned}
\end{equation}
with a constant $C$ that depends on $\tau$. Appendix~\ref{app:n0} shows that
$n\ge n_0$ implies the two sample conditions used in the proofs.

\begin{proposition}[Phase 1: the warm start]\label{prop:pilot}
With probability at least $1-Cn^{-\tau-2}$ every target $x=q+z$ is
assigned an anchor star $S$ with center $x_S$ satisfying $\|x_S-q\|\le Cw_Y$, and
\[
 \|\widehat P_0(x)-P_q\|\le C w_Y/\reach_\M .
\]
The local cylinder of $x$ contains every normal fiber above
$B_\M(q,r_{\rm cyl}/4)$, and the points it contains have base points in
$B_\M(q,C r_{\rm cyl})$, contained in a single graph chart of \Aref{1}.
\end{proposition}

\begin{proposition}[Phase 2: a refinement neighborhood]\label{prop:step}
Fix $1\le k\le k_*$, with $0<\eta\le1/16$, $A\ge1$ and the
initial widths $h_0,b_0$ chosen as in Appendix~\ref{app:propagation},
satisfying \eqref{eq:refinement-resolution}; use the schedule \eqref{eq:schedule}.
With probability at least $1-C(k+1)n^{-\tau-2}$, the following hold
simultaneously at every tube point $x=q+z$: $\Wwin_{x,k}$ contains the complete normal fibers
above $B_\M(q,b_k/16)$, and $\omega_{x,k}\ge1/16$ there. Moreover,
$\E\,\omega_{x,k}(X)\asymp\PP\{X\in\Wwin_{x,k}\}\asymp\pi_-b_k^m$
(Lemma~\ref{lem:cylinder-localization}), and the updated projector satisfies
$R\|\widehat P_k(x)-P_q\|<a_\rho\eta b_k\le\eta b_{k+1}$
throughout the tube.
\end{proposition}

The proof combines the concentration bounds
\eqref{eq:fit-gram-concentration}--\eqref{eq:pool-concentration} with estimates in
three regions: within distance $Ab_k$ of a point where $Df$ is rank deficient
(Lemma~\ref{lem:quadratic-update}); the surrounding annulus up to distance
$r_{\rm sing}$ \eqref{eq:annular-score}; and the regular region, where two
successive updates control the error \eqref{eq:regular-memory}.
Induction in Appendix~\ref{app:propagation} gives the stated probability.

\begin{proposition}[Phase 2: all refinement steps]\label{prop:finite}
With probability at least $1-Cn^{-\tau-1}$, simultaneously for
$1\le k\le k_*$ and every tube point $x=q+z$, $\Wwin_{x,k}$ contains
every normal fiber above $B_\M(q,b_k/16)$, $\omega_{x,k}\ge1/16$ on these
fibers, $R\|\widehat P_k(x)-P_q\|<a_\rho\eta b_k$, and
\begin{equation}
 \|\widehat P_k(x)-P_q\|
 \le C\,\frac{\sigma_N}{R}
   \frac{(b_k/\sigma_N)^2}{d(q)+b_k/\sigma_N}.\label{eq:finite-envelope}
\end{equation}
\end{proposition}

\begin{proposition}[Phase 3: prediction]\label{prop:readout}
Fix the pilot and refinement data and the pilot split on the event of Proposition~\ref{prop:finite}
and put $u=b_*/\sigma_N$. In units of $(\sigma_{f,+}\sigma_N)^2$, the
squared prediction error of the estimator $\widehat F$ defined in \eqref{eq:readout},
averaged over the prediction fold
and an independent query, is at most
\begin{equation}\label{eq:prediction-bound}
 C\left[
 u^4\{1+\ind_{\{m=2,\ \exists q\in\M:\ \operatorname{rank}Df(q)=1\}}|\log u|\}
 +\frac{p\{\sigma_\zeta/(\sigma_{f,+}\sigma_N)\}^2+u^2}
             {n_{\rm pred}\pi_- b_*^m}\right]
\end{equation}
plus the exponentially small contribution \eqref{eq:denominator-risk} from
a small prediction denominator.
The extra logarithm is needed in this bound only for a surface with such a point.
\end{proposition}

The estimator in \eqref{eq:readout} uses the averaging formula \eqref{e:Fhat_def} on the prediction fold $\Srd$, with weights multiplied by the cylinder indicator $\chi_0(x,X_i)$, and projections $\widehat P_{k_*}$ constructed on the separate pilot folds $\Sanc,\Sdes$ and refinement folds $\Sfit,\Spool$, $1\le k\le k_*$.

\begin{theorem}[Learning with rank-deficient $Df$]\label{thm:isolated-rank}
Let $(X,Y)$ satisfy \Aref{1}--\Aref{5} with $D>m$ and $p\ge m$,
for the margins $\varepsilon_{\rm sing}$, $\gamma_{\rm cyl}$ and
$\varepsilon_0$ determined in Appendix~\ref{app:propagation}; these are
positive and depend only on the fixed model parameters.
Fix $0<c<2/3$ and a confidence exponent $\tau>0$.
Let $n_0=C\Theta\log^2(2+\Theta)$ be the sample threshold
\eqref{eq:n0}; $n_0$ and the constants below depend on $\tau$. For $n\ge n_0$,
the MANOR estimator with $s=2$, tuned as in Appendix~\ref{app:defs} with $m$,
$\delta_Y$, the radii $\mathcal R$ and the scales $R$, $\sigma_N$,
$r_{\rm sing}$, $w_Y$, $p_0$, $\pi_-$, $\sigma_\zeta$ and $\sigma_{f,+}$, has mean tangent-space error
\begin{equation}\label{eq:tan-risk}
  \mathcal E_{\rm tan}(\widehat P_{k_*})
    =\widetilde O\bigl(n^{-2/(m+4)}\bigr),
\end{equation}
with the expectation taken over the training data, the pilot split and an
independent query. Its integrated squared prediction error satisfies
\begin{equation}\label{eq:rank-profile-risk}
 \frac{\E\|\widehat F(X)-F(X)\|^2}{(\sigma_{f,+}\sigma_N)^2}
 \ \le\ \frac{C(D,p)\,\polylog(n)}{n^{4/(m+4)}}
 + \frac{C\{\operatorname{diam}(f(\M))^2+p\sigma_\zeta^2\}}
        {n^{\tau+1}(\sigma_{f,+}\sigma_N)^2}.
\end{equation}
The second term is the contribution of the exceptional events of
Propositions~\ref{prop:pilot} and~\ref{prop:finite} and of a small
prediction denominator, which have total probability at most
$Cn^{-\tau-1}$ and on which the prediction error is bounded only by the
response diameter and the noise.
For fixed intrinsic dimension, dimensionless model margins, scale ratios
and $\operatorname{vol}(\M)/\reach_\M^m$, the constant in the tangent bound, $C(D,p)$ and the sample threshold
$n_0$ grow at most polynomially in $D+p$ (Appendix~\ref{app:n0}).
Constructing the estimator requires
\[
 O\bigl(n|\mathcal R|(nD^2+D^3)+k_*\{n^2(D^2+Dp)+n(D^3+D^2p)\}\bigr)
\]
operations, where $k_*=O(\log n)$ is the number of refinement steps, and
each evaluation of $\widehat F$ requires $O(n(Dm+p))$ operations
(Section~\ref{sec:core-algorithm}). For a fixed radius grid $\mathcal R$
and $n\ge D$, the construction cost is $O(n^2(D^2+Dp)\log n)$.
\end{theorem}

The exponent $4/(m+4)$ is that of nonparametric regression with two orders
of smoothness in $m$ dimensions \citep{Stone82}. Appendix~\ref{app:learnability}
establishes this minimax exponent, for fixed positive noise, over the broader
class of Manifold-Index regression functions with the regularity of
\Aref{1} and \Aref{3}. Least squares over a finite net attains it without
\Aref{2} and \Aref{4}, using of \Aref{5} only the sub-Gaussian noise bound; the net has cardinality bounded exponentially
in a power of $n$, but the argument supplies no algorithm for constructing
and evaluating its representatives. The transfer of the minimax conclusion
to the restricted class \Aref{1}--\Aref{5} is conditional on the subclass
conditions stated there. The tube ratio $c$
enters Phase~1 only through a chord inequality that requires $c<2/3$. In
Phase~2 it enters jointly with the bound on root mean square tangential
stretch in \Aref{2}. On a connected manifold the local sampling mass also bounds the
normalized response range (Appendix~\ref{app:main}). When the signal variance
is positive, dividing the squared prediction error by it gives the relative
risk $\mathcal E_{\rm pred}$ in \eqref{eq:intro-metrics}.

\paragraph{Scope of the result.}
Near an isolated rank loss, \Aref{3} and \Aref{4} serve different purposes:
neighboring gradients recover the missing tangent directions, while
localized response packets provide the warm start. Scalar extrema on
closed curves violate the packet condition at sufficiently fine pilot
resolution; vector-valued cusp maps can satisfy it
(Remark~\ref{rem:rank-loss-scope}). Appendix~\ref{app:cusp-example}
gives a fully specified noisy model and explains the restrictions in
the equality case $p=m$.

The Gaussian-noise extension in Corollary~\ref{cor:gaussian-noise}
applies over a finite sample-size range when the noise variance and
pilot response resolution are fixed. Its statement and proof are in
Appendix~\ref{app:gaussian-corollary}.

\paragraph{Uniform tangent-space error with rank-deficient \texorpdfstring{$Df$}{Df}.}
On the event of Proposition~\ref{prop:finite}, setting $d(q_0)=0$ in
\eqref{eq:finite-envelope} bounds the tangent-space error everywhere on the
normal fiber above a singular point $q_0$ by $C b_{k_*}/R$, or
$\widetilde O(n^{-1/(m+4)})$ at the terminal scale. This is also a uniform
bound over the tube. The faster mean rate in Theorem~\ref{thm:isolated-rank}
comes from integrating the distance-dependent bound over the base points
(Appendix~\ref{app:main}). If $Df$ has full rank throughout $\M$, then
$d\equiv1$ and $\widetilde O(n^{-2/(m+4)})$ holds uniformly as well.

\subsection{How the results fit together}\label{sec:theory-roadmap}

\begin{figure}[!htbp]
\centering
\begin{tikzpicture}[
  box/.style={draw, rounded corners, align=center, inner sep=3pt, font=\small,
              minimum height=1.15cm, text width=2.05cm},
  lab/.style={font=\scriptsize, align=center},
  arr/.style={-{Latex[length=2mm]}, thick}]
\node[box] (p1) {Phase 1\\warm start\\(Prop.~\ref{prop:pilot})};
\node[box, right=2.35cm of p1] (p2) {Phase 2\\step $k$\\(Prop.~\ref{prop:step})};
\node[box, right=2.35cm of p2] (p3) {Phase 3\\prediction\\(Prop.~\ref{prop:readout})};
\node[box, text width=2.0cm, right=0.5cm of p3] (th) {Theorem~\ref{thm:isolated-rank}};
\draw[arr] (p1) -- node[lab, above] {initial angle and} node[lab, below] {local cylinder} (p2);
\draw[arr] ([xshift=-4mm]p2.north east) .. controls +(0.45,0.8) and +(-0.45,0.8) .. ([xshift=4mm]p2.north west)
  node[lab, pos=0.5, above] {complete fibers at the next width};
\draw[arr] (p2) -- node[lab, above] {terminal angles} node[lab, below] {(Prop.~\ref{prop:finite})} (p3);
\draw[arr] (p3) -- (th);
\end{tikzpicture}
\caption{From a response-selected warm start to prediction. The successive
neighborhoods retain complete normal fibers. The terminal angle varies with the
distance to the nearest singular point; its integrated effect determines
the prediction bound.}
\label{fig:roadmap}
\end{figure}
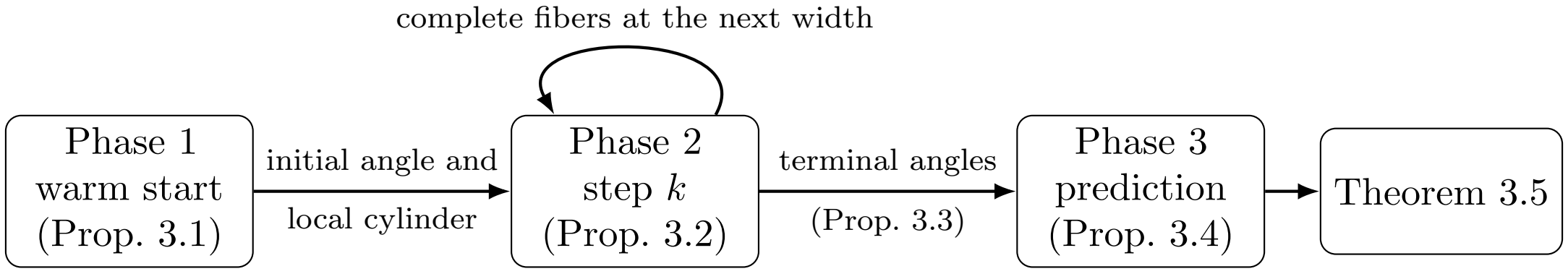

Figure~\ref{fig:roadmap} traces the dependencies of the results leading to the main Theorem. The pilot
gives a single local chart and the first aligned neighborhood. The fitted
slopes keep subsequent neighborhoods aligned, with the weakest tangent signal coming from
a whole ball of regular centers near a point where $Df$ is rank deficient. Prediction uses the stronger
angle bound away from the finitely many points where $Df$ is rank deficient and integrates the remaining
error near them.

\paragraph{Dimension dependence.}
In the sample conditions \eqref{eq:sample-condition} and \eqref{eq:pilotN},
the ambient and response dimensions enter through explicit polynomial
factors, such as $(1+D+p)^4$ and $(1+R/\sigma_N)^{16}$ with
$R/\sigma_N\asymp\sqrt{D-m+2}$. The local mass
$\pi_-\sigma_N^m$ is of the order of the probability that a base point falls in a geodesic
ball of radius $\sigma_N$ on $\M$. It is comparable to
$\sigma_N^m/\operatorname{vol}(\M)$, the inverse of the number of such balls
needed to cover $\M$. Since $R=c\reach_\M$,
\[
 (\pi_-\sigma_N^m)^{-1}\asymp
 \frac{\operatorname{vol}(\M)}{\reach_\M^m}\,c^{-m}(D-m+2)^{m/2}.
\]
For fixed $m$ and $c$, this contribution is polynomial in $D$ provided
$\operatorname{vol}(\M)/\reach_\M^m$ grows at most polynomially.
The theorem retains the local mass explicitly, without imposing this volume bound.
The probability lower bound in \eqref{eq:packet-core} enters through $p_0^{-1}$. Since $p_0\le C\pi_-w_Y^m$,
it also grows at least like $(D-m+2)^{m/2}$, and it is polynomial in $D$
when, in addition, $p_0/(\pi_-w_Y^m)$ and $\sigma_N/w_Y$ are fixed.

\section{Numerical experiments}\label{sec:numerics}

Table~\ref{tab:experiment-overview} summarizes five experiments. The line
segment and Meyer helix have non-monotone scalar links (\(p=m=1\)) and
show how prediction and tangent-space errors vary with sample size and
ambient dimension. The bumpy sphere and coupled Meyer helix have
links whose differentials have rank \(m\). On these surfaces, normal
displacements scale with local reach rather than the global reach used in
the theorem. The bumpy sphere also
shows how the errors change over successive refinement steps. The diffusion
experiment tests prediction when first-hit probabilities vary along a
normal fiber. Appendix~\ref{app:numerics-details} gives the
constructions and sampling distributions, including the non-uniform
normal laws.

\begin{table}[htb]
\centering
\small
\begin{tabularx}{\linewidth}{@{}lcccXc@{}}
\toprule
experiment & $m$ & $D$ & $p$ & what it tests & Section\\
\midrule
single-index (segment) & 1 & 18, 24, 30 & 1 & rates for a flat manifold & \ref{sec:numerics-sim}\\
Meyer helix & 1 & 18, 24, 30 & 1 & rates for a curve of growing complexity & \ref{sec:numerics-helix}\\
bumpy sphere & 2 & 20 & 6 & rates for a curved surface & \ref{sec:numerics-sphere}\\
coupled Meyer helix & 2 & 20 & 3 & rates for an oscillatory surface & \ref{sec:numerics-coupled-helix}\\
first-hit probabilities & 2 & 20 & 4 & robustness to an approximate model & \ref{sec:numerics-reaction}\\
\bottomrule
\end{tabularx}
\caption{Overview of the numerical experiments.}
\label{tab:experiment-overview}
\end{table}

\subsection{Implementation and error measures}\label{sec:numerics-setup}

\paragraph{Implementation.}
The numerical implementation chooses the scan radii from the data and
reuses observations across refinement rounds. Within each scaling
experiment, the parameters governing the scan and the refinement schedule
are held fixed across sample sizes and repetitions.

\paragraph{Error measures.}
For prediction we report the test mean squared error relative to the
response variance (relMSE). The curve experiments use $\Var(Y)$ as the
denominator. The surface scaling experiments use the total empirical
variance of the noiseless test responses, corresponding to the signal
variance in \eqref{eq:intro-metrics}. For the diffusion experiment,
independent Monte Carlo estimates provide the reference responses
(Section~\ref{sec:numerics-reaction}).

For tangent estimation the scaling plots report the mean discrepancy over
independent test queries (devTan). For a curve
$\gamma$, the pointwise error is
$\|\widehat v(X)-\gamma'\|$, where $\widehat v(X)$ is the estimated unit
tangent vector and $\gamma'$ is the unit tangent vector of $\gamma$ at
$\Pi_\M X$, oriented so that $\langle\widehat v(X),\gamma'\rangle\ge0$. For a
surface, let $\theta_1,\theta_2$ be the principal angles between
the estimated and true tangent planes. The pointwise error is
\begin{equation}\label{eq:tangent-error-num}
  \frac{\|\widehat P(X)-P_{\Pi_\M X}\|_F}{\sqrt2}
  =\bigl(\sin^2\theta_1+\sin^2\theta_2\bigr)^{1/2},
\end{equation}
On both curves and surfaces the pointwise
discrepancy lies between the projector operator norm in
\eqref{eq:intro-metrics} and $\sqrt2$ times that norm.

\paragraph{Reference slopes.}
Theorem~\ref{thm:isolated-rank} gives the reference exponents for the two
errors plotted against sample size: $-4/(m+4)$ for relMSE and $-2/(m+4)$
for mean tangent-space error. Thus the reference slopes are $-0.8$ and
$-0.4$ on curves, and $-2/3$ and $-1/3$ on surfaces, respectively.
We fit slopes by least squares on logarithmic axes over the sample-size
ranges specified for each experiment.

\subsection{A single-index model}\label{sec:numerics-sim}

Set the normal standard deviation $\sigma_\gamma=0.5$ and the
segment length $\mathrm{len}_\gamma=\sqrt{12}\,\sigma_\gamma\sqrt D$.
In the \emph{single-index model}, the curve $\gamma$ is the segment
$\{x\in\R^D:\ 0\le x_1\le\mathrm{len}_\gamma,\ x_2=\dots=x_D=0\}$, and the
predictor has independent coordinates
$X_1\sim\mathrm{Unif}(0,\mathrm{len}_\gamma)$ and
$X_2,\dots,X_D\sim\mathcal N(0,\sigma_\gamma^2)$. Thus
$\Var(X_1)=D\sigma_\gamma^2$ is comparable with the total normal variance
$(D-1)\sigma_\gamma^2$, and the regression function $F(X)=f(X_1)$ depends on
$X$ through a single linear index. Writing $t\in[0,\mathrm{len}_\gamma]$ for
the arc length, the link function is $f(t)=f_*(t/\mathrm{len}_\gamma)$, where
\begin{equation}\label{eq:linkf}
  f_*(t)=64t^3-96t^2+36t-2+e^{t}+0.8\cos(\pi t)+2\sin(5\pi t+0.1),
  \qquad 0\le t\le1.
\end{equation}
\noindent
\begin{minipage}[c]{.49\linewidth}
\centering
\begin{tikzpicture}[font=\scriptsize]
  \path[use as bounding box] (0,0) rectangle (7.45cm,5cm);
  \begin{scope}[shift={(0.65cm,2.15cm)},x=6.65cm,y=.48cm]
    \draw[line width=.4pt] (0,-3) rectangle (1,5);

    \foreach \x/\lab in {0/0,.1/{0.1},.2/{0.2},.3/{0.3},.4/{0.4},
                            .5/{0.5},.6/{0.6},.7/{0.7},.8/{0.8},.9/{0.9},1/1} {
      \draw[line width=.35pt] (\x,-3) -- +(0,.12);
      \draw[line width=.35pt] (\x,5) -- +(0,-.12);
      \node[anchor=north] at (\x,-3) {\lab};
    }
    \foreach \y in {-3,-2,...,5} {
      \draw[line width=.35pt] (0,\y) -- +(.012,0);
      \draw[line width=.35pt] (1,\y) -- +(-.012,0);
      \node[anchor=east] at (0,\y) {\y};
    }

    \begin{scope}
      \clip (0,-3) rectangle (1,5);
      \draw[blue!75!black,line width=1pt]
        plot[domain=0:1,samples=201,smooth]
        (\x,{64*(\x)^3-96*(\x)^2+36*\x-2+exp(\x)
             +0.8*cos(deg(pi*\x))+2*sin(deg(5*pi*\x+0.1))});
    \end{scope}

    \node at (.5,5.55) {\small $f_*$};
    \node at (.5,-4) {\small $t$};
  \end{scope}
\end{tikzpicture}
\makeatletter\def\@captype{figure}\makeatother
\caption{Normalized link $f_*$ in \eqref{eq:linkf} for the line segment
and the Meyer helix.}
\label{fig:linkf}
\end{minipage}\hfill
\begin{minipage}[c]{.46\linewidth}
\setlength{\parindent}{0pt}
A response in the range of $f_*$ has between one and five preimages
(Figure~\ref{fig:linkf}). The noise standard deviation is $1\%$ of the
response standard deviation.

\smallskip
At each $n$ we use a $90{:}10$ training--test split and three independent
repetitions. We evaluate prediction against the noiseless response,
normalize by $\Var(Y)$ and set $\delta_Y=(\max_iY_i-\min_iY_i)/50$.
At each $D$ we compare three fixed multipliers $c_\rho$ in
$\rho_*=c_\rho\rho_{k_*}$ (Appendix~\ref{app:helix}).
\end{minipage}\par

\begin{figure}[!htb]
\centering
\begin{minipage}[t]{.32\linewidth}
\centering
{\small $D=18$}\\[2pt]
\includegraphics[viewport=0bp 2bp 260bp 335bp,clip,height=1.25\linewidth]{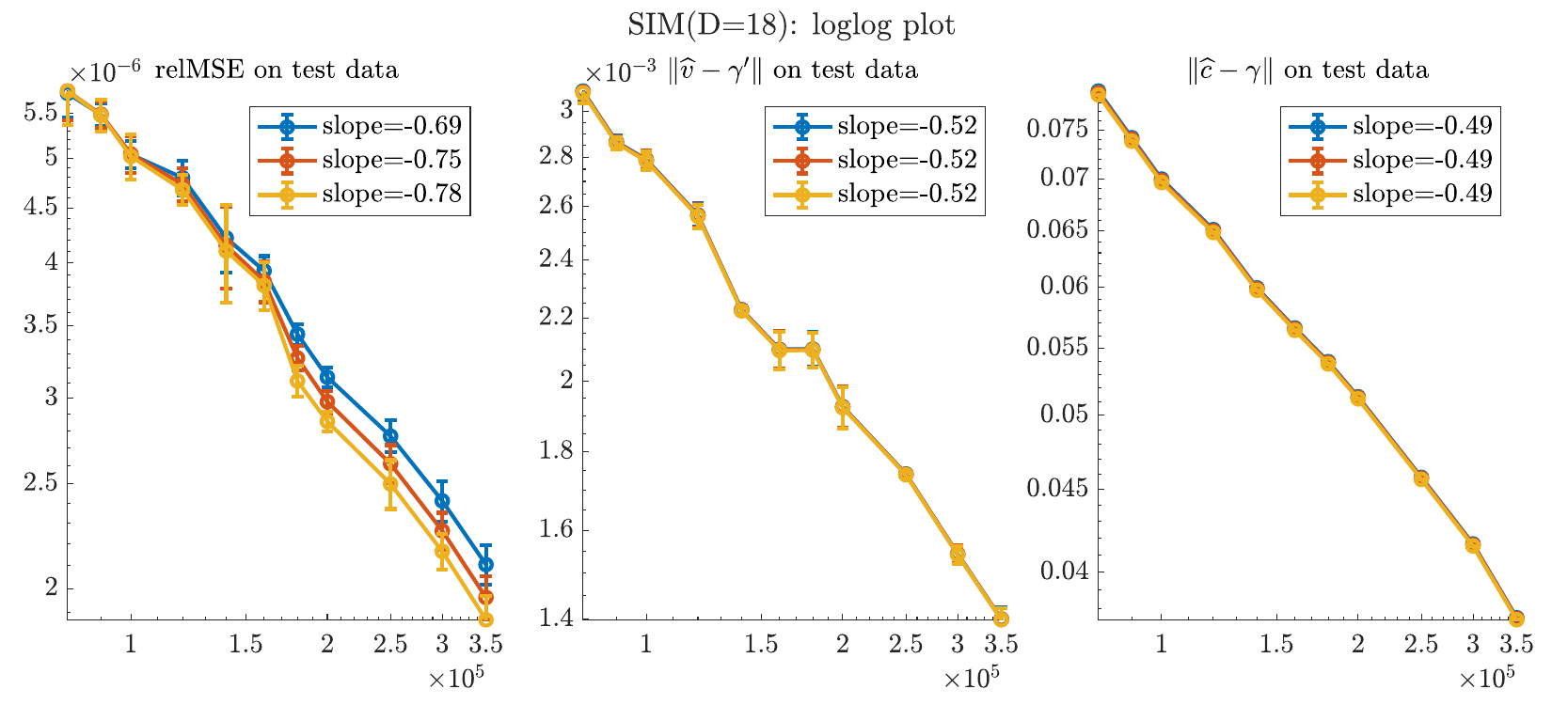}\\[6pt]
\includegraphics[viewport=262bp 2bp 521bp 335bp,clip,height=1.25\linewidth]{Learning_SIM_D18.pdf}
\end{minipage}\hfill
\begin{minipage}[t]{.32\linewidth}
\centering
{\small $D=24$}\\[2pt]
\includegraphics[viewport=0bp 2bp 260bp 335bp,clip,height=1.25\linewidth]{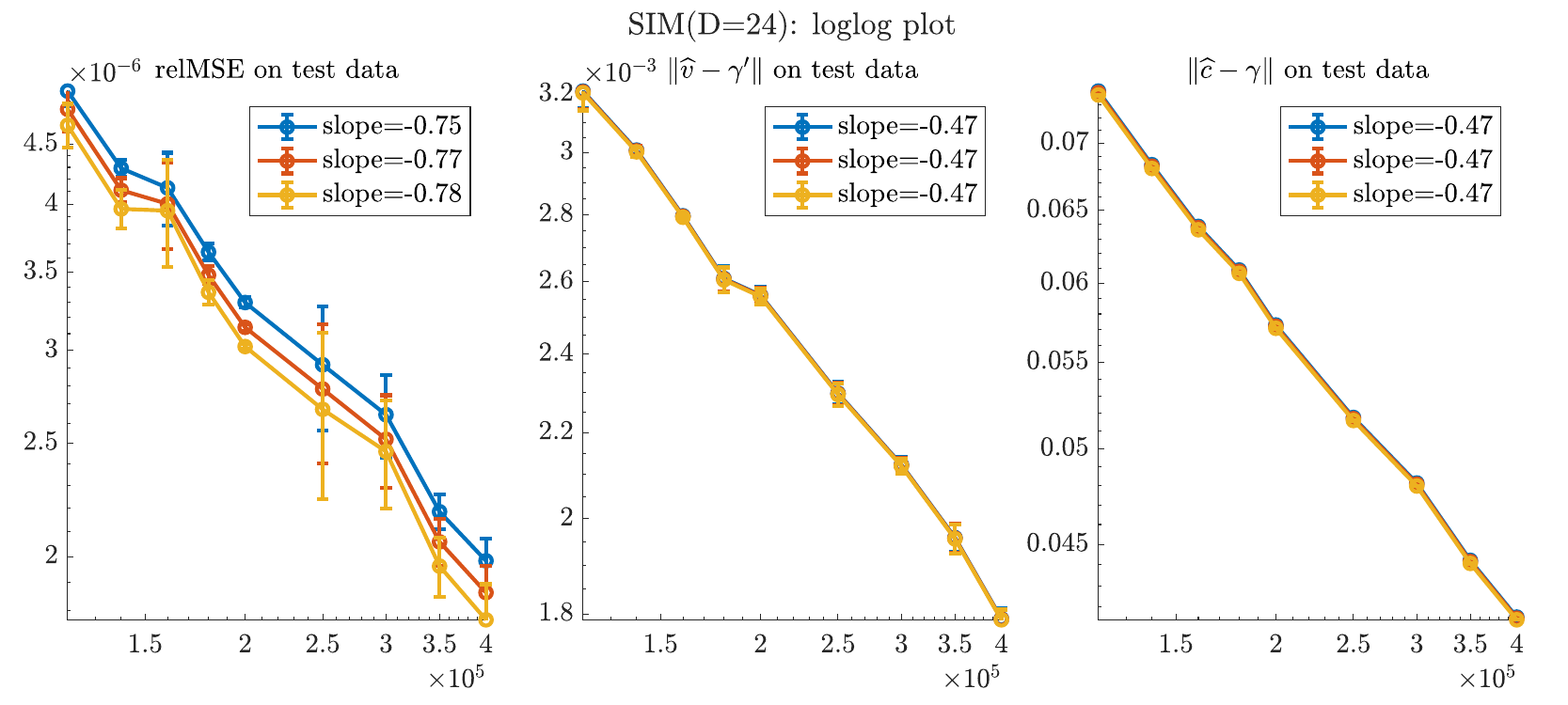}\\[6pt]
\includegraphics[viewport=262bp 2bp 520bp 335bp,clip,height=1.25\linewidth]{Learning_SIM_D24.pdf}
\end{minipage}\hfill
\begin{minipage}[t]{.32\linewidth}
\centering
{\small $D=30$}\\[2pt]
\includegraphics[viewport=0bp 2bp 260bp 335bp,clip,height=1.25\linewidth]{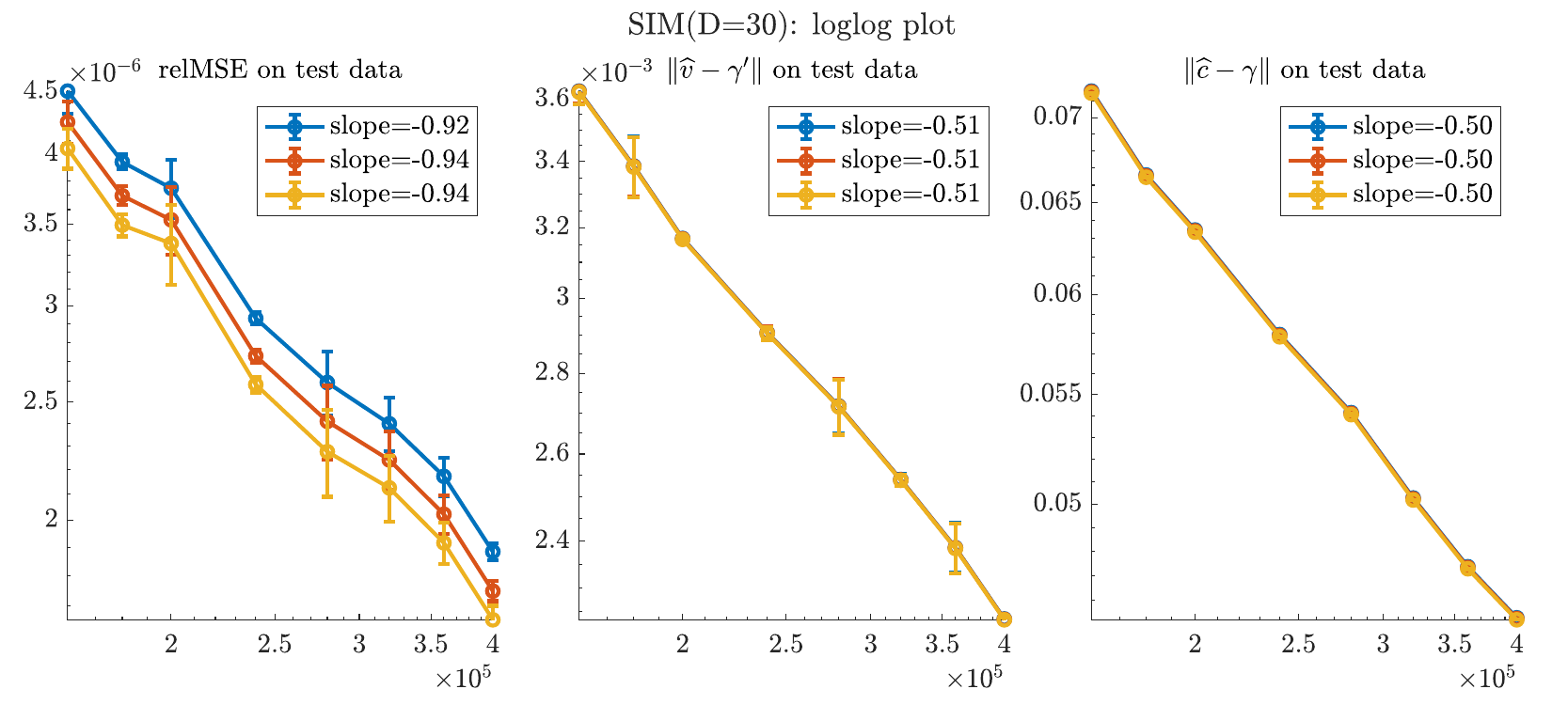}\\[6pt]
\includegraphics[viewport=262bp 2bp 519bp 335bp,clip,height=1.25\linewidth]{Learning_SIM_D30.pdf}
\end{minipage}
\caption{Line segment: relMSE (top row) and mean tangent-direction error
(devTan, bottom row) against sample size $n$ on logarithmic axes.
Columns correspond to $D=18,24,30$, from left to right. The three
curves in each panel use the three fixed $c_\rho$ values in
Appendix~\ref{app:helix}. Bars show variation across three independent
repetitions at each $n$; legends give slopes fitted over the displayed range.}
\label{fig:sim}
\end{figure}

\paragraph{Results.}
On the line segment the tangent direction is constant. Its estimated
direction has mean error close to the parametric rate $n^{-1/2}$ in all
three ambient dimensions (Figure~\ref{fig:sim}). The fitted slopes for
relMSE range from $-0.94$ to $-0.69$, around the reference slope $-0.8$.

\subsection{The Meyer helix}\label{sec:numerics-helix}

The Meyer helix was introduced in our study of coarsely monotone links
\citep{WuMaggioniNSVM}. With its reach normalized to $\sqrt D$, the curve
has diameter of order $\sqrt D$, length of order $D^{3/2}$ and effective linear dimension of order $D$, defined here as the number of singular values of the sampled coordinate matrix exceeding $5\%$ of its largest singular value (Appendix~\ref{app:helix}). Thus a
one-dimensional index moves through an increasing number of ambient
directions. We sample the normal displacement from
$\mathcal N(0,\sigma_\gamma^2 I_{D-1})$, with $\sigma_\gamma=0.5$,
truncated at $\|Z\|<0.9\sqrt D$. The link
$f(t)=\mathrm{len}_\gamma f_*(t/\mathrm{len}_\gamma)$ has the non-monotone
shape of Figure~\ref{fig:linkf}; its arc-length derivative is independent
of the curve length at a fixed relative position. We use the noise level
and test protocol of Section~\ref{sec:numerics-sim}, with response
resolution $\delta_Y=(\max_iY_i-\min_iY_i)/100$.

\begin{figure}[!htb]
\centering
\begin{minipage}[t]{.32\linewidth}
\centering
{\small $D=18$}\\[2pt]
\includegraphics[viewport=0bp 2bp 260bp 335bp,clip,height=1.25\linewidth]{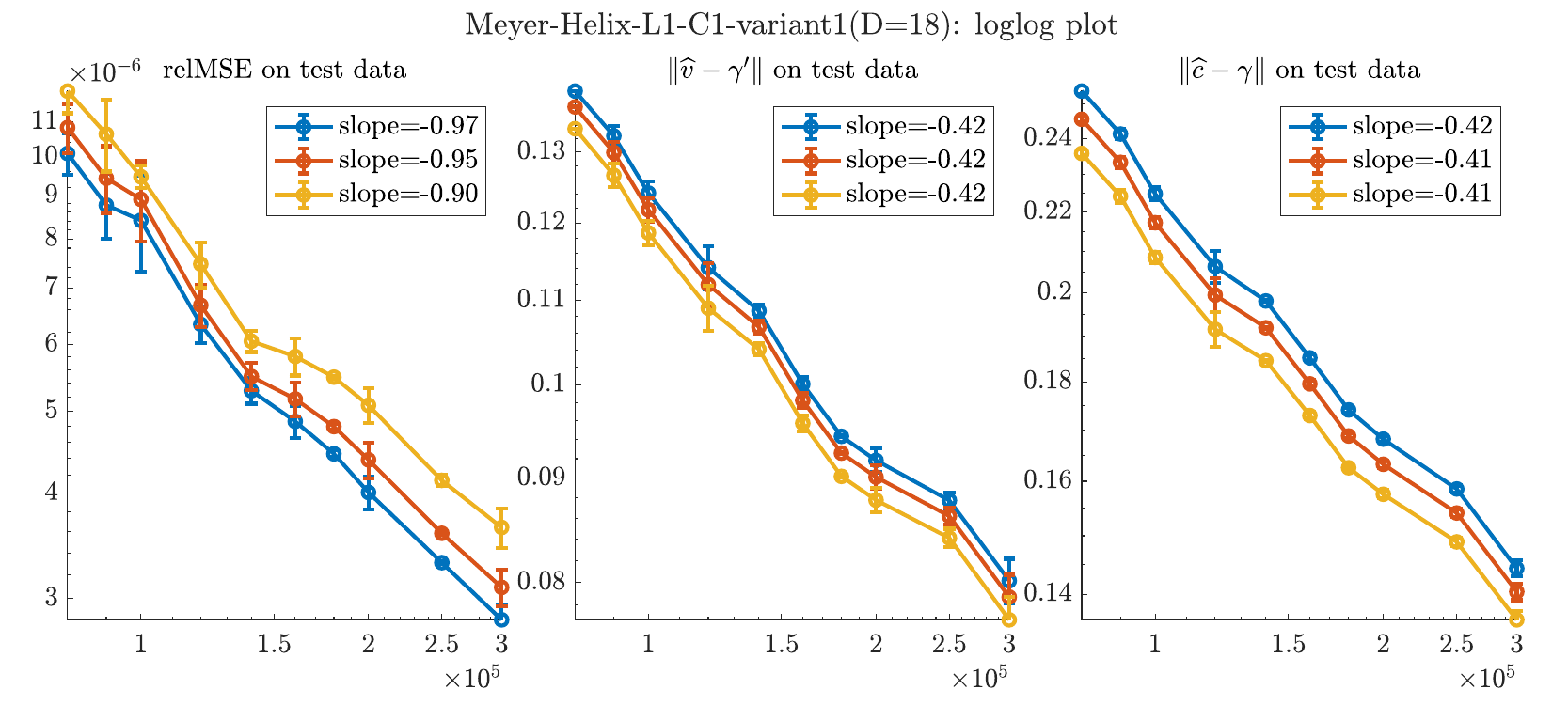}\\[6pt]
\includegraphics[viewport=262bp 2bp 521bp 335bp,clip,height=1.25\linewidth]{Learning_Curve_D18.pdf}
\end{minipage}\hfill
\begin{minipage}[t]{.32\linewidth}
\centering
{\small $D=24$}\\[2pt]
\includegraphics[viewport=0bp 2bp 260bp 335bp,clip,height=1.25\linewidth]{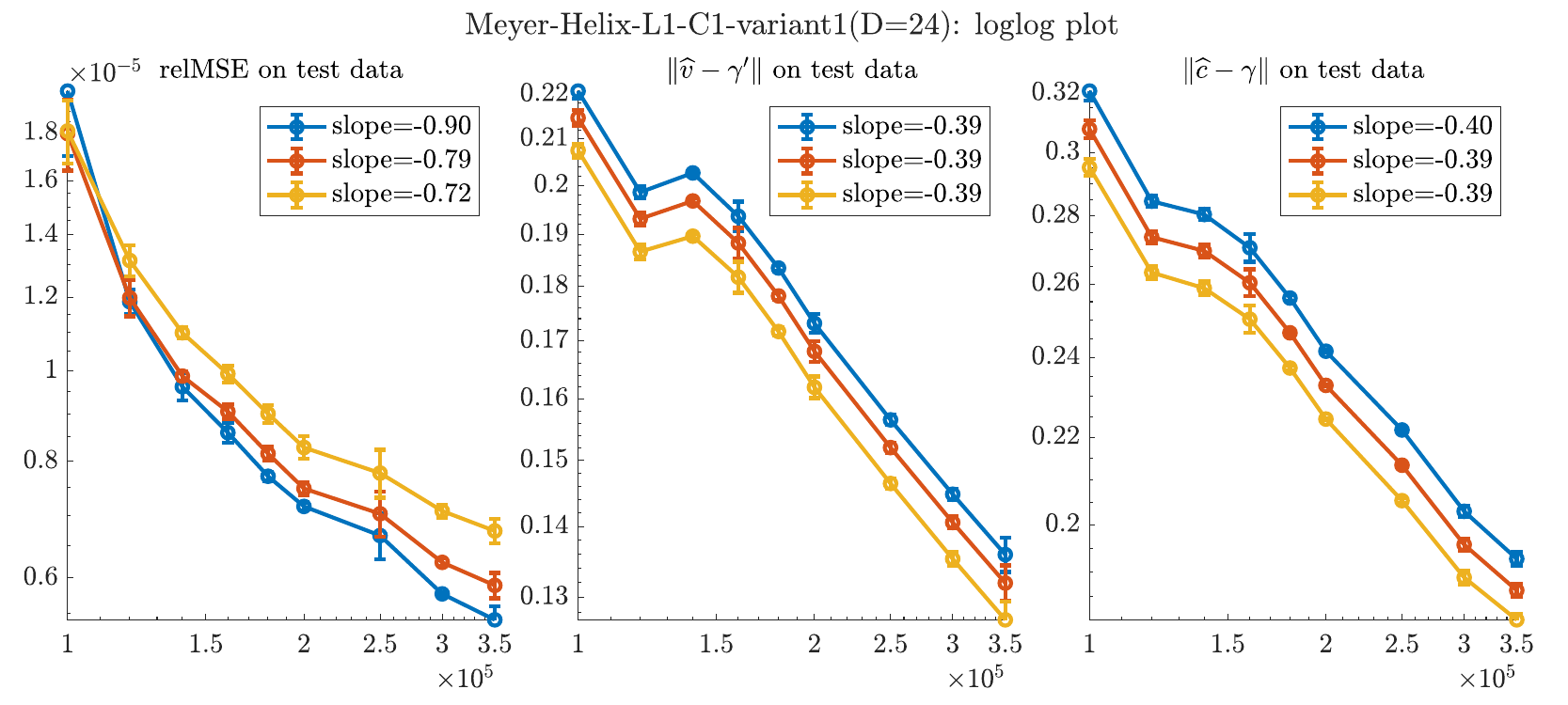}\\[6pt]
\includegraphics[viewport=262bp 2bp 524bp 335bp,clip,height=1.25\linewidth]{Learning_Curve_D24.pdf}
\end{minipage}\hfill
\begin{minipage}[t]{.32\linewidth}
\centering
{\small $D=30$}\\[2pt]
\includegraphics[viewport=0bp 2bp 260bp 335bp,clip,height=1.25\linewidth]{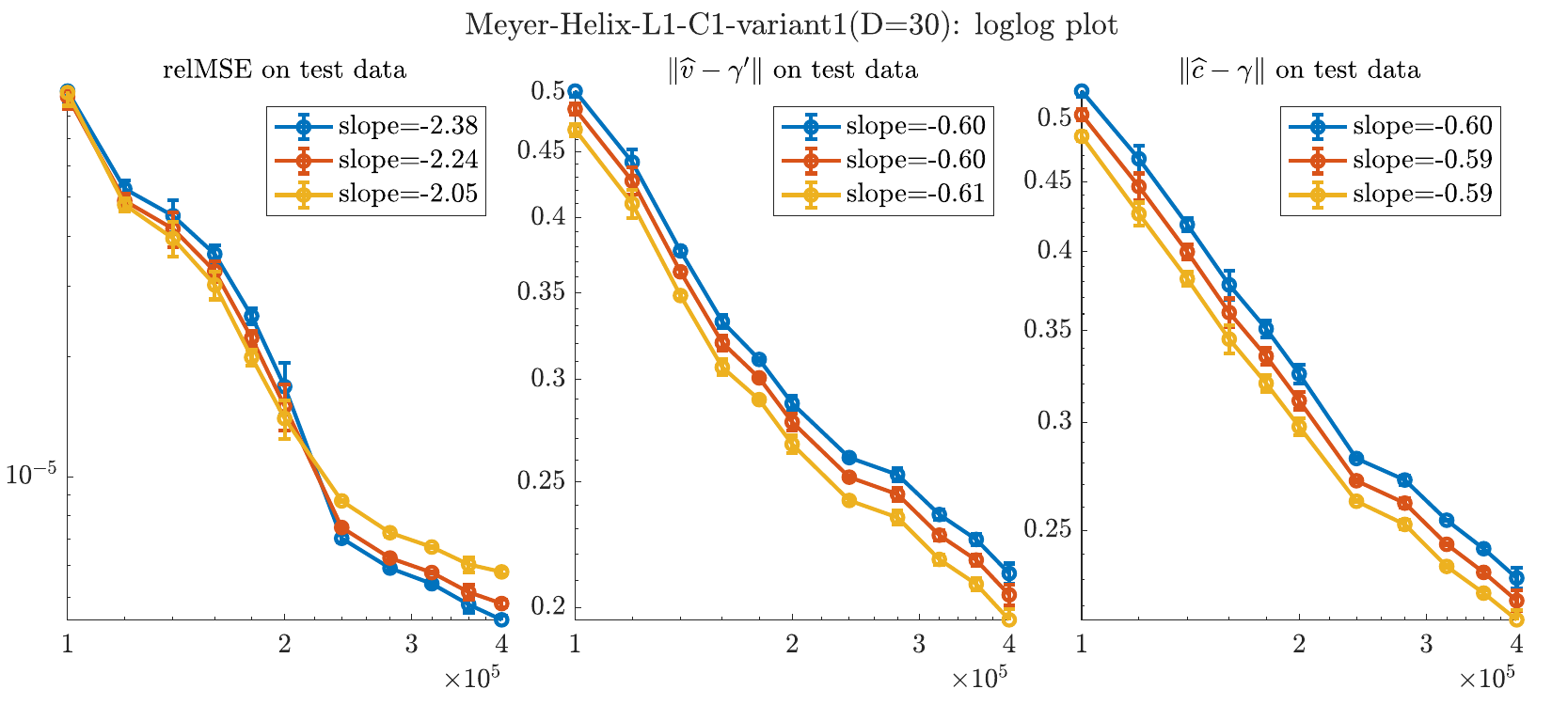}\\[6pt]
\includegraphics[viewport=262bp 2bp 521bp 335bp,clip,height=1.25\linewidth]{Learning_Curve_D30.pdf}
\end{minipage}
\caption{Meyer helix: relMSE (top row) and mean tangent-direction error
(devTan, bottom row) against sample size $n$ on logarithmic axes.
Columns correspond to $D=18,24,30$, from left to right. The three
curves in each panel use the three fixed $c_\rho$ values in
Appendix~\ref{app:helix}. Bars show variation across three independent
repetitions at each $n$; legends give slopes fitted over the displayed range.}
\label{fig:helix}
\end{figure}

\paragraph{Results.}
At $D=18$ and $24$, the fitted prediction slopes lie between $-0.97$ and
$-0.72$, and the tangent-error slopes are close to $-0.4$
(Figure~\ref{fig:helix}). At $D=30$, relMSE falls steeply up to
$n\simeq2.4\times10^5$ and more slowly thereafter; fitting across both
segments gives slopes from $-2.38$ to $-2.05$. Mean tangent-direction error
declines steadily over the same range, with slopes near $-0.60$ for all
three choices of $c_\rho$.

\subsection{The bumpy sphere}\label{sec:numerics-sphere}

The bumpy sphere is a closed two-dimensional surface in $\R^{20}$ with two
$\R^6$-valued link functions differing in phase complexity and the number
of response preimage packets. For both, prediction and tangent-space
errors decrease over the sampled range despite normal displacements
beyond the global tube of \Aref{1} (Figure~\ref{fig:bumpy-iid-rates} and
Table~\ref{tab:bumpy-iid-slopes}).

\paragraph{Manifold and predictors.}
The surface in Figure~\ref{fig:bumpy}(a) is a radial deformation of the
unit sphere. Let $r$ be the smooth radial function of
Appendix~\ref{app:bumpy} and choose the scale $c_\M$ to give the surface
numerically estimated reach $\sqrt{20}$, as for the Meyer helix. Then
\begin{equation}\label{eq:bumpy}
  \M=\bigl\{q(u):\ u\in\mathbb S^2\bigr\},\qquad
  q(u)=c_\M\,r(u)\,(u_1,u_2,u_3,0,\dots,0)\in\R^{20},
\end{equation}
The surface lies in the
first three coordinates, while the estimator recovers its tangent field from
predictors spread over all \(18\) normal directions.  We sample \(Q=q(u)\)
with \(u\) uniform on \(\mathbb S^2\) and use centered normal displacements
scaled by the local reach.  The sampling law and reach calibration are given
in Appendix~\ref{app:bumpy}.

\paragraph{Link functions and noise.}
Write $\xi(u)=O\,r(u)\,u\in\R^3$, where $O$ is a fixed rotation of $\R^3$
(Appendix~\ref{app:bumpy}); thus $\xi(u)$ is the first three coordinates of $q(u)$, rotated by $O$ and rescaled by
$c_\M^{-1}$. For a phase map
$\psi:\R^3\to\R^3$, the link function has the six components
\begin{equation}\label{eq:bumpy-links}
  f_{2i-1}\bigl(q(u)\bigr)=\frac{1+\cos\psi_i(\xi(u))}2,\qquad
  f_{2i}\bigl(q(u)\bigr)=\frac{1+\sin\psi_i(\xi(u))}2,\qquad i=1,2,3.
\end{equation}
We consider two trigonometric link functions: the \emph{linear-phase link function}
$f^{\rm lin}$, with $\psi_i(\xi)=8.75\,\xi_i$, and the
\emph{quadratic-phase link function} $f^{\rm quad}$, with
$\psi_i(\xi)=15\,\xi_i+2\,\xi_j\xi_l$, where
$\{j,l\}=\{1,2,3\}\setminus\{i\}$ (Figure~\ref{fig:bumpy}(b,c)).
Both link functions satisfy \Aref{3},
as verified in Appendix~\ref{app:bumpy}. The phases make several turns over the surface, so
neither link function is injective, and a response preimage can consist of several
separated packets; since the phase of $f^{\rm quad}$ varies faster, its
packets are more numerous. Training and validation responses carry
$1\%$ relative Gaussian noise with the covariance specified in
Appendix~\ref{app:bumpy}. Test errors use the noiseless responses.

\paragraph{Protocol and fixed parameters.}
The constants in Table~\ref{tab:bumpy-iid-params} are selected on validation
data at $n\in\{1,5,9\}\times10^j$, $j=3,4,5$, using the criterion in
Appendix~\ref{app:bumpy}, then fixed across sample sizes and repetitions.
For each link function and each sample size
\[
 n\in\{1,2,3,4,5,7\}\times10^j,\qquad j=4,5,6,
\]
we perform ten independent repetitions, each with fresh training,
validation and test data. Data sets are also independent across sample
sizes. Within a repetition, the two link functions share the predictors and
standardized Gaussian noise draws. Each estimator is evaluated on $5000$
independent test queries.

The relative mean squared prediction error is normalized by the total
empirical variance of the noiseless test responses. The tangent-space error
is \eqref{eq:tangent-error-num}, evaluated after orthonormalizing the
estimated basis. In Figure~\ref{fig:bumpy-iid-rates}, markers give the mean
of the ten repetition-level errors and bars give one sample standard
deviation.

\begin{figure}[p]
\centering
\includegraphics[trim=95bp 48bp 84bp 37bp,clip,width=.34\linewidth]{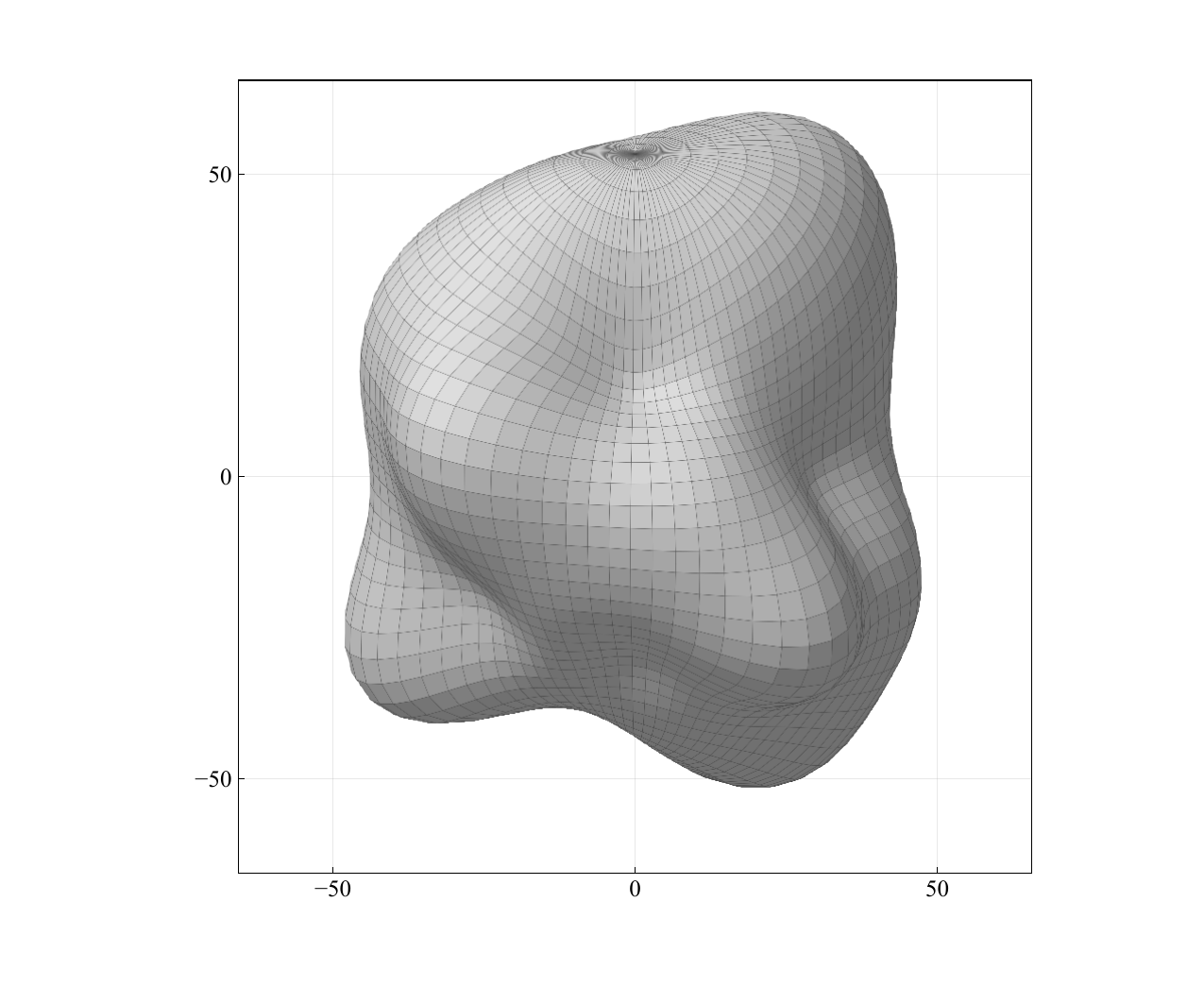}\\[2pt]
{\small (a) The surface $\M$}\\[6pt]
\includegraphics[trim=14bp 14bp 30bp 10bp,clip,width=.94\linewidth]{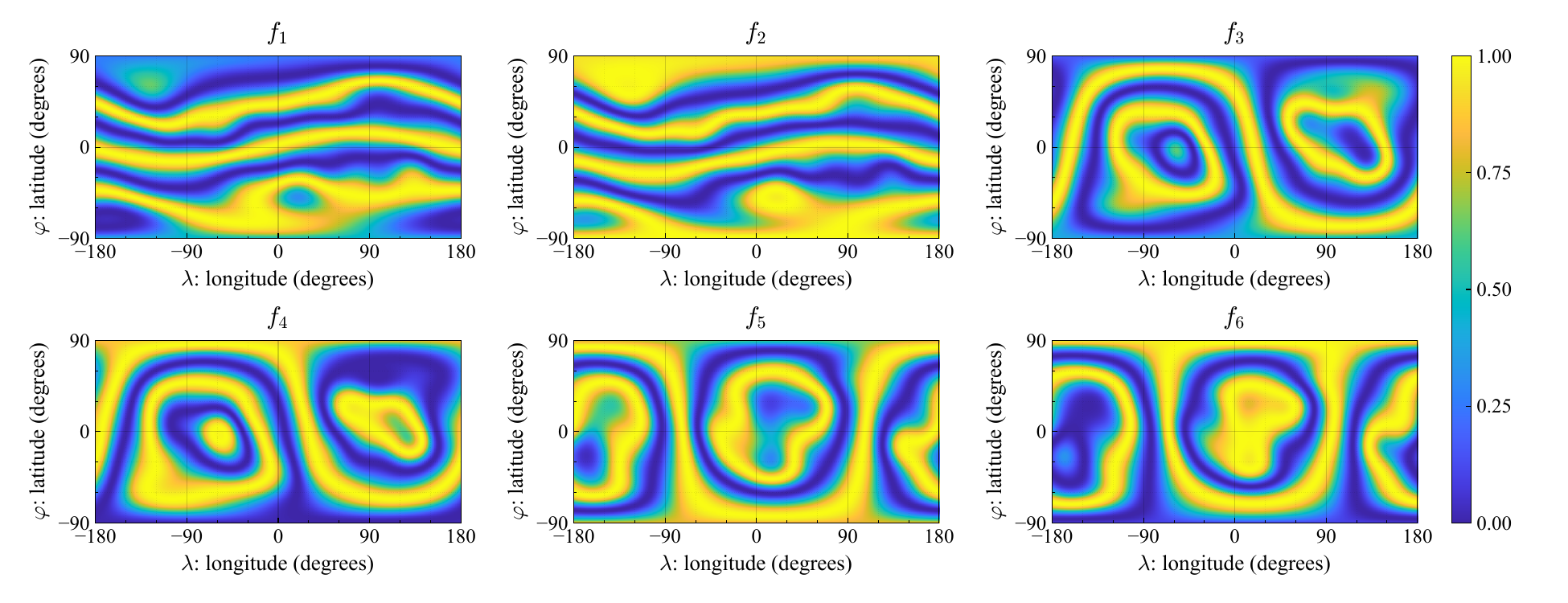}\\[2pt]
{\small (b) The components of $f^{\rm lin}$}\\[6pt]
\includegraphics[trim=14bp 14bp 30bp 10bp,clip,width=.94\linewidth]{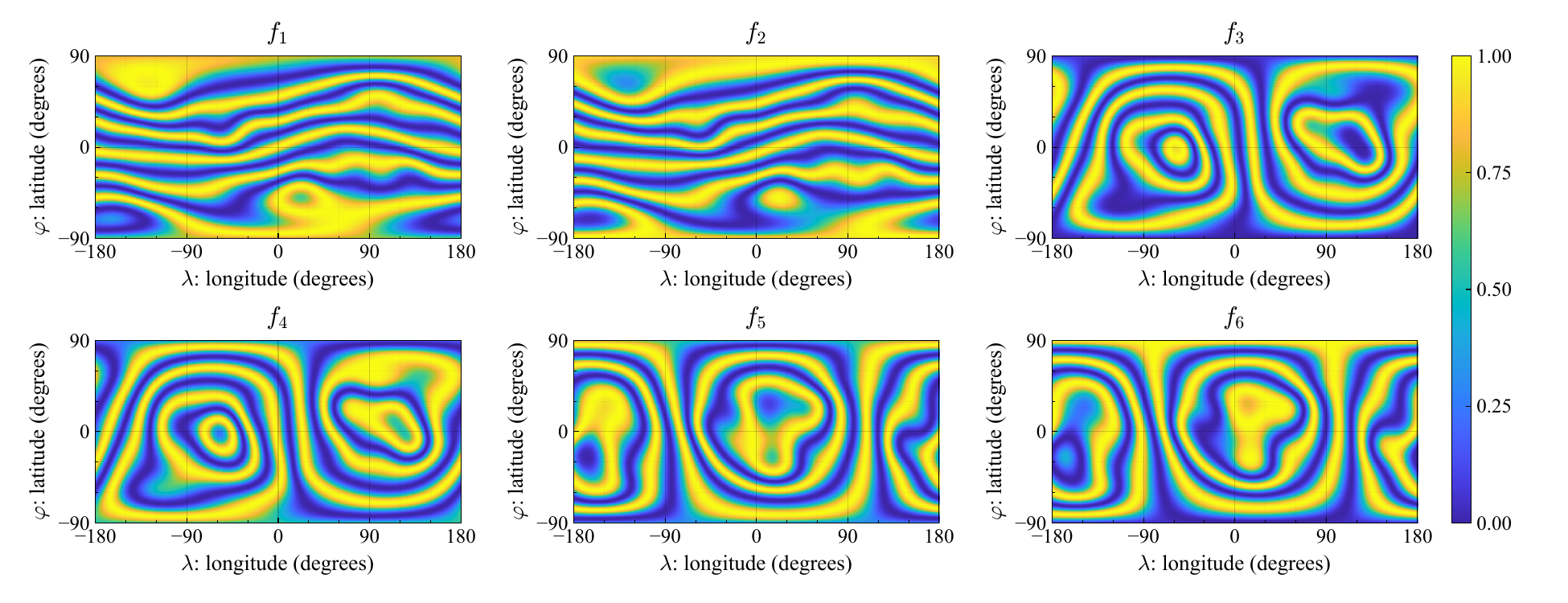}\\[2pt]
{\small (c) The components of $f^{\rm quad}$}
\caption{The bumpy sphere (Section~\ref{sec:numerics-sphere}). (a)
Orthographic view of $\M$ in its first three coordinates; the other $17$
vanish. (b, c) Components $f_1,\dots,f_6$ of the two link functions versus
longitude $\lambda$ and latitude $\varphi$ of $u\in\mathbb S^2$.}
\label{fig:bumpy}
\end{figure}


\begin{table}[!htb]
\centering
\small
\begin{tabular}{@{}llccc@{}}
\toprule
\shortstack{link\\function} & error & \shortstack{$n\ge10^6$\\(six points)} & \shortstack{$n\ge2\times10^6$\\(five points)} & reference\\
\midrule
$f^{\rm lin}$ & relative mean squared error & $-0.815$ & $-0.783$ & $-2/3$\\
 & tangent-space error & $-0.252$ & $-0.239$ & $-1/3$\\
\midrule
$f^{\rm quad}$ & relative mean squared error & $-0.844$ & $-0.815$ & $-2/3$\\
 & tangent-space error & $-0.350$ & $-0.317$ & $-1/3$\\
\bottomrule
\end{tabular}
\caption{Least-squares slopes of the logarithm of the arithmetic mean error
against $\log n$, calculated from the ten independent repetitions at every sample
size. The reference slopes are those for squared prediction risk and mean
tangent-space error in Theorem~\ref{thm:isolated-rank} with $m=2$.}
\label{tab:bumpy-iid-slopes}
\end{table}

\begin{figure}[!htb]
\centering
\begin{minipage}[t]{0.49\linewidth}
\centering
{\small (a) $f^{\rm lin}$}\\[2pt]
\includegraphics[width=\linewidth]{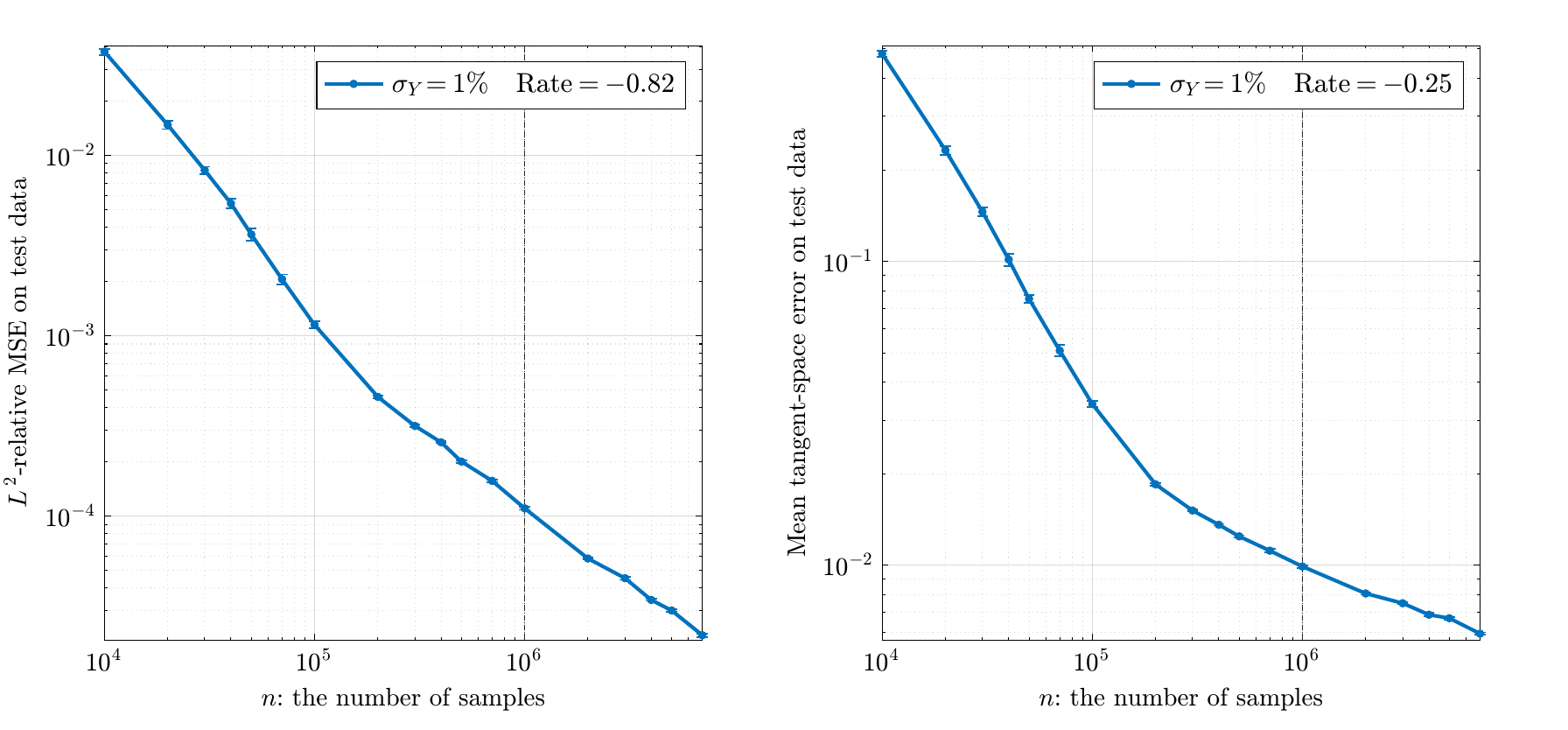}
\end{minipage}\hfill
\begin{minipage}[t]{0.49\linewidth}
\centering
{\small (b) $f^{\rm quad}$}\\[2pt]
\includegraphics[width=\linewidth]{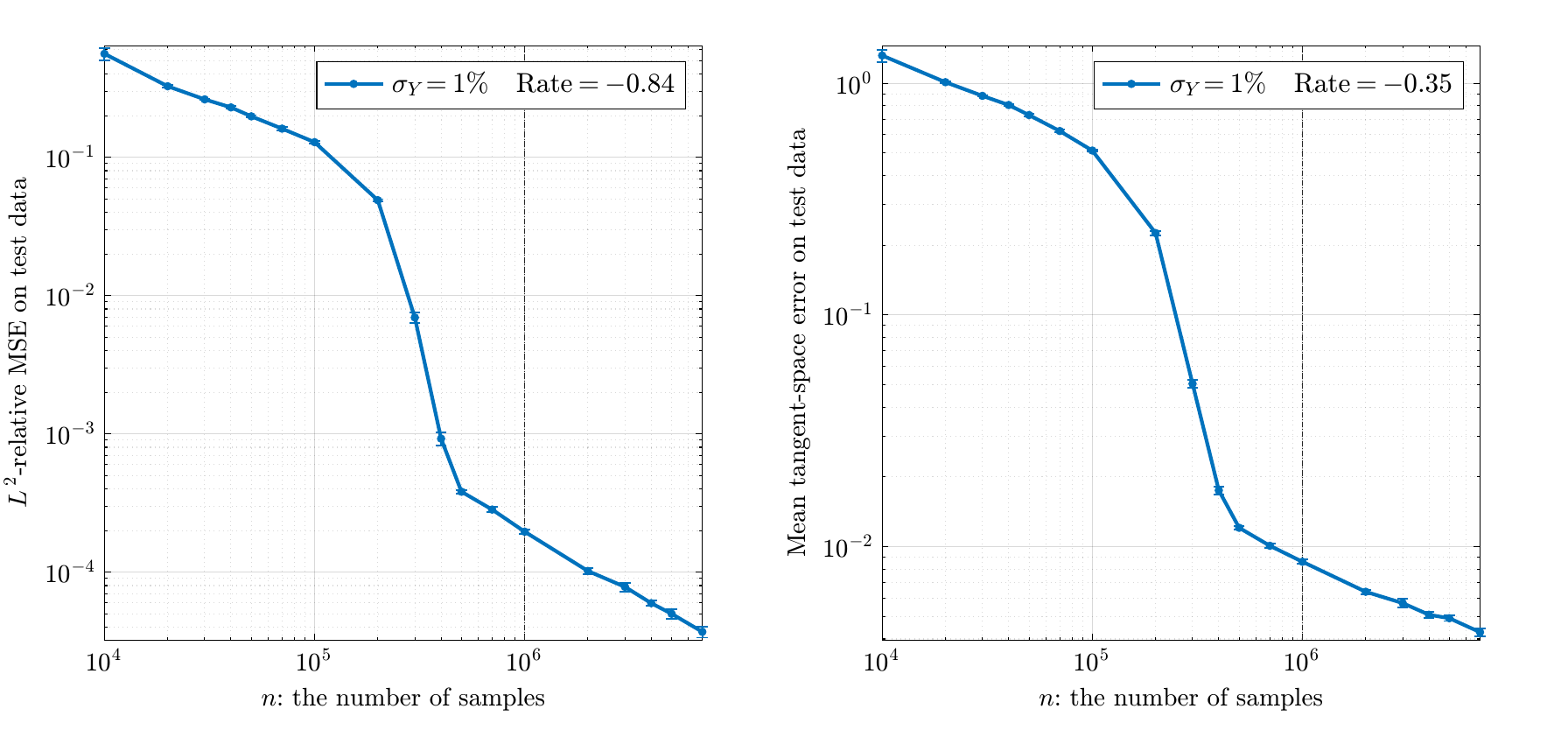}
\end{minipage}
\caption{Bumpy sphere in $D=20$ with the linear-phase link function (a) and
quadratic-phase link function (b), with $1\%$ response noise. Left: relative mean
squared prediction error on noiseless test responses. Right: mean tangent-space
error $(\sin^2\theta_1+\sin^2\theta_2)^{1/2}$ on test queries. Points show means of ten independent repetitions; bars show one sample
standard deviation. The constants of Table~\ref{tab:bumpy-iid-params}
are fixed across all $n$ and repetitions. The dash-dotted line marks $n=10^6$;
the legends give log--log least-squares slopes of the mean errors over
$n\ge10^6$.}
\label{fig:bumpy-iid-rates}
\end{figure}

\Needspace{8\baselineskip}
\paragraph{Results.}
As shown in Figure~\ref{fig:bumpy-iid-rates}, both errors decrease
with the sample size for both link functions. Fitting the mean errors over
$n\ge10^6$ gives prediction-error slopes $-0.815$ for $f^{\rm lin}$ and
$-0.844$ for $f^{\rm quad}$, and tangent-error slopes $-0.252$ and $-0.350$,
respectively (Table~\ref{tab:bumpy-iid-slopes}). All four fits have
coefficients of determination above $0.99$. The tangent-space error slope
for $f^{\rm quad}$ is close to
the reference exponent $-1/3$, whereas the tangent-space error for $f^{\rm lin}$ decreases more
slowly over this range. Both prediction errors decrease faster than the
reference slope $-2/3$ over the fitted range.

\Needspace{5\baselineskip}
At $n=7\times10^6$, the prediction errors are
$(2.19\pm0.05)\times10^{-5}$ for $f^{\rm lin}$ and
$(3.71\pm0.35)\times10^{-5}$ for $f^{\rm quad}$.
The corresponding tangent-space errors are
$(5.96\pm0.06)\times10^{-3}$ and $(4.28\pm0.16)\times10^{-3}$,
where each $\pm$ gives one sample standard deviation across the ten repetitions.
%

\begin{figure}[!t]
\centering
\includegraphics[width=0.75\linewidth]{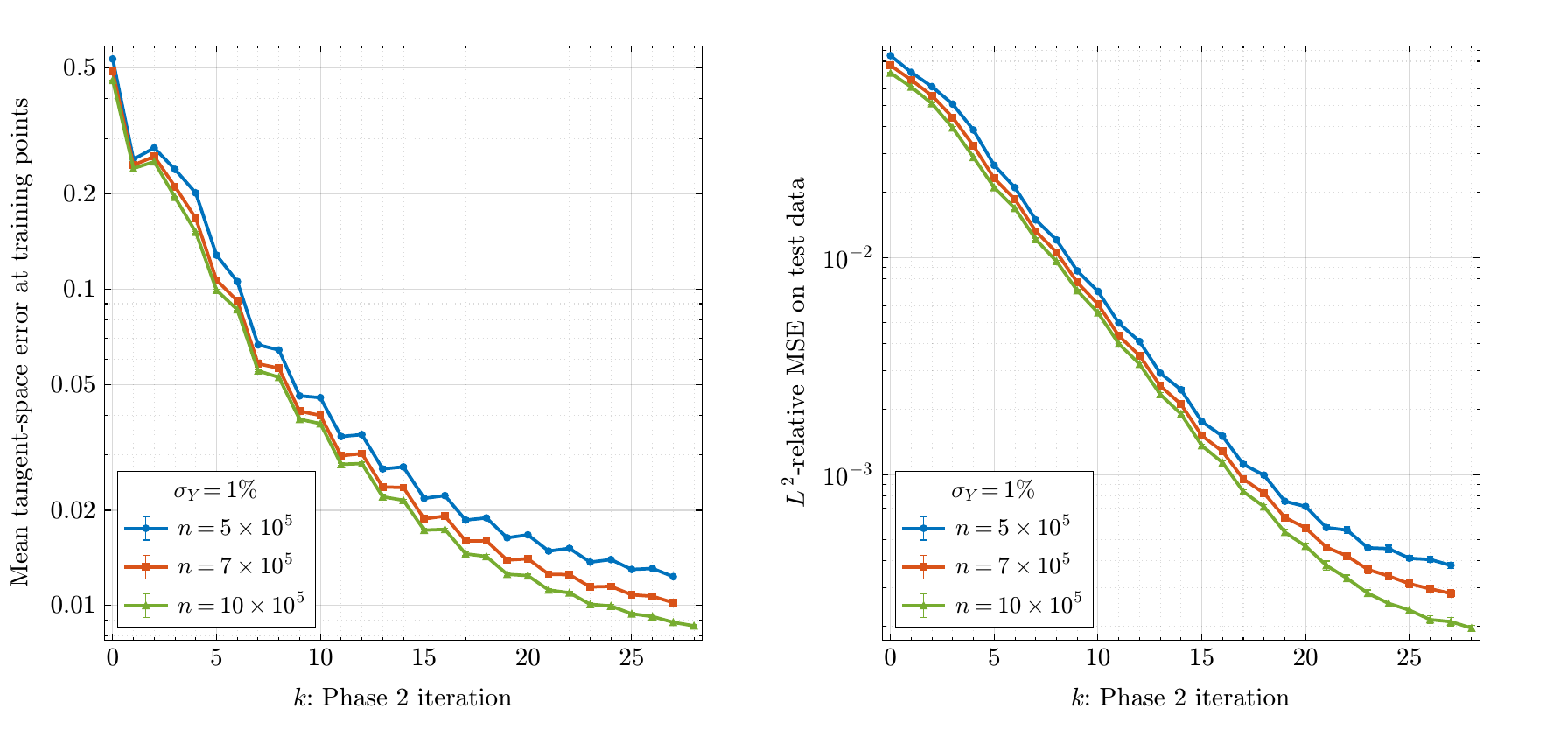}
\caption{Iterative refinement for the quadratic-phase bumpy-sphere experiment.
Left: mean tangent-space error on the training points. Right: relative mean
squared prediction error on $5000$ independent test queries with noiseless
responses. Iteration $k=0$ is the Phase~1 initialization. Curves show means
over ten independent repetitions using the parameters of
Table~\ref{tab:bumpy-iid-params}; bars give one sample standard deviation.}
\label{fig:phase2-refinement-large-n}
\par\vspace{18pt}
\centering
\begin{minipage}{0.8\linewidth}
\centering
\small
\resizebox{\linewidth}{!}{%
\begin{tabular}{@{}rccccc@{}}
\toprule
$n$ & $k_*$ & tangent, initial & tangent, final & prediction, initial & \shortstack{prediction, final\\(units of $10^{-4}$)}\\
\midrule
$5\times10^5$ & 27 & $0.5353\pm0.0008$ & $0.01233\pm0.00010$ & $0.08503\pm0.00087$ & $3.818\pm0.118$\\
$7\times10^5$ & 27 & $0.4885\pm0.0011$ & $0.01020\pm0.00009$ & $0.07659\pm0.00064$ & $2.836\pm0.118$\\
$10^6$ & 28 & $0.4583\pm0.0007$ & $0.008617\pm0.000065$ & $0.07070\pm0.00047$ & $1.961\pm0.060$\\
\bottomrule
\end{tabular}}
\makeatletter\def\@captype{table}\makeatother
\caption{Initial and final iterative-refinement errors: means $\pm$ one
sample standard deviation across ten independent repetitions. Prediction
uses noiseless test responses.}
\label{tab:phase2-large-n}
\end{minipage}
\end{figure}

\Needspace{8\baselineskip}
\paragraph{Performance of iterative refinement.}
Figure~\ref{fig:phase2-refinement-large-n} follows ten independent
quadratic-phase runs at $n=5\times10^5$, $7\times10^5$ and $10^6$ with the
fixed parameters in Table~\ref{tab:bumpy-iid-params}.
Iteration $k=0$ is the Phase~1 initialization. Each trajectory uses the
same training points for tangent-space error and the same $5000$
independent test queries with noiseless responses for prediction error.
The prediction bandwidth follows the prescribed schedule.

From initialization to the final iteration $k_*$, Phase~2 reduces the mean
tangent-space error by factors of $43$--$53$ and the mean relative prediction
error by factors of $223$--$360$ (Table~\ref{tab:phase2-large-n}).
Holding observations, centers and final prediction bandwidth fixed,
replacing the final estimated tangent spaces by the true ones reduces
prediction error by $12$--$14\%$. At the same bandwidth, the initial
tangent-space estimates give errors more than two orders of magnitude
larger than the final estimates.

\subsection{A coupled Meyer helix surface}\label{sec:numerics-coupled-helix}

\begin{figure}[!htbp]
\centering
\includegraphics[width=\linewidth]{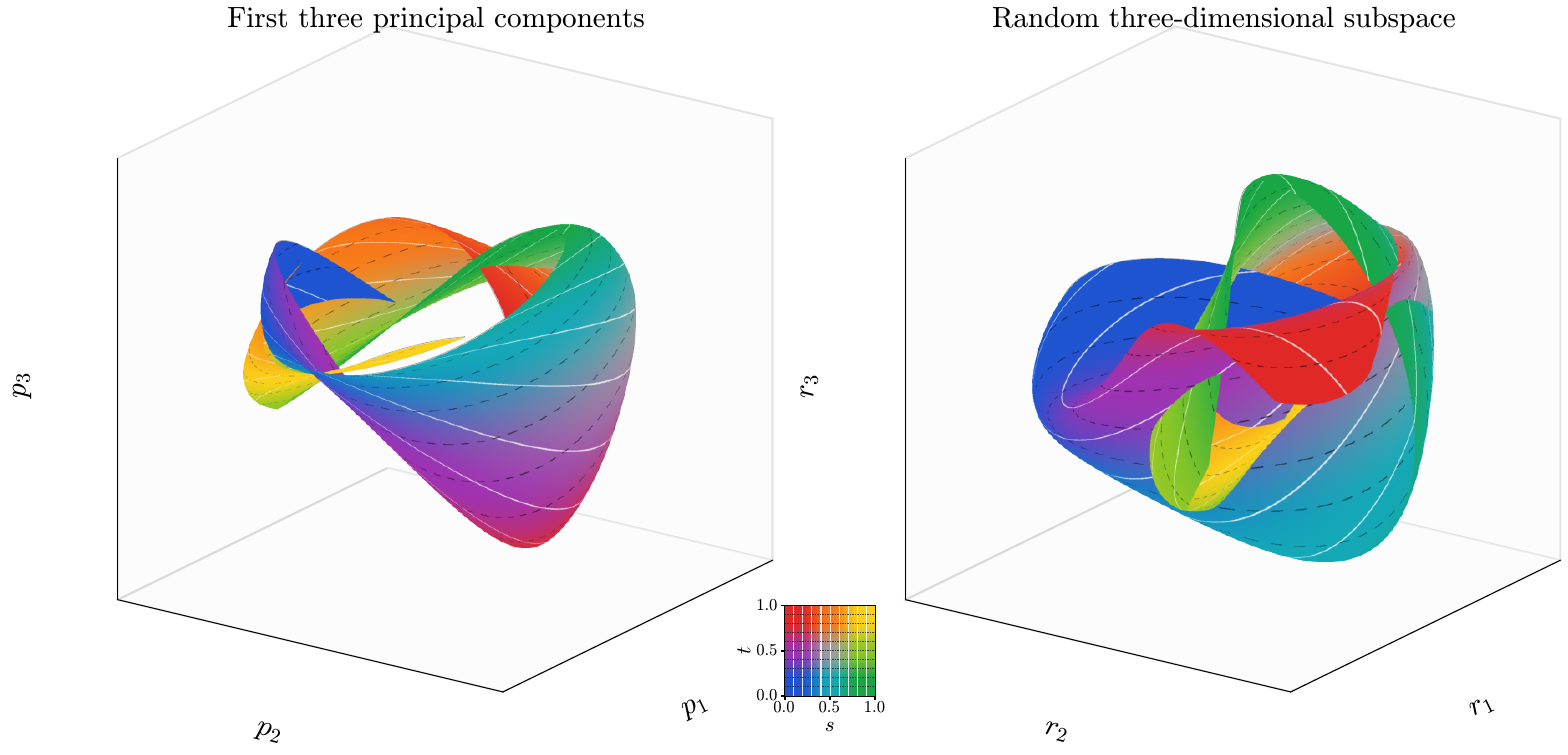}
\caption{The coupled Meyer helix surface in $\R^{20}$, projected onto the first
three principal components of noisy predictors (left) and a random
three-dimensional subspace (right). Only the surface is shown. The central
key encodes $(s,t)$; coordinate curves are $s=j/10$ (white solid) and $t=j/10$ (dark dashed),
$j=1,\ldots,9$.}
\label{fig:coupled-geometry}
\end{figure}

Let $E_i^{(d)}$ and $\phi_i^{(d)}$ be the localized envelopes and
oscillatory phases, and $\mu_\M\in\R^{20}$ and $c_\M>0$ the centering
vector and scale, specified in Appendix~\ref{app:coupled-helix}.
Coupling two Meyer helix coordinate families
(Section~\ref{sec:numerics-helix}) gives the two-dimensional surface
$\M=\{q(s,t):(s,t)\in[0,1]^2\}$ in $\R^{20}$, where
\begin{equation}\label{eq:coupled-surface}
\begin{aligned}
 q(s,t)&=c_\M\bigl(G(s,t)-\mu_\M\bigr),\\
 G_{ij}(s,t)&=E_i^{(5)}(s)E_j^{(4)}(t)
 \cos\bigl(\phi_i^{(5)}(s)+\phi_j^{(4)}(t)\bigr),
 \qquad i=1,\ldots,5,\quad j=1,\ldots,4.
\end{aligned}
\end{equation}
Both parameters enter all twenty
coordinates. Figure~\ref{fig:coupled-geometry} shows two surface projections;
the common color key and coordinate curves identify corresponding regions.

\begin{figure}[!htbp]
\centering
\includegraphics[width=\linewidth]{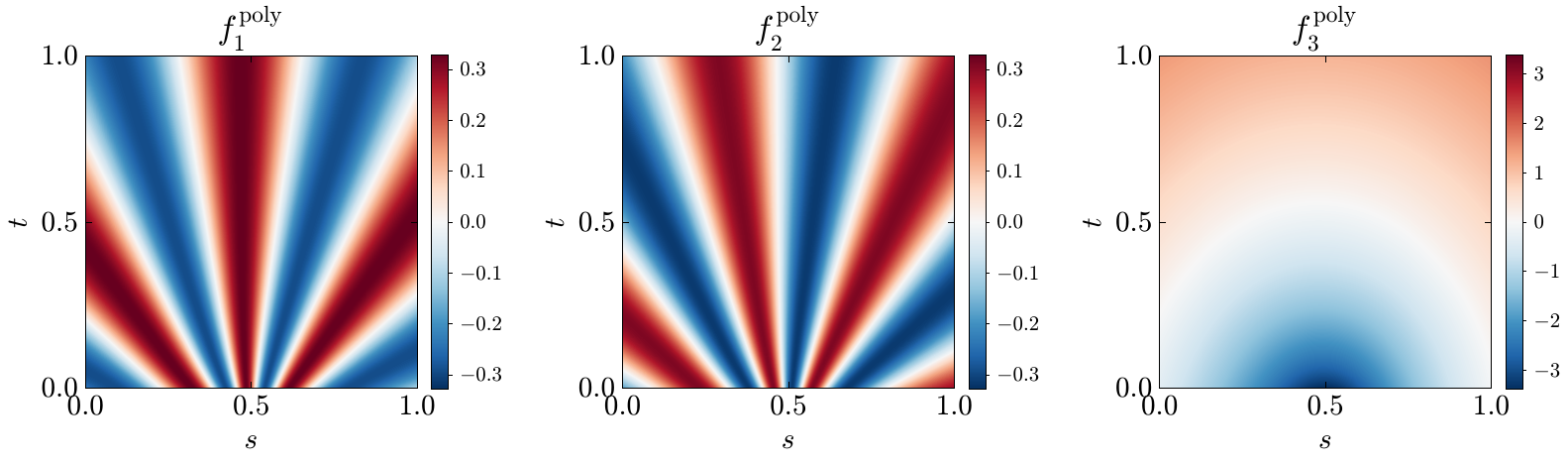}
\caption{Components of $f^{\rm poly}$ on the parameter square. All panels
use the same diverging color map; the first two share the same color scale.}
\label{fig:coupled-response}
\end{figure}

\paragraph{Predictors and link functions.}
We sample $(s,t)$ uniformly on the parameter square and set
$X=Q+Z$, with $Q=q(s,t)$ and $Z$ normal to the surface at $Q$.
As for the bumpy sphere, displacement scales with the numerically estimated
local reach; Appendix~\ref{app:coupled-helix} gives the sampling law. The
first of the two three-component links uses a complex polynomial.
Choose fixed $a_Y>0$ and $\mu_Y\in\R^3$ to center the response and
give it unit total variance on the calibration sample
(Appendix~\ref{app:coupled-helix}):
\begin{equation}\label{eq:coupled-poly-link}
\begin{gathered}
 z=-1+2s+\mathrm i(0.25+1.5t),\qquad
 \mathfrak p(z)=z^7(z+0.6+0.2\mathrm i),\\[3pt]
 f^{\rm poly}(q(s,t))
 =a_Y\left[
 \left(\Re\frac{\mathfrak p(z)}{|\mathfrak p(z)|},\,
       \Im\frac{\mathfrak p(z)}{|\mathfrak p(z)|},\,\log|\mathfrak p(z)|\right)^{\!\T}
       -\mu_Y\right].
\end{gathered}
\end{equation}
The differential has rank two,
while a response can have several preimages. Figure~\ref{fig:coupled-response}
shows the components of the link function: the first two encode the phase of the polynomial,
and the third its log magnitude.

For the same surface and predictor law, define $f^{\rm rat}$ from the rational map
\begin{equation}\label{eq:coupled-rational-map}
 z=-1+2s+\mathrm i(0.3+1.5t),\qquad
 w(z)=\left(\frac{z+1.3-0.1\mathrm i}
                 {z-1.3-1.8\mathrm i}\right)^{10}.
\end{equation}
It uses the phase and log-magnitude features in
\eqref{eq:coupled-poly-link}, with $w$ in place of $\mathfrak p$, followed
by centering and normalization to unit total variance. This link function also has
a rank-two differential and can have several parameter preimages.
Its components are shown in Figure~\ref{fig:coupled-rational-response} in
Appendix~\ref{app:coupled-helix}. Both link functions use $1\%$ relative Gaussian noise in training and validation.

\paragraph{Protocol and results.}
The response-neighbor count and refinement schedule are
\eqref{eq:bumpy-counts}--\eqref{eq:bumpy-schedule}, with parameters selected
separately for each link function on validation data and held fixed across sample
sizes and repetitions (Appendix~\ref{app:coupled-helix}). At each
$n\in\{1,2,3,4,5,7\}\times\{10^4,10^5,10^6\}$ we perform ten independent
repetitions per link, with fresh training data and independent validation
and test sets of $5000$ queries each. Prediction errors use noiseless test
responses. Tangent-space errors compare the estimated planes with the
analytic tangent planes at the sampled base points, using
\eqref{eq:tangent-error-num}.

\begin{figure}[!ht]
\centering
\includegraphics[width=.75\linewidth]{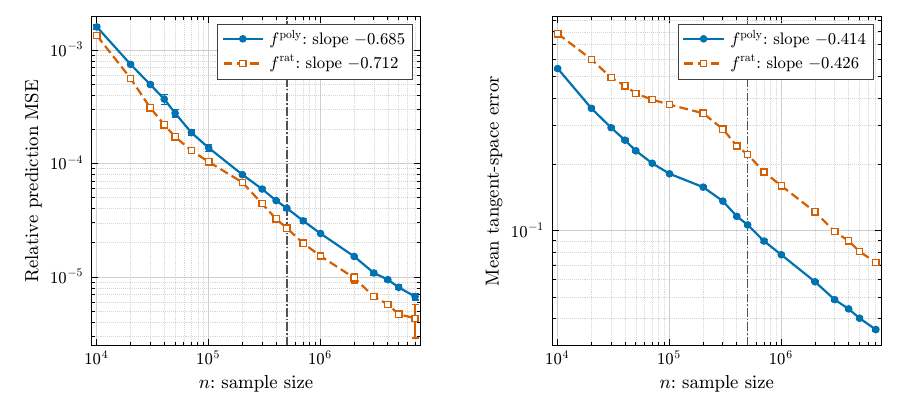}
\caption{Prediction and tangent-space errors for the coupled Meyer helix
surface with the link functions $f^{\rm poly}$ and $f^{\rm rat}$ and $1\%$ response
noise. Each marker is the mean of ten independent repetitions for that
link function; error bars show one sample standard deviation. The vertical line
marks $n=5\times10^5$, the lower end of the fitted range. The legends give
least-squares slopes of the logarithms of the mean errors. Each link function uses the parameters of
Table~\ref{tab:coupled-parameters} for all $n$.}
\label{fig:coupled-scaling}
\end{figure}

Figure~\ref{fig:coupled-scaling} shows decreasing prediction and tangent-space
errors for both link functions across the full sample-size range. Least-squares fits
to the mean errors at the eight sample sizes $n\ge5\times10^5$ give prediction
slopes $-0.685$ and $-0.712$ for $f^{\rm poly}$ and $f^{\rm rat}$,
respectively; the corresponding tangent-space slopes are $-0.414$ and
$-0.426$. All four coefficients of determination exceed $0.993$. The prediction slopes are close
to the reference exponent $-2/3$, while the tangent-space errors decrease
faster than the reference exponent $-1/3$ over the fitted range.
At $n=7\times10^6$, the relative prediction and tangent-space errors are
$(6.72\pm0.47)\times10^{-6}$ and $0.03564\pm0.00058$ for $f^{\rm poly}$,
and $(4.33\pm1.43)\times10^{-6}$ and $0.07193\pm0.00059$ for
$f^{\rm rat}$. Each $\pm$ denotes one sample standard deviation across
the ten repetitions.

\FloatBarrier

\subsection{First-hit probabilities of an overdamped Langevin diffusion}\label{sec:numerics-reaction}

In molecular dynamics and other stochastic systems with several time scales,
trajectories often stay close to a low-dimensional set, and quantities such as
committor or first-hit probabilities are approximately functions of a few slow
variables, the reaction coordinates
\citep{EVandenEijnden2006,CKLMN08,RZMC11,CrosskeyMaggioni2017}. Estimating
such a probability from simulations started at many initial conditions is a
regression problem close to the Manifold-Index Model: the predictor is the
initial condition, the manifold is the low-dimensional set, and the response
is a vector of probabilities. We consider a synthetic example in which the
low-dimensional set is the sphere
$\M=\{(u,0,\dots,0):\ u\in\mathbb S^2\}\subset\R^{20}$, whose reach is $1$.

\begin{figure}[!htb]
\centering
\includegraphics[width=\linewidth]{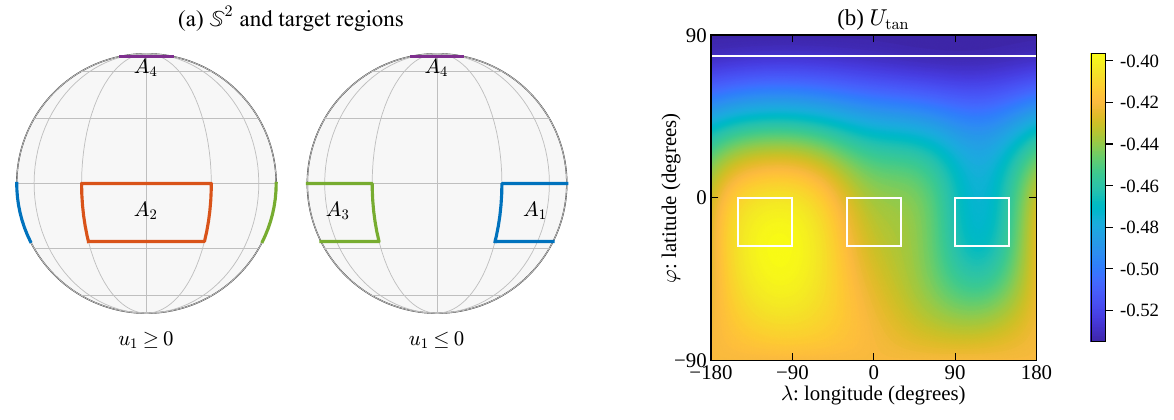}
\caption{The diffusion of Section~\ref{sec:numerics-reaction}. (a) The sphere
$\M$ in its first three coordinates, seen from the hemispheres $u_1\ge0$ and
$u_1\le0$, with the angular parts of the targets $A_1,\dots,A_4$. (b) The
tangential potential $U_{\rm tan}$ as a function of the longitude $\lambda$
and the latitude $\varphi$ of $u$; the white lines bound the angular parts of
the targets.}
\label{fig:reaction-geometry}
\end{figure}

\begin{figure}[!htb]
\centering
\includegraphics[width=.92\linewidth]{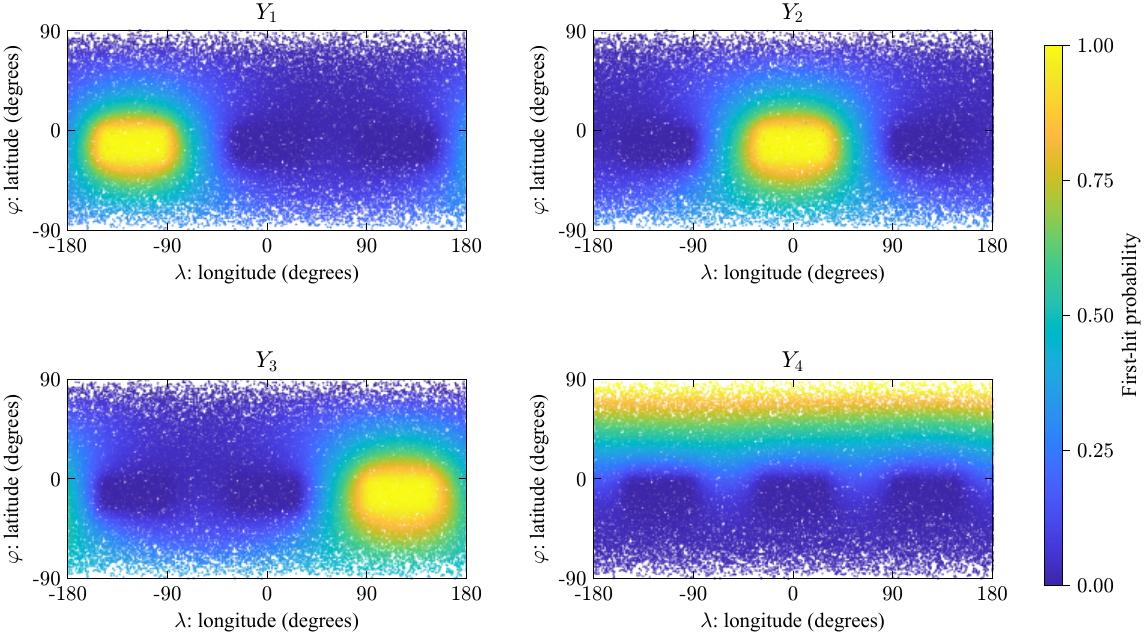}
\caption{Responses of the diffusion of Section~\ref{sec:numerics-reaction} at
$60{,}000$ plotted training initial conditions (a subset of the full training set), as functions of the longitude
$\lambda$ and the latitude $\varphi$. The panel $Y_j$ shows the
$j$-th component of the response, the Monte Carlo estimate of $F_j$ with
$48{,}000$ paths; all panels use the same color scale.}
\label{fig:reaction-responses}
\end{figure}

\paragraph{Diffusion and response.}
Write $x\in\R^{20}$ as $x=(ru,z)$, with $r\ge0$, $u\in\mathbb S^2$ and
$z\in\R^{17}$, so that $\Pi_\M x=(u,0)$ when $r>0$. For an initial condition
$x$, let $B$ be standard Brownian motion in $\R^{20}$, and let $U$
be the potential defined next. Let $X^x(t)$ solve
\begin{equation}\label{eq:reaction-dynamics}
  dX^x(t)=-\nabla U\bigl(X^x(t)\bigr)\,dt+0.4\,dB(t),\qquad X^x(0)=x,
\end{equation}
With tangential potential $U_{\rm tan}$ and positive normal
stiffnesses $\nu_0,\ldots,\nu_{17}$ specified in Appendix~\ref{app:reaction},
\begin{equation}\label{eq:reaction-potential}
  U(x)=U_{\rm tan}(u)+\frac{\nu_0}2(r-1)^2+\frac12\sum_{i=1}^{17}\nu_iz_i^2.
\end{equation}
The tangential potential is shown in Figure~\ref{fig:reaction-geometry}(b),
and the quadratic terms confine the normal coordinates. The four targets
$A_1,\dots,A_4\subset\R^{20}$ combine the angular regions in
Figure~\ref{fig:reaction-geometry}(a) with restrictions on the normal
coordinates. Appendix~\ref{app:reaction} gives the potential, stiffnesses
and target boundaries. The regression function $F=(F_1,\dots,F_4)$ has
the components
\[
  F_j(x)=\PP\bigl(\text{$X^x$ reaches $A_j$ before the other targets}\bigm|
  \text{$X^x$ reaches a target before time $100$}\bigr),
\]
and the response is a Monte Carlo estimate of $F(x)$: among the simulated
paths of \eqref{eq:reaction-dynamics} that reach a target before time $100$,
the fractions that reach $A_1,\dots,A_4$ first. Figure~\ref{fig:reaction-responses} shows the
resulting response fields. Training responses use
$48{,}000$ paths per initial condition, and validation and test responses use
$400{,}000$; the only noise is the Monte Carlo error, whose variance depends
on $F(x)$.

\paragraph{Initial conditions.}
The initial conditions are $X=(ru,z)$, where $u$ is uniform on
$\mathbb S^2$, $r=1+\varepsilon_r$ with $\varepsilon_r\sim\mathcal N(0,0.5^2)$
truncated to $[-1,1]$, and $z\sim\mathcal N(0,0.5^2I_{17})$, all independent;
as for the curves, the normal coordinates have standard deviation $0.5$ before
truncation. Their typical distance $((r-1)^2+\|z\|^2)^{1/2}$ from $\M$ is
about $2.1$, twice the reach, although the base point $(u,0)$ is unique unless
$r=0$. The model is approximate: $F$ depends on the whole initial condition,
but normal relaxation on time scales $1/\nu_i\le0.03$ is much faster than
angular motion over distances of order one, so $F$ is close to a function
of $u$. When $u$ lies in the angular part of $A_j$, $F(u,0)$ is the $j$-th
unit vector, violating \Aref{3}.

\paragraph{Data and parameters.}
The data set contains $4\times10^6$ initial conditions, of which $5000$ are
reserved for validation and $5000$ for testing; the estimator is fitted to
$n=4\times10^5$ of the others. It uses the parametrization
\eqref{eq:bumpy-counts}--\eqref{eq:bumpy-schedule} in
Appendix~\ref{app:bumpy}, with constants selected on validation data
at $n=10^6$ and held fixed for the present experiment
(Appendix~\ref{app:reaction}).

\paragraph{Results.}
On the $5000$ test conditions, the relative mean squared error is $0.048$, and
the tangent-space error \eqref{eq:tangent-error-num}, computed with the tangent
planes of the sphere, is $0.054$. Since $F$ is not available in closed form,
the relative mean squared error is computed against the test responses, which
are independent Monte Carlo estimates of $F(X)$ with $400{,}000$ paths, and
normalized by the sum of the sample variances of their four components.
The estimator recovers the sphere's tangent planes from points typically
about twice its reach away.

\section{Discussion}\label{sec:discussion}
The Manifold-Index Model separates the dimension of the predictor
distribution from the dimension that governs regression. Predictors can
fill the tube in all $D$ directions, yet complete normal fibers contribute
observations without increasing the approximation bias. Multiscale PCA on
response preimages identifies the initial tangent planes, and pooling
outer products of local slopes refines them. For links with Lipschitz
differential ($s=2$), Theorem~\ref{thm:isolated-rank} gives the
$m$-dimensional prediction rate.

MANOR takes the intrinsic dimension $m$ as input. At an isolated point where $Df$ is rank deficient,
inferring dimension from the response differential at that point alone would
underestimate $m$. Slopes at neighboring base points reveal the missing
tangent directions.

In the diffusion experiment of Section~\ref{sec:numerics-reaction}, the
reference map $u\mapsto F(u,0)$ on the sphere equals the $j$th unit vector
throughout the angular patch of target $A_j$. Predicting this common value
does not require identifying the tangent plane: within the patch the
reference response is constant in every tangent direction. Away from the
sphere, the first-hit probabilities also depend on the normal displacement,
so the manifold-index description of the full regression function is approximate.

An open question is whether comparable guarantees hold for $p<m$, in
particular for a scalar response on a manifold of dimension greater than one.
Manifolds with boundary require localization near the edge, while
anisotropic tubes could exploit longer fibers in directions where the
nearest point remains unique. It also remains open whether
$m$-dimensional rates can be obtained without inverse regression when
the intrinsic geometry is concealed by the normal displacement, and
whether fixed response resolution can accommodate unbounded noise as
the sample size grows.

The estimated tangent field defines a spatially varying anisotropic metric.
This suggests using the same neighborhood geometry for broader regression
classes and for metric learning. Models with several layers of composition
raise the further question of how to estimate and refine the geometry at
each layer. Tree-based search could reduce the computational cost of
constructing the neighborhoods.

\acks{This research was partially supported by AFOSR grants FA9550-21-1-0317
and FA9550-23-1-0445. Prisma Analytics Inc.\ provided and managed the computing resources used for the numerical experiments. 
AI tools (ChatGPT, Claude and Grok) assisted with developing and checking
proofs, manuscript preparation, and implementation of the numerical experiments,
including CUDA kernels for the GPU implementation of the estimator.
}

\bibliography{MIM_refs}

@book{Lee2018Riemannian,
  author    = {John M. Lee},
  title     = {Introduction to {Riemannian} Manifolds},
  edition   = {2},
  series    = {Graduate Texts in Mathematics},
  volume    = {176},
  publisher = {Springer},
  address   = {Cham},
  year      = {2018},
  doi       = {10.1007/978-3-319-91755-9}
}

@article{Stone82,
  author  = {Charles J. Stone},
  title   = {Optimal Global Rates of Convergence for Nonparametric Regression},
  journal = {The Annals of Statistics},
  volume  = {10},
  number  = {4},
  pages   = {1040--1053},
  year    = {1982}
}

@article{Stone85Additive,
  author  = {Charles J. Stone},
  title   = {Additive Regression and Other Nonparametric Models},
  journal = {The Annals of Statistics},
  volume  = {13},
  number  = {2},
  pages   = {689--705},
  year    = {1985},
  doi     = {10.1214/aos/1176349548}
}

@book{GKKW02,
  author    = {L{\'a}szl{\'o} Gy{\"o}rfi and Michael Kohler and Adam Krzy{\.z}ak and Harro Walk},
  title     = {A Distribution-Free Theory of Nonparametric Regression},
  publisher = {Springer},
  address   = {New York},
  year      = {2002},
  doi       = {10.1007/b97848}
}

@book{Tsybakov2009,
  author    = {Alexandre B. Tsybakov},
  title     = {Introduction to Nonparametric Estimation},
  publisher = {Springer},
  address   = {New York},
  series    = {Springer Series in Statistics},
  year      = {2009},
  doi       = {10.1007/b13794}
}

@article{DeVore:UniversalAlgorithmsLearningTheoryI,
  author  = {Peter Binev and Albert Cohen and Wolfgang Dahmen and Ronald A. DeVore and Vladimir Temlyakov},
  title   = {Universal Algorithms for Learning Theory. {P}art {I}: Piecewise Constant Functions},
  journal = {Journal of Machine Learning Research},
  volume  = {6},
  pages   = {1297--1321},
  year    = {2005}
}

@article{DeVore:UniversalAlgorithmsLearningTheoryII,
  author  = {Peter Binev and Albert Cohen and Wolfgang Dahmen and Ronald A. DeVore and Vladimir Temlyakov},
  title   = {Universal Algorithms for Learning Theory. {P}art {II}: Piecewise Polynomial Functions},
  journal = {Constructive Approximation},
  volume  = {26},
  number  = {2},
  pages   = {127--152},
  year    = {2007}
}

@misc{DeVore:OptimalLearning,
  author = {Peter Binev and Andrea Bonito and Ronald DeVore and Guergana Petrova},
  title  = {Optimal Learning},
  year   = {2023},
  note   = {arXiv:2203.15994}
}

@inproceedings{BL07,
  author    = {Peter J. Bickel and Bo Li},
  title     = {Local Polynomial Regression on Unknown Manifolds},
  booktitle = {Complex Datasets and Inverse Problems: Tomography, Networks and Beyond},
  series    = {IMS Lecture Notes--Monograph Series},
  volume    = {54},
  pages     = {177--186},
  publisher = {Institute of Mathematical Statistics},
  year      = {2007}
}

@inproceedings{Kpotufe11,
  author    = {Samory Kpotufe},
  title     = {{$k$-NN} Regression Adapts to Local Intrinsic Dimension},
  booktitle = {Advances in Neural Information Processing Systems 24},
  pages     = {729--737},
  year      = {2011}
}

@inproceedings{KG13,
  author    = {Samory Kpotufe and Vikas Garg},
  title     = {Adaptivity to Local Smoothness and Dimension in Kernel Regression},
  booktitle = {Advances in Neural Information Processing Systems 26},
  year      = {2013}
}

@article{LittleMaggioniRosasco2017,
  author  = {Anna V. Little and Mauro Maggioni and Lorenzo Rosasco},
  title   = {Multiscale Geometric Methods for Data Sets {I}: Multiscale {SVD}, Noise and Curvature},
  journal = {Applied and Computational Harmonic Analysis},
  volume  = {43},
  number  = {3},
  pages   = {504--567},
  year    = {2017},
  doi     = {10.1016/j.acha.2015.09.009}
}

@inproceedings{LMV16,
  author    = {Wenjing Liao and Mauro Maggioni and Stefano Vigogna},
  title     = {Learning Adaptive Multiscale Approximations to Data and Functions near Low-Dimensional Sets},
  booktitle = {2016 IEEE Information Theory Workshop (ITW)},
  pages     = {226--230},
  year      = {2016},
  doi       = {10.1109/ITW.2016.7606829}
}

@article{LMV22ME,
  author  = {Wenjing Liao and Mauro Maggioni and Stefano Vigogna},
  title   = {Multiscale Regression on Unknown Manifolds},
  journal = {Mathematics in Engineering},
  volume  = {4},
  number  = {4},
  pages   = {1--25},
  year    = {2022},
  doi     = {10.3934/mine.2022028}
}

@misc{liu2024deepneuralnetworksadaptive,
  author = {Hao Liu and Jiahui Cheng and Wenjing Liao},
  title  = {Deep Neural Networks are Adaptive to Function Regularity and Data Distribution in Approximation and Estimation},
  year   = {2024},
  note   = {arXiv:2406.05320}
}

@article{RosaRousseau2025,
  author  = {Paul Rosa and Judith Rousseau},
  title   = {Nonparametric Regression on Random Geometric Graphs Sampled from Submanifolds},
  journal = {Journal of Machine Learning Research},
  volume  = {26},
  number  = {164},
  pages   = {1--65},
  year    = {2025}
}

@inproceedings{stromberg,
  author    = {Jan-Olov Str{\"o}mberg},
  title     = {Computation with Wavelets in Higher Dimensions},
  booktitle = {Proceedings of the International Congress of Mathematicians},
  volume    = {3},
  pages     = {523--532},
  year      = {1998}
}

@article{Marshall2019MixedHolder,
  author  = {Nicholas F. Marshall},
  title   = {Approximating Mixed {H{\"o}lder} Functions Using Random Samples},
  journal = {The Annals of Applied Probability},
  volume  = {29},
  number  = {5},
  pages   = {2988--3005},
  year    = {2019},
  doi     = {10.1214/19-AAP1471}
}

@misc{Marshall2019HigherDimensions,
  author       = {Nicholas F. Marshall},
  title        = {Randomized Mixed {H{\"o}lder} Function Approximation in Higher-Dimensions},
  year         = {2019},
  howpublished = {arXiv preprint arXiv:1910.04201},
  url          = {https://arxiv.org/abs/1910.04201}
}

@article{Barron93,
  author  = {Andrew R. Barron},
  title   = {Universal Approximation Bounds for Superpositions of a Sigmoidal Function},
  journal = {IEEE Transactions on Information Theory},
  volume  = {39},
  number  = {3},
  pages   = {930--945},
  year    = {1993},
  doi     = {10.1109/18.256500}
}

@article{10.5555/3122009.3122028,
  author  = {Francis Bach},
  title   = {Breaking the Curse of Dimensionality with Convex Neural Networks},
  journal = {Journal of Machine Learning Research},
  volume  = {18},
  number  = {19},
  pages   = {1--53},
  year    = {2017}
}

@inproceedings{10.5555/3540261.3540537,
  author    = {Taiji Suzuki and Atsushi Nitanda},
  title     = {Deep Learning is Adaptive to Intrinsic Dimensionality of Model Smoothness in Anisotropic {B}esov Space},
  booktitle = {Advances in Neural Information Processing Systems 34},
  year      = {2021}
}

@book{ACK87,
  author    = {Gennady I. Arkhipov and Vladimir N. Chubarikov and Anatoly A. Karatsuba},
  title     = {Trigonometric Sums in Number Theory and Analysis},
  series    = {De Gruyter Expositions in Mathematics},
  volume    = {39},
  publisher = {Walter de Gruyter},
  address   = {Berlin},
  year      = {2004},
  note      = {Translated from the 1987 Russian original}
}

@article{BGGIST07,
  author  = {Luca Brandolini and Giacomo Gigante and Allan Greenleaf and Alexander Iosevich and Andreas Seeger and Giancarlo Travaglini},
  title   = {Average Decay Estimates for {F}ourier Transforms of Measures Supported on Curves},
  journal = {The Journal of Geometric Analysis},
  volume  = {17},
  number  = {1},
  pages   = {15--40},
  year    = {2007},
  doi     = {10.1007/BF02922080}
}

@article{JLT09,
  author  = {Anatoli B. Juditsky and Oleg V. Lepski and Alexandre B. Tsybakov},
  title   = {Nonparametric Estimation of Composite Functions},
  journal = {The Annals of Statistics},
  volume  = {37},
  number  = {3},
  pages   = {1360--1404},
  year    = {2009}
}

@article{10.1214/19-AOS1875,
  author  = {Johannes Schmidt-Hieber},
  title   = {Nonparametric Regression Using Deep Neural Networks with {ReLU} Activation Function},
  journal = {The Annals of Statistics},
  volume  = {48},
  number  = {4},
  pages   = {1875--1897},
  year    = {2020},
  doi     = {10.1214/19-AOS1875}
}

@misc{shen2021deepquantileregressionmitigating,
  author = {Guohao Shen and Yuling Jiao and Yuanyuan Lin and Joel L. Horowitz and Jian Huang},
  title  = {Deep Quantile Regression: Mitigating the Curse of Dimensionality Through Composition},
  year   = {2021},
  note   = {arXiv:2107.04907}
}

@misc{Lai2021TheKS,
  author = {Ming-Jun Lai and Zhaiming Shen},
  title  = {The {K}olmogorov Superposition Theorem Can Break the Curse of Dimensionality When Approximating High-Dimensional Functions},
  year   = {2021},
  note   = {arXiv:2112.09963}
}

@misc{liu2024kankolmogorovarnoldnetworks,
  author = {Ziming Liu and Yixuan Wang and Sachin Vaidya and Fabian Ruehle and James Halverson and Marin Solja{\v{c}}i{\'c} and Thomas Y. Hou and Max Tegmark},
  title  = {{KAN}: {K}olmogorov--{A}rnold Networks},
  year   = {2024},
  note   = {arXiv:2404.19756}
}

@article{Shamir_discussion,
  author  = {Ohad Shamir},
  title   = {Discussion of: ``{N}onparametric Regression Using Deep Neural Networks with {ReLU} Activation Function''},
  journal = {The Annals of Statistics},
  volume  = {48},
  number  = {4},
  pages   = {1911--1915},
  year    = {2020},
  doi     = {10.1214/19-AOS1915}
}

@misc{ShenEtAl2025TaskManifold,
  author = {Zhaiming Shen and Alex Havrilla and Rongjie Lai and Alexander Cloninger and Wenjing Liao},
  title  = {Transformers for Learning on Noisy and Task-Level Manifolds: Approximation and Generalization Insights},
  year   = {2025},
  note   = {arXiv:2505.03205},
  url    = {https://arxiv.org/abs/2505.03205}
}

@misc{KokotEtAl2026ContinuousIndex,
  author = {Alex Kokot and Anand Hemmady and Vydhourie Thiyageswaran and Marina Meila},
  title  = {Local {EGOP} for Continuous Index Learning},
  year   = {2026},
  note   = {arXiv:2601.07061v4},
  url    = {https://arxiv.org/abs/2601.07061v4}
}

@inproceedings{ChenMekaPolys,
  author    = {Sitan Chen and Raghu Meka},
  title     = {Learning Polynomials in Few Relevant Dimensions},
  booktitle = {Proceedings of the Thirty Third Conference on Learning Theory},
  series    = {Proceedings of Machine Learning Research},
  volume    = {125},
  publisher = {PMLR},
  year      = {2020}
}

@inproceedings{wang2024learning,
  author    = {Zihao Wang and Eshaan Nichani and Jason D. Lee},
  title     = {Learning Hierarchical Polynomials with Three-Layer Neural Networks},
  booktitle = {The Twelfth International Conference on Learning Representations},
  year      = {2024}
}

@inproceedings{lee2024neuralnetworklearnslowdimensional,
  author    = {Jason D. Lee and Kazusato Oko and Taiji Suzuki and Denny Wu},
  title     = {Neural Network Learns Low-Dimensional Polynomials with {SGD} near the Information-Theoretic Limit},
  booktitle = {Advances in Neural Information Processing Systems 37},
  pages     = {58716--58756},
  year      = {2024}
}

@article{HS89,
  author  = {Wolfgang H{\"a}rdle and Thomas M. Stoker},
  title   = {Investigating Smooth Multiple Regression by the Method of Average Derivatives},
  journal = {Journal of the American Statistical Association},
  volume  = {84},
  number  = {408},
  pages   = {986--995},
  year    = {1989}
}

@book{Horowitz98,
  author    = {Joel L. Horowitz},
  title     = {Semiparametric Methods in Econometrics},
  publisher = {Springer},
  address   = {New York},
  year      = {1998},
  doi       = {10.1007/978-1-4612-0621-7}
}

@article{GL07,
  author  = {St{\'e}phane Ga{\"i}ffas and Guillaume Lecu{\'e}},
  title   = {Optimal Rates and Adaptation in the Single-Index Model Using Aggregation},
  journal = {Electronic Journal of Statistics},
  volume  = {1},
  year    = {2007},
  doi     = {10.1214/07-EJS077}
}

@article{arous2021online,
  author  = {G{\'e}rard {Ben Arous} and Reza Gheissari and Aukosh Jagannath},
  title   = {Online Stochastic Gradient Descent on Non-Convex Losses from High-Dimensional Inference},
  journal = {Journal of Machine Learning Research},
  volume  = {22},
  number  = {106},
  pages   = {1--51},
  year    = {2021}
}

@inproceedings{bietti2022learning,
  author    = {Alberto Bietti and Joan Bruna and Clayton Sanford and Min Jae Song},
  title     = {Learning Single-Index Models with Shallow Neural Networks},
  booktitle = {Advances in Neural Information Processing Systems 35},
  pages     = {9768--9783},
  year      = {2022}
}

@misc{bietti2023learninggaussianmultiindexmodels,
  author = {Alberto Bietti and Joan Bruna and Loucas Pillaud-Vivien},
  title  = {On Learning {G}aussian Multi-Index Models with Gradient Flow},
  year   = {2023},
  note   = {arXiv:2310.19793}
}

@inproceedings{pmlr-v247-damian24a,
  author    = {Alex Damian and Loucas Pillaud-Vivien and Jason Lee and Joan Bruna},
  title     = {Computational-Statistical Gaps in {G}aussian Single-Index Models (Extended Abstract)},
  booktitle = {Proceedings of Thirty Seventh Conference on Learning Theory},
  series    = {Proceedings of Machine Learning Research},
  volume    = {247},
  pages     = {1262},
  publisher = {PMLR},
  year      = {2024}
}

@misc{mousavi2024learning,
  author = {Alireza Mousavi-Hosseini and Denny Wu and Murat A. Erdogdu},
  title  = {Learning Multi-Index Models with Neural Networks via Mean-Field {L}angevin Dynamics},
  year   = {2024},
  note   = {arXiv:2408.07254}
}

@misc{ren2024learningorthogonalmultiindexmodels,
  author = {Yunwei Ren and Jason D. Lee},
  title  = {Learning Orthogonal Multi-Index Models: A Fine-Grained Information Exponent Analysis},
  year   = {2024},
  note   = {arXiv:2410.09678}
}

@article{doi:10.1126/science.adi5639,
  author  = {Adityanarayanan Radhakrishnan and Daniel Beaglehole and Parthe Pandit and Mikhail Belkin},
  title   = {Mechanism for Feature Learning in Neural Networks and Backpropagation-Free Machine Learning Models},
  journal = {Science},
  volume  = {383},
  number  = {6690},
  pages   = {1461--1467},
  year    = {2024},
  doi     = {10.1126/science.adi5639}
}

@article{LMV22B,
  author  = {Alessandro Lanteri and Mauro Maggioni and Stefano Vigogna},
  title   = {Conditional Regression for Single-Index Models},
  journal = {Bernoulli},
  volume  = {28},
  number  = {4},
  pages   = {3051--3078},
  year    = {2022},
  doi     = {10.3150/22-BEJ1482}
}

@article{Ichimura93,
  author  = {Hidehiko Ichimura},
  title   = {Semiparametric Least Squares ({SLS}) and Weighted {SLS} Estimation of Single-Index Models},
  journal = {Journal of Econometrics},
  volume  = {58},
  number  = {1},
  pages   = {71--120},
  year    = {1993},
  doi     = {10.1016/0304-4076(93)90114-K}
}

@article{HHI93,
  author  = {Wolfgang H{\"a}rdle and Peter Hall and Hidehiko Ichimura},
  title   = {Optimal Smoothing in Single-Index Models},
  journal = {The Annals of Statistics},
  volume  = {21},
  number  = {1},
  pages   = {157--178},
  year    = {1993}
}

@article{DH99,
  author  = {Michel Delecroix and Marian Hristache},
  title   = {{M}-estimateurs semi-param{\'e}triques dans les mod{\`e}les {\`a} direction r{\'e}v{\'e}latrice unique},
  journal = {Bulletin of the Belgian Mathematical Society Simon Stevin},
  volume  = {6},
  number  = {2},
  pages   = {161--186},
  year    = {1999}
}

@article{DHH97,
  author  = {Michel Delecroix and Wolfgang H{\"a}rdle and Marian Hristache},
  title   = {Efficient Estimation in Conditional Single-Index Regression},
  journal = {Journal of Multivariate Analysis},
  volume  = {86},
  number  = {2},
  pages   = {213--226},
  year    = {2003}
}

@article{DHP06,
  author  = {Michel Delecroix and Marian Hristache and Valentin Patilea},
  title   = {On Semiparametric {M}-estimation in Single-Index Regression},
  journal = {Journal of Statistical Planning and Inference},
  volume  = {136},
  pages   = {730--769},
  year    = {2006},
  doi     = {10.1016/j.jspi.2004.09.006}
}

@article{CFGW97,
  author  = {Raymond J. Carroll and Jianqing Fan and Ir{\`e}ne Gijbels and Matt P. Wand},
  title   = {Generalized Partially Linear Single-Index Models},
  journal = {Journal of the American Statistical Association},
  volume  = {92},
  number  = {438},
  pages   = {477--489},
  year    = {1997},
  doi     = {10.1080/01621459.1997.10474001}
}

@article{Stoker86,
  author  = {Thomas M. Stoker},
  title   = {Consistent Estimation of Scaled Coefficients},
  journal = {Econometrica},
  volume  = {54},
  number  = {6},
  pages   = {1461--1481},
  year    = {1986}
}

@article{DL91,
  author  = {Naihua Duan and Ker-Chau Li},
  title   = {Slicing Regression: A Link-Free Regression Method},
  journal = {The Annals of Statistics},
  volume  = {19},
  number  = {2},
  pages   = {505--530},
  year    = {1991}
}

@article{Li91,
  author  = {Ker-Chau Li},
  title   = {Sliced Inverse Regression for Dimension Reduction},
  journal = {Journal of the American Statistical Association},
  volume  = {86},
  number  = {414},
  pages   = {316--327},
  year    = {1991}
}

@article{Cook00,
  author  = {R. Dennis Cook},
  title   = {{SAVE}: A Method for Dimension Reduction and Graphics in Regression},
  journal = {Communications in Statistics -- Theory and Methods},
  volume  = {29},
  number  = {9--10},
  pages   = {2109--2121},
  year    = {2000},
  doi     = {10.1080/03610920008832598}
}

@article{LZC05,
  author  = {Bing Li and Hongyuan Zha and Francesca Chiaromonte},
  title   = {Contour Regression: A General Approach to Dimension Reduction},
  journal = {The Annals of Statistics},
  volume  = {33},
  number  = {4},
  pages   = {1580--1616},
  year    = {2005}
}

@article{LW07,
  author  = {Bing Li and Shaoli Wang},
  title   = {On Directional Regression for Dimension Reduction},
  journal = {Journal of the American Statistical Association},
  volume  = {102},
  number  = {479},
  pages   = {997--1008},
  year    = {2007}
}

@article{CLS13,
  author  = {Rapha{\"e}l Coudret and Benoit Liquet and J{\'e}r{\^o}me Saracco},
  title   = {Comparison of Sliced Inverse Regression Approaches for Underdetermined Cases},
  journal = {Journal de la Soci{\'e}t{\'e} Fran{\c{c}}aise de Statistique},
  volume  = {155},
  number  = {2},
  pages   = {72--96},
  year    = {2014}
}

@article{HJS01,
  author  = {Marian Hristache and Anatoli Juditsky and Vladimir Spokoiny},
  title   = {Direct Estimation of the Index Coefficient in a Single-Index Model},
  journal = {The Annals of Statistics},
  volume  = {29},
  number  = {3},
  pages   = {595--623},
  year    = {2001}
}

@article{DalalyanJuditskySpokoiny2008,
  author  = {Arnak S. Dalalyan and Anatoli Juditsky and Vladimir Spokoiny},
  title   = {A New Algorithm for Estimating the Effective Dimension-Reduction Subspace},
  journal = {Journal of Machine Learning Research},
  volume  = {9},
  number  = {53},
  pages   = {1647--1678},
  year    = {2008}
}

@inproceedings{WuMukherjeeLiang2008,
  author    = {Qiang Wu and Sayan Mukherjee and Feng Liang},
  title     = {Localized Sliced Inverse Regression},
  booktitle = {Advances in Neural Information Processing Systems 21},
  year      = {2008}
}

@article{LiArtemiouLi2011,
  author  = {Bing Li and Andreas Artemiou and Lexin Li},
  title   = {Principal Support Vector Machines for Linear and Nonlinear Sufficient Dimension Reduction},
  journal = {The Annals of Statistics},
  volume  = {39},
  number  = {6},
  pages   = {3182--3210},
  year    = {2011},
  doi     = {10.1214/11-AOS932}
}

@article{LeeLiChiaromonte2013,
  author  = {Kuang-Yao Lee and Bing Li and Francesca Chiaromonte},
  title   = {A General Theory for Nonlinear Sufficient Dimension Reduction: Formulation and Estimation},
  journal = {The Annals of Statistics},
  volume  = {41},
  number  = {1},
  pages   = {221--249},
  year    = {2013},
  doi     = {10.1214/12-AOS1071}
}

@inproceedings{BridgesEtAl2019,
  author    = {Robert A. Bridges and Anthony D. Gruber and Christopher Felder and Miki Verma and Chelsey Hoff},
  title     = {Active Manifolds: A Non-linear Analogue to Active Subspaces},
  booktitle = {Proceedings of the 36th International Conference on Machine Learning},
  series    = {Proceedings of Machine Learning Research},
  volume    = {97},
  pages     = {764--772},
  publisher = {PMLR},
  year      = {2019}
}

@article{WuMaggioniNSVM,
  author  = {Yantao Wu and Mauro Maggioni},
  title   = {Conditional Regression for the Nonlinear Single-Variable Model},
  journal = {Journal of Machine Learning Research},
  volume  = {27},
  number  = {155},
  pages   = {1--75},
  year    = {2026}
}

@article{Federer1959,
  author  = {Herbert Federer},
  title   = {Curvature Measures},
  journal = {Transactions of the American Mathematical Society},
  volume  = {93},
  number  = {3},
  pages   = {418--491},
  year    = {1959},
  doi     = {10.1090/S0002-9947-1959-0110078-1}
}

@article{NSV,
  author  = {Partha Niyogi and Stephen Smale and Shmuel Weinberger},
  title   = {Finding the Homology of Submanifolds with High Confidence from Random Samples},
  journal = {Discrete \& Computational Geometry},
  volume  = {39},
  number  = {1--3},
  pages   = {419--441},
  year    = {2008},
  doi     = {10.1007/s00454-008-9053-2}
}

@article{AamariLevrard2019,
  author  = {Eddie Aamari and Cl{\'e}ment Levrard},
  title   = {Nonasymptotic Rates for Manifold, Tangent Space and Curvature Estimation},
  journal = {The Annals of Statistics},
  volume  = {47},
  number  = {1},
  pages   = {177--204},
  year    = {2019},
  doi     = {10.1214/18-AOS1685}
}

@article{PuchkinSpokoiny2022,
  author  = {Nikita Puchkin and Vladimir Spokoiny},
  title   = {Structure-Adaptive Manifold Estimation},
  journal = {Journal of Machine Learning Research},
  volume  = {23},
  number  = {40},
  pages   = {1--62},
  year    = {2022}
}

@article{DavisKahan1970,
  author  = {Chandler Davis and William M. Kahan},
  title   = {The Rotation of Eigenvectors by a Perturbation. {III}},
  journal = {SIAM Journal on Numerical Analysis},
  volume  = {7},
  number  = {1},
  pages   = {1--46},
  year    = {1970},
  doi     = {10.1137/0707001}
}

@book{StewartSun90,
  author    = {Stewart, G. W. and Sun, Ji-guang},
  title     = {Matrix Perturbation Theory},
  series    = {Computer Science and Scientific Computing},
  publisher = {Academic Press},
  address   = {Boston},
  year      = {1990}
}

@article{doi:10.1080/01621459.1985.10478157,
  author  = {Leo Breiman and Jerome H. Friedman},
  title   = {Estimating Optimal Transformations for Multiple Regression and Correlation},
  journal = {Journal of the American Statistical Association},
  volume  = {80},
  number  = {391},
  pages   = {580--598},
  year    = {1985},
  doi     = {10.1080/01621459.1985.10478157}
}

@incollection{Hastie:GeneralizedAdditiveModels,
  author    = {Trevor Hastie and Robert Tibshirani},
  title     = {Generalized Additive Models},
  booktitle = {Wiley StatsRef: Statistics Reference Online},
  publisher = {John Wiley \& Sons},
  year      = {2014},
  doi       = {10.1002/9781118445112.stat03141}
}

@article{Horowitz_generalclassnonparametric,
  author  = {Joel L. Horowitz and Enno Mammen},
  title   = {Rate-Optimal Estimation for a General Class of Nonparametric Regression Models with Unknown Link Functions},
  journal = {The Annals of Statistics},
  volume  = {35},
  number  = {6},
  pages   = {2589--2619},
  year    = {2007}
}

@article{XiaTongLiZhu2002,
  author  = {Yingcun Xia and Howell Tong and W. K. Li and Li-Xing Zhu},
  title   = {An Adaptive Estimation of Dimension Reduction Space},
  journal = {Journal of the Royal Statistical Society: Series B},
  volume  = {64},
  number  = {3},
  pages   = {363--410},
  year    = {2002}
}

@article{Xia2007,
  author  = {Yingcun Xia},
  title   = {A Constructive Approach to the Estimation of Dimension Reduction Directions},
  journal = {The Annals of Statistics},
  volume  = {35},
  number  = {6},
  pages   = {2654--2690},
  year    = {2007}
}

@article{MukherjeeWuZhou2010,
  author  = {Sayan Mukherjee and Qiang Wu and Ding-Xuan Zhou},
  title   = {Learning Gradients on Manifolds},
  journal = {Bernoulli},
  volume  = {16},
  number  = {1},
  pages   = {181--207},
  year    = {2010}
}

@article{AswaniBickelTomlin2011,
  author  = {Anil Aswani and Peter Bickel and Claire Tomlin},
  title   = {Regression on Manifolds: Estimation of the Exterior Derivative},
  journal = {The Annals of Statistics},
  volume  = {39},
  number  = {1},
  pages   = {48--81},
  year    = {2011}
}

@article{ChengWu2013,
  author  = {Ming-Yen Cheng and Hau-Tieng Wu},
  title   = {Local Linear Regression on Manifolds and Its Geometric Interpretation},
  journal = {Journal of the American Statistical Association},
  volume  = {108},
  number  = {504},
  pages   = {1421--1434},
  year    = {2013}
}

@article{FukumizuBachJordan2009,
  author  = {Kenji Fukumizu and Francis R. Bach and Michael I. Jordan},
  title   = {Kernel Dimension Reduction in Regression},
  journal = {The Annals of Statistics},
  volume  = {37},
  number  = {4},
  pages   = {1871--1905},
  year    = {2009}
}

@article{BauerKohler2019,
  author  = {Benedikt Bauer and Michael Kohler},
  title   = {On Deep Learning as a Remedy for the Curse of Dimensionality in Nonparametric Regression},
  journal = {The Annals of Statistics},
  volume  = {47},
  number  = {4},
  pages   = {2261--2285},
  year    = {2019}
}

@article{KohlerKrzyzak2017,
  author  = {Michael Kohler and Adam Krzy{\.z}ak},
  title   = {Nonparametric Regression Based on Hierarchical Interaction Models},
  journal = {IEEE Transactions on Information Theory},
  volume  = {63},
  number  = {3},
  pages   = {1620--1630},
  year    = {2017}
}

@article{EVandenEijnden2006,
  author  = {Weinan E and Eric Vanden-Eijnden},
  title   = {Towards a Theory of Transition Paths},
  journal = {Journal of Statistical Physics},
  volume  = {123},
  number  = {3},
  pages   = {503--523},
  year    = {2006},
  doi     = {10.1007/s10955-005-9003-9}
}

@article{CKLMN08,
  author  = {Ronald R. Coifman and Ioannis G. Kevrekidis and St{\'e}phane Lafon and Mauro Maggioni and Boaz Nadler},
  title   = {Diffusion Maps, Reduction Coordinates, and Low Dimensional Representation of Stochastic Systems},
  journal = {Multiscale Modeling \& Simulation},
  volume  = {7},
  number  = {2},
  pages   = {842--864},
  year    = {2008},
  doi     = {10.1137/070696325}
}

@article{RZMC11,
  author  = {Mary A. Rohrdanz and Wenwei Zheng and Mauro Maggioni and Cecilia Clementi},
  title   = {Determination of Reaction Coordinates via Locally Scaled Diffusion Map},
  journal = {The Journal of Chemical Physics},
  volume  = {134},
  number  = {12},
  pages   = {124116},
  year    = {2011},
  doi     = {10.1063/1.3569857}
}

@article{CrosskeyMaggioni2017,
  author  = {Miles Crosskey and Mauro Maggioni},
  title   = {{ATLAS}: A Geometric Approach to Learning High-Dimensional Stochastic Systems Near Manifolds},
  journal = {Multiscale Modeling \& Simulation},
  volume  = {15},
  number  = {1},
  pages   = {110--156},
  year    = {2017}
}

@article{GoldbergJerrum1995,
  author  = {Paul W. Goldberg and Mark R. Jerrum},
  title   = {Bounding the {Vapnik--Chervonenkis} Dimension of Concept Classes Parameterized by Real Numbers},
  journal = {Machine Learning},
  year    = {1995},
  volume  = {18},
  number  = {2--3},
  pages   = {131--148},
  doi     = {10.1007/BF00993408},
  url     = {https://www.cs.ox.ac.uk/people/paul.goldberg/papers/GJ-ML95.pdf}
}

@article{GineGuillou2002,
  author  = {Evarist Gin{\'e} and Armelle Guillou},
  title   = {Rates of Strong Uniform Consistency for Multivariate Kernel Density Estimators},
  journal = {Annales de l'Institut Henri Poincar{\'e}, Probabilit{\'e}s et Statistiques},
  year    = {2002},
  volume  = {38},
  number  = {6},
  pages   = {907--921},
  doi     = {10.1016/S0246-0203(02)01128-7},
  url     = {https://www.numdam.org/item/AIHPB_2002__38_6_907_0/}
}

@article{Bousquet2002,
  author  = {Olivier Bousquet},
  title   = {A {Bennett} Concentration Inequality and Its Application to Suprema of Empirical Processes},
  journal = {Comptes Rendus Math{\'e}matique},
  year    = {2002},
  volume  = {334},
  number  = {6},
  pages   = {495--500},
  doi     = {10.1016/S1631-073X(02)02292-6}
}

@book{Vershynin2026,
  author = {R. Vershynin},
  title = {High-Dimensional Probability: An Introduction with Applications in Data Science},
  edition = {2}, publisher = {Cambridge University Press}, year = {2026},
  url = {https://www.math.uci.edu/~rvershyn/papers/HDP-book/HDP-2.pdf}
}

@book{BoydVandenberghe2004,
  author    = {Stephen Boyd and Lieven Vandenberghe},
  title     = {Convex Optimization},
  publisher = {Cambridge University Press},
  year      = {2004},
  url       = {https://web.stanford.edu/~boyd/cvxbook/bv_cvxbook.pdf}
}

@article{Caruana1997,
  author  = {Rich Caruana},
  title   = {Multitask Learning},
  journal = {Machine Learning},
  volume  = {28},
  number  = {1},
  pages   = {41--75},
  year    = {1997},
  doi     = {10.1023/A:1007379606734}
}

@article{Baxter2000,
  author  = {Jonathan Baxter},
  title   = {A Model of Inductive Bias Learning},
  journal = {Journal of Artificial Intelligence Research},
  volume  = {12},
  pages   = {149--198},
  year    = {2000},
  doi     = {10.1613/jair.731}
}

@article{AndoZhang2005,
  author  = {Rie Kubota Ando and Tong Zhang},
  title   = {A Framework for Learning Predictive Structures from Multiple Tasks and Unlabeled Data},
  journal = {Journal of Machine Learning Research},
  volume  = {6},
  pages   = {1817--1853},
  year    = {2005}
}

@article{ArgyriouEvgeniouPontil2008,
  author  = {Andreas Argyriou and Theodoros Evgeniou and Massimiliano Pontil},
  title   = {Convex Multi-Task Feature Learning},
  journal = {Machine Learning},
  volume  = {73},
  number  = {3},
  pages   = {243--272},
  year    = {2008},
  doi     = {10.1007/s10994-007-5040-8}
}

@article{MaurerPontilRomeraParedes2016,
  author  = {Andreas Maurer and Massimiliano Pontil and Bernardino Romera-Paredes},
  title   = {The Benefit of Multitask Representation Learning},
  journal = {Journal of Machine Learning Research},
  volume  = {17},
  number  = {81},
  pages   = {1--32},
  year    = {2016}
}

@unpublished{manuscriptNSIM2018,
  author = {Lanteri, Alessandro and Kereta, {\v Z}eljko and Klock, Timo and Maggioni, Mauro and Naumova, Valeriya and Vigogna, Stefano},
  title  = {Thinking Geometrically in Regression with Structured Models: Linear and Nonlinear Single-Index Models},
  note   = {Unpublished manuscript},
  year   = {2018}
}

@article{KolmogorovTikhomirov1959,
  author  = {Kolmogorov, Andrey N. and Tikhomirov, Vladimir M.},
  title   = {$\varepsilon$-entropy and $\varepsilon$-capacity of sets in function spaces},
  journal = {Uspekhi Matematicheskikh Nauk},
  volume  = {14},
  number  = {2},
  pages   = {3--86},
  year    = {1959},
  note    = {English translation: American Mathematical Society Translations, Series 2, 17:277--364, 1961}
}

@article{YangBarron1999,
  author  = {Yang, Yuhong and Barron, Andrew},
  title   = {Information-theoretic determination of minimax rates of convergence},
  journal = {The Annals of Statistics},
  volume  = {27},
  number  = {5},
  pages   = {1564--1599},
  year    = {1999}
}

@article{GenoveseEtAl2012Minimax,
  author  = {Genovese, Christopher R. and Perone-Pacifico, Marco and Verdinelli, Isabella and Wasserman, Larry},
  title   = {Minimax manifold estimation},
  journal = {Journal of Machine Learning Research},
  volume  = {13},
  pages   = {1263--1291},
  year    = {2012}
}

@article{HoffmannLepski2002,
  author  = {Hoffmann, Marc and Lepski, Oleg},
  title   = {Random rates in anisotropic regression},
  journal = {The Annals of Statistics},
  volume  = {30},
  number  = {2},
  pages   = {325--396},
  year    = {2002}
}

@inproceedings{LittleJungMaggioni2009,
  author    = {Little, Anna V. and Jung, Yoon-Mo and Maggioni, Mauro},
  title     = {Multiscale Estimation of Intrinsic Dimensionality of Data Sets},
  booktitle = {AAAI Fall Symposium: Manifold Learning and Its Applications},
  year      = {2009}
}

@inproceedings{LittleLeeJungMaggioni2009,
  author    = {Little, Anna V. and Lee, Jason and Jung, Yoon-Mo and Maggioni, Mauro},
  title     = {Estimation of intrinsic dimensionality of samples from noisy low-dimensional manifolds in high dimensions with multiscale {SVD}},
  booktitle = {IEEE/SP 15th Workshop on Statistical Signal Processing},
  pages     = {85--88},
  year      = {2009}
}

@incollection{ChenLittleMaggioniRosasco2011,
  author    = {Chen, Guangliang and Little, Anna V. and Maggioni, Mauro and Rosasco, Lorenzo},
  title     = {Some recent advances in multiscale geometric analysis of point clouds},
  booktitle = {Wavelets and Multiscale Analysis: Theory and Applications},
  editor    = {Cohen, Jonathan and Zayed, Ahmed I.},
  publisher = {Birkh{\"a}user},
  address   = {Boston},
  pages     = {199--225},
  year      = {2011}
}

\clearpage
\appendix
\section{Proof of Theorem~\ref{thm:isolated-rank}}\label{app:proofs}

As in the curve analysis of \citet{WuMaggioniNSVM}, we first localize the
predictors by response. Here the resulting estimates initialize a moving
tangent field. After the estimator definitions and geometric tools,
Appendices~\ref{app:pilot}--\ref{app:readout} follow the three phases;
Appendices~\ref{app:n0} and~\ref{app:main} give the sample threshold and the
integrated error bounds.

\subsection{Precise definitions}\label{app:defs}

\paragraph{Number of steps.}
The number $k_*$ of refinement steps and the block size $n_f$ are fixed
by a deterministic rule that uses no data. Let $K_0>0$ be the initial
tangential-width multiplier, a fixed constant whose value is chosen in the
paragraph on starting widths below. For $j\ge0$ put
\[
 n_f(j)=\left\lfloor\frac n{4j+8}\right\rfloor,\qquad
 \underline b_j=K_0w_Y\,e^{1/[2(D+4)]}a_\rho^j ,
\]
and choose the largest number of steps $j\ge1$ for which
\begin{equation}\label{eq:sample-condition}
 n_f(j)\,\pi_-\sigma_N^m(\underline b_j/\sigma_N)^{m+4}
 \ \ge\ C_{\rm samp}(1+D+p)^4(1+R/\sigma_N)^{16}
  \left(1+\frac{\sigma_\zeta}{\sigma_{f,+}\sigma_N}\right)^2\log(2n).
\end{equation}
Set $k_*=j$ and $n_f=n_f(k_*)$. This comparison is monotone in $j$. Since
$a_\rho^{m+4}=e^{-1/2}$, $k_*=O(\log n)$ and $n_f\gtrsim n/\log(2n)$. The
multiplier $C_{\rm samp}$, and the multiplier $C_{\rm pil}$ of the pilot
sample condition \eqref{eq:pilotN}, are fixed constants and are part of the
estimator. They depend only on the fixed model parameters, the kernel and
$\tau$; the requirements on them are collected in Appendix~\ref{app:n0}.

\paragraph{Sample splitting.}
With $k_*$ as in Section~\ref{sec:theory-scales} and
$n_f=\lfloor n/(4k_*+8)\rfloor$, the observations are divided into $4k_*+8$
disjoint blocks of $n_f$ or $n_f+1$ observations. At each step $k$, one
block supplies $\Sfit$ and another $\Spool$. Of the remaining blocks,
reserve at least one for prediction ($\Srd$) and at least two for the pilot.
Fix the two pilot fold sizes in advance, each at least $n_f$. Independently
of the data, draw $\Sanc$ uniformly from the pooled pilot indices at its
prescribed size; the complement is $\Sdes$. Conditional on these labels,
the observations in disjoint folds are independent. Write
$n_{\rm pred}=|\Srd|$.
The total pilot size used in Corollary~\ref{cor:gaussian-noise} is
$N_{\rm pil}=|\Sanc|+|\Sdes|$.

\paragraph{Phase 1.}
For $i\in\Sanc$, compute the centered covariance of the observations
$j\in\Sdes$ with $\|Y_j-Y_i\|\le\delta_Y$ and $\|X_j-X_i\|\le r$.
Keep a radius when the number of selected observations is at least
$(p_0/8)|\Sdes|$ and the $(D-m)$-th largest eigenvalue
$\lambda_{D-m}$ of their covariance is at least one eighth of the normal
variance lower bound $\sigma_N^2$ in \Aref{2}. A sharper known lower
bound for $\inf_q\lambda_{\min}(\Sigma_N(q)|_{N_q\M})$, when available, replaces
$\sigma_N^2$ in this test. Resolve ties at the boundary of a selected
eigenspace by a fixed deterministic rule.
An anchor is \emph{eligible} if at least one radius is kept. For
each eligible anchor let $r_i$ be the smallest kept radius maximizing the
score \eqref{eq:hscore}; store the mean of the observations selected at
radius $r_i$, its \emph{selected mean}, and the bottom $m$ eigenvectors
$V_i$ of their covariance. We call $V_iV_i^\T$ its \emph{raw projector}.
Write $\Omega_i^{\rm anc}$ for the eligible anchors
in its response window and spatial ball, its \emph{projector pool}. The
warm start at the anchor is its \emph{pooled projector}
\begin{equation}\label{eq:pilotpool}
 \widehat P_{\rm pil}(X_i)=\Ptop\!\left(\frac1{|\Omega_i^{\rm anc}|}
      \sum_{j\in\Omega_i^{\rm anc}}V_jV_j^\T\right).
\end{equation}
Self-inclusion makes this pool nonempty. If no anchor is eligible, record
a pilot failure and use a fixed rank-$m$ projector.

\paragraph{Assignment to a target.}
Choose $C_{\rm link}$ large enough to join anchors satisfying the event in \eqref{eq:packet-core} for the same base point $q$ (Appendix~\ref{app:pilot}).
Link two eligible anchors if their responses differ by at most
$3\delta_Y/4$ and their selected means differ by at most
$C_{\rm link}w_Y$. An \emph{anchor star} $S$ consists of an
eligible anchor, its \emph{owner}, and the eligible anchors linked to it.
Retain the anchor stars with at least
$(p_0/8)|\Sanc|$ anchors, a threshold below $p_0|\Sanc|$, the lower bound on the expected number of anchors satisfying \eqref{eq:packet-core}. The \emph{center} $x_S$ of an anchor star is the
average of the selected means of its anchors, and $P_S$ is the projector
onto the top $m$ eigenspace of the average of their pooled projectors.
Among the retained anchor stars with $\|x-x_S\|\le2R$, assign to the
target $x$ one minimizing the tangential residual $\|P_S(x-x_S)\|$,
breaking ties by index. Set $\widehat P_0(x)=P_S$. The
\emph{local cylinder} of $x$ is the set \eqref{eq:local-cylinder} defined
by this anchor star.
If no anchor star qualifies, the local cylinder is empty and the
initial projector is a fixed rank-$m$ projector.

\paragraph{Starting widths.}
Take $H_0\ge3$ from Lemma~\ref{lem:quadratic-update}, let $K_0>0$
be the initial tangential-width multiplier chosen below, and set
\begin{equation}\label{eq:goodinterval}
 h_{\rm rescale}=H_0 R/\min\mathcal R,\qquad
 h_0=h_{\rm rescale}\operatorname*{avg}_{i\ {\rm eligible}}r_i,\qquad
 \rho_0=\frac{K_0w_Y}{H_0 R}.
\end{equation}
On a pilot failure, when no anchor is eligible, put
$h_0=h_{\rm rescale}\min\mathcal R=H_0R$.
Thus $h_k/R$ lies between $H_0$ and
$H_0(\max\mathcal R/\min\mathcal R)e^{1/[2(D+4)]}$. The ratio of the
two endpoints is bounded at fixed $c$. The local cylinder \eqref{eq:local-cylinder} restricts the base points
independently of the normal width $h_k$. Choose the fixed $K_0$ large enough that the pilot error is at most $\eta b_1/(2R)$ and its first regular update in \eqref{eq:regular-memory} is at most $a_\rho\eta b_1/(2R)$. Then
\begin{equation}\label{eq:rho0-derived}
 K_0w_Y\le b_0=\rho_0h_0
 \le K_0w_Y\,\max\mathcal R/\min\mathcal R .
\end{equation}
By \Aref{5}, every width in this interval is at most
$K_0\varepsilon_0(\max\mathcal R/\min\mathcal R)\min\{\sigma_N,r_{\rm cyl}\}$. Use the $r_{\rm cyl}$ in \Aref{5},
$h_*=h_{k_*}$, $\rho_*=c_\rho\rho_{k_*}$, and $b_*=\rho_*h_*$.
Choose the fixed $c_\rho\ge1$ so that both the true and estimated prediction kernels satisfy the lower bound in Lemma~\ref{lem:cylinder-localization}, as used in Appendix~\ref{app:readout}; at this resolution the cylinder-restricted
prediction support stays strictly inside the local base chart of
Lemma~\ref{lem:cylinder-localization}.

\paragraph{Phase 2.}
Use the neighborhoods $\Wwin_{x,k}$ and fitting weights
$\omega_{u,k}$ of \eqref{eq:window}.
For each $u=X_j$, $j\in\Spool$, compute and store the slope
\[
 (\widehat\beta_{0,u,k},\widehat G_{u,k})\in\argmin_{\beta_0,G}
 \sum_{\ell\in\Sfit}\omega_{u,k}(X_\ell)
       \|Y_\ell-\beta_0-G^\T(X_\ell-u)\|^2.
\]
The fit uses its own local cylinder and metric and does not depend on the target
that later uses it. Retain the fit only if at least one observation has positive weight and its weighted Gram matrix, including the intercept, is invertible. The concentration bound \eqref{eq:fit-gram-concentration} guarantees this uniformly. Let $\mathcal I_x$ be the pool-fold indices whose
centers lie in $\Wwin_{x,k}$ and whose fits pass. Define
\begin{equation}\label{eq:pool}
 \widehat C_k(x)=\frac1{|\mathcal I_x|}
       \sum_{j\in\mathcal I_x}\widehat G_{X_j,k}\widehat G_{X_j,k}^\T,
 \qquad \widehat P_k(x)=\Ptop\bigl(\widehat C_k(x)\bigr).
\end{equation}
If $\mathcal I_x$ is empty, retain $\widehat P_{k-1}(x)$.
All fits pass on the uniform concentration event, so filtering removes
no centers from the pool.

\paragraph{Population counterparts.}
Let $G_{u,k}$ be the slope of the population affine fit with weights
$\omega_{u,k}$ under the full predictor law. The population pooled matrix
averages its outer product over the target neighborhood:
\begin{equation}\label{eq:popC}
 C_k(x)=\E\bigl[G_{X,k}G_{X,k}^\T\bigm|X\in\Wwin_{x,k}\bigr].
\end{equation}
Both the fitting mass $\E\,\omega_{u,k}(X)$ and the neighborhood
probability $\PP\{X\in\Wwin_{x,k}\}$ are of order $\pi_-b_k^m$ by the kernel and probability bounds in Proposition~\ref{prop:step}, so $N_k=n_f\pi_-b_k^m$ is the effective
local count for fitting and pooling.

\paragraph{Phase 3.}
With the prediction widths $h_*$ and $b_*=\rho_*h_*$, for a query $x$ and $i\in\Srd$
let
\begin{equation}\label{eq:readout}
  \varpi_i(x)=\chi_0(x,X_i)\,K\Bigl(\frac{\|x-X_i\|^2}{h_*^2}
     +\frac{\|\widehat P_{k_*}(X_i)(x-X_i)\|^2}{b_*^2}\Bigr),
  \qquad
  \widehat F(x)=\frac{\sum_{i\in\Srd}\varpi_i(x)Y_i}{\sum_{i\in\Srd}\varpi_i(x)},
\end{equation}
with the prediction-fold mean used if the denominator vanishes. The weights depend only on
predictors and on the final field of tangent spaces.

\paragraph{Targets.}
The estimator defines a field on the whole tube by evaluating
\eqref{eq:pool} at any point $x$ of the tube. Each step uses the same
stored center slopes for all such evaluations. Uniform bounds apply in particular at future pool centers and prediction
predictors. Conditional arguments use only the Phase~1 and completed
refinement folds; the current fit and pool folds and prediction responses
remain independent of this information.

\subsection{Tools}\label{app:tools}

We collect the moment, regression, subspace and integration
estimates used by the 3 phases.

\begin{lemma}[Differential of the projection]\label{lem:dPi}
Under \Aref{1}, for $x=Q+z$ with $Q=\Pi_\M x$, $z\in N_Q\M$ and
$\|z\|\le R=c\reach_\M$,
\[
  D\Pi_\M(Q+z)=E_Q(I-S_z)^{-1}E_Q^\T,
  \qquad
  (1+c)^{-1}I\preceq(I-S_z)^{-1}\preceq(1-c)^{-1}I.
\]
\end{lemma}

\begin{proof}
Choose an orthonormal normal frame $N_q\in\R^{D\times(D-m)}$ near $Q$
whose normal connection vanishes at $Q$, and write $z=N_Q\xi_0$.
Differentiate $(q,\xi)\mapsto q+N_q\xi$, with $\xi\in\R^{D-m}$,
at $(Q,\xi_0)$ in the base direction $E_Qa$, $a\in\R^m$.
The base contributes $E_Qa$; the Weingarten equation gives the frame
contribution $-E_QS_za$, with zero normal component at $Q$.
Thus the derivative is $E_Q(I-S_z)a$.
Differentiating $\Pi_\M(q+N_q\xi)=q$ in base and normal directions gives
$D\Pi_\M(Q+z)E_Q(I-S_z)=E_Q$ and $D\Pi_\M(Q+z)P_Q^\perp=0$, which is
\eqref{eq:dPi}. Finally, the reach bounds the second fundamental form,
$\|S_z\|\le\|z\|/\reach_\M\le c$ \citep{NSV}, so the eigenvalues of the
symmetric matrix $I-S_z$ lie in $[1-c,1+c]$.
\end{proof}

\begin{lemma}[Response events do not select normal displacements]\label{lem:response-events}
Under \Aref{5}, let $E$ be an event that depends only on $(Q,\zeta)$, for instance
$E=\{\|Y-y\|\le\delta\}$ for fixed $y$ and $\delta$. Then for every bounded
function $b$,
\[
  \E[b(Z)\ind_{E}\mid Q]=\E[b(Z)\mid Q]\,\PP(E\mid Q).
\]
Equivalently, conditioning on $E$ in addition to $Q$ leaves the
conditional law of $Z$ unchanged.
\end{lemma}

\begin{proof}
The noise is independent of $X=Q+Z$, so conditionally on $Q$ the variables $Z$
and $\zeta$ are independent, and $\ind_{E}$ is a function of $(Q,\zeta)$. An
independent random anchor (such as $Y_i$ in a response preimage) is handled by
conditioning on it. If $\mathcal V\subset\R^D$ is such that, on $E$, the event
$\{X\in\mathcal V\}$ is determined by $Q$ (for instance, when $\mathcal V$
contains the whole normal fiber over every base point compatible with $E$),
then $E\cap\{X\in\mathcal V\}$ is again determined by $(Q,\zeta)$ and the
identity holds with it in place of $E$.
\end{proof}

\begin{lemma}[Changes of normalized averages]\label{lem:normalized-averages}
Let $\mu_0,\mu_1$ be finite positive measures on the same space, with
$m_j=\mu_j(1)>0$, and put $\bar h_j=\mu_j(h)/m_j$ for a bounded
vector- or matrix-valued function $h$. Write
$\Delta=\int|d\mu_1-d\mu_0|$ and
$\operatorname{osc}(h)=\sup_{x,y}\|h(x)-h(y)\|$ on their joint support.
Then
\begin{equation}\label{eq:normalized-average-change}
 \bar h_1-\bar h_0=\frac{(\mu_1-\mu_0)(h-\bar h_0)}{m_1},\qquad
 \|\bar h_1-\bar h_0\|\le\frac{\Delta}{m_1}\operatorname{osc}(h).
\end{equation}
Replacing $h$ under $\mu_1$ by a function within $\epsilon$ uniformly
adds at most $\epsilon$ to this bound.
\end{lemma}
\begin{proof}
Subtract the two normalized integrals. Since $\bar h_0$ is an average
of $h$, $\|h-\bar h_0\|\le\operatorname{osc}(h)$; integration against
the variation measure gives the bound.
\end{proof}

\begin{lemma}[Local affine fits]\label{lem:affine}
Let $\varpi_\ell\ge0$, $x_\ell\in\R^D$ and $a_\ell\in\R^p$, $\ell=1,\dots,N$, with at least one positive weight, and put
\[
  W=\sum_\ell \varpi_\ell,\quad
  \bar a=W^{-1}\sum_\ell\varpi_\ell a_\ell,\quad
  \bar x=W^{-1}\sum_\ell \varpi_\ell x_\ell,\quad
  M=\sum_\ell \varpi_\ell(x_\ell-\bar x)(x_\ell-\bar x)^\T.
\]
If $M$ is invertible, set
$\gamma_\ell=\varpi_\ell M^{-1}(x_\ell-\bar x)$.
The weighted least-squares affine fit of $a_\ell$ on $x_\ell$ has intercept
$\beta_0$ and slope $G$ with
\begin{equation}\label{eq:affine-centred}
  \beta_0=\bar a-G^\T\bar x,\qquad
  G=M^{-1}\sum_\ell \varpi_\ell(x_\ell-\bar x)a_\ell^\T=\sum_\ell \gamma_\ell a_\ell^\T.
\end{equation}
The coefficients satisfy $\sum_\ell \gamma_\ell=0$ and
$\sum_\ell \gamma_\ell(x_\ell-\bar x)^\T=I_D$. Consequently constants are
removed exactly, an affine response is reproduced exactly, and if
$a_\ell=c_0+G_0^\T x_\ell+e_\ell$, then
\begin{equation}\label{eq:affine-remainder}
  G=G_0+\sum_\ell \gamma_\ell e_\ell^\T.
\end{equation}
Let $\Sigma_a=W^{-1}\sum_\ell\varpi_\ell
(a_\ell-\bar a)(a_\ell-\bar a)^\T$ and
$\operatorname{osc}(a)=\max_{\ell,\ell'}\|a_\ell-a_{\ell'}\|$.
For every orthogonal projector $P$,
\begin{equation}\label{eq:affine-osc}
 \|PG\|\le\bigl(W\|PM^{-1}P\|\,\|\Sigma_a\|\bigr)^{1/2}
 \le\bigl(W\|PM^{-1}P\|\bigr)^{1/2}\operatorname{osc}(a).
\end{equation}
The same statements hold for weighted integrals against a measure in place of
finite sums.
Under the normalized weighted law, suppose $t=\xi-\Gamma(z-y)$, with fixed $\Gamma,y$,
$\Cov(\xi,z)=0$ and invertible coordinate covariances. The fit of a
response $a$ written as $\beta+Bt+Cz$ satisfies
\begin{equation}\label{eq:affine-split}
 B=\Cov(a,\xi)\Cov(\xi)^{-1},\qquad
 C=\Cov(a,z)\Cov(z)^{-1}+B\Gamma.
\end{equation}
\end{lemma}

\begin{proof}
The centered normal equations give \eqref{eq:affine-centred};
$\sum_\ell\gamma_\ell=0$ and
$\sum_\ell\gamma_\ell(x_\ell-\bar x)^\T=I_D$ give
\eqref{eq:affine-remainder}. Under the normalized weighted law,
Cauchy--Schwarz yields, for unit $v\in\Ran P$ and $w\in\R^p$,
\[
 |v^\T Gw|^2\le W(v^\T M^{-1}v)(w^\T\Sigma_a w).
\]
Taking suprema and using $\|\Sigma_a\|\le\operatorname{osc}(a)^2$
proves \eqref{eq:affine-osc}.
For \eqref{eq:affine-split}, regress on the uncorrelated coordinates
$(\xi,z)$, then substitute $\xi=t+\Gamma(z-y)$.
\end{proof}

\begin{lemma}[Perturbation of normal equations]\label{lem:regression-stability}
Let $Q\succeq\lambda I$ with $\lambda>0$, $C=Q^{-1}B$, and
$\widehat C=\widehat Q^{-1}\widehat B$, where $\widehat Q$ is symmetric and
$\|\widehat Q-Q\|\le\lambda/2$. Then
\begin{equation}\label{eq:moment-stability}
 \|\widehat C-C\|\le\frac2\lambda
       \bigl(\|\widehat B-B\|+\|\widehat Q-Q\|\,\|C\|\bigr).
\end{equation}
More specifically, let $T,N$ be linear maps from finite-dimensional
Euclidean spaces into a Hilbert space, with $T$ of full column rank.
Let $H=(T^*T)^{1/2}$, $C=T^\dagger N$, $V=N-TC$, and
$\widehat C=(T+E)^\dagger(N+F)$. If $\epsilon=\|EH^{-1}\|\le1/2$, then
\begin{equation}\label{eq:regression-residual-stability}
 \|H(\widehat C-C)\|
 \le4\bigl\{(1+\epsilon)\|F-EC\|+\epsilon\|V\|\bigr\}.
\end{equation}
\end{lemma}
\begin{proof}
Subtracting the normal equations gives
$\widehat Q(\widehat C-C)=\widehat B-B-(\widehat Q-Q)C$, hence
\eqref{eq:moment-stability}.
For the second assertion, $T^*V=0$ gives the normal-equation identity
\[
 (T+E)^*(T+E)(\widehat C-C)=(T+E)^*(F-EC)+E^*V.
\]
The singular values of $(T+E)H^{-1}$ lie between $1-\epsilon$ and
$1+\epsilon$. Multiply by $H^{-1}$, invert the normalized Gram, and
use $(1-\epsilon)^{-2}\le4$.
\end{proof}

\begin{lemma}[Block eigenspace perturbation]\label{lem:block-spectral}
Let $M$ be symmetric and $P$ an orthogonal projector of rank $m$ with
\[
  PMP\succeq aP,\qquad P^\perp MP^\perp\preceq bP^\perp,\qquad a-b=g>0.
\]
If $E$ is symmetric with $\|PEP\|\le g/4$ and $\|P^\perp EP^\perp\|\le g/4$, then
\begin{equation}\label{eq:block-spectral}
  \bigl\|\Ptop(M+E)-P\bigr\|
    \le\frac2g\bigl(\|P^\perp MP\|+\|P^\perp EP\|\bigr).
\end{equation}
\end{lemma}

\begin{proof}
Write $M'=M+E$ and let $M_0=PM'P+P^\perp M'P^\perp$ be its
block-diagonal part. The diagonal
perturbations only reduce the gap: $PM_0P\succeq(a-g/4)P$ and
$P^\perp M_0P^\perp\preceq(b+g/4)P^\perp$. By the min--max principle, the $m$-th
largest eigenvalue of $M'$ is at least the smallest eigenvalue of its
compression to $\Ran P$, hence at least $a-g/4$, while $M_0$ restricted to
$\Ran P^\perp$ is at most $b+g/4$. These two spectral sets are separated by
$g/2$, and the $\sin\Theta$ theorem \citep{DavisKahan1970,StewartSun90} gives
$\|\Ptop(M')-P\|\le2\|P^\perp M'P\|/g$. Since
$\|P^\perp M'P\|\le\|P^\perp MP\|+\|P^\perp EP\|$, the claim follows. Only the
cross block rotates the eigenspace.
\end{proof}

Applied to $-\Sigma$, Lemma~\ref{lem:block-spectral} controls the eigenspace of
the $m$ smallest eigenvalues of a covariance matrix; this is how it is used in
Phase~1.

\begin{lemma}[Pooled slope subspaces]\label{lem:pooled-graph}
Let $T:\R^m\to\mathcal H$ and $N:\R^k\to\mathcal H$ be linear, where
$\mathcal H$ is a Hilbert space and $s=\sigma_m(T)>0$.
Put $G=T^\dagger N$, $V=N-TG$, and let $P_G$ project onto the graph
of $G^*$. If $\|V\|<s$, then
\begin{equation}\label{eq:pooled-graph}
 \bigl\|\Ptop([T,N]^*[T,N])-P_G\bigr\|
 \le\frac{\|V\|^2\min(1,\|G\|)}{s^2-\|V\|^2}.
\end{equation}
\end{lemma}
\begin{proof}
Since $T^*V=0$,
\[
 [T,N]^*[T,N]=[I,G]^*T^*T[I,G]+\operatorname{diag}(0,V^*V).
\]
The first term has range $\Ran P_G$ and positive spectrum at least $s^2$;
min--max bounds the bottom spectrum of the sum by $\|V\|^2$.
The cross block relative to $P_G$ has norm at most
$\|V\|^2\min(1,\|G\|)$. The eigenspace perturbation bound used in
Lemma~\ref{lem:block-spectral} gives \eqref{eq:pooled-graph}.
\end{proof}

\begin{lemma}[Normal moments under changes of measure]\label{lem:normal-ibp}
Let $\mu$ be a centered law on $\R^k$ and $s>0$, with
$s^2I\preceq\Cov_\mu(Z)\preceq C_Ns^2I$ and
$\Var_\mu g\le C_Ns^2\operatorname{Lip}(g)^2$ for scalar Lipschitz $g$.
For a probability law $\nu$ with $a\le d\nu/d\mu\le b$, $a>0$,
\begin{equation}\label{eq:tilted-moments}
 \begin{gathered}
 a\Cov_\mu(Z)\preceq\Cov_\nu(Z)\preceq b\Cov_\mu(Z),\\
 \|\operatorname{Reg}_\nu g\|
 \le\frac{\|g-\E_\nu g\|_{L^2(\nu)}}{\sqrt a\,s}
 \le\sqrt{\frac{bC_N}{a}}\operatorname{Lip}(g).
 \end{gathered}
\end{equation}
where $\operatorname{Reg}_\nu g=\Cov_\nu(g,Z)\Cov_\nu(Z)^{-1}$.
The first regression bound holds for every $g\in L^2(\nu)$.
The lower covariance bound only requires $\nu\ge a\mu$.
If $\|\epsilon\|_\infty\le\delta$ and $\int Z\epsilon\,d\mu=0$, then,
for every unit vector $e$,
\begin{equation}\label{eq:normal-score-change}
 \left|\int\langle e,Z\rangle g(Z)\epsilon(Z)\,d\mu(Z)\right|
 \le C_N\delta s^2\operatorname{Lip}(g).
\end{equation}
Thus \Aref{2} gives the moment bound \eqref{eq:normal-score-change} by taking
$\mu=\mu_q$, $s=\sigma_N$, $\epsilon=1$; for
$\epsilon=d\widetilde\mu_{q'}/d\mu_q-1$, its bound gains the factor
$C d_\M(q,q')/\sigma_N$. Vector responses are tested in one direction.
\end{lemma}
\begin{proof}
The identity $\Var_\nu h=\inf_c\E_\nu(h-c)^2$ proves both covariance
comparisons and $\Var_\nu g\le b\Var_\mu g$.
Equation~\eqref{eq:affine-osc} with $P=I$ then proves the
regression bounds in \eqref{eq:tilted-moments}. For \eqref{eq:normal-score-change}, subtract
$\E_\mu g$ and apply Cauchy--Schwarz under $\mu$, followed by the
covariance and Lipschitz variance bounds. Both nearby fiber laws are
centered, as required.
\end{proof}

\begin{lemma}[Convex averages on exceptional events]\label{lem:convex-risk}
Let $\mathcal G$ be a sigma-algebra and
$\widehat y=\sum_i a_i(\mu_i+\zeta_i)$, where $a_i\ge0$,
$\sum_i a_i=1$, and $a_i,\mu_i,y$ are measurable with respect to
$\mathcal G$. Suppose $\zeta_i$ are conditionally independent and centered
with $\E[\|\zeta_i\|^2\mid\mathcal G]\le v^2$, and
$\|\mu_i-y\|\le\Delta$. For every $E\in\mathcal G$,
\begin{equation}\label{eq:convex-exceptional-risk}
 \E[\ind_{E}\|\widehat y-y\|^2]\le(\Delta^2+v^2)\PP(E).
\end{equation}
\end{lemma}
\begin{proof}
Conditioning on $\mathcal G$, convexity bounds the squared bias by
$\Delta^2$, and independence bounds the noise variance by
$v^2\sum_i a_i^2\le v^2$. Multiply by $\ind_{E}$ and integrate.
\end{proof}

\begin{lemma}[Integrating distance singularities]\label{lem:distance-integrals}
Under \Aref{1}--\Aref{2}, let $S\subset\M$ be finite and
$d_S(q)=\min\{1,d_\M(q,S)/\sigma_N\}$, with $d_S=1$ if $S$ is empty.
For $\alpha>0$ and $0<u\le1$,
\begin{equation}\label{eq:distance-integrals}
 \E\,(d_S(Q)+u)^{-\alpha}
 \le C\bigl[1+\ind_{\{m=\alpha\}}|\log u|
               +\ind_{\{m<\alpha\}}u^{m-\alpha}\bigr],
\end{equation}
where $C$ depends only on $m,\alpha,|S|$, the chart bounds and
the bound on $\sup_\M\pi/\pi_-$.
\end{lemma}
\begin{proof}
A ball in a graph chart of \Aref{1}, of radius a fixed small multiple of $\sigma_N$, gives
$\pi_-\sigma_N^m\le C$, hence $\sigma_N^m\sup_\M\pi\le C$.
Chart volume bounds and a union bound over $S$ imply
$\PP\{d_S(Q)\le t\}\le Ct^m$ for $0<t<1$.
Integration of the distribution function bounds the expectation by
$1+C\alpha\int_0^1t^m(t+u)^{-\alpha-1}\,dt$.
Splitting at $t=u$ proves \eqref{eq:distance-integrals}.
\end{proof}


\subsection{Phase 1: proof of Proposition~\ref{prop:pilot}}\label{app:pilot}
Condition throughout on the pilot split.
Recall the probability lower bound $p_0$ in \eqref{eq:packet-core}.
The argument first conditions on a retained response window; its moments
are controlled by its support, without division by $p_0$. Only the
empirical concentration step uses this lower bound on probability.

\paragraph{Conditional population moments.}
Fix an anchor $X_i=q_i+z_i$, its noise, and a scan radius $r$.
If an observation is retained, then
\[
 \|f(Q)-f(q_i)\|\le\delta_Y+2(\delta_Y/8)
 <2\delta_Y,\qquad \|Q-q_i\|\le r+2R.
\]
Consequently \eqref{eq:visible-packets} gives
$d_\M(Q,q_i)\le C_{\rm pkt}w_Y$. Since
$r>2R+2C_{\rm pkt}w_Y$, the spatial ball contains the entire normal fiber
above each such base. Conditioning on membership in the response preimage therefore leaves
that fiber law centered, with its original covariance
(Lemma~\ref{lem:response-events}). For $j\in\Sdes$, let $I_j$ indicate
retention and put $p_j=\E I_j$. Write
$\mu_{\rm ret}=(\sum_j p_j)^{-1}\sum_j p_j\mathcal L(Q_j\mid I_j=1)$
for the retention-weighted mixture of base laws. It is an arbitrary probability
measure supported on $B_\M(q_i,C_{\rm pkt}w_Y)$; in particular,
\[
 \int d_\M(q,q_i)^j\,d\mu_{\rm ret}(q)\le(C_{\rm pkt}w_Y)^j,
 \qquad j=1,2,3.
\]
These bounds do not involve $p_0$.

Put $P=P_{q_i}$. By fiber centering, the corresponding predictor
mixture has mean and covariance
\[
 \bar q=\int q\,d\mu_{\rm ret}(q),\qquad
 \Sigma=\Cov_{\mu_{\rm ret}}(Q)+\int\Sigma_N(q)\,d\mu_{\rm ret}(q).
\]
Tangent transport and the normal-chord bound imply
\begin{align*}
 \|P\Sigma P\|&\le Cw_Y^2(1+\sigma_N^2/\reach_\M^2),\\
 P^\perp\Sigma P^\perp&\succeq
 \sigma_N^2(1-Cw_Y^2/\reach_\M^2)P^\perp,\\
 \|P^\perp\Sigma P\|&\le C(w_Y^2+\sigma_N^2)w_Y/\reach_\M.
\end{align*}
For the base term, the tangent chord is $O(w_Y)$ and its normal component
is $O(w_Y^2/\reach_\M)$; for the fiber term use
$\sigma_N^2P_q^\perp\preceq\Sigma_N(q)\preceq C_N\sigma_N^2P_q^\perp$.
Since $w_Y\ll\sigma_N$, Lemma~\ref{lem:block-spectral}, applied to
$-\Sigma$, gives angle $Cw_Y/\reach_\M$ for the bottom $m$ eigenspace,
and $\|\bar q-q_i\|\le Cw_Y$. Under \Aref{4} these bounds hold at every scan radius and hence at
the radius that maximizes the score.

\paragraph{Sampling and averaging.}
The event in \eqref{eq:packet-core} has probability at least
$p_0$. Its observations satisfy the response test, since
$\delta_Y/4+2\delta_Y/8=\delta_Y/2$, and the spatial test: by
\eqref{eq:visible-packets} their bases are within $C_{\rm pkt}w_Y$ of the
anchor's base, so their predictor distance is at most
$2R+C_{\rm pkt}w_Y<r$. Conditional on the anchor,
let $\widehat\mu,\widehat\Sigma$ be the empirical mean and covariance
of the retained predictors. We show that, under \eqref{eq:pilotN},
with probability at least $1-Cn^{-\tau-2}$, simultaneously for all anchors
and scan radii, at least $|\Sdes|p_0/2$ observations are retained and
\[
 \|\widehat\mu-\bar q\|\le Cw_Y,\qquad
 \|\widehat\Sigma-\Sigma\|\le C\sigma_N^2w_Y/\reach_\M .
\]
For a fixed anchor and radius put $N_d=|\Sdes|$,
$\bar p=N_d^{-1}\sum_j p_j\ge p_0$, $r_{\max}=\max\mathcal R$ and
$t=(\tau+4)\log((2+n|\mathcal R|)/p_0)$. Conditional on the anchor,
the variables
$I_j$, $I_j\langle v,X_j-X_i\rangle$ and
$I_j\langle v,X_j-X_i\rangle^2$ are independent across $j$ and, for unit $v$,
bounded by $1,r_{\max},r_{\max}^2$, respectively, with variance sums
at most $N_d\bar p$ times the squared bounds.
Bernstein's inequality on a $1/4$-net of at most $9^D$ unit vectors, normalization by
$N_r=\sum_j I_j$, and empirical centering give
\[
 N_r\ge N_d\bar p/2,\qquad
 \|\widehat\mu-\bar q\|\le Cr_{\max}\epsilon,\qquad
 \|\widehat\Sigma-\Sigma\|\le Cr_{\max}^2\epsilon,
\]
with $\epsilon=\sqrt{(D+t)/(N_d\bar p)}+(D+t)/(N_d\bar p)$, outside an
event of probability $Ce^{-t}$. Condition~\eqref{eq:pilotN} makes the last
two bounds at most $w_Y$ and $\sigma_N^2w_Y/\reach_\M$ and absorbs the linear
term in $\epsilon$. A union bound over all anchors and radii leaves
failure probability $Cn|\mathcal R|e^{-t}\le Cn^{-\tau-2}$.
On this event the empirical eigenvalue $\lambda_{D-m}(\widehat\Sigma)$ exceeds $\sigma_N^2/8$. Thus every anchor is eligible on this
event, and its selected mean and raw projector satisfy the same population
bounds.

For an anchor $i$, every member $j$ of its projector pool has
$d_\M(Q_j,Q_i)\le C_{\rm pkt}w_Y$. Averaging their raw projectors and
extracting the top $m$ eigenspace therefore preserves error
$Cw_Y/\reach_\M$ relative to $P_{Q_i}$. In an anchor star, the selected means differ
by at most $C_{\rm link}w_Y$ from the owner's selected mean. Their base points are
then at ambient distance $Cw_Y$ from the owner's base. The local graph chart of \Aref{1} converts this to intrinsic distance $Cw_Y$. The second projector
average consequently gives, at the owner's base $q_*$,
\begin{equation}\label{eq:star}
 \|x_S-q_*\|\le Cw_Y,\qquad \|P_S-P_{q_*}\|\le Cw_Y/\reach_\M.
\end{equation}
A $w_Y$-net of $\M$ has at most $C/(\pi_-w_Y^m)$ points. At every net point,
the number of anchors satisfying the event in \eqref{eq:packet-core} is binomial with
success probability at least $p_0$, hence at least $|\Sanc|p_0/2$ outside an
event of probability $e^{-|\Sanc|p_0/8}\le e^{-t}$. Since
$p_0\le C\pi_-w_Y^m$ by the bounded density ratio, the net has at most $C/p_0$
points, and the union bound leaves a failure probability at most
$Cn^{-\tau-2}$. They have pairwise response differences at most
$3\delta_Y/4$ and selected means at distance $Cw_Y$, so they form a retained
anchor star when $C_{\rm link}$ is chosen large enough. This supplies the nearby
anchor stars used below. All constants in the population and averaging bounds
are independent of $p_0$; decreasing $p_0$ increases only the required
sample size.

\paragraph{Assignment to a target.}
The preceding net argument supplies a retained anchor star within $Cw_Y$
of each query base. At
$x=q+z$ the tangential residual of a nearby anchor star is $O(w_Y)$ and it is within $2R$.
For the assigned anchor star in \eqref{eq:star}, put
$y=q_*+P_{q_*}^\perp(x-q_*)$. Its smaller measured tangential residual
implies $\|x-y\|\le Cw_Y$ and $\|y-q_*\|\le2R+Cw_Y$. The reach normal-chord
inequality yields
\[
 \left(1-\frac{\|z\|+\|y-q_*\|}{2\reach_\M}\right)\|q-q_*\|\le\|x-y\|.
\]
It follows by taking the inner product of $x-y$ with $q-q_*$ and bounding
each normal component of the chord by $\|q-q_*\|^2/(2\reach_\M)$.
The coefficient is positive for each fixed $c<2/3$ under \Aref{5}.
Thus $\|q-q_*\|\lesssim w_Y$. At this distance, projection of the chord onto
$\M$ gives an intrinsic path of comparable length by Lemma~\ref{lem:dPi}.
Tangent transport proves the asserted angle and center estimates. For base distance at most $r_{\rm cyl}/4$,
normal rotation times $R$ plus the base displacement and the $O(w_Y)$ pilot
residual stay within the bounds of the first two conditions in
\eqref{eq:local-cylinder}, and the third condition follows from the first,
since the tangential residual of the assigned anchor star is $O(w_Y)$ and
$w_Y\le\varepsilon_0r_{\rm cyl}$ by \Aref{5}; all
offsets pass. Conversely, applying the same chord bound to an accepted
point (its normal distance from the plane $x_S+\Ran P_S$ is at most $R+O(r_{\rm cyl})$)
puts its base within $Cr_{\rm cyl}$ of $q$, with $C$ depending only on
$c$. The choice of
$\gamma_{\rm cyl}$ places this ball inside one chart. This proves
Proposition~\ref{prop:pilot}.

\subsection{Phase 2: local affine fits and pooled subspaces}\label{app:phase2}
For the proof only, put $L=\sigma_N$, $u=b/L$, $\nu=L/R$ and $H=h/R$.
We measure tangent lengths in $L$, normal displacements in $R$, and
responses in $\sigma_{f,+}L$. Normals denoted by $x,y,z$ in a fiber integral below have length at most one. We first bound the inverse covariance of the fitting coordinates, then compare the regression moments under the actual and frozen fiber laws. Subspace perturbation bounds convert these estimates into tangent-space errors, separately in the three regions described after Proposition~\ref{prop:step}.

\subsubsection{Population fits and slope outer products}
At a target $x=q+Rz_x$, with normal offset $z_x$ in units of $R$, let $P$ project onto the fitted plane,
write this plane as the graph of $G^*:T_q\M\to N_q\M$, and set
$\Delta=\|P-P_q\|$. In physical coordinates, completing
the square in \eqref{eq:metric} translates the tangent coordinate by
$\Gamma(z-Rz_x)$, with
\[
 \Gamma=(I+(b/h)^2(I+GG^*))^{-1}G.
\]
The tangent quadratic form in units $b$ differs from $I$ by
$O(\Delta^2+(b/h)^2)$, and the remaining normal quadratic form in units
$h$ by $O(\Delta^2)$ from $I$. When $\Delta=O(b/R)$ and $h\asymp R$, we have $\|G-\Gamma\|+|\|G\|-\Delta|=O((b/R)^3)$. Thus a strict bound on $\|\Gamma\|$ gives the corresponding bound on $\Delta$ after decreasing the initial resolution.

\begin{lemma}[Localization by the pilot cylinder]\label{lem:cylinder-localization}
On the event of Proposition~\ref{prop:pilot}, fix $x=q+Rz_x$ and put
$\mathcal U_q=B_\M(q,r_{\rm cyl}/4)$. Take $h\ge3R$,
$0<\eta\le1/16$ and $b\le\min\{R/8,r_{\rm cyl}/64\}$.
Let $\widehat P$ be a field of rank-$m$ orthogonal projectors.
For a tube point $x'=q'+z'$, let
$t_x(x')=d_{\widehat P(x),b,h}(x'-x)^2$ when the window uses the projector at its center $x$,
or $t_x(x')=d_{\widehat P(x'),b,h}(x'-x)^2$ for prediction.
Assume $R\|\widehat P(x)-P_q\|\le\eta b$ in the first case, and
$R\|\widehat P(x')-P_{q'}\|\le\eta b$ throughout the cylinder in the second.
For the fixed sufficiently small $\gamma_{\rm cyl}$ in \Aref{5},
\begin{equation}\label{eq:cylinder-localization}
 \chi_0(x,x')\varphi(t_x(x'))
 =\ind_{\{q'\in\mathcal U_q\}}\varphi(t_x(x')),
 \qquad \varphi(t)=K(t)\ \text{or}\ \ind_{\{t\le1\}}.
\end{equation}
The set $\{x^{\prime}:\chi_0(x,x^{\prime})=1,\ t_x(x^{\prime})\le1\}$ has bases at distance less than $8b$ from $q$,
and $K(t_x(x'))\ge1/16$ on every normal fiber above
$B_\M(q,b/16)$. Hence both its probability and the restricted kernel
mass are $\asymp\pi_-b^m$.
\end{lemma}
\begin{proof}
The pilot proof puts the base of every point in the cylinder within
$C r_{\rm cyl}$ of $q$. Choose $\gamma_{\rm cyl}$ so this distance is
at most $\reach_\M/5$, and write $s=d_\M(q,q')$.
Along the minimizing geodesic, the reach bounds the tangent-chord remainder
by $s^2/(2\reach_\M)$ and tangent transport by $s/\reach_\M$.
Since $\|Rz_x\|,\|z'\|\le R=c\reach_\M$, both
$\|P_q(x'-x)\|$ and $\|P_{q'}(x'-x)\|$ are at least
$(1-c-s/(2\reach_\M))s\ge(1-c)s/2$.
If $t_x(x')\le1$, the relevant estimated projection has norm at most $b$,
whereas $\|x'-x\|\le s+2R$. Thus, in either case,
\[
 \left(\frac{1-c}{2}-\frac{\eta b}{R}\right)s
 \le(1+2\eta)b,
\]
which gives $s<8b\le r_{\rm cyl}/8$. Conversely, every fiber above
$\mathcal U_q$ lies in the cylinder by Proposition~\ref{prop:pilot};
this proves \eqref{eq:cylinder-localization}.
For $s\le b/16$, tangent transport also bounds either true projection by
$(1+c)s$. Consequently
\[
 t_x(x')\le
 \left(\frac{1+c}{16}+2\eta+\frac{\eta b}{16R}\right)^2
 +\left(\frac{2R+b/16}{h}\right)^2<\frac34.
\]
Then $K(t_x(x'))\ge1/16$. Integrating the full normal laws over the
inner and outer base balls proves the mass bounds under \Aref{2}.
\end{proof}

The intercept lets us use $z$ in place of $z-Rz_x$. On a subset of the fibers retained in Lemma~\ref{lem:cylinder-localization}, the kernel is bounded below and the translated tangent coordinate in $b$ units ranges over a fixed ball for every $z$. Write $X(w,z)$ for the tube point with base-chart coordinate
$w$ and transported normal offset $z$. The base slice map
$w\mapsto P(X(w,z)-x)$ has smallest linear singular value at least
$(1-c)\sqrt{1-\Delta^2}$ at the origin and derivative variation
$O(\|w\|/\reach_\M)$. Its restriction to the local support therefore
has bounded Jacobian and inverse in $m$ dimensions.
In the normal frame at the target, \eqref{eq:fiber-variation}
bounds the sampling law on this subset above and below by fixed multiples
of the product of horizontal volume and the target fiber law. Under
that product the centered horizontal and normal coordinates are orthogonal;
the former have covariance comparable to $I$ and the latter to $\nu^2 I$.
The observed normal coordinate differs from $z$ by $O(\nu^2u^2)$ in norm.
After division by $\nu$, this changes any unit linear test's
standard deviation by $O(\nu u^2)$. The variational covariance comparison
of Lemma~\ref{lem:normal-ibp}, applied on this subset, gives a uniformly bounded inverse covariance in horizontal units $b$ and normal units $\sigma_N$. The argument applies
to every center's own fit.

\paragraph{Pooling outer products.}
Write $T_v,N_v$ for the tangent and normal columns of each stored
slope's transpose, and $T,N$ for their stacks in the Hilbert space of
responses over centers, under the conditional pool law or its empirical analog.
For $G=T^\dagger N$ in mixed units, the pooled matrix in common length
units is $[T,\nu N]^*[T,\nu N]$ and its comparison graph is $\nu G$.
If $\|N\|/\sigma_m(T)=O(u)$ and $\|G\|=O(u)$,
\eqref{eq:pooled-graph} gives projector error $O(\nu^3u^3)$.
Graph coordinates are Lipschitz in the projector near the tangent plane;
dividing by $\nu$ gives $O(\nu^2u^3)=O(u^3)$ in mixed units.

\paragraph{Comparison of the actual and frozen normal equations.}\label{app:normalcompare}
The population normal equations use $f(Q)$ as the response since
$\E[\zeta\mid X]=0$. At each fit center, first apply
\eqref{eq:cylinder-localization} to restrict the base integration to its
own $\mathcal U_q$. All actual and frozen measures below use this same
base chart and common normal coordinates. At a fit center of offset $y$, freeze the tangent frame and curvature
at its base $q$. First transport the law $\mu_q$ over the fit support,
keeping all observed coordinates and weights. This gives a reference
fit with a fixed normal law in the chart. Write $A_z=(I-S_{q,Rz})^{-1}$ and
$\theta=R\Gamma/b$. At fixed translated horizontal variable $s$ in units
$b$, the frozen base displacement in $L$ units is $w=uA_z(s-\theta(z-y))$.
In $(s,z)$ the frozen section is even in $s$. Its base Jacobian includes
$\det A_z$. Since $\|A_z\|\le(1-c)^{-1}$ and the section translation
is bounded, its bases also lie within $Cb$ of the center. The resolution
choice in Appendix~\ref{app:propagation} keeps both supports strictly inside
$\mathcal U_q$. We may therefore omit the cylinder within this chart and
extend the squared kernel by zero; the extension vanishes near the chart
boundary. Take $s$ in a fixed bounding ball.
If $g_0,g$ denote the weights times normalized base densities in
the frozen and curved geometries, both with the reference fiber law, and $s',n'$ the observed sheared horizontal and normal
design coordinates, use
$\|a\|_{C^{0,1}_z}=\sup_z\|a(z)\|+\operatorname{Lip}_z(a)$.
Then
\begin{align}
 \|g-g_0\|_{C^{0,1}_z}+\|s'-s\|_{C^{0,1}_z}&\le C u,&
 \|g\|_{C^{0,1}_z}+\|g_0\|_{C^{0,1}_z}&\le C,\notag\\
 \|n'-z\|_{C^{0,1}_z}&\le C\nu^2u^2 .\label{eq:mixed-design}
\end{align}
Writing $g_{\rm gr}$ for the local graph map, use the isometric normal frame
$(-Dg_{\rm gr}^*,I)(I+Dg_{\rm gr}Dg_{\rm gr}^*)^{-1/2}$.
This is the polar isometry in \eqref{eq:fiber-variation}: projection of an
origin normal onto the graph normal space has this isometric factor. Taylor
expansion using \Aref{1} gives horizontal error
$O((L/\reach_\M)\|w\|^2)$ in $L$ units and vertical error
$O((L^2/(R\reach_\M))\|w\|^2)$ in $R$ units, also after a normal
derivative since $D_z w=O(u)$. The density comparison uses
$L\|\nabla\log\pi\|\le C$, and the squared kernel has Lipschitz derivative.
Together with the separation from the chart boundary, these facts
give \eqref{eq:mixed-design} without a cylinder or chart boundary term.

Subtract the exactly reproduced ambient affine response at the center;
this is the affine-reproduction identity of Lemma~\ref{lem:affine}.
The first remaining term is
$u\,Df(q)(A_z-I)(s-\theta(z-y))$ in normalized link units.
The actual response differs from it by $O(u^2)$ in $C^{0,1}_z$.
The notation $L^2_z$ refers to integration against the reference normal law, reweighted by the section's kernel and Jacobian when present; the density ratio is bounded above and below.
The first term has Lipschitz norm $O(u)$ and directional $L^2_z$ norm
$O(\nu u)$: the $m$-dimensional entries of $S_{q,Rz}$ have second moments
$O(\nu^2)$ by \Aref{1}--\Aref{2}. Its frozen horizontal coefficient in
$s$ units and intercept are $O(\nu u)$ by conditional evenness and
directional moments. Its normal coefficient is $O(u)$ because regression
of a Lipschitz scalar on $z$ costs $C\operatorname{Lip}$ under this reweighted law (Lemma~\ref{lem:normal-ibp}). After subtracting that frozen
fit, the residual has zero mean and zero covariance with the fitting coordinates under the frozen law, $L_z^2$ norm $O(\nu u)$ in each response direction, and $C^{0,1}_z$ norm $O(u)$.

For both comparisons, let $r_0$ be the response minus the reference
affine fit, evaluated in the compared coordinates, and
let $\Delta a_s,\Delta a_n$ be
the changes from the reference fit's tangent and normal coefficients.
The preceding covariance comparison gives $\Cov(s')\asymp I$ and
$\Cov(n')-\Cov(n',s')\Cov(s')^{-1}\Cov(s',n')\asymp\nu^2I$;
conditional evenness, \eqref{eq:mixed-design} and
Lemma~\ref{lem:normal-ibp} give $\|\Cov(s',n')\|\le C\nu^2u$.
The Schur-complement inverse \citep[Appendix~A.5.5]{BoydVandenberghe2004},
applied to the centered normal equations of Lemma~\ref{lem:affine}, therefore gives
\begin{equation}\label{eq:score-schur}
 \|\Delta a_s\|+\|\Delta a_n\|
 \le C\bigl(\|\Cov(s',r_0)\|+\nu^{-2}\|\Cov(n',r_0)\|\bigr).
\end{equation}
All bounds hold in operator norm by testing scalar response directions.
For the geometry comparison, \eqref{eq:mixed-design} gives
$\Cov(s',r_0)=O(u^2)$ and $\Cov(n',r_0)=O(\nu^2u^2)$. Indeed, the frozen normal equations vanish, so both covariances are integrals of $r_0$
against the changes $s'g-sg_0$ and $n'g-zg_0$ of the weighted design
coordinates. Since $\|r_0\|_{C^{0,1}_z}=O(u)$, \eqref{eq:mixed-design} bounds
the first by $O(u^2)$ and the part of the second that contains $n'-z$ by
$O(\nu^2u^3)$. For the remaining part, the function
$\int r_0\,(g-g_0)\,ds$ of $z$ has Lipschitz seminorm $O(u^2)$, and
Lemma~\ref{lem:normal-ibp} with $\Cov(z)\preceq C_N\nu^2I$ bounds its
covariance with $z$ by $C\nu^2u^2$.
The normal mean is $O(\nu^2)$, and replacing $z$ by
$n'=z+O_{C^{0,1}_z}(\nu^2u^2)$ preserves this order after centering.
Thus \eqref{eq:score-schur} bounds both coefficient changes by $O(u^2)$.

To restore the varying fiber law, work at fixed base $q'$ and put
$\epsilon=d\widetilde\mu_{q'}/d\mu_q-1$. By \eqref{eq:fiber-variation}
and centering of both laws,
$\|\epsilon\|_\infty\le Cu$ and
$\int\epsilon\,d\mu_q=\int z\epsilon\,d\mu_q=0$.
At fixed base, the
weighted residual mean and its uncentered cross moment with the tangent coordinates change by $O(u\|r_0\|_{C^{0,1}_z})$, whereas Lemma~\ref{lem:normal-ibp}, applied to $Kr_0$, bounds the change of each normal cross moment by $C\nu^2u\|r_0\|_{C^{0,1}_z}$. The base restriction to
$\mathcal U_q$ is independent of the normal variable, and the local kernel
is separated from its boundary. Thus the integrands in these weighted moments
have the normal Lipschitz bounds used here after zero extension, without
differentiating a cylinder indicator.
Integration over the base ball, normalization by the kernel mass $\asymp\pi_-b^m$,
centering with normal mean $O(\nu^2)$, and replacing $z$ by $n'$ preserve
these orders. Hence \eqref{eq:score-schur} bounds both coefficient changes
by $O(u\|r_0\|_{C^{0,1}_z})$. The preceding residual bounds give $\|r_0\|_{C^{0,1}_z}=O(u)$.
Dividing the horizontal coefficient by $u$ and undoing the bounded shear
give tangent- and normal-slope errors $O(u)$ and $O(u^2)$, respectively.

\paragraph{The actual response in a singular chart.}
Fix an isolated singularity $q_0$. For a target within $A b$ of it,
where $A$ is fixed, Lemma~\ref{lem:cylinder-localization}, applied
to the target pool and then to each center's own fit, puts all contributing
bases within $C_A b$ of $q_0$. We express these localized integrals in
the common chart at $q_0$; the fixed chart boundaries do not cut their
supports. In a graph chart centered there write the base
as \((Lv,g(Lv))\), \(\Upsilon(v)=Dg(Lv)\), \(\Xi(v)=(I+\Upsilon(v)\Upsilon(v)^*)^{-1/2}\).
Use the orthonormal normal frame \((-\Upsilon^*,I)\Xi\). In these fixed tangent and
normal coordinates the \emph{observed} predictor is exactly
\[
 \begin{aligned}
 t&=v+\ell(v,z),&\ell(v,z)&=-(R/L)\Upsilon(v)^*\Xi(v)z,\\
 \tilde z&=z+g(Lv)/R+(\Xi(v)-I)z,& |z|&\le1 .
 \end{aligned}
\]
Use the transported law at $q_0$ as reference throughout this chart;
the actual fiber weights differ from it by $O_A(u)$.
Set $J=Df(q_0)/\sigma_{f,+}$, the normalized link differential
at the singularity. For \(|v|\le C_Au\),
\begin{align*}
 \|\ell\|+\|D_z\ell\|&\le C\varepsilon_{\rm sing} |v|,&
 \|D_v\ell\|&\le C\varepsilon_{\rm sing},\\
 \|J\ell\|+\|JD_z\ell\|&\le C(\varepsilon_{\rm sing}+|v|)|v|^2,&
 \|JD_v\ell\|&\le C(\varepsilon_{\rm sing}+|v|)|v|.
\end{align*}
Here and below \(z\)-derivatives are in the unit normal ball. To verify the
second line, the difference of differentials in \eqref{eq:opening-margin}
multiplied on the right by \(I-S_{q,Rz}\) is \(Df(q)S_{q,Rz}\). Thus \(\|Df(q)S_{q,Rz}\|\le C\varepsilon_{\rm sing} |v|\), since the weakest singular
value vanishes at \(0\) and \(Df\) is Lipschitz. At regular points \eqref{eq:opening-margin} also
implies \(\|S_{q,Rz}\|\le(1+c)\varepsilon_{\rm sing}\), and continuity extends this to the
center. Parallel transporting to the origin now gives
\(\|J S_{q,Rz}\|\le C\varepsilon_{\rm sing} |v|\).
In graph coordinates the shape matrix is
\((I+\Upsilon^*\Upsilon)^{-1}R\langle D^2g,\Xi z\rangle\); this follows by pairing the
second graph derivative with \((-\Upsilon^*,I)\Xi z\). Metric and tangent-frame
corrections are \(O((L|v|/\reach_\M)^2)\), and differentiating the \(\Xi\) in
\(\Upsilon^*\Xi\) costs \(O(|v|^2)\) after scaling. This proves the derivative
bounds. Integrating radially gives the bounds for \(\ell\); it is linear in
\(z\). Also \(\tilde z-z\), its \(z\)-derivative and \(|v|D_v(\tilde z-z)\) are
\(O((L/R)^2 |v|^2)\) by \Aref{1}.

Subtract the \emph{ambient} affine response \(f(q_0)+\sigma_{f,+}L\,Jt\),
which every local fit reproduces exactly by Lemma~\ref{lem:affine}.
Define the symmetric bilinear map
$B[v,w]=(L/\sigma_{f,+})(\nabla_v Df)(q_0)w$, where $\nabla Df$ is the
covariant derivative in parallel tangent frames, and use
$\|B\|=\sup_{\|v\|,\|w\|\le1}\|B[v,w]\|$.
The remaining response in observed horizontal coordinate $t$ is
$\tfrac12B[t,t]+e(t,z)$, where
\begin{equation}
 \|e\|+u\|D_t e\|+\|D_z e\|
 \ \le\ C_Au^2\{\varepsilon_{\rm sing}+u+\omega(C_Au)\}.                 \label{eq:critical-chart-error}
\end{equation}
Indeed \(v\mapsto t\) is invertible with \(D_tv=I+O(\varepsilon_{\rm sing})\) and
\(D_zv=O(\varepsilon_{\rm sing} |v|)\). Subtract the two quadratics at \(v\) and \(t\),
apply the second line above to the affine part, and apply \Aref{3} to the intrinsic
Taylor remainder. The input plane at each center, expressed in the
fixed chart plane rather than its own tangent, changes its shear in units
\(b/R\) by \(O_A(\varepsilon_{\rm sing})\). Changing the output back costs the same amount;
its component tested by \(J\) costs only \(O_A((\varepsilon_{\rm sing}+u)(L/R)u^2)\).
Thus the three contributions are the frame and chart comparison
$O_A((\varepsilon_{\rm sing}+u)u^2)$ from $J\ell$, the change of quadratic
coordinates $-B[\ell,t]+\tfrac12B[\ell,\ell]=O_A(\varepsilon_{\rm sing}u^2)$,
and the link Taylor remainder $O_A(u^2\omega(C_Au))$. Their $t$- and $z$-derivatives have the same
scaled order, which gives the full $C^{0,1}_z$ bound in
\eqref{eq:critical-chart-error}.

\paragraph{The quadratic response over the full tangent ball.}
Now measure the base positions in units of $b$. For the quadratic
response in \eqref{eq:critical-chart-error} write $x,y,z$ for the unit-ball
normal offsets of the target, pool center and fitting observation. In the space of square-integrable responses over pool centers, let $\Pi$ project onto the \emph{constant functions} with values in $\Ran J$. This projection leaves response variation within $\Ran J$. The input shear is $\theta=R\Gamma/b$ with $\Gamma$ in physical units;
$\theta_x$ and $\theta_{\rm ctr}$ refer to the full target and fit center.
Let $a$ be the target base's displacement from $q_0$ in $b$ units,
put $v_y(z)=1-\|z-y\|^2/H^2$, and let $\xi$ denote the centered
horizontal displacement in the hard outer window.
In the leading horizontal section the outer base point has coordinates
\[
 w=a+\theta_x(x-y)+\xi,\qquad
 \E[\xi\mid y]=0,\quad \Cov(\xi\mid y)=\frac{v_x(y)}{m+2}I .
\]
Below $Z$ is the unit-ball normal variable. Subscripts $F,y$ and $T,y$
denote, respectively, its reference fit law (the law at $q_0$ tilted by
$v_y^{m/2+2}$) and law used for tangent second moments (reweighted by $v_y^{m/2+3}$),
each normalized to mass one. Expectations and covariances carry the same
subscripts. Write $\operatorname{Reg}_{F,y}\phi
=\Cov_{F,y}(\phi,Z)\Cov_{F,y}(Z)^{-1}$ for the regression coefficient with
intercept, and let $(e_j)$ be an orthonormal tangent basis.

The resulting tangent coefficients restricted to $\ker J$, divided by $u$, and normal coefficients, divided by $u^2$, in mixed units, are
\begin{align}
 Uv&=B[w+\theta_{\rm ctr}(y-\E_{T,y}Z),v],\qquad v\in\ker J,\label{eq:weak-T}\\
 V&=\frac{\sum_jB[e_j,e_j]}{2(m+6)H^2}
           (2y-\operatorname{Reg}_{F,y}\|Z\|^2)^\T
   +\tfrac12\operatorname{Reg}_{F,y} B[\theta_{\rm ctr} Z,\theta_{\rm ctr} Z]\notag\\
  &\quad{}-B[\theta_{\rm ctr}\E_{T,y}Z,\theta_{\rm ctr}\,\cdot\,].\label{eq:weak-N}
\end{align}
A scalar regression row is identified with its vector.
Translate the inner tangent coordinate by $\theta_{\rm ctr}(z-y)$.
Its conditional law is even, with covariance $v_y(z)I/(m+6)$ for
kernel $(1-s)_+^2$. Apply \eqref{eq:affine-split} to the quadratic:
expanding $\|z-y\|^2$ and canceling the terms linear in
$\theta_{\rm ctr}y$ gives \eqref{eq:weak-T}--\eqref{eq:weak-N}, separately
at each center; the incoming field may vary arbitrarily between centers.

\Needspace{7\baselineskip}
\begin{lemma}[Pooled slopes for a quadratic response]\label{lem:quadratic-update}
Under \Aref{2}, let $g_B$ be the least value of
$(\sum_{j=1}^m|B[e_j,v]|^2)^{1/2}$ over unit $v\in\ker J$, where $(e_j)$
is an orthonormal tangent basis. If $g_B>0$, one can choose $0<\eta\le1/16$ and a finite $H_0\ge3$, depending only on $m$, $C_N$, $a_\rho$ and $\|B\|/g_B$, such that for every $H\ge H_0$, every target and every incoming field with $\|\theta\|\le\eta$,
\[
 \big\|[(I-\Pi)U]^\dagger (I-\Pi)V\big\|\le a_\rho\eta/4 .
\]
\end{lemma}
\begin{proof}
For $H\ge3$, the two reweighted normal laws just defined have density ratios, relative to the original fiber law, bounded above and below in terms of $m$. Lemma~\ref{lem:normal-ibp}, with
$s=\nu=\sigma_N/R$, gives
$\|\operatorname{Reg}_{F,y} \phi\|\le C\operatorname{Lip}\phi$ at every actual
center. Equation~\eqref{eq:weak-N} therefore gives
$\|V\|\le C\|B\|(H^{-2}+\eta^2)$, with $C$ depending only on $m$ and $C_N$.
For the term $B[w,v]$, subtracting any constant cannot remove its conditional fluctuation $B[\xi,v]$. Thus
\[
 \sigma_{\min}((I-\Pi)U)\ge
 \sqrt{(1-4/H^2)/(m+2)}\,g_B-3\|B\|\eta .
\]
The triangle inequality in $L^2$ permits the incoming field to depend on $\xi$. First choose $\eta\le1/16$ so that the subtracted term is less than half the positive lower bound at $H=3$, and the $\eta^2$ contribution after inversion is at most $a_\rho\eta/8$. Then choose $H_0$ so the $H^{-2}$ contribution costs at most another $a_\rho\eta/8$. The constants used in making these two choices are uniform for $H\ge3$.
\end{proof}

The signal $g_B$ is supplied by \eqref{eq:rank-opening}. For a unit
$v\in\ker J$, differentiate along the geodesic from $q_0$ in any unit
direction $e$, transporting $v$ in parallel. Dividing the lower bound
for $\|Df\,v\|$ by the geodesic distance and passing to the center gives
$\|B[e,v]\|\ge c_{\rm op}$ in normalized units. Thus
$g_B\ge\sqrt m\,c_{\rm op}$; the conditional horizontal
variance in Lemma~\ref{lem:quadratic-update} supplies this signal even
after projecting out the constant functions with values in $\Ran J$.

In the translated tangent section the displacement of \(t\) per unit \(z\) is
\(O_A(u)\). Thus the integrated remainder \eqref{eq:critical-chart-error} costs its displayed order
also in the reference normal slope. The base density and horizontal Jacobian in this
section differ from their constant values relatively by \(O_A(\varepsilon_{\rm sing}+u)\),
including their normal Lipschitz errors: differentiate
\(\det(I+D_v\ell)^{-1}\), using \(D_zD_v\ell=O(\varepsilon_{\rm sing})\),
\(D_v^2\ell=O(1)\) by \Aref{1}, and \(D_zv=O_A(u)\). The fit kernel retains a uniform positive lower bound on the subset used in the covariance comparison. Replacing the normal predictor \(z\) by \(\tilde z\) adds at most
\(O_A((b/L)^2)\) to the scaled normal error: its coordinate discrepancy is
\(O_A((L/R)^2u^2)\), the residual is \(O_A(u^2)\), and the normal covariance
is of order \((L/R)^2\). After removing the common affine response, the
remaining response and the reference fit have size $O_A(u^2)$ on this
chart: the horizontal coefficient in $b$ units and the normal coefficient
in $R$ units are both $O_A(u^2)$ by
\eqref{eq:critical-chart-error}--\eqref{eq:weak-N} and the preceding
bound for the normal regression coefficient. At fixed base the response is constant in $z$
before subtraction of the affine part, whose remaining normal variation
is bounded by the estimate for $JD_z\ell$. The residual therefore has
$C^{0,1}_z$ norm $O_A(u^2)$ in the preceding comparison of fiber laws. The changing law thus perturbs the
tangent slope by $O_A(u^2)$ and the normal slope by $O_A(u^3)$.
For the hard outer section, only the ellipsoidal boundary moves:
both sections stay inside the fixed base chart by the preceding localization.
The relative boundary-shell mass is $O_A(\varepsilon_{\rm sing}+u)$ and the fiber density changes by $O_A(u)$
on the intersection. Lemma~\ref{lem:normalized-averages} therefore
compares bounded moments at the same centers, without transporting the
incoming field.

The section metric correction $O((b/h)^2+\Delta^2)$ is absorbed
in the same comparison, since the normal section radius stays positive.
Put $\chi=\varepsilon_{\rm sing}+u+\omega(C_Au)$.
Divide tangent coefficients in $\ker J$ by $u$ and normal coefficients
by $u^2$. For the complementary tangent columns,
$T_+=J|_{(\ker J)^\perp}+O_A(u)$, where the restriction of $J$ to
$(\ker J)^\perp$ is constant over
centers and has nonzero singular values bounded below by \Aref{3}.
Thus $\|P_{\Ran T_+}-\Pi\|\le C_Au$ in each corresponding pool space;
both projections vanish if $J=0$.

Set $Q_0=U^*(I-\Pi)U$ and $B_0=U^*(I-\Pi)V$. Let $Q,B_{\rm act}$ be the actual rescaled Gram and
cross moment after projecting onto $(\Ran T_+)^\perp$.
The rescaled columns are bounded, with pointwise errors $O_A(\chi)$.
Lemma~\ref{lem:normalized-averages} and the preceding pool-law comparison
give $\|Q-Q_0\|+\|B_{\rm act}-B_0\|\le C_A\chi$. Indeed, both are
normalized averages of products of bounded columns. Replacing the actual
columns by the model ones changes each product by $O_A(\chi)$; replacing
the projection onto $(\Ran T_+)^\perp$ by $I-\Pi$ changes it by $O_A(u)$;
and the change of pool law, of relative mass
$O_A(\varepsilon_{\rm sing}+u)$, changes the average by the same order, by
\eqref{eq:normalized-average-change}. Lemma~\ref{lem:quadratic-update}
gives a fixed positive lower bound for the smallest eigenvalue of $Q_0$.
Lemma~\ref{lem:regression-stability} therefore bounds the change of
$Q^{-1}B_{\rm act}$ by $C_A\chi$ when $C_A\chi$ is at most half that
lower bound.
The complementary graph block is $O_A(u^2)$: its right-hand side is
$O_A(u^2)$, as is the product of the tangent columns restricted to $\ker J$, of size $O_A(u)$, and the corresponding graph coefficients, also of size $O_A(u)$, while $T_+$ is uniformly invertible
on its range. Consequently $\|N\|=O_A(u^2)$ and $\sigma_m(T)\gtrsim_A u$;
Lemma~\ref{lem:pooled-graph} adds only $O_A(u^3)$ in mixed units.

Choose $H_0=C/\sqrt\eta$. Taking $C$ large and then $\eta$ small makes
$H^{-2}/\eta$ small for every $H\ge H_0$, as the proof requires.
At fixed $A$, the scaled normal graph differs from that in
Lemma~\ref{lem:quadratic-update} by $O_A(\chi)$ plus sampling error.
The rotations back to individual frames are included; no continuity of
the estimated projector field is used.

\paragraph{The annulus between the singular and regular regions.}
Write $d=d(q)$.
For a tube point $v$, write $q_v=\Pi_\M v$ and
$z_v=(v-q_v)/R$. From here on, $\Gamma$ denotes $R/L$ times the map defined by completing the square above, so it maps normal coordinates in $R$ units to tangent coordinates in $L$ units, with a subscript or argument for the point at which it is taken. The frozen normal laws
$F,y$ and $T,y$ now also include the base Jacobian $\det A_z$;
the expectation, covariance and regression notation remains the same.
For $Au\le d\le r_{\rm sing}/L$, \eqref{eq:rank-opening} gives
$\sigma_m(Df(q))/\sigma_{f,+}\gtrsim d$. By \eqref{eq:opening-margin},
$\|Df(q)S_{q,Rz}\|\lesssim\varepsilon_{\rm sing}d$ in normalized units.
The coefficient $Df(q)(A_z-I)=Df(q)S_{q,Rz}A_z$ has the same normal
Lipschitz bound: differentiate the latter product in $z$ and use linearity
of the shape operator and the bounded resolvent.
Let $T,N$ be the actual stacked tangent and normal fit columns in mixed units, $H_T=(T^*T)^{1/2}$, and $\delta$ the supremum of $\|\Gamma\|$ over the target and its fit centers. The linear frozen tangent coefficient at normal offset $y$ is $T_y^0=Df(q)\E_{T,y}(I-S_{q,Rz})^{-1}$, where the weight defining $\E_{T,y}$ includes $\det(I-S_{q,Rz})^{-1}$. The preceding bound gives $\|T_y^0-Df(q)\|\le C\varepsilon_{\rm sing} d$. Subtract the reproduced center affine response; for a center $v$ of base $q_v$ the frozen residual is $Df(q_v)((I-S_{q_v,Rz})^{-1}-I)$ applied to the tangent displacement. Its coefficient and normal Lipschitz seminorm are $O(\varepsilon_{\rm sing} d)$, since $d(q_v)=d+O(u)$. Lemma~\ref{lem:normal-ibp} bounds its normal regression after the
section translation by $C\varepsilon_{\rm sing}d\delta$. The nonlinear response remainder and the change from frozen to actual coordinates contribute $Cu^2$ by the centered-moment estimate following \eqref{eq:mixed-design}. Tangentially the fit differs from $T_y^0$ by $O(u)$: moving its base by
$O(b)$ changes the differential and geometry by $O(u)$, and
\eqref{eq:fiber-variation} changes the normalized integral of the bounded
resolvent by $O(u)$; the tilt changes the quadratic form in the normal coordinates and tangent second-moment weight by $O(\|\Gamma\|^2)$ relative to the aligned fit. Testing a target normal in a center frame adds $\mathcal K_q(w)n=S_{q,Rn}w$, with $w$ measured in $L$ units. Thus
\[
 T_v=T_y^0+O(u),\qquad N_v=T_v\mathcal K_q(w)+O(u^2+\varepsilon_{\rm sing} d\delta),
 \qquad \|N_v\|\le C(du+u^2).
\]
For conditioning uniform in $A$, the fit comparisons give
$\sup_v\|T_v-Df(q)\|\le C_1(\varepsilon_{\rm sing}d+u)$ on $Cb$
neighborhoods, with $C_1$ independent of $A$.
Fix
$\varepsilon_{\rm sing}\le c_{\rm op}/(8C_1)$.
For every $A\ge\max\{1,8C_1/c_{\rm op}\}$,
$\sigma_m(T)\ge c_{\rm op}d/2$ and
$\|T_vH_T^{-1}\|+\|T_y^0H_T^{-1}\|+\|Df(q)H_T^{-1}\|\le C$.

Use these estimates before applying the last inverse to the normal equation. Replacing $T_v H_T^{-1}$ by $T_y^0 H_T^{-1}$ in $H_T^{-1}T^*N$ costs $O((u/d)du)=O(u^2)$; replacing the other tangent factor inside $T_v\mathcal K_q(w)$ costs $O(u^2)$. Conditional on $y$, the centered part of the frozen $w$ integrates to zero, since its multiplier depends only on $y$. Its mean is $(I-S_{q,Ry})^{-1}\Gamma_x(z_x-y)$ in mixed units. Applying $T_y^0\mathcal K_q$ to this mean costs $O(\varepsilon_{\rm sing} d\delta)$ by the product bound on $Df(q)S_{q,Rn}$ at this target. Replacing the true law by the frozen law for this oracle integrand costs $O(u^2)$, by the section-boundary and density comparison above; the input field is kept at the same centers in the error terms. Consequently for $G=T^\dagger N$,
\begin{equation}
 \|H_T G\|\le C(u^2+\varepsilon_{\rm sing} d\delta),\qquad
 (d+u)\|G\|\le C(u^2+\varepsilon_{\rm sing} d\delta).                     \label{eq:annular-score}
\end{equation}
Only $Cb$ neighborhoods enter these estimates, with $C$
independent of $A$. Thus the coefficient of $u^2$ in \eqref{eq:annular-score}
is fixed before $A$ is enlarged in Appendix~\ref{app:propagation}.
The second inequality follows from the first because
$G=H_T^{-2}T^*N$ and the comparison above gives
$\sigma_m(T)\gtrsim d+u$.

\paragraph{Two successive updates on the regular region.}\label{app:regular}
Suppose the target is at least $r_{\rm sing}$ from $\Sigma_{\rm sing}$.
Its target and center neighborhoods in two successive rounds stay outside
radius $r_{\rm sing}/2$ since $b_0\ll r_{\rm sing}$. The derivation below
uses only this property, so it also applies with a step-$k$ pool center of
the target in place of the target, as it is used for
\eqref{eq:regular-memory}. For a base displacement
$w$ in $L$ units, let $\mathcal K_q(w)n=S_{q,Rn}w$.
It maps a unit normal test to the first-order rotation of the tangent
frame and has norm at most $c\|w\|$.
Let $\delta$ be the supremum of $\|\Gamma\|$ over the target and contributing centers.

Recall the regular condition number $\kappa_f$ of \Aref{2}. In the bounds below, up to
\eqref{eq:regular-memory}, the constants $C$ multiplying $\omega_{\rm bend}$ depend only on
$m$, $c$, $C_N$ and $\kappa_f$; Remark~\ref{rem:regular-constant}
also gives uniformity as $c$ decreases.

At a center with base $r$ and unit normal offset $y$, put
$A_z=(I-S_{r,Rz})^{-1}$ and let $\Gamma$ be its incoming mixed shear.
Conditional symmetry of the horizontal coordinate gives frozen normal coefficient
\[
 Df(r)\left\{\E_{T,y}[A_z]\,\Gamma
 -\Cov_{F,y}(A_z\Gamma(z-y),z)\Cov_{F,y}(z)^{-1}\right\}.
\]
Indeed, set $J=Df(r)$ and $\xi=t+\Gamma(z-y)$, where $t$ is
the observed tangent coordinate in $L$ units. Conditional symmetry gives
$\E[\xi\mid z]=0$. Apply \eqref{eq:affine-split} to
$JA_z\xi-JA_z\Gamma(z-y)$: its $\xi$ coefficient is
$J\E_{T,y}A_z$ and its $z$ coefficient is
$-J\operatorname{Reg}_{F,y}[A_z\Gamma(z-y)]$.
Returning to $t$ gives the displayed $p\times(D-m)$ normal coefficient;
its transpose is the stored normal-gradient block.
Write $S_Z=S_{r,RZ}$ in the
unit normal coordinates. Regression reproduces
$S_Z\Gamma y=\mathcal K_r(\Gamma y)Z$ exactly. Thus the center's
normal coefficient is $J\{\mathcal K_r(\Gamma y)+\mathcal R_y\}$, where
\[
 \mathcal R_y=(\E_{T,y}A_Z-I)\Gamma
       -\operatorname{Reg}_{F,y}[(A_Z-I)\Gamma Z]
       +\operatorname{Reg}_{F,y}[S_Z^2 A_Z\Gamma y].
\]
Here we used $A_Z-I=S_Z+S_Z^2A_Z$. Under either section tilt,
\eqref{eq:normalcov}, the covariance lower bound and polarization give
\[
 \|S_Z\|_{L^4}\le C\omega_{\rm bend},\qquad
 \|\Gamma Z\|_{L^4}\le C\nu\|\Gamma\|.
\]
Indeed each entry of $S_Z$ is a normal linear functional of norm
at most $\omega_{\rm bend}/\nu$; summing squared entries gives the first bound.
The second follows by summing the $m$ rows of $\Gamma$.
Since $\|A_Z-I\|\le\|S_Z\|/(1-c)$,
$\|\E_{T,y}A_Z-I\|\le C\omega_{\rm bend}$ by \eqref{eq:bend}.
Lemma~\ref{lem:normal-ibp} bounds regression by $C/\nu$
times the response $L^2$ norm. Since $\|S_Z^2A_Z\|\le
\|S_Z\|^2/(1-c)$, Cauchy--Schwarz and
$\omega_{\rm bend}/\nu\le\sqrt{C_N}c$ give
the bound $C\omega_{\rm bend}\|\Gamma\|$ for each of the three terms of
$\mathcal R_y$. Hence
\begin{equation}\label{eq:regular-center}
 \|\mathcal R_y\|\le
 C\omega_{\rm bend}\|\Gamma\|,
 \qquad \|\E_{T,y}A_Z-I\|\le C\omega_{\rm bend} .
\end{equation}

Freeze the outer pool at $q$, and write $\E_0$ for expectation in
that pool. Put $J=Df(q)$, $T_y^0=J\E_{T,y}A_z$ and
$Q_y=(T_y^0)^*T_y^0$. The eigenvalues of $\E_{T,y}A_z$ lie in
$[(1+c)^{-1},(1-c)^{-1}]$. Therefore
\[
 \|(\E_0 Q_y)^{-1}\|\le\frac{(1+c)^2}{\sigma_m(J)^2},
 \qquad
 \|Q_y-J^*J\|\le C\|J\|^2\omega_{\rm bend} ,
\]
where the second bound follows by subtracting one factor at a time.

Write $\Gamma_j(v)$ for the mixed shear at $v$ after step $j$.
Let $a_v$ be the parallel transport of
$\Gamma_{k-1}(v)z_v$ to $T_q\M$, extended by zero outside the actual
pool in the common base and normal coordinates. Put
$U_v=\mathcal K_q(w)+\mathcal K_q(a_v)$: the two terms account for
transport to the target and the tilt of the center fit, respectively.
With normal column $T_y^0U_v$, the frozen normal equation gives $G_0=(\E_0 Q_y)^{-1}\E_0[Q_yU_v]$. Hence
\[
 G_0-\E_0U_v=(\E_0 Q_y)^{-1}
    \E_0\!\left[(Q_y-J^*J)
       \bigl(\E_0[U_v\mid y]-\E_0U_v\bigr)\right].
\]
The centered horizontal section has
$\E_0[w\mid y]=A_y\Gamma_x(z_x-y)$, whereas $\|a_v\|\le\delta$
pointwise. Hence $\|\E_0[U_v\mid y]\|\le
c(2/(1-c)+1)\delta$. The last three bounds give the
absolute error
\begin{equation}\label{eq:regular-pool-a}
 \|G_0-\E_0U_v\|
 \le C\kappa_f^2\omega_{\rm bend}\delta .
\end{equation}

To evaluate the remaining mean, the frozen outer normal marginal
has bounded density relative to the unselected law of $Z/R$.
Polarization of \eqref{eq:bend} gives
$\E_0\|S_{q,Ry}\|\le C\omega_{\rm bend}$.
Moreover $\|\E_0 y\|\le C\nu$ by testing unit normal
directions, and $\nu\|\mathcal K_q\|\le\omega_{\rm bend}$: the covariance lower bound gives
$\sigma_N\|\II_q(v,v)\|\le\omega_{\rm bend}$ for each unit tangent $v$.
In $\E_0[A_y\Gamma_x(z_x-y)]$ use $A_y-I=A_y S_{q,Ry}$ to obtain
\begin{equation}\label{eq:regular-pool-b}
 \|\E_0 U_v-\mathcal K_q(\Gamma_x z_x+\E_0 a_v)\|
 \le C\omega_{\rm bend}\delta .
\end{equation}

The center formula adds $J\mathcal R_y$ to the normal column.
Writing its leading term as $T_y^0\mathcal K_q(a_v)$ instead of
$J\mathcal K_q(a_v)$ costs at most $C\|J\|\omega_{\rm bend}\delta$.
The two errors together cost at most $C\|J\|\omega_{\rm bend}\delta$
in the column, hence at most
\begin{equation}\label{eq:regular-pool-c}
 C\kappa_f\omega_{\rm bend}\delta
\end{equation}
after the pooled inverse $T^\dagger$
for the stack of $T_y^0$.

The actual and frozen sections differ by relative mass $O(u)$:
within the common local chart, whose boundary is separated from both
supports, at each normal offset the rescaled base boundary moves by $O(u)$, its radius
is bounded below, and the density ratio on the intersection is $1+O(u)$. The tangent
column changes by $O(u)$; after transport to the target, the normal
column changes by $O(u^2)$ by \eqref{eq:mixed-design} and the
comparison of the normal regression coefficients above. Here $\|w\|+\|a_v\|=O(u)$, so the
Gram error is $O(u)$ and the cross-moment error is $O(u^2)$
(Lemma~\ref{lem:normalized-averages}). The frozen graph is $O(u)$;
Lemma~\ref{lem:regression-stability} gives an $O(u^2)$ graph error.
Explicitly, the two Gram matrices differ by $O(u)$ and have bounded
inverses, and the two cross moments differ by $O(u^2)$ and have size $O(u)$;
a graph is the product of an inverse Gram matrix and a cross moment, so the
two graphs differ by $O(u^2)+O(u)\,O(u)$.
The same change of measure costs $O(u^2)$ in $\E_0 a_v$, with the
incoming field held at the same centers.
The three contributions \eqref{eq:regular-pool-a},
\eqref{eq:regular-pool-b} and \eqref{eq:regular-pool-c} sum to at most
$C\omega_{\rm bend}\delta$.
Consequently the outgoing mixed shear has leading term
\[
 \mathcal K_q\bigl(\Gamma_{k-1}(x)z_x+\E\,a_v\bigr),
\]
where the expectation is over the centers $v$ of the target's actual pool,
with remainder of norm at most $C\omega_{\rm bend}\delta+Cu^2$
at population level.

Let $\delta_j$ be the supremum of $\|\Gamma_j\|$ over the neighborhoods used by the update, and put $u_j=b_j/L$.
Evaluating the preceding leading term on its own normal offset
introduces a shape operator. Its pool average is controlled by the mass bounds in Lemma~\ref{lem:cylinder-localization} and \eqref{eq:bend}, giving
\begin{equation}\label{eq:regular-memory}
 \|\Gamma_k(x)\|
 \le (c+C_{\rm reg}\omega_{\rm bend})\delta_{k-1}
       +C_{\rm reg}\omega_{\rm bend}\delta_{k-2}
       +C(u_k^2+u_{k-1}^2).
\end{equation}
Indeed, the vector inside $\mathcal K_q$
in the preceding leading-term formula has norm at most $2\delta_{k-1}$.
Evaluation at the center's own offset gives
\[
 \|\Gamma_{k-1}(v)z_v\|
 \le 2\|S_{\Pi_\M v,Rz_v}\|\delta_{k-2}
       +C\omega_{\rm bend}\delta_{k-2}+Cu_{k-1}^2.
\]
At fixed base the selected normal mass is bounded above by the original
fiber law. Lemma~\ref{lem:cylinder-localization} bounds the ratio of the mass above $B_\M(q,8b)$ to the pool mass by a constant (up to $1+O(u)$). The fiber mean of
$\|S_{\Pi_\M v,Rz_v}\|$ is at most $m\omega_{\rm bend}$ by polarization; its
pool mean is therefore at most $C\omega_{\rm bend}+O(u_k)$; the last term
multiplies $\delta_{k-2}=O(u_{k-1})$ and is part of the quadratic remainder.
Applying $\|\mathcal K_q\|\le c$ to that averaged
vector and adding the current remainder gives a coefficient
$c+C\omega_{\rm bend}$ for $\delta_{k-1}$ and a coefficient
$C\omega_{\rm bend}$ for $\delta_{k-2}$.
Taking $C_{\rm reg}$ to be the larger of these two constants proves
\eqref{eq:regular-memory}; the direct first update has
coefficient at most $2c+C_{\rm reg}\omega_{\rm bend}$.
Fresh-fold errors add to the remainders on the concentration events
of both rounds (only step $1$ for the first update).
The pilot multiplier in \eqref{eq:rho0-derived} covers the first update.

\begin{remark}[Dependence of the regular-region constant]
\label{rem:regular-constant}
The preceding proof gives $C_{\rm reg}=C_{\rm reg}(m,c,C_N,\kappa_f)$,
independent of $D,p,n$. It can be chosen uniformly for
$0<c\le\bar c<2/3$: the section density ratios and the ratios of
volumes of the outer and inner base balls in Lemma~\ref{lem:cylinder-localization} are bounded in terms of
$m,\bar c$, since $1-4/H^2\ge5/9$, $\eta\le1/16$ and
$(1+\bar c)^{-m}\le\det A_z\le(1-\bar c)^{-m}$; the normal-moment
estimates and pooled inverses introduce only $C_N$ and $\kappa_f$.
For curves, $\kappa_f=1$ even when the nonzero derivative is small;
the coefficient of $u_k^2+u_{k-1}^2$ in \eqref{eq:regular-memory} may depend on $c_{\rm op}^{-1}$.
In the circle example of Appendix~\ref{app:cusp-example},
$\omega_{\rm bend}=c/\sqrt3$, so this uniform bound gives
\eqref{eq:bend-margin} for all sufficiently small positive $c$.
That example also verifies the remaining assumptions.
\end{remark}

\subsubsection{Sampling the fits and their outer products}\label{app:concentration}
Condition on all earlier folds. At width $b=b_k$, with the effective
local count $N_k=n_f\pi_-b_k^m$ of Appendix~\ref{app:defs}, put
\[
 e_k=C(1+R/\sigma_N)^6
       \left(1+\frac{\sigma_\zeta}{\sigma_{f,+}\sigma_N}\right)
       \left(\sqrt{\frac{(1+D+p)^2\log(2n)}{N_k}}
             +\frac{(1+D+p)^2\log(2n)}{N_k}\right).
\]
Put $\ell=C_\tau(1+D+p)^2\log(2n)$ and
$\beta_k=\sqrt{\ell/N_k}+\ell/N_k$. Condition on the earlier folds and
pool locations. For each stored fit with weight $\omega$ from
\eqref{eq:window}, use the augmented predictor
$\psi=(1,s',n'/\nu)$ and response $y_0=(f(Q)-f(q))/(\sigma_{f,+}L)$;
on its support, $\|\psi\|\le C(1+\nu^{-1})$ and $\|y_0\|\le Cu$.
Normalize population and empirical moments by $\E\omega\asymp\pi_-b^m$,
which cancels from the normal equations.
Set $H_\psi=\E[\omega\psi\psi^*]/\E\omega$. Write $\Theta$ for the
population coefficients for $y_0$ and $\widehat\Theta$ for the empirical
coefficients for $(Y-f(q))/(\sigma_{f,+}L)$, both regressed on $\psi$.
Let $\widehat H_\psi$ be the empirical counterpart of $H_\psi$.
The covariance comparison above gives $C^{-1}I\preceq H_\psi\preceq CI$.
Least-squares projection bounds the population coefficient matrix by
$\|\Theta\|\le Cu$, hence $\|y_0-\Theta^*\psi\|\le C(1+\nu^{-1})u$.
Bernstein's inequality \citep[Theorem~2.9.5]{Vershynin2026}, applied on
$1/4$-nets, bounds the difference between $H_\psi$ and its empirical
counterpart on the fit fold by $C(1+\nu^{-1})^2\beta_k$ and the
empirical average of $\omega\psi(y_0-\Theta^*\psi)^*/\E\omega$ by
$C(1+\nu^{-1})^2u\beta_k$;
the latter has zero expectation.
Condition~\eqref{eq:sample-condition} gives
\begin{equation}\label{eq:fit-gram-concentration}
 \|\widehat H_\psi-H_\psi\|\le C(1+\nu^{-1})^2\beta_k
 \le\lambda_{\min}(H_\psi)/2.
\end{equation}
Conditional on the predictors, \Aref{5} and $\omega^2\le\omega$ bound the noise
contribution to the coefficients by
$C\{\sigma_\zeta/(\sigma_{f,+}L)\}\beta_k$.
Lemma~\ref{lem:regression-stability}, with $Q=H_\psi$, gives
\begin{equation}\label{eq:fit-concentration}
 \|\widehat\Theta-\Theta\|
 \le C\left((1+\nu^{-1})^2u+\frac{\sigma_\zeta}{\sigma_{f,+}L}\right)\beta_k
 \le e_k,
\end{equation}
since $u\le1$ and $1+\nu^{-1}=1+R/\sigma_N$. A union bound over at most $n$ stored centers
makes these bounds simultaneous with failure probability at most
$Cn^{-\tau-2}$ by the choice of $C_\tau$, despite reuse of the fit fold.
Dividing the tangent and normal coefficients by $u$ and $\nu$ gives errors
$Ce_k/u$ and $Ce_k/\nu$ in mixed slope units. A center-to-target rotation
adds $O(u)$ times the tangent error to the mixed normal error.

For the outer sample, first pool the \emph{population} fits as in \eqref{eq:popC}, which are
fixed functions of the predictor conditional on the earlier folds.
Let $H_T=(T^*T)^{1/2}$ under the conditional target window and suppose the
normal matrices satisfy $\sup_v\|N_v\|\le M_N$ on this window. In all three regions
$\sup_v\|T_v H_T^{-1}\|\le C$: use the lower bound \eqref{eq:rank-opening} on the regular region, the annular
comparison to $Df(q)$, or, for $d(q)\le Au$, the comparison to $J+O_A(u)$
and the bound $\sigma_m(T)\gtrsim_A u$. In the last case the varying part is
$O_A(u)$ and for any unit vector $a$ the difference between
$\|T_v a\|$ and $\|Ta\|_{L^2}$ is $O_A(u)$.
For $M_N>0$, the normalized matrices $H_T^{-1}T_v^*T_vH_T^{-1}$ and
$H_T^{-1}T_v^*N_v/M_N$ have bounded operator norms on the window;
when $M_N=0$, the cross moment vanishes. Multiplying their scalar directional
tests by the window indicator gives variances $O(\pi_-b^m)$.
Their uniform relative deviations are $O(\beta_k)$, as follows.
For uniformity, recall the population slope
$G_{v,k}\in\R^{D\times p}$ at a center $v$ from Appendix~\ref{app:defs}.
Conditional on earlier folds,
$M(v)=G_{v,k}G_{v,k}^\T$ is fixed. Rescale the scalar tests
above by their uniform bound and extend them by zero outside their
window $W=\Wwin_{x,k}$, the target neighborhood \eqref{eq:window},
which lies in the local cylinder; call the resulting functions $h_W$.
We retain these observable windows for the uniform concentration argument;
Lemma~\ref{lem:cylinder-localization} supplies their support and mass bounds,
without replacing the indexed windows by geodesic balls.
Then $|h_W|\le\ind_W$, and $h_W$ is a linear functional of $M(v)$ on $W$.
The subgraphs $\{(v,t):t<h_W(v)\}$ are Boolean combinations of a fixed
number of linear inequalities in the $O(D^2)$ coordinates
$1,t,v_i,v_iv_j,M(v)_{ij}$. Thus their Vapnik--Chervonenkis (VC) dimension
is at most $CD^2$ \citep[Theorem~2.2]{GoldbergJerrum1995}, without requiring
regularity of $M$.

Write $\PP_{n_f}$ for the empirical measure of the pool fold. At fixed
width $b$, all window masses $p_W=\PP(W)$ are comparable to $\pi_-b^m$.
Applying \citet[inequality~(2.2)]{GineGuillou2002} and
\citet[Theorem~2.3]{Bousquet2002} to the tests and window indicators,
whose variances are at most $C\pi_-b^m$, gives simultaneously
\begin{equation}\label{eq:pool-concentration}
 |(\PP_{n_f}-\PP)h_W|+|\PP_{n_f}(W)-p_W|
 \le C\left(\sqrt{p_W\ell/n_f}+\ell/n_f\right),
\end{equation}
with $\ell$ as above after enlarging $C_\tau$, since the VC dimension
is at most $C(1+D+p)^2$.
The failure probability is at most $Cn^{-\tau-2}$.
Condition~\eqref{eq:sample-condition} gives
$\PP_{n_f}(W)\ge p_W/2$. The centering identity in
Lemma~\ref{lem:normalized-averages}, with $|\PP h_W|\le p_W$, bounds the
conditional-mean error by $2/p_W$ times the displayed deviation.
Condition~\eqref{eq:sample-condition} makes this deviation, after
the fit and pooling normalizations, a small multiple of $u_k^2$.

Replacing the population fits by the simultaneously estimated ones now
perturbs the normalized tangent matrix $T_vH_T^{-1}$ by $O(e_k/(u(d+u)))$ in a singular
chart, where $d=d(q)$. Its product with the normal bound
$M_N=O_A(u(d+u))$ costs $O_A(e_k)$, and the normal perturbation costs
$O(e_k/\nu)$. The tangent Gram normalized on both sides by the population matrix $H_T^{-1}$ stays bounded
above and below when $e_k/\nu$ is a sufficiently small multiple of $u^2$.
Apply \eqref{eq:regression-residual-stability} with $E$ and $F$
the errors in the stacked tangent and normal columns, and $V=N-TG$.
The preceding bounds give
$\|H_T(\widehat G-G)\|=O(e_k/\nu)$, the additional term in
\eqref{eq:annular-score}.
For $d(q)\le Au$, the error after division by the weakest singular
value is at most $C_A e_k/(\nu u)$, which preserves the strict inequality in Lemma~\ref{lem:quadratic-update} by the third requirement on $C_{\rm samp}$. On a regular patch the weakest
singular value is bounded below; this gives the sampling contribution
used in \eqref{eq:regular-memory}. The two events for pool sampling and
for the common fit fold are intersected, without treating stored slopes as
independent. Intersecting with the pilot event and summing the
conditional failure probabilities over $O(\log n)$ rounds gives the event
of Proposition~\ref{prop:finite}.

\subsubsection{Preservation of alignment over successive rounds}\label{app:propagation}
The constants can be chosen in the following order. The lower bound \eqref{eq:rank-opening} and the fixed scale bounds of \Aref{3} give $\eta,H_0$ in
Lemma~\ref{lem:quadratic-update}; the regular constants in
\eqref{eq:regular-memory} determine the stretch condition
\eqref{eq:bend-margin}. Choose $A$
large enough that dividing the term $Cu^2$ in \eqref{eq:annular-score}
by $u(d+u)$ for $d\ge Au$ costs at most $a_\rho\eta/8$. This coefficient uses the uniform chart bounds and the annular signal lower
bound with $A$ above a fixed conditioning threshold; it does not grow
with $A$. Only the local comparison terms carry the notation $C_A$; the coefficient of $u^2$ used to choose $A$ is fixed. Next take
$\varepsilon_{\rm sing}$ small enough to absorb both
the annular term $C\varepsilon_{\rm sing}\delta$ and the comparison error
$C_A\varepsilon_{\rm sing}$ for $d\le Au$. Fix $K_0$ for the pilot and its first regular update and $c_\rho$ for the kernel lower bound in the prediction step. Then choose $\varepsilon_0$ in \Aref{5} so the initial
width is a small enough fraction of $\sigma_N$, $r_{\rm sing}$ and
$r_{\rm cyl}$, with
\begin{equation}\label{eq:refinement-resolution}
 c_\rho b_0\le\min\{R/8,r_{\rm cyl}/64\},\qquad
 \max\{C_A,Cc_\rho\}\,b_0<\min\{\sigma_N/8,r_{\rm cyl}/16\}.
\end{equation}
The constants here bound the actual and frozen support radii. These fixed
choices give Lemma~\ref{lem:cylinder-localization} at every refinement and
prediction width, separate all compared supports from their chart boundaries,
and cover every use of \eqref{eq:fiber-variation}.
Finally the third requirement on $C_{\rm samp}$ in
Appendix~\ref{app:n0} makes
$e_k/\nu$ a sufficiently small fixed multiple of $u_k^2$.

The pilot then starts with the required incoming angle. At $d\le Au_k$,
the strict bound in Lemma~\ref{lem:quadratic-update}, the chart comparison
and the sampling event imply that the outgoing physical angle is less
than $a_\rho\eta b_k/R$. The complementary graph block is already $O_A(u_k^2)$ by the
same chart comparison. On $Au_k\le d\le r_{\rm sing}/L$, the same bound
follows from \eqref{eq:annular-score} by the choice of $A$ and
$\varepsilon_{\rm sing}$. On $d\ge r_{\rm sing}/L$, divide \eqref{eq:regular-memory}
by the next width. The sum of its two predecessor coefficients is bounded
by the stricter squared-width sum in \eqref{eq:bend-margin}; the remaining terms divided by $u_k$ are $O(u_k)$. Its first step is covered by the initial slack in \eqref{eq:rho0-derived}.
For an empty singular set use this regular estimate everywhere, with fixed
$0<\eta\le1/16$ and $H_0\ge3$ as required by Lemma~\ref{lem:cylinder-localization}.
These estimates leave slack for the cubic graph-to-angle conversion.
Since $b_{k+1}/b_k\ge a_\rho$, each window contains the normal fibers
specified in Proposition~\ref{prop:step} at every round. Induct from the pilot event. On its intersection with
the concentration events through step $k-1$, the induction hypothesis
supplies the incoming angular bounds. Conditional on these earlier folds,
all three concentration bounds
\eqref{eq:fit-gram-concentration}--\eqref{eq:pool-concentration} hold at step $k$
with probability at least $1-Cn^{-\tau-2}$; on this event, the
preceding estimates give the updated angle bound.
Including the pilot failure, the conclusions hold simultaneously through
step $k$ with probability at least $1-C(k+1)n^{-\tau-2}$, proving
Proposition~\ref{prop:step}. Taking $k=k_*=O(\log n)$ gives total failure
probability at most $Cn^{-\tau-1}$ and the first claim of
Proposition~\ref{prop:finite}.

For the sharper bound \eqref{eq:finite-envelope}, take the supremum over targets of
$(d+u_k)\|\Gamma_k\|/u_k^2$. For $d\le Au_k$ it is
bounded using Proposition~\ref{prop:step}. In the annular region a predecessor is at
base distance $O(b_k)$, so its coefficient after division by squared
width is at most $C\varepsilon_{\rm sing}(1+C/A)/a_\rho^2<1$.
For a regular target, the sum in \eqref{eq:bend-margin} acquires a factor at most
$1+C b_0/r_{\rm sing}$ over the two rounds and remains below one.
The remaining terms are uniformly bounded. Thus the supremum is at most the maximum of the initial two bounds, the bound for $d\le Au_k$, and a bound for the remaining terms divided by one minus the larger sum of coefficients multiplying previous errors. The resulting constant may depend on $L/b_0$. Returning to physical
angles gives \eqref{eq:finite-envelope} and proves
Proposition~\ref{prop:finite}.

\subsection{Phase 3: proof of Proposition~\ref{prop:readout}}\label{app:readout}
Fix the pilot and refinement data and the pilot split on the event of Proposition~\ref{prop:finite} and set $u=b_*/\sigma_N$.
Fix a query $x=q+Rz_x$, with $z_x$ in units of $R$ as in Appendix~\ref{app:phase2}. Apply Lemma~\ref{lem:cylinder-localization}
with the projector evaluated at the predictor, for both the true and estimated kernels;
the bound $R\|\widehat P_{k_*}(x)-P_q\|<a_\rho\eta b_{k_*}$
in Proposition~\ref{prop:finite} and
$b_*=c_\rho b_{k_*}$ give the hypothesis of the lemma. Both integrals are thereby restricted to the same
$\mathcal U_q=B_\M(q,r_{\rm cyl}/4)$, and their supports have base radius
$O(b_*)$. All comparisons below are between these localized measures.
First use the true projector $P_{\Pi_\M X_i}$ at each predictor $X_i$ in \eqref{eq:readout}.
If the query is $q+Rz_x$ and the predictor base has displacement $w$ from $q$,
its tangent projection of the query-minus-predictor vector, transported
back to $T_q\M$, is
\[
 -(I-S_{q,Rz_x})w+O(\|w\|^2/\reach_\M)
\]
in physical units: the predictor's own offset drops out exactly; differentiating
the moving tangent projection gives the $S_{q,Rz_x}w$ term. For a fixed
parallel normal offset $z$ the Euclidean squared distance has constant term
$\|z-Rz_x\|^2$ and no term linear in $w$. The leading oracle kernel is
therefore even in $w$ for each $z$. Its support is strictly inside
$\mathcal U_q$ by the lower singular-value bound for $I-S_{q,Rz_x}$ and
the fixed resolution choice, so the base restriction does not truncate
this even section and the frozen average of the linear link term vanishes.
On the $O(b_*)$ support, chart, density,
kernel and fiber-law changes, in the same base and normal coordinates,
have total variation $O(u)$ times the kernel mass.
Lemma~\ref{lem:normalized-averages}, applied to the $O(u)$ response
oscillation, gives bias $O(u^2)$ in units $\sigma_{f,+}\sigma_N$.
The $C^{1,1}$ remainder has the same order. No cylinder or chart boundary
contributes to this comparison.

Lemma~\ref{lem:cylinder-localization} bounds both kernels below by
$1/16$ on every normal fiber above $B_\M(q,b_*/16)$ and places their supports
above $B_\M(q,8b_*)$. On that ball,
$d(\Pi_\M X_i)+u\asymp d(q)+u$, and replacing $P_{\Pi_\M X_i}$ changes
the tangent argument by at most
$C R\|\widehat P_{k_*}(X_i)-P_{\Pi_\M X_i}\|$. Lipschitz continuity of
$t\mapsto(1-a-t^2)_+^2$ uniformly in $a\ge0$ and
\eqref{eq:finite-envelope} bound the weight difference by $Cu/(d(q)+u)$.
Both masses are $\asymp\pi_-b_*^m$. Lemma~\ref{lem:normalized-averages}
therefore bounds the bias change by $Cu^2/(d(q)+u)$.
Near a point where $Df=0$, the response oscillation is
$O(u(d(q)+u))$, giving bias $O(u^2)$ instead.
Outside the singular charts the bias is also $O(u^2)$.
Partial rank deficiency requires $m\ge2$; applying
Lemma~\ref{lem:distance-integrals} with $\alpha=2$ to those points gives
integrated squared bias $O(u^4)$, with an extra $1+|\log u|$ only for
$m=2$ when such points exist.

For a fixed query, Bernstein's inequality gives a prediction denominator
$\gtrsim n_{\rm pred}\pi_-b_*^m$ except with probability
$\exp(-c' n_{\rm pred}\pi_-b_*^m)$. Center the responses at the population
weighted mean and use this denominator in the ratio. Independence of
the prediction observations and $0\le K^2\le K$ give conditional mean
square fluctuation, in normalized units, at most
\[
 \frac{C\,\{p(\sigma_\zeta/(\sigma_{f,+}\sigma_N))^2+u^2\}}
      {n_{\rm pred}\pi_- b_*^m}.
\]
On denominator failure, including the fallback to the mean of the prediction sample,
apply Lemma~\ref{lem:convex-risk} conditional on the pilot and refinement data, query and prediction locations, with $\Delta=\operatorname{diam}(f(\M))$
and $v^2=Cp\sigma_\zeta^2$. Its normalized risk contribution is at most
\begin{equation}\label{eq:denominator-risk}
 C\exp(-c'n_{\rm pred}\pi_-b_*^m)\,
 \frac{\operatorname{diam}(f(\M))^2+p\sigma_\zeta^2}
      {(\sigma_{f,+}\sigma_N)^2}.
\end{equation}

\subsection{Stopping and the sample threshold}\label{app:n0}
The rule \eqref{eq:sample-condition} of Appendix~\ref{app:defs} for
the number of steps makes the polynomial sample requirement explicit.
By \eqref{eq:rho0-derived} the actual terminal width for a run of $j$ steps
lies between $\underline b_j$ and $(\max\mathcal R/\min\mathcal R)\underline b_j$.
The width decreases within a run because
$a_\rho a_h<1$ for $D>m$. At each preceding round, the left-hand side of \eqref{eq:sample-condition} with the actual width $b_k$ is no smaller. The multiplier $C_{\rm samp}$ is taken large enough for
four requirements: the Gram perturbations in
Appendix~\ref{app:concentration} are at most half the smallest eigenvalue;
the empirical window masses there are at least half their population
values; substitution into the definition of $e_k$ makes $e_k/\nu$ at most
the fixed multiple of $u_k^2$ required in Appendix~\ref{app:propagation};
and the prediction count satisfies the bound at the end of this
subsection. The third requirement also makes the pool deviation in
Appendix~\ref{app:concentration} a small multiple of $u_k^2$, since
$\beta_k\le e_k/\nu$. The multiplier $C_{\rm pil}$ in \eqref{eq:pilotN} is taken large
enough for the Bernstein bounds of Appendix~\ref{app:pilot}.

It suffices that \eqref{eq:sample-condition} hold for $j=1$, and that the
smaller of the two pilot folds satisfy
\begin{equation}\label{eq:pilotN}
\begin{aligned}
 \min\{|\Sanc|,|\Sdes|\}\,p_0\ \ge\ C_{\rm pil}
 &\left[D+\log\frac{2+n|\mathcal R|}{p_0}\right]\\[-1pt]
 &{}\times
 \left(1+\frac{\max\mathcal R}{w_Y}
       +\frac{(\max\mathcal R)^2\reach_\M}{\sigma_N^2 w_Y}\right)^2 .
\end{aligned}
\end{equation}
Since each pilot fold has at least $n_f$ observations, both
requirements are conditions on $n$ alone. Both hold for every
$n\ge n_0$, with the threshold $n_0=C\Theta\log^2(2+\Theta)$ of
\eqref{eq:n0}, when the constant there is large. Indeed,
$\underline b_1/\sigma_N$ is a fixed multiple of $w_Y/\sigma_N$ and
$n_f(1)\ge n/24$, so \eqref{eq:sample-condition} with $j=1$ follows from
$n\ge C\Theta\log(2n)$. Since $\pi_-\sigma_N^m$ and
$\underline b_0/\sigma_N$ are bounded, the rule for the number of steps
gives $k_*\le2\log n+C$, hence $n_f\ge n/(C\log(2n))$. Because
$D+\log((2+n|\mathcal R|)/p_0)\le(D+\log(2|\mathcal R|/p_0))(1+\log(2n))$,
\eqref{eq:pilotN} with $n_f$ in place of
$\min\{|\Sanc|,|\Sdes|\}$ follows from $n\ge C\Theta\log^2(2n)$. Both
inequalities hold for $n\ge C\Theta\log^2(2+\Theta)$. The threshold
depends on $\tau$ through the multipliers $C_{\rm samp}$ and
$C_{\rm pil}$. Up to logarithmic factors it is polynomial in $D+p$, in
the inverse local masses $(\pi_-\sigma_N^m)^{-1}$ and $p_0^{-1}$, and in the
scale ratios $R/\sigma_N$, $\sigma_N/w_Y$,
$\sigma_\zeta/(\sigma_{f,+}\sigma_N)$, $\max\mathcal R/w_Y$ and
$(\max\mathcal R)^2\reach_\M/(\sigma_N^2w_Y)$.
Since $b_*\ge\underline b_{k_*}$, $\underline b_{k_*}\le\sigma_N$ and
$n_{\rm pred}\ge n_f$, \eqref{eq:sample-condition} gives
$n_{\rm pred}\pi_-b_*^m\ge C\log(2n)$. By the fourth requirement on
$C_{\rm samp}$, the constant here is as large as the prescribed confidence
exponent $\tau$ requires, so the denominator failure
probability on the good training event is at most $n^{-\tau-1}$.
Failure of \eqref{eq:sample-condition} at $k_*+1$, the bounded ratio of
neighboring widths and $n_f(k_*+1)\asymp n_f$ give
\begin{equation}\label{eq:terminal-width}
 \left(\frac{b_*}{\sigma_N}\right)^{m+4}
 \lesssim
 \frac{(1+D+p)^4(1+R/\sigma_N)^{16}
   (1+\sigma_\zeta/(\sigma_{f,+}\sigma_N))^2\log^2(2n)}
      {n\pi_-\sigma_N^m}.
\end{equation}
All displayed dimension costs are polynomial since
$R/\sigma_N\asymp\sqrt{D-m+2}$. The local probability and the initial
resolution remain explicit. The localization requirements of
Lemma~\ref{lem:cylinder-localization} constrain only the fixed initial
resolution in Appendix~\ref{app:propagation}; they introduce no additional
power of $n$ or dimension factor into these sample bounds.

\subsection{Proof of Theorem~\ref{thm:isolated-rank}}\label{app:main}
Fix the pilot and refinement data and the pilot split on the event of
Proposition~\ref{prop:finite}, and put $u=b_{k_*}/\sigma_N$.
Equation~\eqref{eq:finite-envelope} and
Lemma~\ref{lem:distance-integrals}, with $S=\Sigma_{\rm sing}$ and
$\alpha=1$, give conditional mean tangent-space error
$C(\sigma_N/R)u^2[1+\ind_{\{m=1\}}|\log u|]$.
On the complementary event the projector error is at most one, contributing
at most $Cn^{-\tau-1}$ to its expectation. The terminal width bound
\eqref{eq:terminal-width} applies since $b_*=c_\rho b_{k_*}$ in the theorem;
it gives the stated rate for $\mathcal E_{\rm tan}$.

For prediction, \eqref{eq:sample-condition} bounds the variance in
\eqref{eq:prediction-bound} by a constant times $(b_*/\sigma_N)^4$ on the
good training event. A point with partially rank-deficient $Df$ in dimension $m=2$ contributes the logarithm
in the integrated bias. Equation~\eqref{eq:terminal-width} then gives the
first term of \eqref{eq:rank-profile-risk}.

Training failure and conditional denominator failure each have
probability at most $Cn^{-\tau-1}$ by Proposition~\ref{prop:finite} and
Appendix~\ref{app:n0}. Lemma~\ref{lem:convex-risk} gives the
exceptional-event term of \eqref{eq:rank-profile-risk}.

If $\M$ is connected, place points $\sigma_N/8$ apart along a minimizing
geodesic. Their disjoint intrinsic balls of radius $\sigma_N/32$ each have
probability at least $C^{-1}\pi_-\sigma_N^m$ by \Aref{1}--\Aref{2}.
The length of the geodesic is therefore at most
$C\sigma_N[1+(\pi_-\sigma_N^m)^{-1}]$. Integrating $Df$ along it yields
$\operatorname{diam}(f(\M))/(\sigma_{f,+}\sigma_N)
\le C[1+(\pi_-\sigma_N^m)^{-1}]$.

\subsection{Gaussian response noise}
\label{app:gaussian-corollary}

\begin{corollary}[Gaussian response noise]\label{cor:gaussian-noise}
Assume \Aref{1}--\Aref{4} and the spatial resolution condition in
\eqref{eq:A5}, and let the response noises be independent
$N(0,\overline{\sigma}_\zeta^2I_p)$ vectors, independent of the predictors, with
$\overline{\sigma}_\zeta>0$. Use the estimator, scales and sample-size requirements of
Theorem~\ref{thm:isolated-rank} with $\sigma_\zeta=2\overline{\sigma}_\zeta$.
Let $N_{\rm pil}=|\Sanc|+|\Sdes|$ be the total number of observations in
the pilot anchor and design folds (Appendix~\ref{app:defs}). If
\begin{equation}\label{eq:gaussian-resolution}
 \delta_Y\ge8\overline{\sigma}_\zeta\left(\sqrt p+
                  \sqrt{2\log(N_{\rm pil}n^{\tau+1})}\right),
\end{equation}
then \eqref{eq:tan-risk} and \eqref{eq:rank-profile-risk} hold, with possibly larger constants.
\end{corollary}

\begin{remark}[Fixed Gaussian variance and response resolution]
\label{rem:gaussian-fixed-resolution}
With $p$, $\overline{\sigma}_\zeta>0$ and $\delta_Y$ fixed, the
probability that every pilot noise vector has norm at most $\delta_Y/8$
tends to zero as $N_{\rm pil}\to\infty$. At this variance and resolution,
condition \eqref{eq:gaussian-resolution} holds only over a finite range
of sample sizes.
\end{remark}

\paragraph{Proof of Corollary~\ref{cor:gaussian-noise}.}
Set $\epsilon=n^{-\tau-1}$ and let
$E_G=\{\|\zeta_i\|\le\delta_Y/8\text{ for all }
i\in\Sanc\cup\Sdes\}$. For $G\sim N(0,I_p)$, the Gaussian norm bound
$\PP\{\|G\|>\sqrt p+\sqrt{2t}\}\le e^{-t}$, $t\ge0$, and a union
bound give, by \eqref{eq:gaussian-resolution},
\[
 \PP(E_G^c)\le N_{\rm pil}
 \exp\left[-\frac12\left(\frac{\delta_Y}{8\overline{\sigma}_\zeta}-\sqrt p\right)^2\right]
 \le\epsilon.
\]
The union of the pilot blocks is reserved before sampling, so $E_G$ is
independent of their random anchor/design split. Conditional on $E_G$,
the pilot noises remain independent of one another and of all predictors,
and each has its Gaussian law conditioned on a centered ball. These laws
are centered by symmetry. Radial truncation cannot increase
$\E\exp(\langle v,\zeta_i\rangle^2/(4\overline{\sigma}_\zeta^2))\le\sqrt2$ for a unit
$v$: Gaussian radius and direction are independent, and the integrand
is nondecreasing in the radius for each direction. Thus their directional
$\psi_2$ norms are at most $2\overline{\sigma}_\zeta$. The remaining folds retain their
original independent Gaussian laws. All hypotheses of
Theorem~\ref{thm:isolated-rank} therefore hold conditionally with the same
geometric and sampling constants; in particular
$2\overline{\sigma}_\zeta/\delta_Y$ is bounded by
\eqref{eq:gaussian-resolution}.

On $E_G^c$, the projector error is at most one, and
Lemma~\ref{lem:convex-risk} bounds the prediction estimator's conditional squared risk
by $\operatorname{diam}(f(\M))^2+p\overline{\sigma}_\zeta^2$:
$E_G$ and the weights are independent of the prediction noises.
Since $E_G$ depends only on pilot noises, splitting each expectation
over $E_G$ and $E_G^c$ adds at most $\epsilon$ to the mean tangent error
and at most
\[
 \epsilon\,\frac{\operatorname{diam}(f(\M))^2+p\overline{\sigma}_\zeta^2}
                    {(\sigma_{f,+}\sigma_N)^2}
\]
to the normalized prediction risk. Since $\tau+1>2/(m+4)$, the first
term is absorbed into the stated tangent rate. With
$\epsilon=n^{-\tau-1}$ and $\sigma_\zeta=2\overline{\sigma}_\zeta$, the second is
absorbed into the exceptional-event term of \eqref{eq:rank-profile-risk}
by enlarging its constant.

\subsection{Examples and response dimension}\label{app:cusp-example}

\begin{example}[A noisy annulus with rank-deficient $Df$]\label{ex:noisy-cusps}
Let $m=1$, $D=p=2$, $\M=\{q(\theta)=(\cos\theta,\sin\theta):
0\le\theta<2\pi\}$, and $\sigma=c/\sqrt3$. Define
\begin{equation}\label{eq:cusp-link-example}
 f(q(\theta))=
 \left(\arcsin\frac{\cos\theta}{\sqrt{1+\sigma^2}},
       \frac{\sin^3\theta}{\sigma^2+\sin^2\theta}\right).
\end{equation}
Take independent $\Theta\sim\operatorname{Unif}[0,2\pi)$,
$U\sim\operatorname{Unif}[-c,c]$, and
$\zeta\sim\operatorname{Unif}B_{\R^2}(0,\delta_Y/16)$, and observe
\[
 X=(1+U)q(\Theta),\qquad Y=f(q(\Theta))+\zeta.
\]
The predictor law fills the annulus of radius $c$ around the circle, and
both noise variables have positive variance. The link is injective and
smooth, with derivative of rank zero exactly at $q(0)$ and $q(\pi)$.
For the strictly positive choices of $c,w_Y,\delta_Y$ specified in
the verification below, all assumptions of
Theorem~\ref{thm:isolated-rank} hold. In particular the theorem yields
mean tangent error $\widetilde O(n^{-2/5})$ and squared prediction error
$\widetilde O(n^{-4/5})$ for this model.
\end{example}

\paragraph{Verification of Example~\ref{ex:noisy-cusps}.}
All constants in this verification are uniform for $0<c\le1/16$.
The circle has reach one and bounded $C^3$ graph charts, and its base
density is constant, $\pi=\pi_-=1/(2\pi)$. The normal law is uniform on the
full interval $[-c,c]$, so
\[
 \sigma_N=\sigma=c/\sqrt3,\qquad
 \E U^4=c^4/5=(9/5)\sigma^4,\qquad
 \Var g(U)\le\sigma^2\operatorname{Lip}(g)^2.
\]
The last inequality follows by applying the Lipschitz bound to two
independent copies of $U$. Polar transport carries one radial normal
line to the next and preserves this law; \eqref{eq:fiber-variation}
therefore has left side zero. We may take $C_N=2$.

Write $s=\sin\theta$. Differentiation of \eqref{eq:cusp-link-example}
gives, in the unit arclength coordinate,
\[
 f'(\theta)=\left(-\frac{s}{\sqrt{\sigma^2+s^2}},
 \cos\theta\,\frac{s^2(3\sigma^2+s^2)}{(\sigma^2+s^2)^2}\right).
\]
The second scalar fraction has maximum $9/8$, so $\|f'\|\le2$.
The first coordinate vanishes only at $0,\pi$; at those points the second
coordinate also vanishes and
$f''(0)=(-1/\sigma,0)$, $f''(\pi)=(1/\sigma,0)$.
On each open semicircle the first component of $f$ is strictly monotone,
and the sign of the second distinguishes the semicircles. Thus $f$ is
injective. Direct differentiation gives
$\|f''\|\le10/\sigma$ and, within distance $\sigma/4$ of a point where $Df$ is rank deficient,
$\|f'''\|\le100/\sigma^2$.
Take $\sigma_{f,+}=2$, $r_{\rm sing}=\sigma/8$ and
$c_{\rm op}=1/64$.
For distance $t\le\sigma/8$ from a point where $Df$ is rank deficient, the first derivative component
has magnitude at least $t/(2\sigma)$; outside distance $\sigma/16$
its magnitude is at least $1/32$. These inequalities give
\eqref{eq:rank-opening} with $c_{\rm op}=1/64$. The normalized Taylor modulus can be taken
as $\omega(t)=100t$. In particular all normalized link bounds are
independent of $c$, and at either point where $Df$ is rank deficient $\|B\|/g_B=1$.

The geometric margins can be fixed uniformly before choosing the response resolution.
For $m=1$, $C_N=2$ and $\kappa_f=1$,
Remark~\ref{rem:regular-constant} gives a finite bound
$\overline C_{\rm reg}$ valid for every $0<c\le1/4$.
Use the uniform chart, design and link bounds just verified to choose
$\eta,H_0,A,\varepsilon_{\rm sing}$ as in
Appendix~\ref{app:propagation}, using $1/4$ as an upper bound on the tube ratio. Fix any
\begin{equation}\label{eq:cusp-tube-choice}
 0<c\le\min\left\{\frac1{16},\frac{\varepsilon_{\rm sing}}2,
       \frac{a_\rho^4}{8(1+\overline C_{\rm reg})}\right\}.
\end{equation}
This is a nonempty interval, since $\overline C_{\rm reg}$ and the
margins used to choose its endpoints are independent of $c$.
The shape operator in a unit radial direction has
magnitude one. Hence the left side of \eqref{eq:opening-margin} is at
most $c\|Df(q)\|/(1-c)\le\varepsilon_{\rm sing}\lambda(q)$, since
$m=1$. Moreover $\omega_{\rm bend}=\sigma=c/\sqrt3$, and
\eqref{eq:cusp-tube-choice} gives \eqref{eq:bend-margin}.
This verifies both curvature restrictions.

Choose a finite scan grid in $[2.1c,2.2c]$, containing both endpoints,
and set $C_{\rm pkt}=2$. After fixing $K_0$ and the multiplier $\gamma_{\rm cyl}$ of the local cylinder,
choose $\epsilon>0$ sufficiently small for \Aref{5} and
Appendix~\ref{app:propagation}, with additionally
$\epsilon<1/(100C_{\rm pkt})$. Put
\begin{equation}\label{eq:cusp-resolution-choice}
 \begin{gathered}
 w_Y=w=\epsilon\sigma,\qquad
 \delta_Y=2^{-13}w^3/\sigma^2,\qquad
 p_0=\pi_-\delta_Y/4.
 \end{gathered}
\end{equation}
All scan radii satisfy the strict bounds in \Aref{4}.
Since $\|Df\|\le2$, every point in the intrinsic interval of radius $\delta_Y/8$ about a base point satisfies the two inequalities in \eqref{eq:packet-core}. Its probability is $\pi_-\delta_Y/4=p_0$, proving \eqref{eq:packet-core}.
For the diameter condition, any pair visible to the scan has intrinsic
distance $d\le8\sigma$, by the circle chord formula and $c\le1/16$.
For these pairs we claim
\begin{equation}\label{eq:cusp-inverse-bound}
 \|f(q)-f(q')\|\ge d^3/(1024\sigma^2).
\end{equation}
If the shorter arc avoids the $\sigma$-neighborhoods of the two points where $Df$ is rank deficient,
the first response coordinate has derivative of fixed sign and
magnitude at least $1/2$, which proves the claim. Otherwise both
endpoints are within $9\sigma$ of the same point where $Df$ is rank deficient. On the same side of
that point, integration of $|f'_1(t)|\ge |t|/(10\sigma)$ gives
$|f_1(t)-f_1(t')|\ge d^2/(20\sigma)\ge d^3/(160\sigma^2)$.
On opposite sides, $|f_2(t)|\ge |t|^3/(100\sigma^2)$ and the two signs
are opposite; the inequality $a^3+b^3\ge(a+b)^3/4$ gives
$d^3/(400\sigma^2)$. These establish \eqref{eq:cusp-inverse-bound}.
Thus response distance at most $2\delta_Y=2^{-12}w^3/\sigma^2$ implies
$d\le w\le C_{\rm pkt}w$, proving \eqref{eq:visible-packets}.

Finally the independent response noise in Example~\ref{ex:noisy-cusps}
is centered, bounded by $\delta_Y/16$, and sub-Gaussian, with positive
covariance $(\delta_Y/16)^2I_2/4$. It satisfies \Aref{5}.
At the chosen resolution, \eqref{eq:cusp-resolution-choice} gives
$p_0=2^{-15}\epsilon^2\pi_-w>0$. In units of the local mass $\pi_-w$,
the lower bound $p_0$ in \eqref{eq:packet-core} decreases as the resolution is refined. It enters the pilot sample-size requirement
\eqref{eq:pilotN}; the angular and width multipliers are already fixed.
For this fixed model both sample-size conditions hold for all sufficiently
large $n$.

\begin{remark}[Which rank deficiencies of $Df$ are covered]\label{rem:rank-loss-scope}
Theorem~\ref{thm:isolated-rank} allows both $0<\operatorname{rank}Df(q_0)<m$
and $Df(q_0)=0$, provided the points $q_0\in\M$ where $Df(q_0)$ is rank
deficient are isolated, satisfy the lower bound \eqref{eq:rank-opening},
and satisfy the response-packet condition \Aref{4}, together with the
remaining hypotheses.

For a scalar $C^2$ link on a curve, a nondegenerate extremum
$f'(0)=0$, $f''(0)\ne0$ satisfies \eqref{eq:rank-opening} but has two local
inverse branches with exactly equal responses. At a sufficiently fine
pilot resolution, there are such pairs visible to the scan whose
intrinsic separation exceeds $C_{\rm pkt}w_Y$, contradicting
\eqref{eq:visible-packets} for every positive response-window width.
Thus ordinary scalar minima and maxima, locally of type $f(0)\pm t^2$,
are not covered, including when higher-order terms make $f$ asymmetric.

Other response components can distinguish the branches while preserving
rank deficiency of $Df$. For the local cusp model
$f(t)=(t^2,t^3)$,
\[
 \|Df(t)\|\asymp |t|,\qquad
 \|f(t)-f(s)\|\ge |t-s|^3/4.
\]
The second inequality follows from
$t^2+ts+s^2\ge(t-s)^2/4$. Consequently a sufficiently small fixed
multiple of $w_Y^3$ for $\delta_Y$ gives the local
packet bound, although $Df(0)=0$. Example~\ref{ex:noisy-cusps} verifies
all the hypotheses for this type of rank deficiency of $Df$ on a closed curve with noise.
For $f(t)=(t^2,t)$, $Df(0)$ instead has full rank: a critical point of
one component need not make $Df$ rank deficient.

Nonisolated points where $Df$ is rank deficient and higher-order vanishing of its smallest singular value are excluded by \Aref{3}; for example, the injective local map
$t\mapsto(t^3,t^4)$ has $\|Df(t)\|\asymp |t|^2$ and fails the linear
condition \eqref{eq:rank-opening}.
\end{remark}

\begin{remark}[The equality case $p=m$]\label{rem:response-dimension}
Theorem~\ref{thm:isolated-rank} includes $p=m$ formally, but the other
assumptions impose the following obstructions. For $p=m\ge2$, put
$V=f(\Sigma_{\rm sing})$, a finite set by \Aref{3}. The inverse function
theorem and compactness imply that
$f(\M)\setminus V$ is nonempty and both open and closed in
$\R^m\setminus V$. Since the latter is connected, the two sets must
coincide, contradicting boundedness of $f(\M)$. Thus this case is
vacuous under \Aref{1} and \Aref{3}.
For $p=m=1$, topology permits finitely many critical points, but a scalar
link on a closed curve must have extrema, which \Aref{3} makes
nondegenerate. Their two local inverse branches have equal responses and
violate \Aref{4} at sufficiently fine pilot resolution
(Remark~\ref{rem:rank-loss-scope}), so the scalar equality case is also
vacuous in that regime. This conclusion depends on the fixed resolution
in \Aref{4}, not on topology alone. Both arguments use the absence of
boundary in \Aref{1}.
\end{remark}

\section{Details on the setup of the numerical experiments}\label{app:numerics-details}
\subsection{The curve experiments}\label{app:helix}

\paragraph{Sample sizes and tuning.}
In order of dimension $D=18,24,30$, the single-index experiment uses
$11$, $9$ and $8$ sample sizes, with $n/10^5$ in $[0.8,3.5]$, $[1.2,4]$
and $[1.6,4]$, respectively. The Meyer helix uses $10$, $9$ and $11$
sample sizes with $n/10^5$ in $[0.8,3]$, $[1,3.5]$ and $[1,4]$.
For $k_*=\lceil2\log n+k_0\rceil$, the offsets $k_0$ are $(-10,-10,-7)$
for the segment and $(-4,-4,-1)$ for the helix. At each dimension, the three choices of the prediction multiplier
$c_\rho$ are spaced by a factor of $2^{1/3}$. The middle values at the
three dimensions are
$(2^{-5},2^{-14/3},2^{-14/3})$ for the segment and $(1,2^{2/3},2^{4/3})$
for the helix.

\paragraph{Meyer helix.}
We use the construction of \citet[Appendix~C]{WuMaggioniNSVM}.
The Meyer helix combines localized envelopes with oscillations whose
frequencies and phases vary across coordinates.
Set $a=10$, $\ell_{D,k}=(1+0.3\cos(ak))/D$,
$\ell'_{D,k}=(1+0.3\sin(ak))/D$, and $G(z)=\exp(-z^2/(1+z))$. Then
\begin{equation}\label{eq:meyer-helix}
    x_k(t)=\frac1{\sqrt[4]{2\pi}\sqrt{\ell_{D,k}}}
  \cos\Bigl(ak+\frac{t-k/D}{\ell_{D,k}'}\Bigr)\,
  G\Bigl(\frac{|k/D-t|}{\ell_{D,k}}\Bigr),
  \qquad k=1,\dots,D,\quad t\in[0,1],
\end{equation}
The oscillatory factors produce the curve's winding geometry.
For $u=t-k/D$ and $\ell=\ell_{D,k}$, the envelope satisfies
\[
 \exp\!\left(-\frac{u^2}{\ell(\ell+|u|)}\right)
 =1-\frac{u^2}{\ell^2}+\frac{|u|^3}{\ell^3}+O(u^4).
\]
It is $C^{2,1}$, but its third derivative jumps at $u=0$. The resulting
third-derivative jump of the curve generally has a nonzero normal
component, so the curve need not be a $C^3$ manifold, as required by
\Aref{1}.
By scale invariance of the regression problem, we normalize its reach to
$\sqrt D$; normal displacements of order one per coordinate remain in the
tube, and curvature does not grow with $D$.

After this normalization, the length grows like $D^{1.5}$, the diameter and
the reach like $D^{0.5}$, the curvature decays like $D^{-0.5}$, and the
effective linear dimension grows like $D$. These scaling properties, including the singular-value count defined in Section~\ref{sec:numerics-helix}, are documented in our earlier paper
\citep[Appendix~C.0.2, Eq.~(27) and Fig.~13]{WuMaggioniNSVM}.

\FloatBarrier
\subsection{The bumpy sphere}\label{app:bumpy}

\paragraph{Neighborhood counts and refinement schedule.}
For the surface experiments, Phase~1 selects the $n_Y$ observations whose responses are closest to the anchor response in Euclidean distance. Let $\pi_Y$ be the prescribed fraction of responses to retain, $k_0$ the iteration offset, and $k_h$ the cap on the number of normal-width growth factors. We set
\begin{equation}\label{eq:bumpy-counts}
  n_Y=\min\{n,\max\{234,\lceil\pi_Yn\rceil\}\},\qquad
  k_*=\lceil2\log n+k_0\rceil .
\end{equation}
Phase~2 uses
\begin{equation}\label{eq:bumpy-schedule}
  \rho_0a_\rho^{k-1}\quad\text{and}\quad
  h_0a_h^{\min\{k-1,k_h\}},
  \qquad a_\rho=e^{-1/12},\quad a_h=e^{1/40}.
\end{equation}
Table~\ref{tab:bumpy-iid-params} gives the parameters selected by validation for each link.

\begin{table}[t]
\centering
\small
\setlength{\tabcolsep}{4pt}
\begin{tabular}{@{}lccccccc@{}}
\toprule
 link function & $\pi_Y$ & $k_0$ & $h_{\rm rescale}$ & $\rho_0$ & $k_h$ & $c_\rho$ & $c_h$\\
\midrule
$f^{\rm lin}$ & $0.00152$ & $12$ & $2.45$ & $2.62413$ & $6$ & $1.125$ & $0.316860$\\
$f^{\rm quad}$ & $0.00152$ & $0$ & $7$ & $0.344828$ & $4$ & $1.53960$ & $0.148148$\\
\bottomrule
\end{tabular}
\caption{Constants held fixed in the ten independent repetitions. The response
preimage size is $n_Y=\min\{n,\max\{234,\lceil\pi_Yn\rceil\}\}$, and
the number of Phase~2 iterations is $k_*=\lceil2\log n+k_0\rceil$.
The width and aspect-ratio schedules are given in
\eqref{eq:bumpy-schedule}; prediction uses the normal width
$c_hh_{k_*}$, with a tuning multiplier $c_h$ in place of the value one of
Section~\ref{sec:core-phase3}, and the aspect ratio $c_\rho\rho_{k_*}$.}
\label{tab:bumpy-iid-params}
\end{table}

\paragraph{Radial function.}
Let $\bar u_j,a_j,w_j$ denote the centers, amplitudes and widths of
six broad bumps, and $\hat u_i,A_i,v_i,o_i$ those of $32$ smaller bumps,
with $o_i$ their offsets; the values are specified immediately below.
For $u\in\mathbb S^2$, the radial function in \eqref{eq:bumpy} is
\begin{equation}\label{eq:bumpy-radius}
\begin{aligned}
  r(u)={}&1+0.75\Bigl[0.035\,u_1-0.025\,u_2+0.045\,u_3
     +\sum_{j=1}^{6}a_j\exp\Bigl(\frac{u^\T\bar u_j-1}{w_j^2}\Bigr)\Bigr]\\
  &+\sum_{i=0}^{31}A_i\Bigl[\exp\Bigl(\frac{u^\T\hat u_i-1}{v_i^2}\Bigr)-o_i\Bigr].
\end{aligned}
\end{equation}
The six broad terms have the centers $\bar u_j=\tilde u_j/\|\tilde u_j\|$,
amplitudes $a_j$ and widths $w_j$ of Table~\ref{tab:bumpy-broad}. The $32$
smaller terms have centers on a spiral, widths, amplitudes and offsets
\[
\begin{gathered}
  \hat u_i=\Bigl(\sqrt{1-t_i^2}\cos\phi_i,\ \sqrt{1-t_i^2}\sin\phi_i,\ t_i\Bigr),
  \qquad t_i=1-\frac{2i+1}{32},\qquad \phi_i=i\pi(3-\sqrt5)+0.73,\\
  v_i=0.26+0.055\,\frac{(5i+1)\bmod8}{7},\qquad
  o_i=\frac{v_i^2}2\bigl(1-e^{-2/v_i^2}\bigr),\\
  A_i=0.36\,\frac{\tilde a_i-\bar a}{\max_{0\le k\le31}|\tilde a_k-\bar a|},\qquad
  \tilde a_i=0.72\sin(1.73i+0.21)+0.28\cos(2.91i+0.47),
\end{gathered}
\]
for $i=0,\dots,31$, where $\bar a=32^{-1}\sum_{i=0}^{31}\tilde a_i$.
The offset $o_i$ is the mean of the $i$-th exponential over the sphere,
so all $32$ terms have mean zero. On a fine sphere grid, $r$ ranges from
$0.688$ to $1.401$. Each exponential in \eqref{eq:bumpy-radius} lies in
$(0,1]$, so the coefficient signs give the uniform bound $r(u)\le4.26$ for
all $u$, used below to verify the full rank of $Df^{\rm quad}$.

\begin{table}[!ht]
\centering
\resizebox{\textwidth}{!}{%
\small
\setlength{\tabcolsep}{4pt}%
\begin{tabular}{@{}lcccccc@{}}
\toprule
$j$ & $1$ & $2$ & $3$ & $4$ & $5$ & $6$\\
\midrule
$\tilde u_j$ & $(0.83,-0.22,0.51)$ & $(-0.31,0.91,0.27)$ & $(0.15,-0.62,0.77)$
 & $(-0.76,-0.44,-0.48)$ & $(0.55,0.39,-0.74)$ & $(-0.18,0.10,0.98)$\\
$a_j$ & $0.18$ & $-0.12$ & $0.10$ & $0.075$ & $-0.065$ & $0.055$\\
$w_j$ & $0.36$ & $0.52$ & $0.43$ & $0.61$ & $0.39$ & $0.56$\\
\bottomrule
\end{tabular}}
\caption{The six broad terms in \eqref{eq:bumpy-radius}: unnormalized
centers, amplitudes and widths.}
\label{tab:bumpy-broad}
\end{table}

\paragraph{Scale, reach and normal displacement.}
The numerically estimated reach of the unscaled surface
$\{r(u)\,u:\ u\in\mathbb S^2\}$ is $0.09475149$, and the factor
$c_\M=47.19858183760044$ in \eqref{eq:bumpy} sets it to $\sqrt{20}$. The scaled
surface has radius between $32.5$ and $66.1$. The reach estimate combines a screen of $65{,}536$ sphere points with
continuous searches for the smallest radius of curvature. The local reach
$\reach_\M(Q)$ is the largest radius of a ball around $Q$ in which every point
has a unique nearest point on $\M$, so that
$\reach_\M=\inf_Q\reach_\M(Q)$. It is estimated from $50{,}000$ scrambled Sobol points of the
sphere, using pairwise normal-projection distances and the analytic radii of
curvature; at a new point, the estimate interpolates the $32$ nearest of these
points with weights proportional to the inverse squared distance, and it is
capped by the analytic radius of curvature at that point. Let $V$ be uniform on $\mathbb S^{19}$ and independent of a
$\chi^2$ variable $G$ with $20$ degrees of freedom, conditioned on
$G\le64.8$. The normal displacement is
\[
  W=\frac{0.49\,\reach_\M(Q)}{\sqrt{20}}\,\sqrt G\,V,\qquad
  Z=P_Q^\perp W.
\]

\paragraph{Rotation.}
The rotation in the definition of $\xi(u)$ is
{\small
\[
  O=\begin{pmatrix}
  \phantom{-}0.04400330497343763 & 0.23930491044768210 & -0.96994683822725110\\
  -0.30239263248978127 & 0.92853374787354440 & \phantom{-}0.21536892736838617\\
  \phantom{-}0.95216721481434900 & 0.28382783319388520 & \phantom{-}0.11322259552381468
  \end{pmatrix}.
\]}

\paragraph{Regularity of the link function.}
Here $\sigma_{f,-}=\inf_{Q\in\M}\sigma_m(Df(Q))$ and
$\sigma_{f,+}=\sup_{Q\in\M}\|Df(Q)\|$.
Since
each pair of components in \eqref{eq:bumpy-links} traces a circle of radius
$1/2$, $\|Df(Q)v\|=\frac12\|D(\psi\circ\xi)\,v\|$ for tangent vectors $v$, where
$\psi\circ\xi$ is viewed as a function on $\M$. For $f^{\rm lin}$ the
differential $Df(Q)$ is therefore a multiple of an isometry, with
$\sigma_{f,-}=\sigma_{f,+}=8.75/(2c_\M)\approx0.0927$ and $\sigma_{f,+}/\sigma_{f,-}=1$. For
$f^{\rm quad}$, the derivative matrix of the phase map $\psi$ is $15I+2A(\xi)$, where $A(\xi)$ is
symmetric with zero diagonal and $A_{ij}(\xi)=\xi_l$ for distinct
$i,j,l\in\{1,2,3\}$, so that
$\|A(\xi)\|\le\sqrt2\|\xi\|$; since $\|\xi(u)\|=r(u)\le4.26$ by the bound above, its eigenvalues are at least
$15-2\sqrt2\cdot4.26>2.9$, and $Df^{\rm quad}(Q)$ has full rank. Numerically,
$\sigma_{f,-}\approx0.131$, $\sigma_{f,+}\approx0.187$ and $\sigma_{f,+}/\sigma_{f,-}\approx1.43$
for $f^{\rm quad}$.

\paragraph{Response noise.}
Let $\widehat\Sigma$ be the covariance of the noiseless responses
of the first $\min\{n,10^6\}$ training samples.
Training and validation responses carry Gaussian noise
$\zeta\sim\mathcal N(0,10^{-4}\widehat\Sigma)$.

\paragraph{Phase~1.}
The response-neighbor count and the number of Phase~2 steps are
\eqref{eq:bumpy-counts}.
These observations are sorted by predictor distance to the anchor.
The scan uses $128$ truncation positions, from $3D$ to $\lfloor0.8\,n_Y\rfloor$.
For $D=20$, the lower bound
$234=\max\{9D,\lceil(3D+127)/0.8\rceil\}$ ensures $128$ distinct positions.
For surfaces, candidate subspaces are compared on two halves of the
Phase~1 data. The returned radius is the root-mean-square scale
of the selected covariance. The initial width is
$h_0=h_{\rm rescale}\,r_{\rm med}$, where $r_{\rm med}$ is the median of the
positive returned radii over the retained anchors, floored at $10^{-6}$.

\paragraph{Refinement and prediction.}
The aspect ratio and normal width at step $k$ are \eqref{eq:bumpy-schedule},
with the constants of Table~\ref{tab:bumpy-iid-params}.

\paragraph{Tangent-space error.}
We orthonormalize each estimated basis with two passes of modified
Gram--Schmidt in double precision before computing its tangent-space error.

\paragraph{Selection of the parameters.}
The validation search evaluated $78$ parameter vectors for $f^{\rm lin}$
and $66$ for $f^{\rm quad}$ at the sample sizes $\mathcal A=\{1,5,9\}\times10^k$, $k=3,4,5$. Let
$L_n(\mathbf p)$ be the validation relative mean squared error of the
parameter vector $\mathbf p$ at sample size $n$, and $L_n^*$ the smallest
value over the tested vectors. The selected vector minimizes
\[
  \max_{n\in\mathcal A}\Bigl\{\frac{L_n(\mathbf p)}{L_n^*}-1\Bigr\},
\]
with ties broken by the geometric mean of $L_n(\mathbf p)/L_n^*$ over
$\mathcal A$.

\clearpage
\subsection{The diffusion}\label{app:reaction}

\paragraph{Potential.}
Let $q_1,\dots,q_4$ be the vectors in
Table~\ref{tab:reaction-directions}. The tangential term in
\eqref{eq:reaction-potential} is
\begin{equation}\label{eq:reaction-tangential-potential}
  U_{\rm tan}(u)=-\frac1{10}\sum_{j=1}^4e^{(0.425+0.17j)\,q_j^\T u},
\end{equation}
The first three vectors have length $\sqrt{31}/6\approx0.928$ and are not normalized. With
$z_i=x_{i+3}$, the stiffnesses in \eqref{eq:reaction-potential} are
\[
  \nu_0=\frac{736}{15},\quad
  \nu_1=\dots=\nu_4=\frac{3680}{81},\quad
  \nu_5=\dots=\nu_{10}=\frac{115}3,\quad
  \nu_{11}=\dots=\nu_{17}=\frac{14720}{343},
\]
approximately $49.07$, $45.43$, $38.33$ and $42.92$. The normal coordinates
therefore relax on time scales between $1/\nu_0\approx0.020$ and
$1/\nu_5\approx0.026$.

\begin{table}[h]
\centering
\small
\begin{tabular}{@{}lcccc@{}}
\toprule
$j$ & $1$ & $2$ & $3$ & $4$\\
\midrule
$(q_j)_1$ & $-\sqrt3/4$ & $\sqrt3/2$ & $-\sqrt3/4$ & $0$\\
$(q_j)_2$ & $-3/4$ & $0$ & $3/4$ & $0$\\
$(q_j)_3$ & $-1/3$ & $-1/3$ & $-1/3$ & $1$\\
\bottomrule
\end{tabular}
\caption{The vectors of the tangential potential $U_{\rm tan}$ in
\eqref{eq:reaction-tangential-potential}.}
\label{tab:reaction-directions}
\end{table}

\paragraph{Targets.}
For $u\in\mathbb S^2$, let $\lambda(u)\in(-\pi,\pi]$ and
$\varphi(u)\in[-\pi/2,\pi/2]$ denote longitude and latitude, so
$u=(\cos\varphi\cos\lambda,\cos\varphi\sin\lambda,\sin\varphi)$. For
$x=(ru,z)$ (Section~\ref{sec:numerics-reaction}), the targets are
\[
  A_j=\bigl\{x:\ u\in D_j,\ |r-1|<0.8,\ \|z\|<0.8\bigr\},\qquad j=1,\dots,4,
\]
with the angular parts
\[
\begin{aligned}
  D_j&=\Bigl\{u:\ \Bigl|\lambda(u)-\frac{2\pi(j-2)}3\Bigr|\le\frac\pi6,\
     -\arctan\frac12\le\varphi(u)\le0\Bigr\},\qquad j=1,2,3,\\
  D_4&=\{u:\ u_3>\cos0.2\}.
\end{aligned}
\]

\paragraph{Simulation.}
The paths of \eqref{eq:reaction-dynamics} are simulated by the
Euler--Maruyama scheme with time step $0.002$ and $50{,}000$ steps, which
gives the horizon $100$.

\paragraph{Selection of the parameters.}
Validation at $n=10^6$ selected the constants in two rounds. The first
compared the $24$ combinations of $\pi_Y\in\{0.005,0.010,0.020\}$,
$h_{\rm rescale}\in\{2.45,3.50\}$, $\rho_0\in\{0.5,1\}$ and $k_h\in\{2,4\}$,
with $k_0=-16$ and $c_\rho=c_h=0.25$. The second round kept the selected values
of these four constants and compared the $12$ combinations of
$k_0\in\{-20,-16,-12\}$, $c_\rho\in\{0.25,0.5\}$ and $c_h\in\{0.25,0.5\}$. The
selected constants are
\[
  (\pi_Y,k_0,h_{\rm rescale},\rho_0,k_h,c_\rho,c_h)
  =(0.005,\,-12,\,2.45,\,0.5,\,2,\,0.5,\,0.25),
\]
and are used without change at $n=4\times10^5$, giving $n_Y=2000$
observations per response preimage and $k_*=14$ Phase~2 steps.

\clearpage
\subsection{The coupled Meyer helix surface}\label{app:coupled-helix}

\paragraph{Surface.}
For $d\in\{4,5\}$ and $i=1,\ldots,d$, the envelopes and phases in
\eqref{eq:coupled-surface} are
\[
 c_i^{(d)}=\frac{i-1}{d-1},\qquad
 w_i^{(d)}=\frac{1+0.3\cos(10i)}d,\qquad
 v_i^{(d)}=\frac{1+0.3\sin(10i)}d,
\]
\[
 E_i^{(d)}(u)=
 \frac{\exp\!\left\{-\dfrac{(u-c_i^{(d)})^2}
 {2w_i^{(d)}(2w_i^{(d)}+|u-c_i^{(d)}|)}\right\}}
 {(2\pi)^{1/4}(w_i^{(d)})^{1/4}},\qquad
 \phi_i^{(d)}(u)=10i+\frac{u-c_i^{(d)}}{v_i^{(d)}}.
\]
At an envelope center, writing $w=w_i^{(d)}$ and $c_i=c_i^{(d)}$,
\[
 E_i^{(d)}(c_i+u)=E_i^{(d)}(c_i)
 \left[1-\frac{u^2}{4w^2}+\frac{|u|^3}{8w^3}+O(u^4)\right].
\]
As for the curve in Appendix~\ref{app:helix}, these envelopes are
$C^{2,1}$ and generally fail to give the $C^3$ geometry in \Aref{1}.
The coordinates are ordered first by $i$, then by $j$. The vector $\mu_\M$
is the mean of $G$ on a fixed $100{,}000$-point calibration sample, and
$c_\M=74.10524209977933$. Sampling $(s,t)$ uniformly on $[0,1]^2$ gives
uniform parameter area. The resulting patch has boundary.

\paragraph{Local reach calibration and normal displacement.}
At calibration point $Q$ with analytic $P_Q$, use the minimum of the
curvature radius and tangent--chord quotients
\[
 \frac{\|Q'-Q\|^2}{2\|(I-P_Q)(Q'-Q)\|},
\]
over calibration points $Q'$ at parameter distance at least $0.02$ from
$Q$. In an orthonormal tangent frame, let $\|\II_Q\|_F$ be the
Frobenius norm of the second fundamental form and put
$e_j=(\cos(j\pi/128),\sin(j\pi/128))$, $0\le j<128$.
The curvature radius is estimated by
\[
 \left[\max_{0\le j<128}\|\II_Q(e_j,e_j)\|
       +\frac{\pi}{128}\|\II_Q\|_F\right]^{-1}.
\]
The scaling makes the smallest calibrated value equal to $\sqrt{20}$;
the minimum, median and maximum are $4.472$, $30.448$ and $63.784$.
At a new parameter point, we interpolate the $32$ nearest calibration
values with weights proportional to inverse squared parameter distance,
using distance floor $10^{-12}$, and cap the result by the curvature radius
there. Write $\ell(Q)$ for this numerical local-reach estimate. The
calibration is fixed across all repetitions.

Draw $W\sim\mathcal N(0,I_{20})$ conditional on
$\|W\|\le1.8\sqrt{20}$, independently of $Q$, and set
\begin{equation}\label{eq:coupled-normal-noise}
 Z=\frac{0.49\ell(Q)}{\sqrt{20}}(I-P_Q)W,
 \qquad X=Q+Z.
\end{equation}
Thus $\|Z\|\le0.882\ell(Q)$.

\paragraph{Polynomial link function and noise.}
The normalization constants in \eqref{eq:coupled-poly-link} are
\[
 a_Y=0.31029722321272146,\qquad
 \mu_Y=\begin{pmatrix}-0.058782804805174\\0.013768230206499\\0.916843578667272\end{pmatrix}.
\]
On the fixed calibration sample, the covariance of the normalized response is
\[
 \Sigma_f\simeq
 \begin{pmatrix}
 0.0478891&0.000157187&-0.0283818\\
 0.000157187&0.0480453&0.00918279\\
 -0.0283818&0.00918279&0.9040656
 \end{pmatrix},\qquad \operatorname{tr}(\Sigma_f)=1.
\]
Training and validation responses are $Y=f^{\rm poly}(Q)+\zeta$, where
$\zeta\sim\mathcal N(0,10^{-4}\Sigma_f)$ is independent of $X$; testing
uses the noiseless response. Both zeros of $\mathfrak p$ and all zeros of
$\mathfrak p'(z)=z^6(8z+4.2+1.4\mathrm i)$ lie outside the parameter
rectangle, so the feature differential in \eqref{eq:coupled-poly-link}
has rank two. Inverting $\mathfrak p$ at calibration responses yields
one to four parameter preimages.

\begin{figure}[!htbp]
\centering
\includegraphics[width=\linewidth]{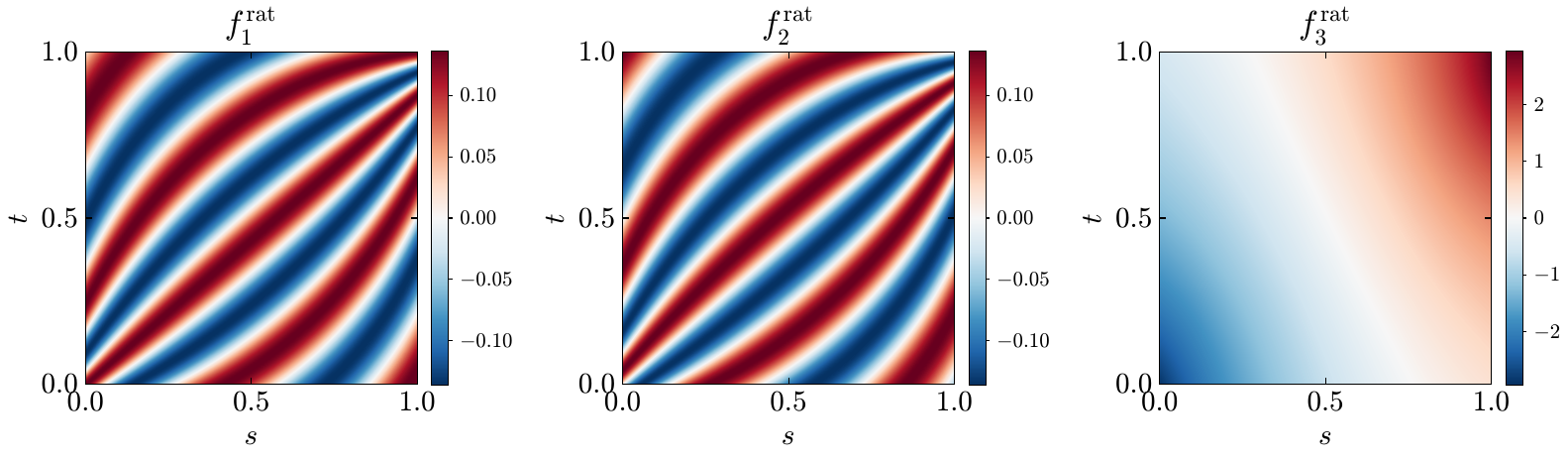}
\caption{Components of $f^{\rm rat}$ on the parameter square. The first two
panels encode the phase of the rational map and share the same color scale;
the third encodes its log magnitude. All panels use the same diverging
color map.}
\label{fig:coupled-rational-response}
\end{figure}

\paragraph{Link function constructed from a rational map.}
For $f^{\rm rat}$, replace $a_Y,\mu_Y$ in \eqref{eq:coupled-poly-link} by
\[
 a_{\rm rat}=0.1353656776435608,\qquad
 \mu_{\rm rat}=\begin{pmatrix}
 -0.00308426830608437\\-0.00465022983878112\\0.8002766360491473
 \end{pmatrix}.
\]
On the same fixed calibration sample, the normalized response has covariance
\[
 \Sigma_{\rm rat}\simeq
 \begin{pmatrix}
 0.00920537&0.0000133033&0.00123311\\
 0.0000133033&0.00911811&0.000516827\\
 0.00123311&0.000516827&0.98167652
 \end{pmatrix},\qquad \operatorname{tr}(\Sigma_{\rm rat})=1.
\]
Training and validation responses are $Y=f^{\rm rat}(Q)+\zeta$, with
$\zeta\sim\mathcal N(0,10^{-4}\Sigma_{\rm rat})$ independent of $X$;
testing uses the noiseless response. On the parameter rectangle, $w$ in \eqref{eq:coupled-rational-map}
has no zero or pole and $w'(z)\ne0$, so the response differential has rank two.
Inverting $w$ at calibration responses gives one to four parameter
preimages. Figure~\ref{fig:coupled-rational-response} shows the response components.

\paragraph{Parameters and validation.}
The implementation uses counts and schedule
\eqref{eq:bumpy-counts}--\eqref{eq:bumpy-schedule} and the constants in
Table~\ref{tab:coupled-parameters}. The initial normal width is
$h_{\rm rescale}$ times the median positive Phase~1 radius; prediction
scales are $c_\rho,c_h$ times the final aspect ratio and normal width.

\begin{table}[!htbp]
\centering
\small
\setlength{\tabcolsep}{4pt}
\begin{tabular}{@{}lccccccc@{}}
\toprule
link function & $\pi_Y$ & $k_0$ & $h_{\rm rescale}$ & $\rho_0$ & $k_h$ & $c_\rho$ & $c_h$\\
\midrule
$f^{\rm poly}$ & $0.001013$ & $-8$ & $3.5$ & $0.525625$ & $4$ & $0.5625$ & $1/3$\\
$f^{\rm rat}$ & $0.001013$ & $-4$ & $1.715$ & $1$ & $6$ & $0.421875$ & $1/2$\\
\bottomrule
\end{tabular}
\caption{Selected parameters for the two link functions on the coupled Meyer helix
surface. Each set is fixed across all sample sizes and repetitions.}
\label{tab:coupled-parameters}
\end{table}
\FloatBarrier

For each link function, the final comparison used six parameter vectors and independent validation sets of $5000$
queries at each $n\in\{1,2,3,4,5,7\}\times\{10^3,10^4,10^5\}$.
For each sample size, we divide a candidate's validation relative mean
squared error by the smallest error among the six candidates. The selected
candidate minimizes the largest ratio across sample sizes, with ties broken
by the geometric mean of the ratios.

\paragraph{Evaluation.}
Prediction error is the mean squared Euclidean error divided by the sum
of the component sample variances of the $5000$ noiseless test responses
(divisor $4999$). Estimated tangent bases are orthonormalized before
computing \eqref{eq:tangent-error-num}. Error bars are sample standard
deviations across ten repetitions (divisor $9$). The slopes are unweighted
least-squares fits of log arithmetic mean error against $\log n$ over the
eight displayed sample sizes $n\ge5\times10^5$.

\section{Statistical learnability and minimax rate}\label{app:learnability}

We show that regression functions $F=f\circ\Pi_\M$, with $\M$ as regular as
in \Aref{1} and $f$ as regular as in \Aref{3}, are learnable at the rate
$n^{-4/(m+4)}$ of Theorem~\ref{thm:isolated-rank}, with constants linear in
$D+p$, and that its exponent in $n$ is minimax for the class below,
at fixed positive noise. The upper bound uses neither \Aref{2} nor
\Aref{4}, of \Aref{5} only the sub-Gaussian noise bound, and no rank condition on $Df$. Its estimator is a
least-squares fit over a finite family whose cardinality is bounded
exponentially in a power of $n$; the construction is an existence argument,
and it uses standard arguments from learning theory, borrowing from standard
complexity arguments for learning in the function class
\citep{GKKW02,YangBarron1999}, and perhaps slightly less standard but rather
well-established arguments for studying the complexity of having to also
estimate the manifold well enough to be able to attain the desired rate for
regression \citep{GenoveseEtAl2012Minimax,AamariLevrard2019}.
We do not know whether an estimator computable in polynomial time attains the same bound under these assumptions alone;
MANOR attains the rate in polynomial time under \Aref{1}--\Aref{5}.

\subsection{The class and the two bounds}\label{app:learnability-class}
Fix integers $m<D$ and $p$, a tube radius $R>0$, a ratio $0<c<1$ and the
unit $\sigma_N=R/\sqrt{D-m+2}$, of the order prescribed in \Aref{2}. Fix a regularity constant
$C_{\mathcal F}\ge1$, a volume bound $v_{\max}$, an ambient radius $\varrho$ and a
response bound $B_f$. Let $\mathfrak M$ be the set of compact
$m$-dimensional $C^3$ submanifolds $\M\subset B_D(\varrho)$ without
boundary, with $\reach_\M\ge R/c$ and
$\operatorname{vol}(\M)\le v_{\max}$, whose graph charts of radius $R/(5c)$,
written as in \Aref{1}, satisfy $\|Dg\|\le1/4$ and
\[
 (R/c)\,\|D^2g\|+(R/c)^2\|D^3g\|\le C_{\mathcal F} .
\]
For $\M\in\mathfrak M$ let $\mathcal L(\M)$ be the set of $C^{1,1}$ links
$f:\M\to\R^p$ with $\sup_\M\|f\|\le B_f$ and $\|Df\|\le\sigma_{f,+}$ whose
differential is Lipschitz in parallel frames with constant at most
$C_{\mathcal F}\sigma_{f,+}/\sigma_N$. A manifold as in \Aref{1} with $R=c\reach_\M$
and a link as in \Aref{3} satisfy these conditions for suitable values of
$C_{\mathcal F}$, $v_{\max}$, $\varrho$ and $B_f$. Write
$\M_R=\{x:\operatorname{dist}(x,\M)\le R\}$ for the tube and
\[
 \mathcal F=\bigl\{(\M,F):\ \M\in\mathfrak M,\ F=f\circ\Pi_\M
 \ \text{on}\ \M_R,\ f\in\mathcal L(\M)\bigr\}.
\]
We assume $\mathfrak M\ne\emptyset$. Then $\varrho\ge R/c$, since a compact
submanifold of $B_D(\varrho)$ has reach at most $\varrho$, and
$v_{\max}\ge c_m(R/c)^m\ge c_m\sigma_N^m$, since a compact manifold without
boundary and with reach at least $R/c$ has volume at least $c_m(R/c)^m$
\citep{NSV}.
We observe $(X_i,Y_i)$, $i\le n$, where $X_1,\dots,X_n$ are independent
copies of a random vector $X\in\M_R$ with otherwise arbitrary law,
$Y_i=F(X_i)+\zeta_i$, and the noise vectors $\zeta_i$ are centered,
independent of one another and of the predictors, and satisfy
$\|\langle v,\zeta_i\rangle\|_{\psi_2}\le\sigma_\zeta$ for unit $v$.
Put
\[
 \Lambda=1+\frac{\sigma_\zeta^2+B_f^2}{(\sigma_{f,+}\sigma_N)^2},
 \qquad
 \ell_n=\log\Bigl(e\,n\max\Bigl\{1,\frac{\varrho}{\sigma_N},
 \frac{B_f}{\sigma_{f,+}\sigma_N}\Bigr\}\Bigr).
\]

\begin{proposition}[Upper bound]\label{prop:learn}
There are an estimator $\widehat F_{\rm LS}$ and a constant $C$ depending
only on $m$, $c$ and $C_{\mathcal F}$ such that, for every $(\M,F)\in\mathcal F$,
every law of $X$ on $\M_R$ and every
$n\ge\Lambda(D+p)\,v_{\max}/\sigma_N^m$,
\[
 \frac{\E\|\widehat F_{\rm LS}(X)-F(X)\|^2}{(\sigma_{f,+}\sigma_N)^2}
 \ \le\ C\left(\frac{\Lambda\,(D+p)\,v_{\max}/\sigma_N^m}{n}\right)^{4/(m+4)}
 \ell_n .
\]
\end{proposition}

\begin{proposition}[Lower bound]\label{prop:lower}
Let $\M_0\in\mathfrak M$ have volume $v_0$, and suppose
$B_f\ge\sigma_{f,+}\sigma_N$. There are constants $c_0,n_1>0$ depending
only on $m$, $c$ and $C_{\mathcal F}$ such that, for
$n\ge n_1(\sigma_\zeta/(\sigma_{f,+}\sigma_N))^2v_0/\sigma_N^m$, every
estimator $\widehat F$ satisfies
\[
 \sup\frac{\E\|\widehat F(X)-F(X)\|^2}{(\sigma_{f,+}\sigma_N)^2}
 \ \ge\ c_0\left(\frac{\sigma_\zeta^2}{(\sigma_{f,+}\sigma_N)^2}\,
 \frac{v_0/\sigma_N^m}{n}\right)^{4/(m+4)},
\]
where the supremum is over $(\M,F)\in\mathcal F$, the laws of $X$ on
$\M_R$ and the admissible noise laws.
\end{proposition}

For fixed positive noise, the two bounds match in their exponents in
$n$ and in the inverse local mass $v_{\max}/\sigma_N^m$, up to the logarithm $\ell_n$,
when $v_0\asymp v_{\max}$. The upper bound contains $\Lambda$ in place of
$\sigma_\zeta^2/(\sigma_{f,+}\sigma_N)^2$; the noise dependence therefore
matches when $\sigma_\zeta^2\gtrsim(\sigma_{f,+}\sigma_N)^2+B_f^2$, but not
uniformly as $\sigma_\zeta\to0$. In particular, the lower bound is vacuous
at zero noise. The factor $p$ is matched in the noise-dominated regime:
the construction of Appendix~\ref{app:learnability-lower}, with independent
bumps in each of the $p$ response coordinates and amplitudes divided by
$\sqrt p$, gives the lower bound with $p\sigma_\zeta^2$ in place of
$\sigma_\zeta^2$, provided the sample threshold in
Proposition~\ref{prop:lower} is also multiplied by $p$. The factor $D^{4/(m+4)}$ remains unmatched:
a fixed manifold cannot detect the ambient dimension, so
Proposition~\ref{prop:lower} does not address the sharpness of the
dependence on $D$. Compared with
Theorem~\ref{thm:isolated-rank}, Proposition~\ref{prop:learn} has the same
exponent in $n$ and in the inverse local mass, since
$\operatorname{vol}(\M)/\sigma_N^m\le1/(\pi_-\sigma_N^m)$. Inside that power
it has the factor $D+p$ where \eqref{eq:terminal-width} has
$(1+D+p)^4(1+R/\sigma_N)^{16}$, and its sample-size requirement is linear in
$D+p$. The response bound enters $\Lambda$ here, whereas in
\eqref{eq:rank-profile-risk} the response diameter enters only the
exceptional-event term.

The same exponent is minimax under \Aref{1}--\Aref{5} conditionally
on the existence of a subclass with fixed strict margins that admits the
small bump perturbations used below on a regular region of volume comparable
to $\operatorname{vol}(\M)$. The perturbations must preserve all five
assumptions with the same constants, and the subclass must admit a fixed
compactly supported response-noise density satisfying \Aref{5}, whose
square root has a square-integrable derivative in the perturbation direction.
Under these conditions, a Hellinger version of Assouad's argument gives the
same exponent with bounded noise. We do not construct such a subclass here;
the unconditional minimax conclusion of this appendix concerns the broader
class $\mathcal F$.

\subsection{Proof of the upper bound}\label{app:learnability-upper}

\paragraph{Reduction to entropy.}
Call a finite family $\mathcal G_\varepsilon$ of maps
$\R^D\to B_p(B_f)$ an $\varepsilon$-net of $\mathcal F$ if every
$(\M,F)\in\mathcal F$ has a $G\in\mathcal G_\varepsilon$ with
$\sup_{x\in\M_R}\|F(x)-G(x)\|\le\varepsilon$. The least-squares estimator
$\widehat F_{\rm LS}$, a minimizer of $\sum_i\|Y_i-G(X_i)\|^2$ over
$G\in\mathcal G_\varepsilon$, satisfies
\begin{equation}\label{eq:ls}
 \E\|\widehat F_{\rm LS}(X)-F(X)\|^2\le
 C\Bigl[\varepsilon^2+(\sigma_\zeta^2+B_f^2)\,
 \frac{1+\log|\mathcal G_\varepsilon|}{n}\Bigr].
\end{equation}
Indeed, let $G^\circ\in\mathcal G_\varepsilon$ be within $\varepsilon$ of $F$.
Conditionally on the predictors, $\sum_i\langle\zeta_i,G(X_i)-G^\circ(X_i)\rangle$
has $\psi_2$ norm at most $C\sigma_\zeta(\sum_i\|G(X_i)-G^\circ(X_i)\|^2)^{1/2}$
for each $G$. A union bound over $\mathcal G_\varepsilon$ in the basic
inequality of least squares bounds the empirical mean squared error of
$\widehat F_{\rm LS}$ by
$C[\varepsilon^2+\sigma_\zeta^2(1+\log|\mathcal G_\varepsilon|+t)/n]$ with
probability at least $1-e^{-t}$. Bernstein's inequality for the variables
$\|G(X)-F(X)\|^2\le4B_f^2$, with a union bound over $\mathcal G_\varepsilon$,
passes to the population mean squared error at the cost of
$CB_f^2(1+\log|\mathcal G_\varepsilon|+t)/n$, and integrating over $t$ gives
\eqref{eq:ls} \citep[compare][Chapter~11]{GKKW02}. As in \citet{YangBarron1999},
the rate is then governed by the metric entropy of $\mathcal F$.
It remains to bound $\log|\mathcal G_\varepsilon|$.

\paragraph{Stability of the nearest-point projection.}
\begin{lemma}\label{lem:stab}
Let $\M,\widetilde\M$ be compact manifolds without boundary and with reach at least $R/c$. Assume every $q\in\M$ has a corresponding $q'\in\widetilde\M$ satisfying
\[
 \|q-q'\|\le\delta,\qquad \|P_q-\widetilde P_{q'}\|\le\theta,
 \qquad \delta\le(1-c)R/(2c),
\]
where $P_q,\widetilde P_{q'}$ project onto the respective tangent spaces.
For every $x\in\M_R$, with $q=\Pi_\M x$ and corresponding $q'$,
\[
 \|\Pi_{\widetilde\M}x-q'\|\le\frac{2(\delta+R\theta)}{1-c}.
\]
If moreover $f\in\mathcal L(\M)$, $\tilde f\in\mathcal L(\widetilde\M)$,
$\|f(q)-\tilde f(q')\|\le\varepsilon_1$ for all corresponding pairs $q,q^{\prime}$, and the right
side of the display is at most $R/(2c)$, then on $\M_R$
\[
 \|f(\Pi_\M x)-\tilde f(\Pi_{\widetilde\M}x)\|
 \le\varepsilon_1+\frac{4\sigma_{f,+}(\delta+R\theta)}{1-c}.
\]
\end{lemma}

\begin{proof}
Since $\|x-q'\|\le R+\delta<R/c$, the point
$\tilde q=\Pi_{\widetilde\M}x$ is defined and
$\|x-\tilde q\|\le R+\delta$. Put $u=\tilde q-q'$. For a set of reach at
least $R/c$, points $a,b$ of the set and a vector $\nu$ normal at $a$,
\[
 \langle\nu,b-a\rangle\le\frac{c}{2R}\,\|\nu\|\,\|b-a\|^2
\]
\citep[Theorems~4.8(7) and~4.18]{Federer1959}. Decompose $x-q'=\nu'+t'$
into its normal and tangential parts at $q'$. Since $P_q(x-q)=0$,
\[
 \|t'\|\le\|(\widetilde P_{q'}-P_q)(x-q)\|+\|q-q'\|\le R\theta+\delta .
\]
Applying the inequality at $q'$ with $\nu'$ and at $\tilde q$ with
$x-\tilde q$,
\[
 \|u\|^2=\langle x-q',u\rangle+\langle x-\tilde q,q'-\tilde q\rangle
 \le\frac{c(R+\delta)}{R}\|u\|^2+\|t'\|\,\|u\| ,
\]
which gives the first claim because $c(R+\delta)/R\le(1+c)/2$. For the
second,
$\|f(q)-\tilde f(\tilde q)\|\le\varepsilon_1+
\|\tilde f(q')-\tilde f(\tilde q)\|$,
and the last term is at most $\sigma_{f,+}$ times the geodesic distance
from $q'$ to $\tilde q$ on $\widetilde\M$, which is at most $2\|u\|$ when
$\|u\|\le R/(2c)$ \citep[Proposition~6.3]{NSV}.
\end{proof}

\paragraph{Degrees of freedom.}
Put $\varepsilon_{\rm rel}=\varepsilon/(\sigma_{f,+}\sigma_N)\le1$. By
Lemma~\ref{lem:stab} it suffices to describe $(\M,f)$ up to
$\delta\asymp\varepsilon/\sigma_{f,+}$ in position, up to
$\theta\asymp\varepsilon/(\sigma_{f,+}R)$ in tangent planes, and up to
$\varepsilon$ in the values of $f$. Cover $\M$ by $N\le Cv_{\max}/R^m$
charts as in \Aref{1}, centered at a maximal $R/(10c)$-separated subset of
$\M$; the bound on $N$ is the volume lower bound for balls in manifolds
with positive reach \citep[Lemma~5.3]{NSV}. Quantize each center to
accuracy $\delta$ and each tangent plane to accuracy $\theta$, and fix for
every quantized plane an orthonormal basis of it and of its orthogonal
complement. Over the quantized plane $\M$ is still a graph, of a map $g$
with $\|Dg\|\le1/4+C\theta$ and the same bounds on $D^2g$ and $D^3g$ up to
constants; in the fixed bases $g$ has values in $\R^{D-m}$, and we write
$\varphi(w)=f(q+w+g(w))$ for the link in the chart. The pair $(\M,f)$ is
then described
by the following data, all in classical classes on Euclidean $m$-balls;
a count of this kind, for the Hausdorff distance and without tracking
the ambient dimension, is in
\citet{GenoveseEtAl2012Minimax}, and the chart model is that of
\citet{AamariLevrard2019}. Write
$\ell(\varepsilon)=\log(e\max\{1,\varrho/\sigma_N,
B_f/(\sigma_{f,+}\sigma_N)\}/\varepsilon_{\rm rel})$.
The counts below bound the logarithm of the number of quantized values.
\begin{itemize}\setlength\itemsep{1pt}
\item The $N$ centers in $B_D(\varrho)$ and tangent planes, to accuracy
$\delta$ and $\theta$: at most $CND\,\ell(\varepsilon)$, by volumetric
bounds.
\item The $N$ graph maps $g$, which lie in a ball of $C^3$ maps with values
in $\R^{D-m}$, to accuracy $\theta$ in $C^1$. Degree-two Taylor polynomials
on a grid of mesh $R\sqrt\theta$ suffice, as in
\citet{KolmogorovTikhomirov1959} for smoothness $3-1=2$; quantizing their
coefficient vectors in $\R^{D-m}$ by volumetric nets costs at most
$CN(D-m)\,\theta^{-m/2}\log(e/\theta)$.
\item The $N$ link maps $\varphi$, which lie in a ball of $C^{1,1}$ maps
with values in $\R^p$, to accuracy $\varepsilon$ in the uniform norm.
Degree-one Taylor polynomials on a grid of mesh
$\sigma_N\sqrt{\varepsilon_{\rm rel}}$ suffice, with coefficient
vectors in $\R^p$ quantized in the same way, at a cost of at most
$CNp\,(R/\sigma_N)^m\varepsilon_{\rm rel}^{-m/2}\ell(\varepsilon)$.
\end{itemize}
If two pairs have the same quantized data, corresponding points with the same chart coordinate satisfy the position and tangent-projector bounds of Lemma~\ref{lem:stab} with $C\delta,C\theta$ in place of $\delta,\theta$; their link values differ by at most $C\varepsilon$. Choose one pair
$(\widetilde\M,\tilde f)$ of $\mathcal F$ for each value of the quantized
data, and let its net function be $\tilde f\circ\Pi_{\widetilde\M}$ on the
tube of radius $(1+c)R/(2c)$ around $\widetilde\M$, where the projection is
still defined, and zero elsewhere. By Lemma~\ref{lem:stab} these functions
form a $C\varepsilon$-net. Since $\varrho\ge R/c$,
$\log(e/\theta)\le C\ell(\varepsilon)$. Since $R\ge\sigma_N$ and
\[
 N\theta^{-m/2}\lesssim\frac{v_{\max}}{\sigma_N^m}
 \Bigl(\frac{\sigma_N}{R}\Bigr)^{m/2}\varepsilon_{\rm rel}^{-m/2},
\]
we obtain
\begin{equation}\label{eq:entropy}
 \log|\mathcal G_\varepsilon|\le
 C\,(D+p)\,\frac{v_{\max}}{\sigma_N^m}\,\varepsilon_{\rm rel}^{-m/2}\,
 \ell(\varepsilon).
\end{equation}
The manifold and the link have the same exponent $m/2$: the manifold is
needed only through its tangent planes, and for a $C^3$ manifold these are
as smooth as a $C^{1,1}$ link. Inserting \eqref{eq:entropy} in
\eqref{eq:ls} with
\[
 \varepsilon_{\rm rel}^{(m+4)/2}=\frac{\Lambda(D+p)\,v_{\max}}{\sigma_N^m\,n}
\]
proves Proposition~\ref{prop:learn}. At this resolution, \eqref{eq:entropy} gives
\[
 \log|\mathcal G_\varepsilon|\le
 C\bigl((D+p)\,v_{\max}/\sigma_N^m\bigr)^{4/(m+4)}
 (n/\Lambda)^{m/(m+4)}\ell_n .
\]
This bounds the logarithm of the number of candidates. Direct evaluation
of all empirical losses uses $n|\mathcal G_\varepsilon|$ pointwise function
evaluations, assuming access to the representatives. A runtime bound would
additionally require an algorithm to construct these representatives and
evaluate their nearest-point projections; the entropy argument provides
only their existence.

\subsection{Proof of the lower bound}\label{app:learnability-lower}
Restrict the supremum to $\M=\M_0$, to $Q$ uniform on $\M_0$ with $Z$
uniform on the normal ball of radius $R$ given $Q$, and to Gaussian noise
$N(0,\sigma^2I_p)$ with $\sigma$ a fixed fraction of $\sigma_\zeta$. In
this subproblem $\M_0$ is known, so $Q=\Pi_{\M_0}X$ is observed and $Z$
carries no information about $f$: the subproblem is regression of $f$ on
$\M_0$ with uniform design, since averaging an estimator over the
normal fiber does not increase its risk, by Jensen's inequality. For $b\le\sigma_N/6$, which is less than
$R/(10c)$, take $N_0\asymp v_0/b^m$ disjoint geodesic balls of radius $b$ in
$\M_0$ with centers $q_k$, a fixed smooth bump $\psi$ on the unit ball, a
unit vector $e\in\R^p$, a small fixed $\lambda_0>0$, and
\[
 f_\vartheta=\sum_{k\le N_0}\vartheta_k\Phi_k,\qquad
 \Phi_k(q)=\lambda_0\frac{\sigma_{f,+}}{\sigma_N}b^2\,
 \psi\bigl(2P_{q_k}(q-q_k)/b\bigr)\,e,\qquad\vartheta\in\{0,1\}^{N_0} ,
\]
with $\Phi_k$ given by this formula on the $k$-th ball and zero
elsewhere.
For $\lambda_0$ small, depending only on $m$, $c$ and $C_{\mathcal F}$, every
$f_\vartheta$ belongs to $\mathcal L(\M_0)$, by the chart bound. If
$\vartheta$ and $\vartheta'$ differ in one coordinate,
$\|f_\vartheta-f_{\vartheta'}\|_{L^2}^2$ under the uniform law is of order
\[
 (\lambda_0\sigma_{f,+}b^2/\sigma_N)^2\,b^m/v_0
\]
and the Kullback--Leibler divergence of the samples is $n/(2\sigma^2)$
times this quantity. Choose $b$ so that this divergence is one, that is
\[
 b^{m+4}\asymp\frac{\sigma^2\sigma_N^2v_0}{n\lambda_0^2\sigma_{f,+}^2};
\]
the condition on $n$ ensures $b\le\sigma_N/6$. Assouad's lemma
\citep[Theorem~2.12(iv)]{Tsybakov2009}, applied to the bumps of
disjoint supports, bounds the risk below by a constant
times $N_0$ such squared distances, that is by a constant times
$(\lambda_0\sigma_{f,+}/\sigma_N)^2b^4$, which is the claim. This is the
argument of \citet{Stone82} and of \citet[Theorem~3.2]{GKKW02} in a chart.

\subsection{Two remarks on the approximating family}
\label{app:learnability-remarks}

\begin{remark}[Known manifold and anisotropic classes]
If $\M$ is known, the degrees of freedom are those of the link alone. In
tubular coordinates $x=q+z$ the function $F(q+z)=f(q)$ does not depend on
the $D-m$ normal variables, so it belongs to an anisotropic H\"older class
with smoothness $s=2$ in $m$ variables and arbitrary smoothness in the
others. Its effective smoothness $\bar s$, given by
$1/\bar s=\sum_i1/s_i=m/s$ with $s_i$ the smoothness in the $i$th variable, yields the classical anisotropic rate
$n^{-2\bar s/(2\bar s+1)}=n^{-2s/(2s+m)}$ \citep{HoffmannLepski2002}. The
content of Lemma~\ref{lem:stab} and of the count above is that the unknown
change of coordinates costs no more entropy than the link. For general
smoothness, a $C^s$ link with $s\ge1$ on a $C^{s+1}$ manifold gives the exponent
$2s/(2s+m)$ by the same argument.
\end{remark}

\begin{remark}[Flat pieces can lose one order]
Pieces that are polynomial in the tangent coordinates $P_{q_j}(x-q_j)$ at
points $q_j$ of a net of $\M$, and constant along the normal space at
$q_j$, in general approximate $F$ only to first order in the spacing $b$ of the net.
For $x=q+z$ with $z$ normal at $q$,
\[
 P_{q_j}(x-q_j)=P_{q_j}(q-q_j)+(P_{q_j}-P_q)z,
\]
and the last term has size up to $c\,d_\M(q,q_j)$: the normal fibers over
nearby base points are tilted against the normal space at $q_j$. This is
the amplification $(I-S_z)^{-1}$ of the tangential gradient in
Section~\ref{sec:core-gradient}. In this worst case, a construction
with a fixed normal space on each patch needs
$b\asymp\varepsilon$, hence $\varepsilon^{-m}$ pieces instead of
$\varepsilon^{-m/2}$, and gives only the rate $n^{-2/(m+2)}$. Projecting
onto an approximating manifold, as in Lemma~\ref{lem:stab}, follows the
variation of the normal spaces with the base point; each normal fiber
itself is an affine disk.
\end{remark}

\end{document}